%% file: main.tex
\documentclass{article}

\usepackage[nonatbib,final]{neurips_2023}

\usepackage[utf8]{inputenc} 
\usepackage[T1]{fontenc}    
\usepackage{hyperref}       
\usepackage{url}            
\usepackage{booktabs}       
\usepackage{amsfonts}       
\usepackage{nicefrac}       
\usepackage{microtype}      
\usepackage{xcolor}         

\usepackage{caption}
\usepackage{subcaption}
\usepackage{adjustbox}
\usepackage{wrapfig}
\usepackage{float}
\usepackage{array}
\usepackage{booktabs}
\newcolumntype{N}{>{\centering\arraybackslash}m{.5in}}
\newcolumntype{G}{>{\centering\arraybackslash}m{2in}}
\usepackage{pifont}
\usepackage[square,numbers]{natbib}
\title{Centralized Training with Hybrid Execution in Multi-Agent Reinforcement Learning}

\author{%
  Pedro P. Santos, Diogo S. Carvalho, Miguel Vasco, \\
  \textbf{Alberto Sardinha, Pedro A. Santos, Ana Paiva \& Francisco S. Melo} \\
  INESC-ID \& Instituto Superior Técnico, University of Lisbon\\
  \texttt{pedro.pinto.santos@tecnico.ulisboa.pt} \\
}

\begin{document}

\maketitle

\begin{abstract}
We introduce \emph{hybrid execution} in multi-agent reinforcement learning (MARL), a new paradigm in which agents aim to successfully complete cooperative tasks with arbitrary communication levels at execution time by taking advantage of information-sharing among the agents. Under hybrid execution, the communication level can range from a setting in which no communication is allowed between agents (fully decentralized), to a setting featuring full communication (fully centralized), but the agents do not know beforehand which communication level they will encounter at execution time. To formalize our setting, we define a new class of multi-agent partially observable Markov decision processes (POMDPs) that we name hybrid-POMDPs, which explicitly model a communication process between the agents. We contribute MARO, an approach that makes use of an auto-regressive predictive model, trained in a centralized manner, to estimate missing agents' observations at execution time. We evaluate MARO on standard scenarios and extensions of previous benchmarks tailored to emphasize the negative impact of partial observability in MARL. Experimental results show that our method consistently outperforms relevant baselines, allowing agents to act with faulty communication while successfully exploiting shared information.
\end{abstract}

\section{Introduction}
Multi-agent reinforcement learning (MARL) aims to learn utility-maximizing behavior in scenarios involving multiple agents. In recent years, deep MARL methods have been successfully applied to multi-agent tasks such as game-playing \citep{papoudakis_2020}, traffic light control \citep{wei_2019}, or energy management \citep{fang_2020}. Despite recent successes, the multi-agent setting happens to be substantially harder than its single-agent counterpart \citep{canese_2021} because multiple concurrent learners can create non-stationarity conditions that hinder learning; the curse of dimensionality obstructs centralized approaches to MARL due to the exponential growth in state and action spaces with the number of agents; and agents seldom observe the true state of the environment. 

As a way to deal with the exponential growth in the state/action space and with environmental constraints, both in perception and actuation, existing methods aim to learn decentralized policies that allow the agents to act based on local perceptions and partial information about other agents' intentions. The paradigm of \emph{centralized training with decentralized execution} is undoubtedly at the core of recent research in the field \citep{oliehoek_2011,rashid_2018,foerster_2016}; such paradigm takes advantage of the fact that additional information, available only at training time, can be used to learn decentralized policies in a way that the need for communication is alleviated.

While in some settings partial observability and/or communication constraints require learning fully decentralized policies, the assumption that agents cannot communicate at execution time is often too strict for a great number of real-world application domains such as robotics, game-playing or autonomous driving~\citep{ho2019multi,yurtsever2020survey}. In such domains, learning fully decentralized policies should be deemed too restrictive since such policies do not take into account the possibility of communication between the agents. Other MARL strategies, which do take advantage of additional information shared among the agents, can surely be developed \citep{zhu_2022}.

In this work, we propose RL agents that are able to exploit the benefits of centralized training while, simultaneously, taking advantage of information-sharing at execution time. We introduce the paradigm of \textit{hybrid execution}, in which agents act in scenarios with arbitrary (but unknown) communication levels that can  range from no communication (fully decentralized) to full communication between the agents (fully centralized). In particular, we consider scenarios with faulty communication during execution, in which agents passively share their local observations to perform partially observable cooperative tasks. To formalize our setting, we start by defining \emph{hybrid partially observable Markov decision process} (H-POMDP), a new class of multi-agent POMDPs that explicitly considers a communication process between the agents. We then propose a novel method that allows agents to solve H-POMDPs regardless of the communication process encountered at execution time. Specifically, we propose \emph{multi-agent observation sharing under communication dropout} (MARO). MARO can be easily integrated with current deep MARL methods and comprises an auto-regressive model, trained in a centralized manner, that explicitly predicts non-shared information from past observations of the agents.

We evaluate the performance of MARO across different communication levels, in different MARL benchmark environments and using multiple RL algorithms. Furthermore, we introduce novel MARL environments that explicitly require communication during execution to successfully perform cooperative tasks, currently missing in the literature. Experimental results show that our method consistently outperforms the baselines, allowing agents to exploit shared information during execution and perform tasks under various communication levels.

In summary, our contributions are three-fold: (i) we propose and formalize the setting of hybrid execution in MARL, in which agents must perform partially-observable cooperative tasks across all possible communication levels; (ii) we propose MARO, an approach that makes use of an autoregressive predictive model of agents' observations; and (iii) we evaluate MARO in multiple environments using different RL algorithms, showing that our approach consistently allows agents to act with different communication levels.

\section{Hybrid Execution in Multi-Agent Reinforcement Learning}
\label{sec:hybrid_MARL}
A fully cooperative multi-agent system with Markovian dynamics can be modeled as a decentralized partially observable Markov decision process (Dec-POMDP) \citep{oliehoek2016concise}. A Dec-POMDP is a tuple $([n], \mathcal{X}, \mathcal{A}, \mathcal{P}, r, \gamma, \mathcal{Z}, \mathcal{O})$, where $[n] = \{1, \ldots, n\}$ is the set of indexes of $n$ agents, $\mathcal{X}$ is the set of states of the environment, $\mathcal{A} = \times_i \mathcal{A}_i$ is the set of joint actions, where $\mathcal{A}_i$ is the set of individual actions of agent $i$, $\mathcal{P}$ is the set of probability distributions over next states in $\mathcal{X}$, one for each state and action in $\mathcal{X}\times \mathcal{A}$, $r: \mathcal{X}\times\mathcal{A} \to \mathbb{R}$ maps states and actions to expected rewards, $\gamma \in [0, 1[$ is a discount factor, $\mathcal{Z} = \times_i \mathcal{Z}_i$ is the set of joint observations, where $\mathcal{Z}_i$ is the set of local observations of agent $i$, and $\mathcal{O}$ is the set of probability distributions over joint observations in $\mathcal{Z}$, one for each state and action in $\mathcal{X} \times \mathcal{A}$. A decentralized policy for agent $i$ is $\pi_i: \mathcal{Z}_i \rightarrow \mathcal{A}_i$ and the joint decentralized policy is $\pi:\mathcal{Z}\to\mathcal{A}$ such that $\pi(z_1, \ldots, z_n) = \big(\pi_1(z_1), \ldots, \pi_n(z_n)\big)$.

Fully decentralized approaches to MARL directly apply standard single-agent RL algorithms for learning each agent's policy $\pi_i$ in a decentralized manner. In independent $Q$-learning (IQL)~\citep{tan1993multi}, each agent treats other agents as being part of the environment, ignoring the influence of other agents' observations and actions. Similarly, independent proximal policy optimization (IPPO), an adaptation of the PPO algorithm~\cite{schulman2017proximal}, learns fully decentralized critic and actor networks, neglecting the influence of other agents. More recently, under the paradigm of centralized training with decentralized execution, QMIX~\citep{rashid_2018} aims at learning decentralized policies with centralization at training time while fostering cooperation among the agents. Multi-agent PPO (MAPPO)~\cite{yu_2021} learns decentralized actors using a centralized critic during training. Finally, if we know that all agents can share their local observations among themselves at execution time, we can use any of the approaches above to learn fully centralized policies.

None of the aforementioned classes of methods assumes, however, that agents may sometimes have access to other agents' observations and sometimes not. Therefore, decentralized agents are unable to take advantage of the additional information that they may receive from other agents at execution time, and centralized agents are unable to act when the sharing of information fails. In this work, we introduce hybrid execution in MARL, a setting in which agents act regardless of the communication process while taking advantage of additional information they may receive during execution. To formalize this setting, we define a new class of multi-agent POMDPs that we name hybrid-POMDPs (H-POMDPs), which explicitly considers a specific communication process among the agents.

\subsection{Hybrid Partially Observable Markov Decision Processes}
\label{sec:H-POMDPs}
We define a hybrid-POMDP (H-POMDP) as a tuple $([n], \mathcal{X}, \mathcal{A}, \mathcal{P}, r, \gamma, \mathcal{Z}, \mathcal{O}, C)$ where, in addition to the tuple that describes the Dec-POMDP, we consider a $n \times n$ communication matrix $C$ such that $[C]_{i, j}=p_{i, j}$ is the probability that, at a certain time step, agent $i$ has access to the local observation of agent $j$ in $\mathcal{Z}_j$. 
H-POMDPs generalize both the notion of decentralized execution and centralized execution in MARL. Specifically, for a given Dec-POMDP, we can consider $C$ as the identity matrix to capture fully decentralized execution or as a matrix of ones to capture fully centralized execution. 

In our setting, we assume that at execution time agents will face an H-POMDP with an unknown communication matrix $C$, sampled from a set $\mathcal{C}$ according to an unknown probability distribution $\mu$. The performance of the agent is measured as $J_\mu(\pi) = \mathbb{E}_{C \sim \mu}\left[J(\pi;C)\right]$, where $J(\pi;C)$ denotes the expected discounted cumulative reward under an H-POMDP with communication matrix $C$. At training time, agents may have access to the fully centralized H-POMDP. Therefore, the setting we consider is one of centralized training with hybrid execution and an unknown communication process. 

We note here that every H-POMDP has a corresponding Dec-POMDP, which can be obtained by adequately changing the observation space $\mathcal{Z}$ and the set of emission probability distributions $\mathcal{O}$. Consequently, any reinforcement learning method can be trained to solve a specific H-POMDP, with a specific communication matrix $C$, by solving the corresponding Dec-POMDP. However, we seek to find a method that takes explicit advantage of the characteristics of hybrid execution to be able to act on H-POMDPs regardless of the matrix $C$ that models the communication process at execution time. To the best of our knowledge, there exists no method that addresses our problem.

\section{Multi-Agent Observation Sharing under Communication Dropout}

\begin{figure*}[t]
    \centering
    \begin{subfigure}[b]{0.49\textwidth}
        \centering
        \includegraphics[height=4.2cm]{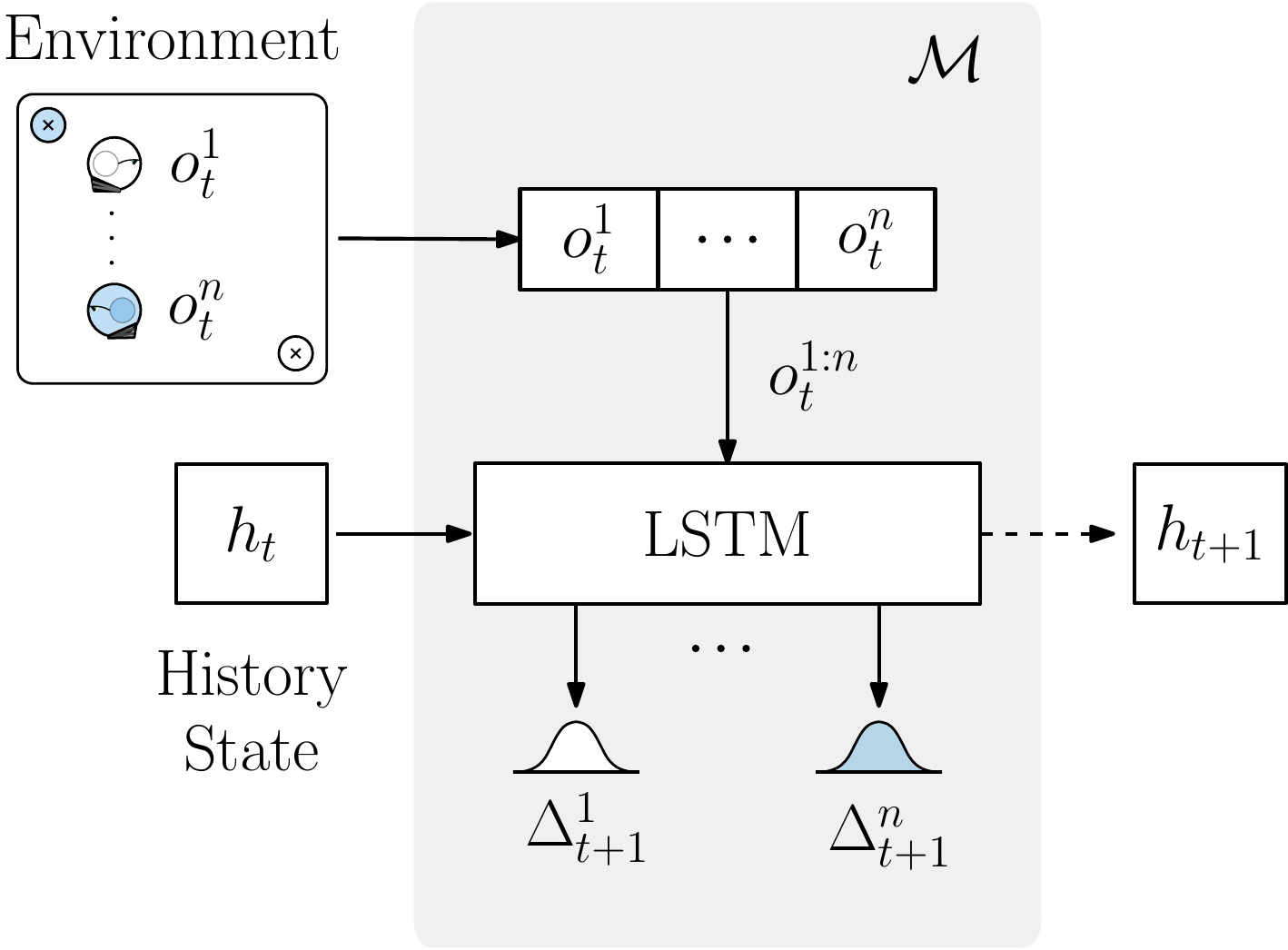}
        \caption{Training time.}
        \label{fig:methods:maro:training}
    \end{subfigure}
    \hfill
    \begin{subfigure}[b]{0.49\textwidth}
        \centering
        \includegraphics[height=3.4cm]{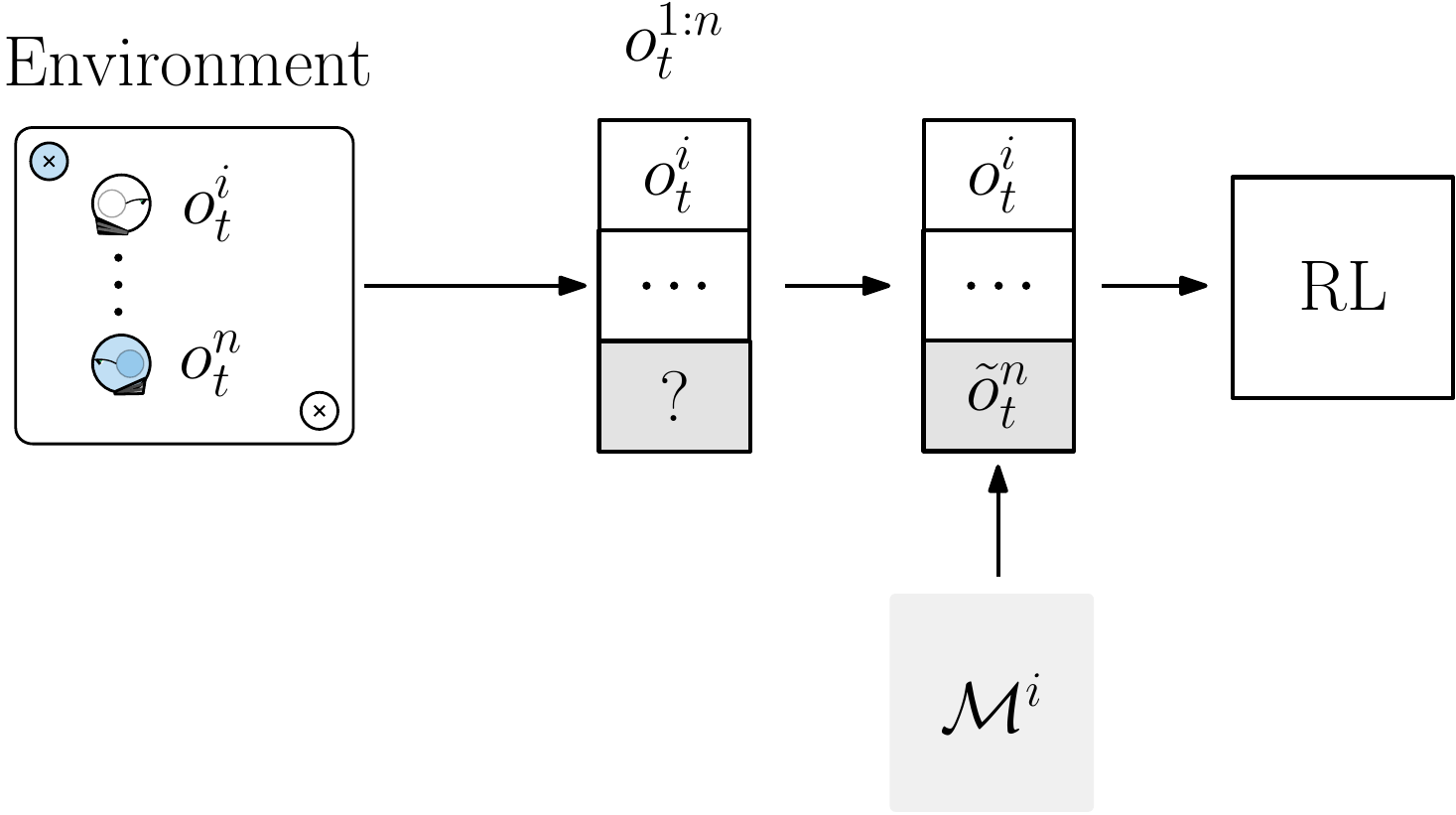}
        \caption{Execution time.}
        \label{fig:methods:maro:execution}
    \end{subfigure}
    \caption{MARO approach for hybrid execution: (a) at training time, an autoregressive predictive model $\mathcal{M}$ learns to estimate observation deltas ${p(\Delta^{1:n}_{t} \mid o^{1:n}_{t}, h_{t})}$ from previous observations $o^{1:n}_{t}$ and a history variable $h_{t}$; and (b) at execution time, an agent-specific predictive model, $\mathcal{M}^i$, predicts missing agents' observations. More details in the main text.}
    \label{fig:methods:maro}
\end{figure*}

While acting on an H-POMDP, agents may not have access to the perceptual information of all agents due to a faulty communication process. We propose MARO, a novel approach to exploit shared information and overcome communication issues during task execution. MARO comprises an autoregressive predictive model that estimates missing information from previous observations.

We set up the RL controller of each agent, i.e., the $Q$-network associated with each agent for the IQL and QMIX algorithms, and the actor network associated with each agent for the IPPO and MAPPO algorithms, to receive as input the joint observation ${o^{1:n}_t = \{o^1_t, \ldots,o^n_t\}}$, where $o^i_t$ is the observation of the $i$-th agent at timestep $t$. In order to overcome communication failures during execution, we train a predictive model $\mathcal{M}$ to impute the non-shared observations $\tilde{o}_t^{i}$, $i \in [n]$.

\paragraph{Training time}
We learn a transition model, ${p(\Delta^{1:n}_{t} \mid o^{1:n}_{t}, h_{t})}$, depicted in Fig. \ref{fig:methods:maro:training}, that given the current observations $o^{1:n}_{t}$ and some history variable $h_{t}$ is able to predict the next-step observations as $o^{1:n}_{t+1} = o^{1:n}_{t} + \Delta^{1:n}_{t}$, where $\Delta^{1:n}_{t}$ corresponds to the predicted deltas of the observations. We learn a single predictive model in a fully centralized and supervised fashion. We instantiate ${p_{\theta}(\Delta^{1:n}_{t} \mid o^{1:n}_{t}, h_{t})}$ as an LSTM, parameterized by $\theta$, with:
\begin{equation}
p_{\theta}(\Delta^{1:n}_{t} \mid o^{1:n}_{t}, h_{t}) = \prod_{i=1}^{n} p_{\theta}(\Delta^{i}_{t} \mid o^{1:n}_{t}, h_{t}),
\end{equation}
where $p_{\theta}(\Delta^{i}_{t} \mid o^{1:n}_{t}, h_{t})$ is the Gaussian distribution of the predicted deltas for the $i$-th agent. We train the predictive model and RL controllers simultaneously: we consider single-step transitions $(o_{t}^{1:n}, \Delta_{t}^{1:n})$, with $\Delta_{t}^{1:n} = o_{t+1}^{1:n} - o_{t}^{1:n}$, and evaluate the negative log-likelihood of the target next-step deltas $\Delta_{t}^{1:n}$, given the estimated next-step deltas distribution $p_{\theta}(\cdot \mid o^{1:n}_{t}, h_t)$:
\begin{equation}
    \mathcal{L}_{\mathcal{M}}(o^{1:n}_t, \Delta_{t}^{1:n}) = - \sum_{i=1}^{n} \log p_{\theta}(\Delta^{i}_{t} \mid o^{1:n}_{t}, h_{t}).
\end{equation}

\paragraph{Execution time}
We provide each agent with an independent instance of the predictive model $\mathcal{M}^i$, which updates the estimated joint-observations in the perspective of the agent ${\tilde{o}_t^{1:n, i} = \{\tilde{o}_t^{1, i}, \ldots, \tilde{o}_t^{n, i}\}}$ and maintains an agent-specific history state $h_{t}^{i}$. As depicted in Fig. \ref{fig:methods:maro:execution}, we use the predictive model $\mathcal{M}^i$ to impute missing observations.

\section{Evaluation}
\label{sec:evaluation}
In this section, we evaluate our approach for hybrid execution against relevant baselines under multiple MARL algorithms. We show that the core component of MARO, i.e., the predictive model, allows the execution of tasks across multiple communication levels, outperforming baselines. We start by describing our experimental scenarios and baselines in Sec.~\ref{sec:evaluation:scenarios} and Sec.~\ref{sec:evaluation:baselines}, respectively. In Sec.~\ref{sec:evaluation:results}, we present our main experimental results.

\subsection{Experimental Scenarios}
\label{sec:evaluation:scenarios}
We focus our evaluation on multi-agent cooperative environments. As discussed by \citet{papoudakis2021benchmarking}, the main challenges in current MARL benchmark scenarios majorly involve coordination, large action space, sparse reward and non-stationarity. Thus, in order to emphasize the impact of information sharing among the agents, we contribute the following environments (adapted from \citep{lowe2017multi}):
\begin{itemize}
    \item \textbf{HearSee} (HS): Two heterogeneous agents cover a single landmark in a 2D map. The ``Hear'' agent observes the absolute position of the landmark, but it does not have access to its own position in the environment. The ``See'' agent observes the position and velocities of both agents, yet does not have access to the position of the landmark.
    \item \textbf{SpreadXY-2} (SXY-2): Two heterogeneous agents cover two designated landmarks in a 2D map while avoiding collisions. In this scenario, one of the agents has access to the X-axis position and velocity of both agents, while the other agent has access to the Y-axis position and velocity of both agents. Both agents observe the landmarks' absolute position;
    \item \textbf{SpreadXY-4} (SXY-4): Similar to the scenario above but with two teams of two agents;
    \item \textbf{SpreadBlindfold} (SBF): Three agents cover three designated landmarks in a 2D map while avoiding collisions. Each agent's observation only includes its own position and velocity and the absolute position of all landmarks;
\end{itemize}

In addition to the proposed environments, we evaluate our approach in the standard \textbf{SpeakerListener} (SL) environment from \cite{lowe2017multi}, as well as the \textbf{Level-Based Foraging} (Foraging-2s-15x15-2p-2f-coop-v2) (LBF) environment \cite{papoudakis2021benchmarking}, which we modified to comprise the absolute positions of the agents. For some scenarios in standard benchmarks, such as the Multi-Agent Particle Environment \cite{lowe2017multi},
or Level-Based Foraging \cite{papoudakis2021benchmarking},
we observed no advantage in allowing observation sharing between the agents even without considering communication failures (more details in Appendix \ref{appendix:experimental_evaluation:scenarios}). Thus, we did not consider such environments in this work. For a complete description of the scenarios, as well as additional details regarding the choice of the environments used, we refer to Appendix \ref{appendix:experimental_evaluation:scenarios}.

Finally, we consider H-POMDPs with communication matrices such that each agent $i$ can always access its own local observation, i.e., $p_{i, i}=1$, and the communication matrix is symmetric between agents $i$ and $j$, i.e., $p_{i, j}=p_{j, i}$. To simplify the exposition and the evaluation, we use the same $p_{i, j}=p$ for all pairs of different agents $i$, $j$. Therefore, we use $p$ to unambiguously denote the communication level of a given H-POMDP. Nevertheless, we perform a comparative study between different sampling schemes for the communication matrix in Sec.~\ref{sec:evaluation:results:comm_protocols}, highlighting the robustness of MARO under different communication settings.

\subsection{Baselines and Experimental Methodology}
\label{sec:evaluation:baselines}
We compare MARO against the following baselines, which do not make use of a predictive model and perform constant imputation of missing observations:
\begin{itemize}
    \item \textbf{Observation} (Obs.): Agents only have access to their own observations and are unable to communicate with other agents during execution. Corresponds to standard MARL algorithms designed for decentralized execution.
    \item \textbf{Masked Joint-Observation} (Masked j. obs.): During the centralized training phase, the RL controllers receive as input the concatenation of the observations of all agents. At execution-time, missing observations are replaced with a vector of zeros.
    \item \textbf{Message-Dropout} (MD): During the centralized training phase, the RL controllers receive as input the concatenation of the observations of all agents, but a dropout-based mechanism randomly drops some of the observations (i.e., replaces them with a vector of zeros) according to $p \sim \mathcal{U}(0,1)$. At execution-time, missing observations are replaced with a vector of zeros. This baseline is adapted from~\cite{kim_2019_dropout}.
    \item \textbf{Message-Dropout w/ masks} (MD w/ masks): This baseline is similar to the MD baseline, but additionally appends to the input of the RL controllers a set of binary flags encoding whether the observations of the agents are missing or not. The masks give additional context to the RL agent regarding the validity of the entries in the vector of observations.
\end{itemize}

All baselines above can be used in the context of hybrid execution. Additionally, we consider an \textbf{Oracle} baseline under which all agents have access to the observations of all agents both during training and execution. Such oracle baseline corresponds to standard MARL algorithms designed for centralized execution, however, it is unable to perform when communication fails. We use the Oracle baseline to better contextualize the performance of the methods developed for hybrid execution against an optimal setting featuring no communication failures.

We employ the same RL controller networks across all evaluations. The RL networks include recurrent layers to mitigate the effects of partial observability. We consider four different MARL algorithms: IQL, QMIX, IPPO, and MAPPO. We perform 3 training runs for each experimental setting and 100 evaluation rollouts for each training run. We report, both in tables and plots, the 95\% bootstrapped confidence interval alongside the corresponding scalar mean value. We assume that $p=1$ at $t=0$ for all algorithms. The algorithms are evaluated for $p \sim \mathcal{U}(0,1)$ whenever the communication level is not explicitly referred, or for a given fixed communication level $p$ when explicitly specified. The Oracle baseline is always evaluated with $p=1$. We refer to Appendix~\ref{appendix:experimental_evaluation:exprimental_methodology} for a complete description of the experimental methodology, including hyperparameters of the predictive model and the RL controllers, as well as the code used for this work.

\subsection{Results}
\label{sec:evaluation:results}

\begin{table}[t]
\caption{Mean episodic returns for the value-based algorithms in all scenarios. Higher is better.}
\label{table:evaluation:main_results_value_based}
\centering
\noindent
\resizebox{\linewidth}{!}{%
\begin{tabular}{N N N N N N N N N N N}\toprule
\multicolumn{1}{N }{\textbf{}} & \multicolumn{5}{c }{\textbf{IQL}} & \multicolumn{5}{c }{\textbf{QMIX}} \\  
\cmidrule(lr){2-6}
\cmidrule(ll){7-11}
\multicolumn{1}{ l }{\textbf{Env.}} & \textbf{Obs.} & \textbf{Masked j. obs.} & \textbf{MD} & \textbf{MD w/ masks} & \textbf{MARO} & \textbf{Obs.} & \textbf{Masked j. obs.} & \textbf{MD} & \textbf{MD w/ masks} & \textbf{MARO} \\ 
\cmidrule(rr){1-1}\cmidrule(lr){2-6}\cmidrule(ll){7-11}
\multicolumn{1}{ l }{SL} & -40.0 \tiny{(-0.4,+0.4)} & -45.3 \tiny{(-1.2,+1.9)} & \textbf{-25.4} \tiny{(-0.6,+1.1)} & -25.5 \tiny{(-0.6,+1.1)} & \textbf{-25.3} \tiny{(-0.6,+1.0)} & -24.9 \tiny{(-0.1,+0.0)}  & -40.5 \tiny{(-0.8,+0.8)} & \textbf{-25.2} \tiny{(-0.6,+1.1)} & \textbf{-25.2} \tiny{(-0.6,+1.1)} & \textbf{-25.1} \tiny{(-0.6,+1.2)} \\
\cmidrule(rr){1-1}\cmidrule(lr){2-6}\cmidrule(ll){7-11}
\multicolumn{1}{ l }{HS} & -114.5 \tiny{(-2.0,+1.6)} & -64.0 \tiny{(-2.1,+2.5)} & -34.8 \tiny{(-1.6,+1.7)} & -34.1 \tiny{(-1.3,+2.5)} & \textbf{-29.6} \tiny{(-1.0,+0.7)} & -62.2 \tiny{(-1.9,+1.4)} & -67.4 \tiny{(-4.6,+3.0)} & \textbf{-29.2} \tiny{(-1.3,+0.9)} & \textbf{-29.1} \tiny{(-0.9,+0.8)} & \textbf{-28.8} \tiny{(-1.2,+1.9)} \\
\cmidrule(rr){1-1}\cmidrule(lr){2-6}\cmidrule(ll){7-11}
\multicolumn{1}{ l }{SXY-2} & -199.6 \tiny{(-0.9,+0.9)} & -202.9 \tiny{(-2.3,+3.7)} & -165.2 \tiny{(-0.5,+0.7)} & -160.7 \tiny{(-1.7,+1.6)} & \textbf{-148.0} \tiny{(-0.4,+0.5)} & -177.8 \tiny{(-7.6,+4.1)}  & -201.2 \tiny{(-2.3,+2.3)} & -157.2 \tiny{(-1.4,+0.7)} & -154.5 \tiny{(-0.5,+1.0)} & \textbf{-145.7} \tiny{(-0.9,+0.5)} \\  
\cmidrule(rr){1-1}\cmidrule(lr){2-6}\cmidrule(ll){7-11}
\multicolumn{1}{ l }{SXY-4} & -1225.5 \tiny{(-4.2,+4.9)} & -1161.2 \tiny{(-7.6,+9.4)} & -1157.0 \tiny{(-1.2,+1.0)} & -1164.5 \tiny{(-10.8,+13.3)} & \textbf{-988.3} \tiny{(-21.7,+37.4)} & -1132.6 \tiny{(-6.6,+5.9)} & -1146.4 \tiny{(-12.7,+22.1)} & -1024.6 \tiny{(-39.5,+54.9)} & -1014.0 \tiny{(-40.4,+40.6)} & \textbf{-850.0} \tiny{(-22.5,+17.0)} \\
\cmidrule(rr){1-1}\cmidrule(lr){2-6}\cmidrule(ll){7-11}
\multicolumn{1}{ l }{SBF} & -425.1 \tiny{(-1.1,+1.4)} & -415.5 \tiny{(-7.5,+4.1)} & \textbf{-401.2} \tiny{(-6.5,+8.5)} & \textbf{-400.5} \tiny{(-5.5,+8.2)} & \textbf{-399.3} \tiny{(-5.2,+6.4)} & -416.1 \tiny{(-10.0,+7.7)} & -407.3 \tiny{(-1.9,+1.5)} & -401.4 \tiny{(-4.9,+3.5)} & -398.3 \tiny{(-3.0,+2.3)} & \textbf{-382.3} \tiny{(-5.2,+5.7)} \\ 
\cmidrule(rr){1-1}\cmidrule(lr){2-6}\cmidrule(ll){7-11}
\multicolumn{1}{ l }{LBF} & 0.38 \tiny{(-0.03,+0.01)} & 0.19 \tiny{(-0.02,+0.01)} & \textbf{0.53} \tiny{(-0.03,+0.02)} & \textbf{0.52} \tiny{(-0.02,+0.02)} & 0.35 \tiny{(-0.01,+0.02)} & 0.55 \tiny{(-0.01,+0.0)} & 0.25 \tiny{(-0.05,+0.03)} & \textbf{0.58} \tiny{(-0.01,+0.02)} & \textbf{0.59} \tiny{(-0.02,+0.02)} & 0.44 \tiny{(-0.02,+0.02)} \\
\bottomrule
\end{tabular}
}
\end{table}

\begin{table}[t]
\caption{Mean episodic returns for the actor critic-based algorithms in all scenarios. Higher is better.}
\label{table:evaluation:main_results_actor_critic}
\centering
\noindent
\resizebox{\linewidth}{!}{%
\begin{tabular}{N N N N N N N N N N N}\toprule
\multicolumn{1}{N }{\textbf{}} & \multicolumn{5}{c }{\textbf{IPPO}} & \multicolumn{5}{c }{\textbf{MAPPO}} \\  
\cmidrule(lr){2-6}
\cmidrule(ll){7-11}
\multicolumn{1}{ l }{\textbf{Env.}} & \textbf{Obs.} & \textbf{Masked j. obs.} & \textbf{MD} & \textbf{MD w/ masks} & \textbf{MARO} & \textbf{Obs.} & \textbf{Masked j. obs.} & \textbf{MD} & \textbf{MD w/ masks} & \textbf{MARO} \\ 
\cmidrule(rr){1-1}\cmidrule(lr){2-6}\cmidrule(ll){7-11}
\multicolumn{1}{ l }{SL} & -33.3 \tiny{(-11.6,+5.9)} & -39.5 \tiny{(-0.8,+0.7)} & -45.2 \tiny{(-5.0,+5.0)} & -47.0 \tiny{(-3.2,+4.0)} & \textbf{-25.2} \tiny{(-0.1,+0.1)} & -59.6 \tiny{(-0.5,+0.6)} & -39.3 \tiny{(-1.2,+1.0)} & -28.2 \tiny{(-0.3,+0.3)} & -27.5 \tiny{(-0.4,+0.5)} & \textbf{-25.2} \tiny{(-0.1,+0.1)} \\
\cmidrule(rr){1-1}\cmidrule(lr){2-6}\cmidrule(ll){7-11}
\multicolumn{1}{ l }{HS} & -114.1 \tiny{(-37.7,+27.7)} & -81.8 \tiny{(-5.8,+7.0)} & -102.9 \tiny{(-20.9,+22.5)} & -101.9 \tiny{(-19.4,+20.1)} & \textbf{-31.4} \tiny{(-1.4,+1.2)} & -70.5 \tiny{(-18.3,+10.1)} & -82.2 \tiny{(-6.4,+3.3)} & -32.4 \tiny{(-0.0,+0.0)} & \textbf{-30.9} \tiny{(-3.5,+2.6)} & \textbf{-31.4} \tiny{(-1.4,+1.2)} \\
\cmidrule(rr){1-1}\cmidrule(lr){2-6}\cmidrule(ll){7-11}
\multicolumn{1}{ l }{SXY-2} & -235.6 \tiny{(-0.6,+0.6)} & -214.8 \tiny{(-5.4,+4.6)} & -184.9 \tiny{(-5.8,+3.6)} & -175.9 \tiny{(-3.0,+2.7)} & \textbf{-160.7} \tiny{(-1.3,+1.0)} & -212.6 \tiny{(-13.7,+24.5)} & -221.5 \tiny{(-2.3,+3.2)} & -181.1 \tiny{(-3.9,+2.2)} & \textbf{-163.5} \tiny{(-2.3,+2.7)} & \textbf{-161.2} \tiny{(-0.9,+0.8)} \\  
\cmidrule(rr){1-1}\cmidrule(lr){2-6}\cmidrule(ll){7-11}
\multicolumn{1}{ l }{SXY-4}  & -1133.2 \tiny{(-7.1,+8.4)} & -1162.1 \tiny{(-33.8,+33.8)} & -1124.7 \tiny{(-27.6,+16.9)} & -1177.9 \tiny{(-29.4,+38.5)} & \textbf{-920.6} \tiny{(-50.1,+92.7)} & -1116.9 \tiny{(-43.2,+79.0)} & -1196.5 \tiny{(-29.8,+20.3)} & -1112.1 \tiny{(-28.9,+26.4)} & -1149.7 \tiny{(-12.5,+16.3)} & \textbf{-827.4} \tiny{(-7.8,+5.8)} \\
\cmidrule(rr){1-1}\cmidrule(lr){2-6}\cmidrule(ll){7-11}
\multicolumn{1}{ l }{SBF} & -436.3 \tiny{(-74.5,+38.4)} & \textbf{-403.1} \tiny{(-2.1,+2.7)} & -446.6 \tiny{(-5.6,+6.4)} & -472.7 \tiny{(-9.7,+12.5)} & \textbf{-401.7} \tiny{(-0.5,+0.6)} & -420.3 \tiny{(-0.4,+0.4)} & -403.3 \tiny{(-0.8,+1.2)} & -407.3 \tiny{(-1.8,+2.8)} & -412.5 \tiny{(-1.9,+3.3)} & \textbf{-399.9} \tiny{(-1.2,+0.9)} \\
\cmidrule(rr){1-1}\cmidrule(lr){2-6}\cmidrule(ll){7-11}
\multicolumn{1}{ l }{LBF} & 0.31 \tiny{(-0.0,+0.0)} & 0.3 \tiny{(-0.01,+0.01)} & 0.03 \tiny{(-0.02,+0.05)} & 0.01 \tiny{(-0.01,+0.02)} & \textbf{0.37} \tiny{(-0.01,+0.02)} & 0.36 \tiny{(-0.0,+0.0)} & 0.31 \tiny{(-0.02,+0.03)} & 0.02 \tiny{(-0.01,+0.02)} & 0.02 \tiny{(-0.02,+0.03)} & \textbf{0.38} \tiny{(-0.03,+0.04)} \\ 
\bottomrule
\end{tabular}
}
\end{table}

\begin{figure*}[t]
    \centering
    \begin{subfigure}[b]{0.99\textwidth}
        \centering
        \includegraphics[width=0.99\textwidth]{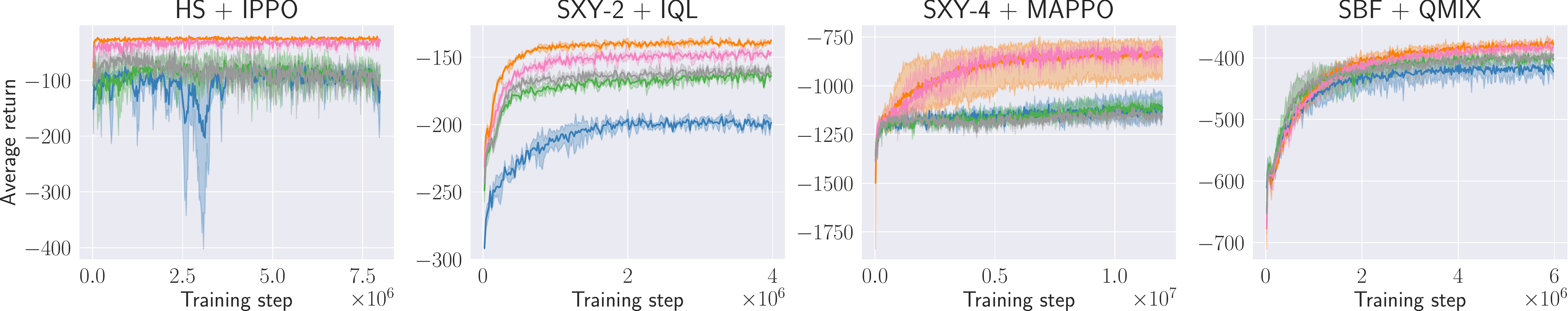}
    \end{subfigure}

    \vspace{0.1cm}
    
    \begin{subfigure}[b]{0.99\textwidth}
        \centering
        \includegraphics[width=0.5\textwidth]{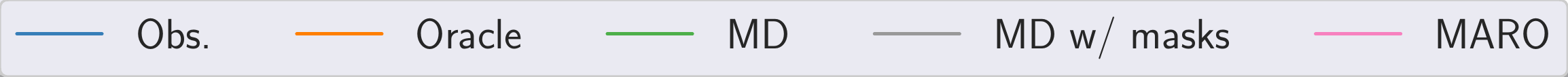}
    \end{subfigure}

    \caption{Mean episodic returns during training for $p \sim \mathcal{U}(0,1)$.}
    \label{fig:evaluation:results:training}
\end{figure*}

\begin{figure*}[t]
    \centering
    \begin{subfigure}[b]{0.99\textwidth}
        \centering
        \includegraphics[width=0.99\textwidth]{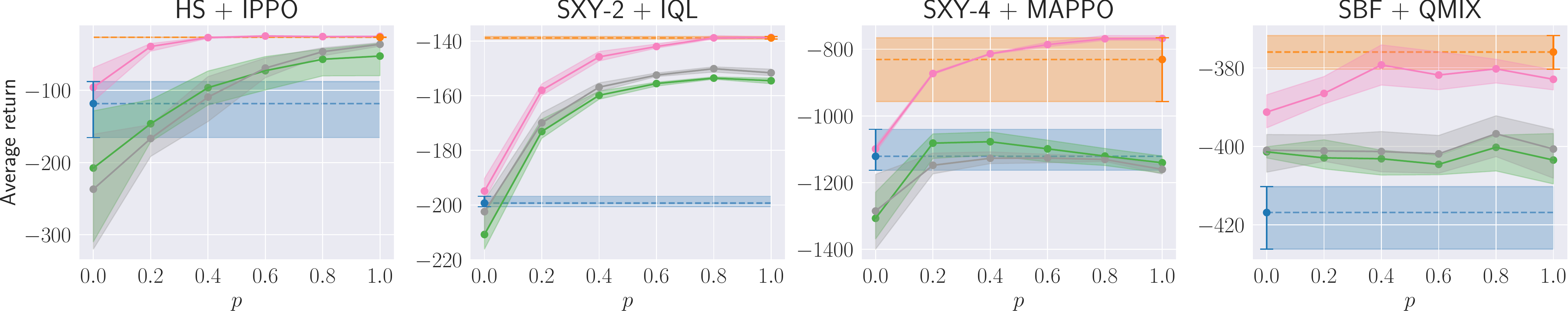}
    \end{subfigure}

    \vspace{0.1cm}
    
    \begin{subfigure}[b]{0.99\textwidth}
        \centering
        \includegraphics[width=0.5\textwidth]{Images/evaluation/legend.pdf}
    \end{subfigure}

    \caption{Mean episodic returns at execution time for different communication levels $p$.}
    \label{fig:evaluation:results:execution}
\end{figure*}

We present the main evaluation results in Tables~\ref{table:evaluation:main_results_value_based} and ~\ref{table:evaluation:main_results_actor_critic} for the value-based and actor critic-based algorithms respectively. For each environment, RL algorithm and method, we present the values of the accumulated rewards obtained, for $p \sim \mathcal{U}(0,1)$. The values that are not significantly different than the highest are presented in bold. The results show that MARO is the best-performing method overall. In particular, out of the 24 algorithm-environment combinations considered, MARO performed equal to or better than all baselines for 22 of them (we further dissect the two settings for which MARO did not perform the best in Sec.~\ref{sec:evaluation:results:training_impact}). Overall, MARO is able to exploit the information provided by the other agents, in contrast with the fully decentralized approaches (Obs.). The obtained results show that the use of a predictive model of other agents' observations, as in MARO, allows for improved performance over the MD baselines.

Additionally, we recall that the RL controllers of MARO are trained in a fully centralized manner, i.e., without considering communication failures at training time. Thus, an interesting observation is that MARO can be seen as a method that provides robustness to centralized execution methods when performing tasks in settings with potential faulty communication, despite never being trained to execute in such conditions. This is clear when comparing the performance of the two methods that consider a centralized training of the RL controllers (without considering communication failures during training), i.e., Masked j. obs. and MARO. As can be seen in Tabs.~\ref{table:evaluation:main_results_value_based} and \ref{table:evaluation:main_results_actor_critic}, MARO's predictive model allows for zero-shot multi-agent execution with respect to communication failures, significantly outperforming the Masked j. obs baseline.

In Fig.~\ref{fig:evaluation:results:training}, we highlight the training curves of MARO and the MD baselines, for $p \sim \mathcal{U}(0,1)$ under some experimental configurations. We display the training curves for all methods, RL algorithms and environments in Appendix~\ref{appendix:experimental_evaluation:experimental_results}. As can be seen, MARO achieves higher returns throughout training as compared to the other baselines. Additionally, MARO attains a performance similar to that of the Oracle baseline. In Fig.~\ref{fig:evaluation:results:execution}, we display the episodic returns of MARO and the MD baselines for different communication levels at execution time for some experimental configurations (more in Appendix~\ref{appendix:experimental_evaluation:experimental_results}). As can be seen, MARO outperforms the MD baselines across the different communication levels. For the displayed environment-algorithm configurations, MARO always performs better or equal to the decentralized (Obs.) baseline, even when $p=0$. Contrarily, this is not always the case for the MD baselines, as can be seen, for example, for the SXY4 + MAPPO configuration. Moreover, the performance of MARO improves as the level of communication in the environment increases, showing that our approach is able to efficiently make use of all provided information, contrary to the standard fully-decentralized approaches (Obs.).

\begin{figure}[t]
\centering

    \begin{subfigure}[b]{0.24\linewidth}
        \centering
        \includegraphics[width=0.7\linewidth]{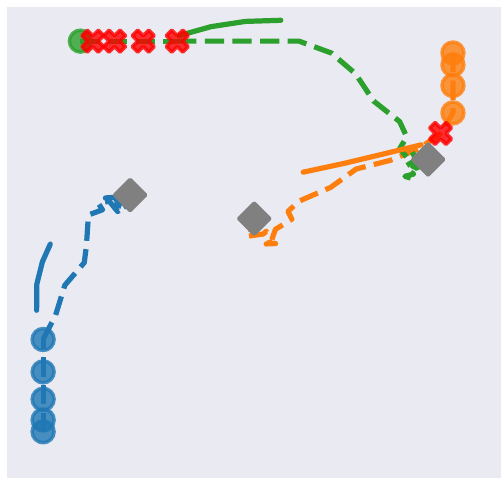}
        \caption{Blue agent.}
        \label{fig:evaluation:results:trajectory_prediction:a}
    \end{subfigure}
    \hfill
    \begin{subfigure}[b]{0.24\linewidth}
        \centering
        \includegraphics[width=0.7\linewidth]{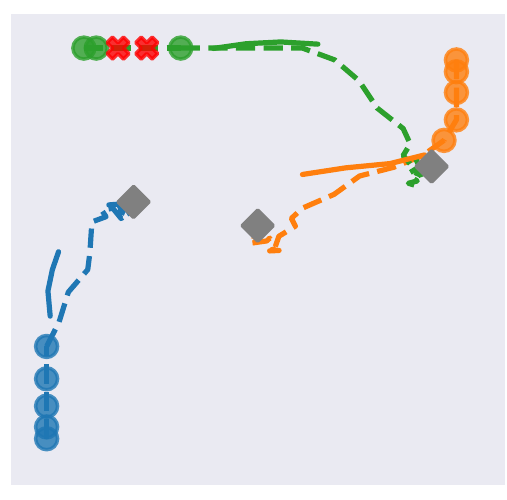}
        \caption{Orange agent.}
        \label{fig:evaluation:results:trajectory_prediction:b}
    \end{subfigure}
    \hfill
    \begin{subfigure}[b]{0.24\linewidth}
        \centering
        \includegraphics[width=0.7\linewidth]{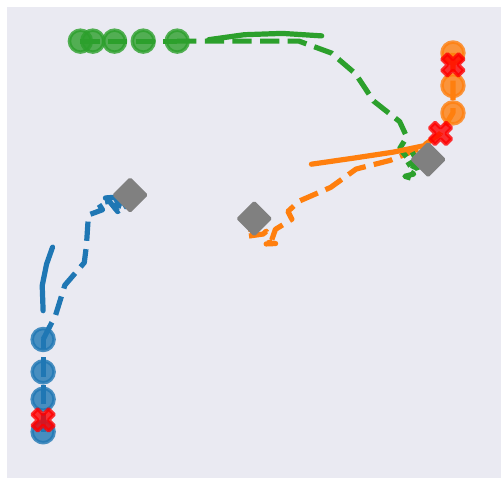}
        \caption{Green agent.}
        \label{fig:evaluation:results:trajectory_prediction:c}
    \end{subfigure}
    \hfill
    \begin{subfigure}[b]{0.24\linewidth}
        \centering
        \raisebox{1.0cm}{\includegraphics[width=0.75\linewidth]{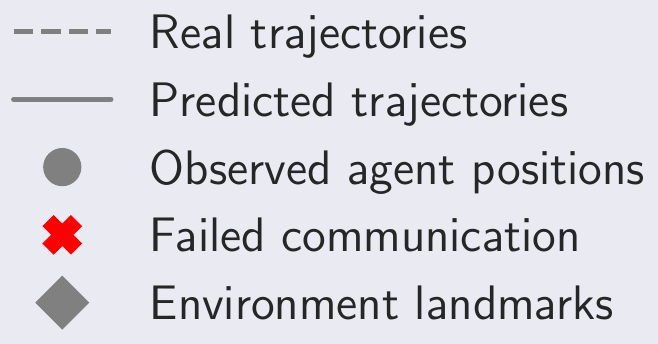}}
    \end{subfigure}
    
  \caption{Estimation of the agents' trajectories made by the predictive model at $t=4$ (SBF, QMIX). The plots are computed, from the perspective of each agent, by computing the estimated trajectories of all agents for the next 4 timesteps. The predictions are computed in a fully auto-regressive manner.}
  \label{fig:evaluation:results:trajectory_prediction}

\end{figure}

As the previous results point out, the use of a predictive model to estimate missing agents' observations, the core component of MARO and the main distinguishing feature over the remainder baselines, is key for the improved performance of MARO. MARO outperforms the recurrent dropout-based baselines, irrespective of whether dropout is incorporated during training (MD baseline) and/or additional context is given to the RL controllers regarding the validity of the entries in the joint-observation vector (MD w/ masks baseline). While previous results attest that the predictions of the predictive model are useful for control, we now take a look at the quality of the predictions themselves. In Fig.~\ref{fig:evaluation:results:trajectory_prediction}, we show the predicted trajectories of all agents from the perspective of each of the agents. As seen, the predicted trajectories are close to the real trajectories of the agents (more in Appendix~\ref{appendix:experimental_evaluation:experimental_results:trajectory_prediction}). The predictive model is thus able to perform accurate agent modeling with faulty communication, providing an interpretable insight into the decision-making process of the agents. 

\begin{table}[t]
\centering
\begin{minipage}{.4\textwidth}
    \caption{MARO w/ drop. mean episodic returns ($p \sim \mathcal{U}(0,1)$). Higher is better.}\label{tab:maro_vs_maro_drop}
    \centering
    \begin{tabular}{llc}\\\toprule  
    \textbf{Env.} & \textbf{Algo.} & \textbf{MARO w/ drop.} \\\midrule
    LBF & IQL & 0.52 \tiny{(-0.01,+0.01)}\\ \midrule
    LBF & QMIX & 0.60 \tiny{(-0.01,+0.02)}\\ \midrule
    HS & MAPPO & -28.7 \tiny{(-1.4,+1.2)} \\ \midrule
    SXY-2 & MAPPO & -159.6 \tiny{(-0.3,+0.3)} \\ \bottomrule
    \end{tabular}
\end{minipage}%
\hspace{0.6cm}
\begin{minipage}{.55\textwidth}

  \centering
    \begin{subfigure}[b]{0.99\textwidth}
        \centering
        \includegraphics[width=0.99\textwidth]{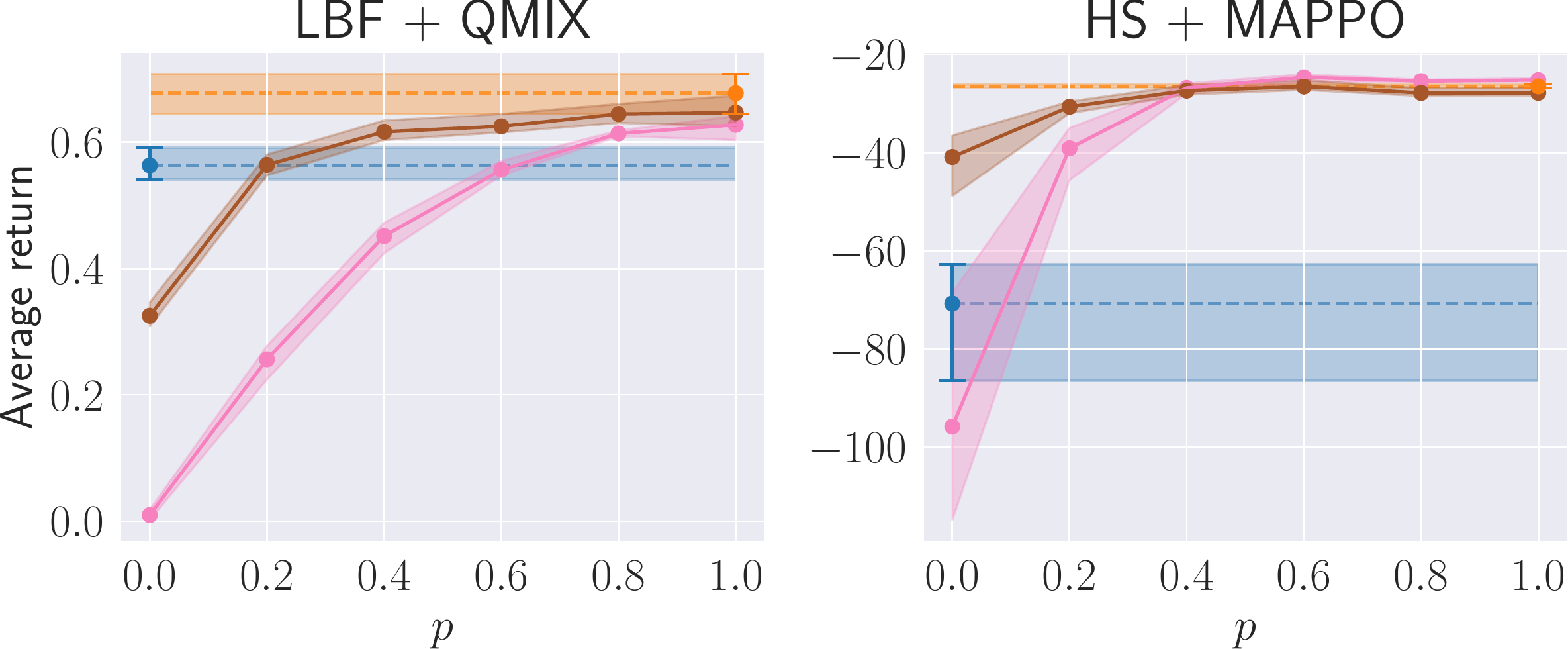}
    \end{subfigure}

    \vspace{0.1cm}
    
    \begin{subfigure}[b]{0.99\textwidth}
        \centering
        \includegraphics[width=0.9\textwidth]{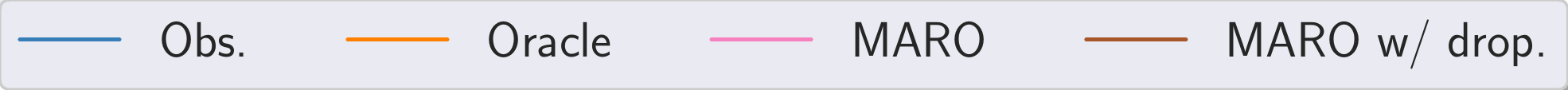}
    \end{subfigure}

    \vspace{0.1cm}
    
  \captionof{figure}{Mean episodic returns for the MARO and MARO w/ drop. methods. Higher is better.}
  \label{fig:evaluation:results:comm_dropout_training}

\end{minipage}
\end{table}

In the next sections, we: (i) assess the impact of considering communication failures at training time, investigating the experimental settings for which MARO was not the best method, in Sec.~\ref{sec:evaluation:results:training_impact}; and (ii) evaluate the performance of MARO under different communication protocols, in Sec.~\ref{sec:evaluation:results:comm_protocols}.

\subsubsection{On the impact of communication dropout at training time}
\label{sec:evaluation:results:training_impact}
%

We now focus our attention on the two settings from Tables~\ref{table:evaluation:main_results_value_based} and \ref{table:evaluation:main_results_actor_critic} under which the MD baselines outperformed MARO. In Fig.~\ref{fig:evaluation:results:comm_dropout_training}, we display the performance of MARO for different communication levels under the LBF + QMIX setting, as well as the HS + MAPPO experimental configuration. As can be seen, the performance of MARO deteriorates for low levels of communication, likely because the RL controllers never had to rely almost entirely on estimations from the predictive model during training. Therefore, we now investigate the impact of considering communication failures also during the training of MARO. To do so, we consider MARO w/ drop., a new version of MARO under which a dropout-based mechanism randomly drops some of the observations of the agents during training according to $p \sim \mathcal{U}(0,1)$. The dropped observations are replaced with predictions from the predictive model. We emphasize that it is fair to compare the performance of MARO w/ drop. against the other baselines, as the MD and MD w/ masks baselines already consider a dropout-based training scheme.

As displayed in Fig.~\ref{fig:evaluation:results:comm_dropout_training}, it is clear that the inclusion of a dropout-based training mechanism to MARO greatly contributed to improve the performance of the method, especially for the settings featuring a low degree of communication (low values of $p$ in the plot). In Table~\ref{tab:maro_vs_maro_drop}, we display the mean episodic returns for MARO under four experimental configurations. As can be seen, when comparing with the values obtained by MARO in Tables~\ref{table:evaluation:main_results_value_based} and \ref{table:evaluation:main_results_actor_critic}, the performance of MARO further improved for the aforementioned experimental settings by considering dropout at training time. In particular, we highlight that, for the two experimental settings under which the MD baselines outperformed MARO, MARO w/ drop. is now able to equal the performance of the best-performing method. 

The results above show that the inclusion of a dropout-based mechanism can impact the performance of the algorithm for hybrid execution. We display the experimental results obtained for MARO w/ drop. for all environments and RL algorithms in Appendix~\ref{appendix:experimental_evaluation:experimental_results}. We note that, according to our results, the improvement of MARO over MARO w/ drop. was not consistent over all tested algorithm-environment configurations. In fact, for some experimental settings, MARO outperformed MARO w/ drop. Therefore, we decided to treat the use of the dropout-based mechanism in the training of MARO as an hyperparemeter that should be tuned according to each experimental scenario. Nevertheless, we highlight that MARO and MARO w/ drop. were, overall, among the best-performing methods.

\subsubsection{On the impact of the sampling scheme of the communication matrix at execution time}
\label{sec:evaluation:results:comm_protocols}
Finally, in order to better understand MARO's performance under different communication protocols, we evaluate different sampling schemes of the communication matrix: (i) our default setting, $p_{\textrm{default}}$, under which $p_{i,j} = p_{j,i} = p$ with $p \sim \mathcal{U}(0,1)$, sampled at the beginning of each episode; (ii) $p_{\textrm{asymmetric}}$, featuring communication matrices $C$ such that $p_{i, j} \neq p_{j, i} \sim \mathcal{U}(0,1)$, sampled at the beginning of each episode; and (iii) $p_{\textrm{dynamic}}$, similar to (ii) but with $C$'s sampled every $5$ time steps. We present the results in Table~\ref{table:evaluation:results:comm_protocols} for a subset of all RL algorithm-environment combinations. As can be seen, MARO continues to perform equal to or better than the remainder baselines, independently of the sampling scheme. We display the complete results in Appendix~ \ref{appendix:experimental_evaluation:experimental_results}; across all results, either MARO or MARO w/ drop. performed equal to or better than all other baselines.

\begin{table}[t]
\caption{Mean episodic returns for different communication protocols. Higher is better.}
\label{table:evaluation:results:comm_protocols}
\centering
\noindent
\resizebox{\linewidth}{!}{%
\begin{tabular}{N N N N N N N N N N}\toprule
\multicolumn{2}{N }{\textbf{}} & \multicolumn{4}{c }{$p_{\textrm{asymmetric}}$} & \multicolumn{4}{c }{$p_{\textrm{dynamic}}$} \\  
\cmidrule(lr){3-6}
\cmidrule(ll){7-10}
\multicolumn{1}{ l }{\textbf{Env.}} & \textbf{Algo.} & \textbf{Masked j. obs.} & \textbf{MD} & \textbf{MD w/ masks} & \textbf{MARO} & \textbf{Masked j. obs.} & \textbf{MD} & \textbf{MD w/ masks} & \textbf{MARO} \\
\cmidrule(rr){1-2} \cmidrule(lr){3-6} \cmidrule(ll){7-10}
SXY-2 & IQL & -200.2 \tiny{(-0.5,+0.4)} & -164.1 \tiny{(-1.1,+0.7)} & -160.5 \tiny{(-0.8,+0.6)} & \textbf{-149.4} \tiny{(-0.2,+0.3)} & -196.7 \tiny{(-2.3,+2.2)} & -161.2 \tiny{(-0.4,+0.3)} & -157.6 \tiny{(-0.8,+0.8)} & \textbf{-145.3} \tiny{(-1.0,+1.5)}\\ 
\cmidrule(rr){1-2} \cmidrule(lr){3-6} \cmidrule(ll){7-10}
SBF & QMIX & -407.3 \tiny{(-6.8,+6.6)} & -403.7 \tiny{(-8.4,+4.4)} & -401.0 \tiny{(-6.6,+5.0)} & \textbf{-380.8} \tiny{(-2.1,+3.3)} & -404.6 \tiny{(-2.7,+2.0)} & -402.7 \tiny{(-6.7,+3.5)} & -398.6 \tiny{(-4.2,+4.5)} & \textbf{-381.9} \tiny{(-5.0,+5.6)} \\
\cmidrule(rr){1-2} \cmidrule(lr){3-6} \cmidrule(ll){7-10}
LBF & IPPO & 0.3 \tiny{(-0.01,+0.0)} & 0.03 \tiny{(-0.03,+0.05)} & 0.01 \tiny{(-0.01,+0.01)} & \textbf{0.36} \tiny{(-0.01,+0.01)} & 0.31 \tiny{(-0.02,+0.01)} & 0.03 \tiny{(-0.03,+0.06)} & 0.02 \tiny{(-0.01,+0.02)} & \textbf{0.4 }\tiny{(-0.03,+0.03)} \\
\cmidrule(rr){1-2} \cmidrule(lr){3-6} \cmidrule(ll){7-10}
SXY-4 & MAPPO & -1066.9 \tiny{(-23.2,+28.7)} & -1128.1 \tiny{(-26.2,+20.1)} & -1160.7 \tiny{(-12.0,+13.1)} & \textbf{-823.9} \tiny{(-1.9,+1.8)} & -1068.3 \tiny{(-2.7,+4.4)} & -1101.7 \tiny{(-32.2,+23.5)} & -1136.7 \tiny{(-13.5,+16.8)} & \textbf{-813.0} \tiny{(-3.4,+2.9)} \\
\bottomrule
\end{tabular}
}
\end{table}

\section{Related Work}
\label{sec:related_work}
In this section, we connect our work with other lines of research, discussing the similarities and differences between our approach and previous works in the field. Due to space limitations, we discuss only the most relevant works and provide an extended discussion of related work in Appendix~\ref{sec:extended_rw}.

Closely related to our work are papers that address the problem of partial observability in MARL. As an example, \citet{omidshafiei_2017} propose a decentralized MARL algorithm that uses RNNs to improve the agents' observability. \citet{mao_2020} use an RNN to compress the agents' histories, helping to improve agents' observability. The commonly used paradigm of centralized training with decentralized execution also contributes to alleviating partial observability at training time \citep{oliehoek_2011,rashid_2018,foerster_2016,foerster_2017}. Other lines of research investigate communication techniques for MARL \citep{zhu_2022}, focusing on how \citep{niu_2021,kim_2019}, when \citep{singh_2018,hu_2020}, and what \citep{foerster_2016} to communicate to foster cooperation. Previous works focused on the sharing of (encoded) local observations and actions among agents \citep{foerster_2016} in a proxy-like manner \citep{wang_2019}. Others consider learning robust communication protocols under missing information. Some approaches learn mechanisms that improve communication efficiency by either limiting the variance of exchanged messages \citep{zhang_2019} or temporally smoothing information shared between agents \citep{zhang_2020}. \citet{kim_2019_dropout} propose message-dropout, which aims at making learning robust against communication errors. Message-dropout drops the messages received from other agents independently at random during training before inputting them into the RL algorithm. In a similar fashion to message dropout, \citet{wang_2020} propose a recurrent actor-critic algorithm for handling multi-agent coordination under partial observability with limited communication, showing that recurrency successfully contributes to robust performance under communication failures.

In contrast, we assume that agents have no control over when and with whom to communicate. Hence, they should robustly perform under any type of communication policy/level at execution time. Also, we do not learn the content of the messages and consider a rather passive communication setting in which agents share local observations and actions. For this reason, we did not include the works of \citet{zhang_2019} and \citet{zhang_2020} as baselines since the comparison between methods would not be meaningful. Instead, following both \citet{kim_2019_dropout} and \citet{wang_2020}, we use message-dropout with recurrent learners as a baseline. Finally, none of the aforementioned works proposed the use of predictive models to account for missing information at execution time as we do in our work, nor mathematically formalized hybrid execution as we present in Sec.~\ref{sec:H-POMDPs}.
 
Other lines of research are also relevant to our work. As opposed to agent/opponent modelling \cite{papoudakis2021agent,xie2020learning,he2016opponent}, we aim at learning policies for multiple agents concurrently. In contrast to multi-agent trajectory prediction \citep{alahi_2016,yeh_2019,hauri_2020,omidshafiei_2021}, we consider a rather broader setting in which agents' observations can correspond to any type of information collected by the agents and use the predictive model with the objective of being robust to missing information. Finally, while our method can be categorized as a model-based MARL method \citep{bargiacchi_2021,kim_2021,wang_2022}, as opposed to previous works which are mainly focused on increasing sample efficiency, we use the predictive model to estimate missing observations.

\section{Conclusion}
\label{sec:conclusion}
In this work, we introduce \emph{hybrid} execution, a new paradigm in which agents act under any communication level at execution time, while exploiting information-sharing among the agents. To formalize our setting, we define hybrid-POMDPs, a new class of POMDPs that explicitly considers a communication process between the agents. To allow for hybrid execution we propose MARO, a novel approach that consists of an autoregressive predictive model to estimate missing observations. We show that MARO's predictive model allows for successful agent trajectory modeling across different communication levels, successfully exploiting available shared information and contributing to improved performance over the remainder baselines. Future work could comprise: (i) studying other neural network architectures, such as graph neural networks, for better multi-agent trajectory prediction; (ii) studying MARO under learned communication protocols such as in \cite{foerster_2016}; and (iii) studying MARO under scenarios comprising a higher number of agents.




\newpage
\bibliography{biblio}

\newpage
\appendix
\input{appendix}


\end{document}

%% file: appendix.tex
\section{Extended Related Work}
\label{sec:extended_rw}
In this section, we connect our work with other lines of research, discussing the similarities and differences between our study and previous works in the field. The discussion herein presented corresponds to an extended version of Sec. \ref{sec:related_work}.

\subsection{Partial Observability in MARL}
\label{sec:related_work:partial_observability}
Closely related to our work are studies that address the problem of partial observability in MARL. As an example, \citet{omidshafiei_2017} propose a decentralized MARL algorithm that uses RNNs to improve the agents' observability. \citet{mao_2020} use an RNN to first compress the agents' histories into embeddings that are posteriorly fed into deep Q-networks, helping to improve agents' observability. The commonly used paradigm of centralized training with decentralized execution also contributes to alleviating partial observability at train time \citep{oliehoek_2011,rashid_2018,foerster_2016,foerster_2017}. Under such paradigm, the calculation of value functions or policy gradients can exploit the centralization of information, thus alleviating partial observability. 

Another way to alleviate the problem of partial observability in MARL, especially at test time, is to consider communication between the agents. We review such setting in the following section.


\subsubsection{Communication in MARL}
\label{sec:related_work:partial_observability:communication}
Different lines of research focus their attention on the development of communication techniques for MARL \citep{zhu_2022}, focusing on how the sharing of information between the agents can be used to improve the RL agents' learning. Early works addressed communication under partially observable cooperative MARL tasks: \citet{sukhbaatar_2016} share the outputs of the hidden layers of a shared neural network among the agents; \citet{foerster_2016} explicitly learn the content of the messages transmitted between agents by following an end-to-end approach in which gradients are back-propagated through the communication variables. Recent works in the field study how \citep{niu_2021,kim_2019}, when \citep{singh_2018,hu_2020}, and what \citep{foerster_2016} should be communicated among the agents in order to foster cooperation.

Similarly to our work, previous studies focused on the sharing of (encoded) local observations and actions among agents \citep{foerster_2016} in a proxy-like manner \citep{wang_2019}. However, as opposed to previous works, we assume that agents have no control over when and with whom to communicate and, instead, should robustly perform under any type of communication policy. We also emphasize that we are not focused on learning the content of the messages being communicated (as in \citep{foerster_2016}), focusing our attention on a rather ``passive'' communication setting by considering the sharing of local observations and actions among the agents.
Other works focus on learning how to combine received information with local information before feeding it into the RL model \citep{jiang_2018,wang_2019}. In our work, we concatenate received information alongside local information. However, the methods developed by previous studies, which can be seen as orthogonal contributions in comparison to our work, can be readily incorporated into our method.

Finally, some works consider learning robust communication protocols under failing/missing information. Previous studies learn mechanisms that improve communication efficiency by either limiting the variance of exchanged messages \citep{zhang_2019}, or temporally smoothing information shared between agents \citep{zhang_2020}. Due to the decreased variability of the messages exchanged throughout timesteps, such methods achieve improved robustness against transmission loss. However, since in this work we are not focusing our attention on methods that learn the contents of the messages being exchanged, we did not use the aforementioned methods as baselines in our work as the comparison between methods would be deemed inappropriate. \citet{kim_2019_dropout} propose a learning technique for MARL called message-dropout, which aims at: (i) effectively handling the increased input dimension in MARL with communication; and (ii) making learning robust against communication errors in the execution phase. Message-dropout drops the messages received from other agents independently at random during training before inputting them into the RL algorithm. In a similar fashion to message dropout, \citet{wang_2020} propose a recurrent actor-critic algorithm for handling multi-agent coordination under partial observability with limited communication, showing that recurrency successfully contributes to robust performance when communication fails. Following both \citet{kim_2019_dropout} and \citet{wang_2020}, we use message-dropout with recurrent learners as a baseline in our work.



We refer to \citet{zhu_2022} for an extensive discussion of the different works that propose communication protocols for MARL.

\subsection{Modelling Other Agents}
\label{sec:related_work:modeling_other_agents}
In contexts where a single agent learns in an environment where other agents are also present, some works have explored ways to model information about the other agents, such as their actions and observations, based on local information available to the learning agent. The work of \cite{papoudakis2021agent} uses a recurrent neural network to predict the other agents' actions and observations in order to make better action selections in a centralized training with decentralized execution setting. At execution time, the agent then uses its learned model to make explicit predictions about other agents' observations and actions. \cite{xie2020learning} does similarly in a latent space. The work of \cite{he2016opponent} uses, instead, the other agents' observations to predict their actions, which assumes centralization will be available at execution time.

Contrarily to the mentioned settings, in ours, we aim at learning policies for multiple agents. To that end, we use a model of the agents' observations and actions to make predictions about other agents and improve their performance in cooperative tasks.

\subsection{Model-based MARL}
\label{sec:related_work:model_based_MARL}
Recently, different works addressed model-based MARL, being mostly focused on improving the sample efficiency of MARL methods by leveraging the knowledge of the learned environment dynamics in policy optimization \citep{willemsen_2021,bargiacchi_2021}. Closely related to our work are studies that propose model-based approaches to MARL while considering communication among the agents. As an example, \citet{kim_2021} propose a communication protocol that encodes into the message an agent's imagined trajectory computed by performing rollouts using an opponent model and a dynamics function.

As opposed to previous works, in our work we focus our attention on the study of methods that allow for robust execution under different communication degrees. While our method can be categorized as a model-based MARL method that implicitly models both the dynamics function as well as the other agents' policies, we use it with a rather different objective than the aforementioned articles.

We refer to \citet{wang_2022} for an extensive discussion of model-based approaches to MARL.


\subsection{Multi-agent Trajectory Prediction}
\label{sec:related_work:trajectory_prediction}
There exists a number of works that address trajectory prediction under multi-agent settings using sequence models \citep{alahi_2016,yeh_2019,hauri_2020,omidshafiei_2021}. As an example, \citet{alahi_2016} use an RNN to learn and predict the trajectory of pedestrians. \citet{hauri_2020} propose an uncertainty-aware multi-modal deep learning model to predict multiple future trajectories of basketball players. \citet{omidshafiei_2021} propose a method based on graph neural networks and bi-directional RNNs to predict unobserved parts of football players' trajectories. 

Our work resembles some similarities with the aforementioned studies since our predictive method MARO solves a similar problem to that of the aforementioned works if we consider that the observations correspond to agents' coordinates. However, in our study, we consider a rather broader setting in which agents' observations can correspond to any type of information collected by the agents. Importantly, we focus our attention on control settings whereas the previous works only deal with predictive settings. Also, we use the predictive model with the objective of being robust to missing information during agency. Nevertheless, we note that our method MARO can possibly benefit from techniques proposed by the aforementioned works.

\section{Experimental Evaluation}
\label{appendix:experimental_evaluation}
In this section, we present supplementary materials for Sec. \ref{sec:evaluation}. In Sec.~\ref{appendix:experimental_evaluation:scenarios}, we provide additional results to support the choice of environments used in this work and describe the proposed MARL scenarios in detail. In Sec.~\ref{appendix:experimental_evaluation:exprimental_methodology}, we describe our experimental methodology. Finally, we present our complete set of experimental results in Sec.~\ref{appendix:experimental_evaluation:experimental_results}.

\subsection{MARL Scenarios}
\label{appendix:experimental_evaluation:scenarios}

\subsubsection{On the selection of the MARL scenarios}
As described in the main text, in this work we are focused on developing methods for hybrid execution, i.e., methods that aim to successfully perform tasks under any degree of centralization. Implicit to our study is the idea that observation-sharing between the agents provides additional information that allows each agent to make better informed choices, thus leading to a better overall performance in comparison to fully decentralized approaches. As observed, however, this is not always the case for environments from standard benchmarks.

\begin{figure}
    \centering
    \begin{subfigure}[b]{0.24\textwidth}
        \centering
        \includegraphics[width=0.97\linewidth]{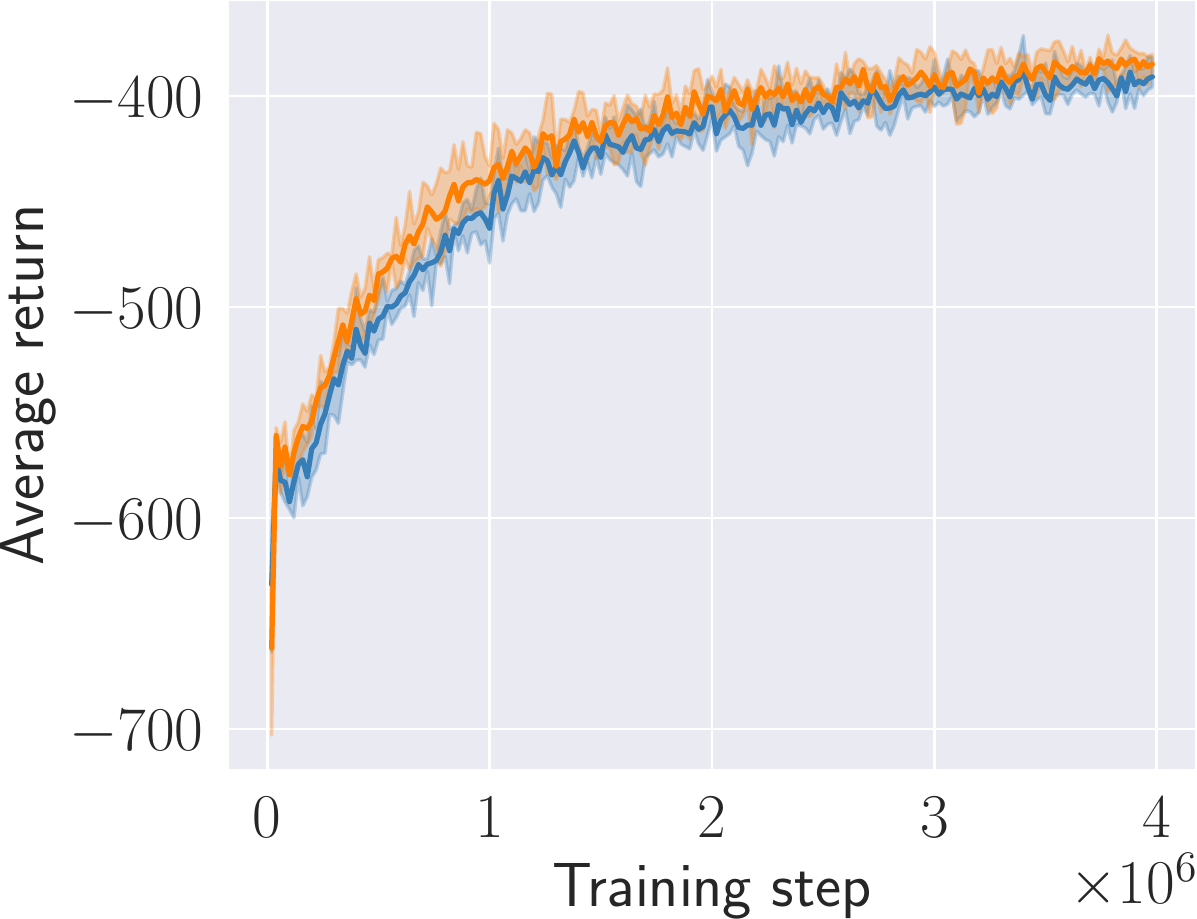}
        \caption{IQL.}
    \end{subfigure}
    \begin{subfigure}[b]{0.24\textwidth}
        \centering
        \includegraphics[width=0.97\linewidth]{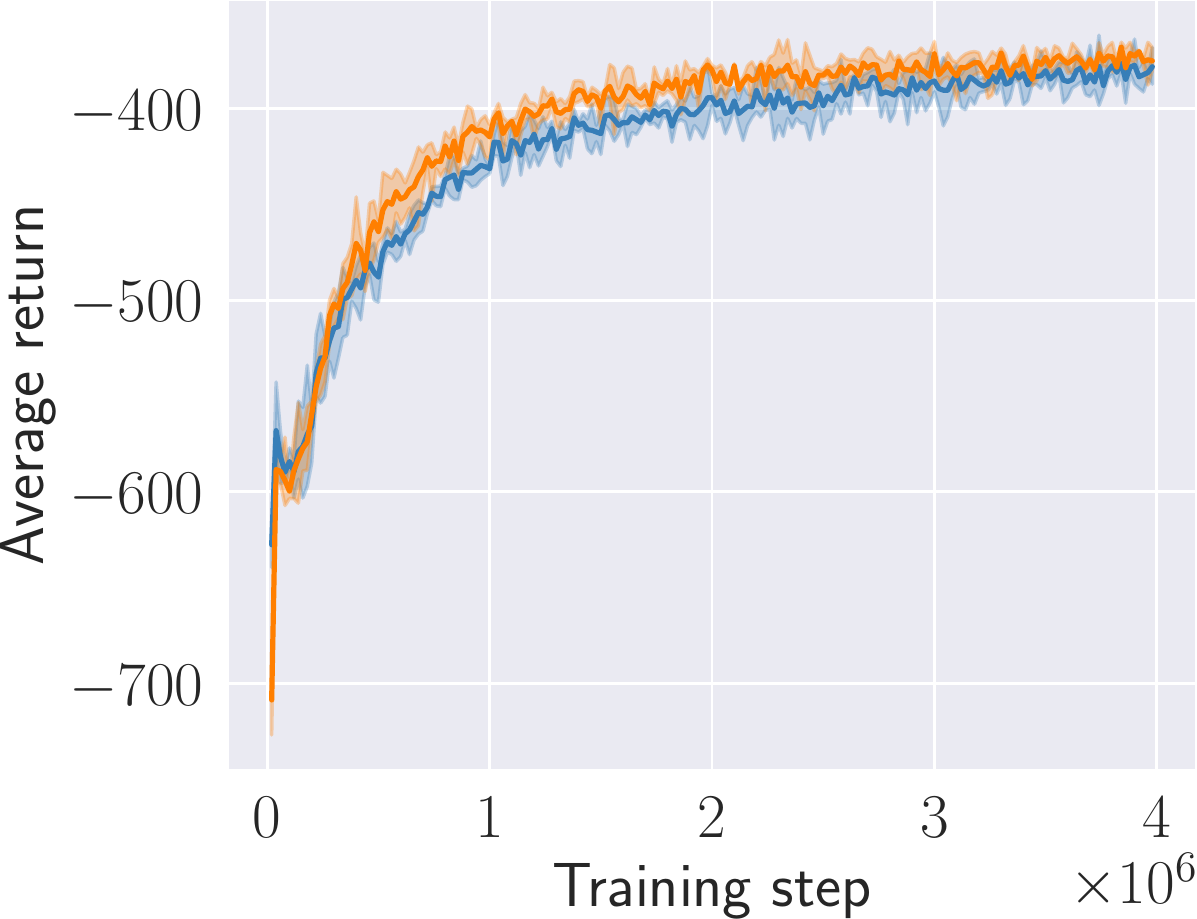}
        \caption{QMIX.}
    \end{subfigure}
    \begin{subfigure}[b]{0.24\textwidth}
        \centering
        \includegraphics[width=0.97\linewidth]{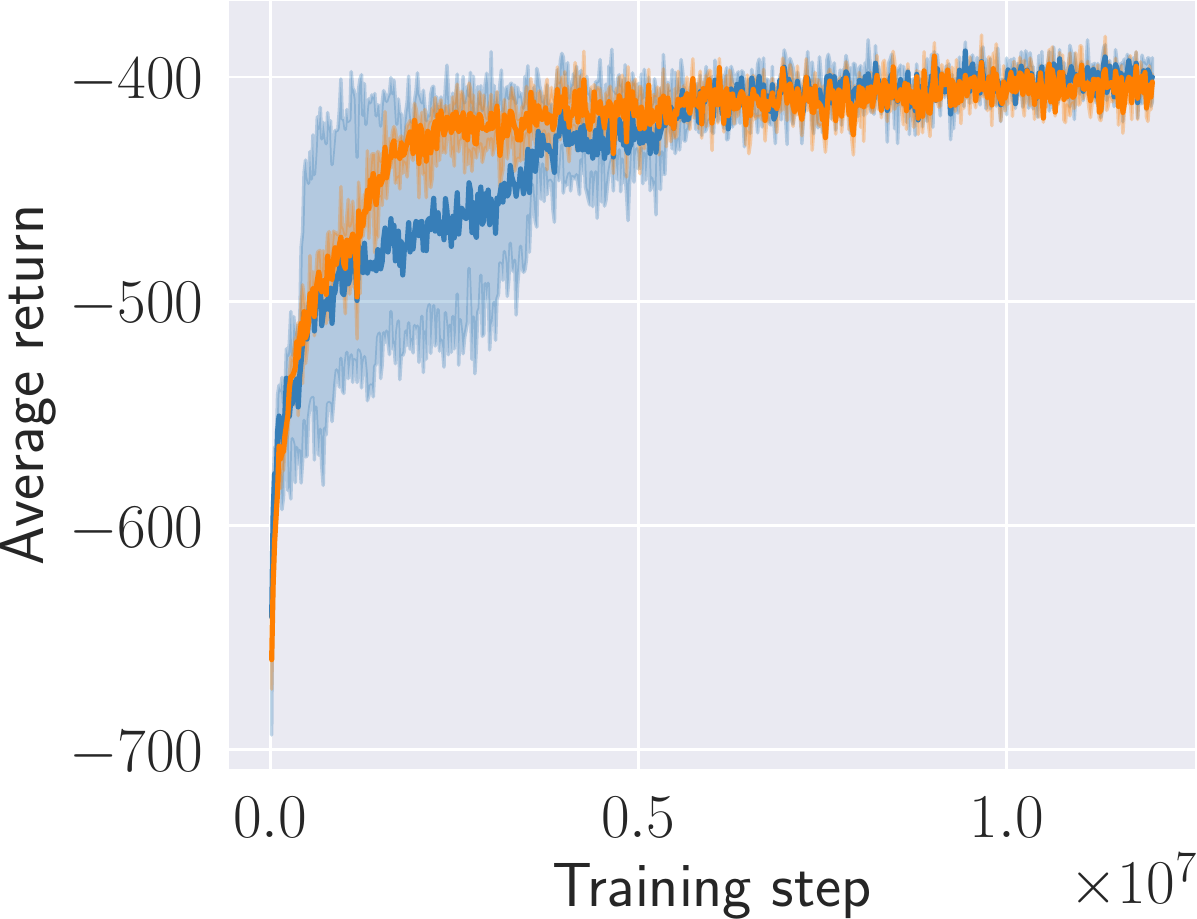}
        \caption{IPPO.}
    \end{subfigure}
    \begin{subfigure}[b]{0.24\textwidth}
        \centering
        \includegraphics[width=0.97\linewidth]{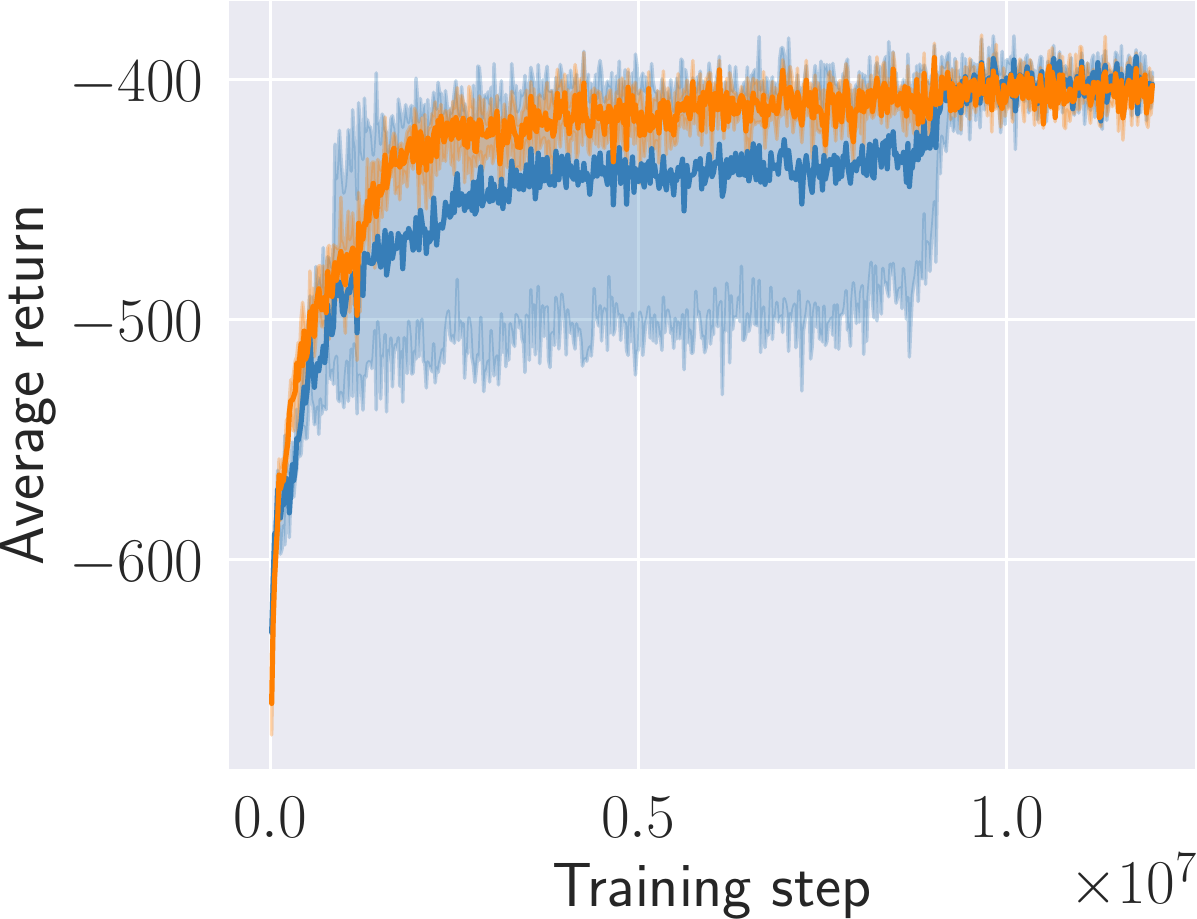}
        \caption{MAPPO.}
    \end{subfigure}
    \caption{(SimpleSpread-v0) Mean episodic returns for $p \sim \mathcal{U}(0,1)$ during training.}
    \label{fig:appendix:no_gap_envs:simple_spread}
\end{figure}

\begin{figure}
    \centering
    \begin{subfigure}[b]{0.24\textwidth}
        \centering
        \includegraphics[width=0.97\linewidth]{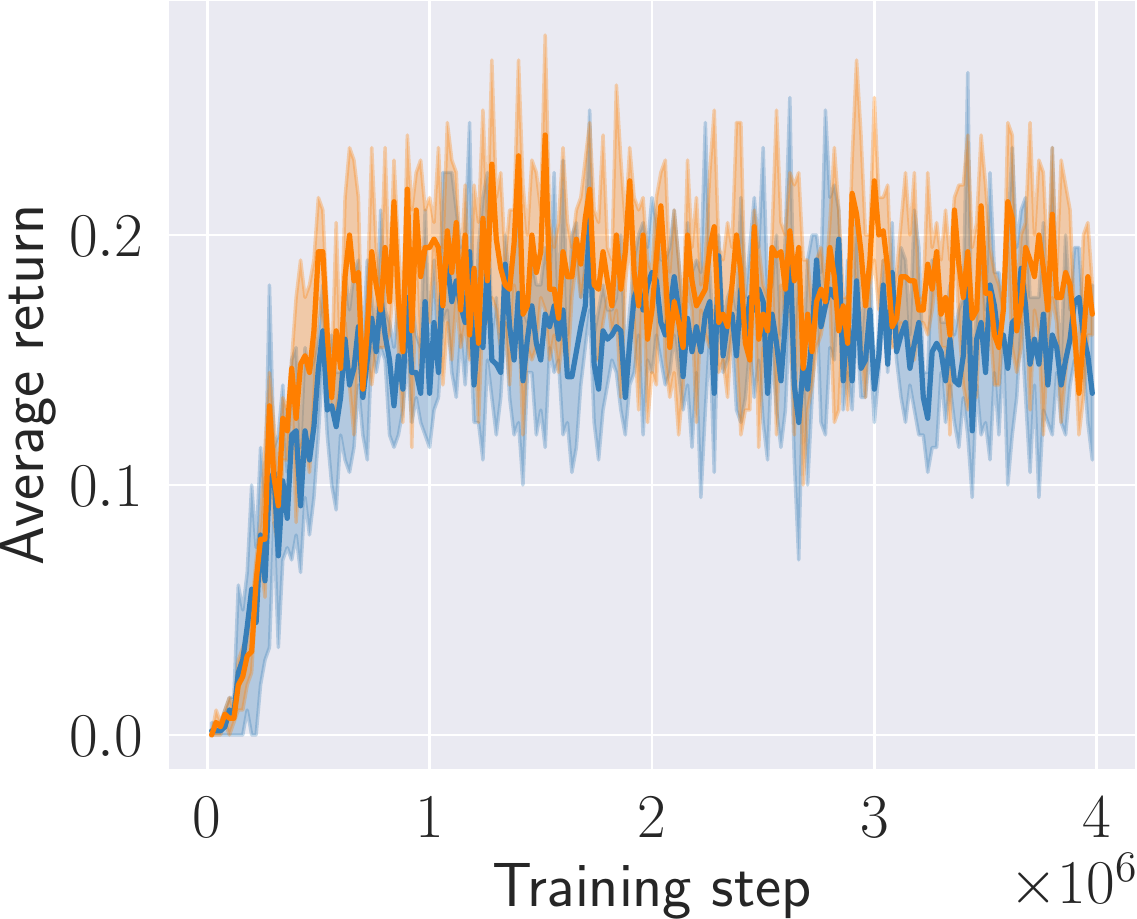}
        \caption{IQL.}
    \end{subfigure}
    \begin{subfigure}[b]{0.24\textwidth}
        \centering
        \includegraphics[width=0.97\linewidth]{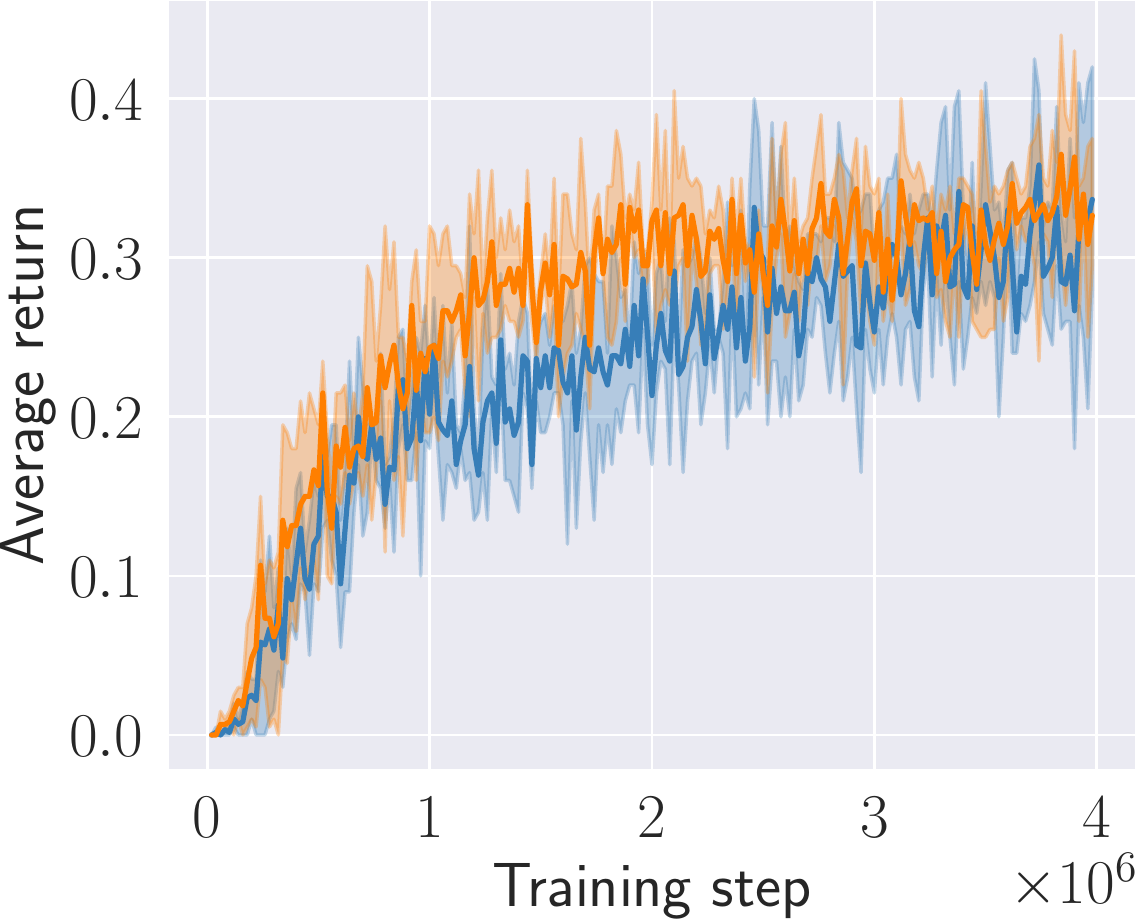}
        \caption{QMIX.}
    \end{subfigure}
    \begin{subfigure}[b]{0.24\textwidth}
        \centering
        \includegraphics[width=0.97\linewidth]{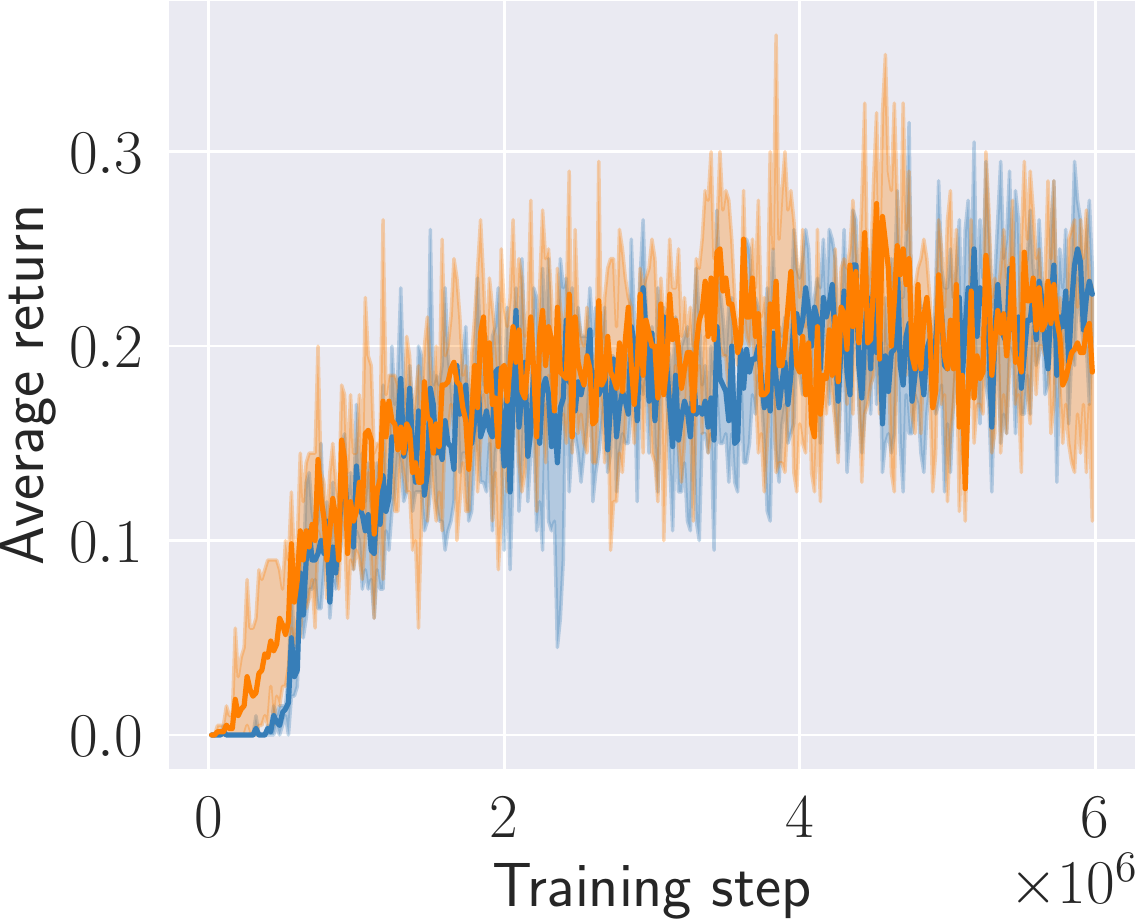}
        \caption{IPPO.}
    \end{subfigure}
    \begin{subfigure}[b]{0.24\textwidth}
        \centering
        \includegraphics[width=0.97\linewidth]{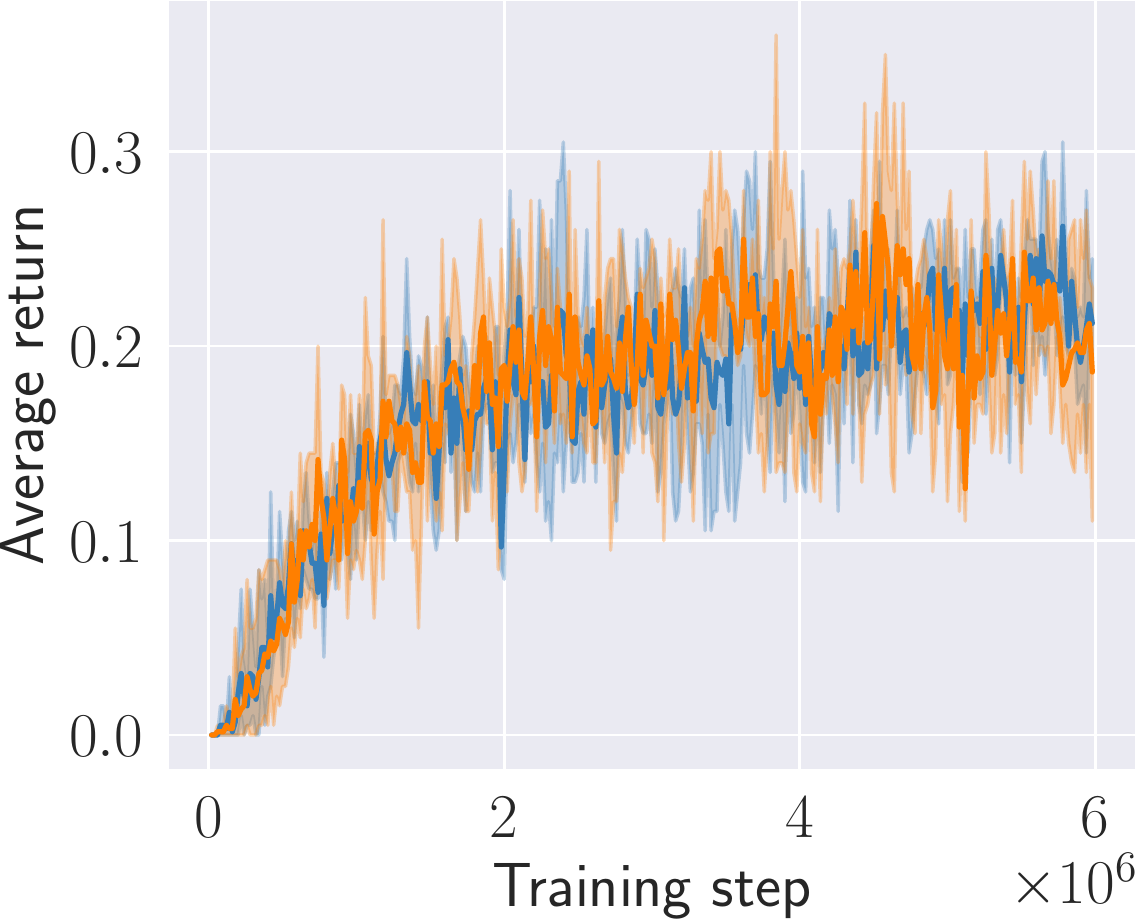}
        \caption{MAPPO.}
    \end{subfigure}
    \caption{(Foraging-2s-15x15-2p-2f-coop-v2, Original) Mean episodic returns for $p \sim \mathcal{U}(0,1)$ during training.}
    \label{fig:appendix:no_gap_envs:lbf_original}
\end{figure}

\begin{figure}
    \centering
    \begin{subfigure}[b]{0.24\textwidth}
        \centering
        \includegraphics[width=0.97\linewidth]{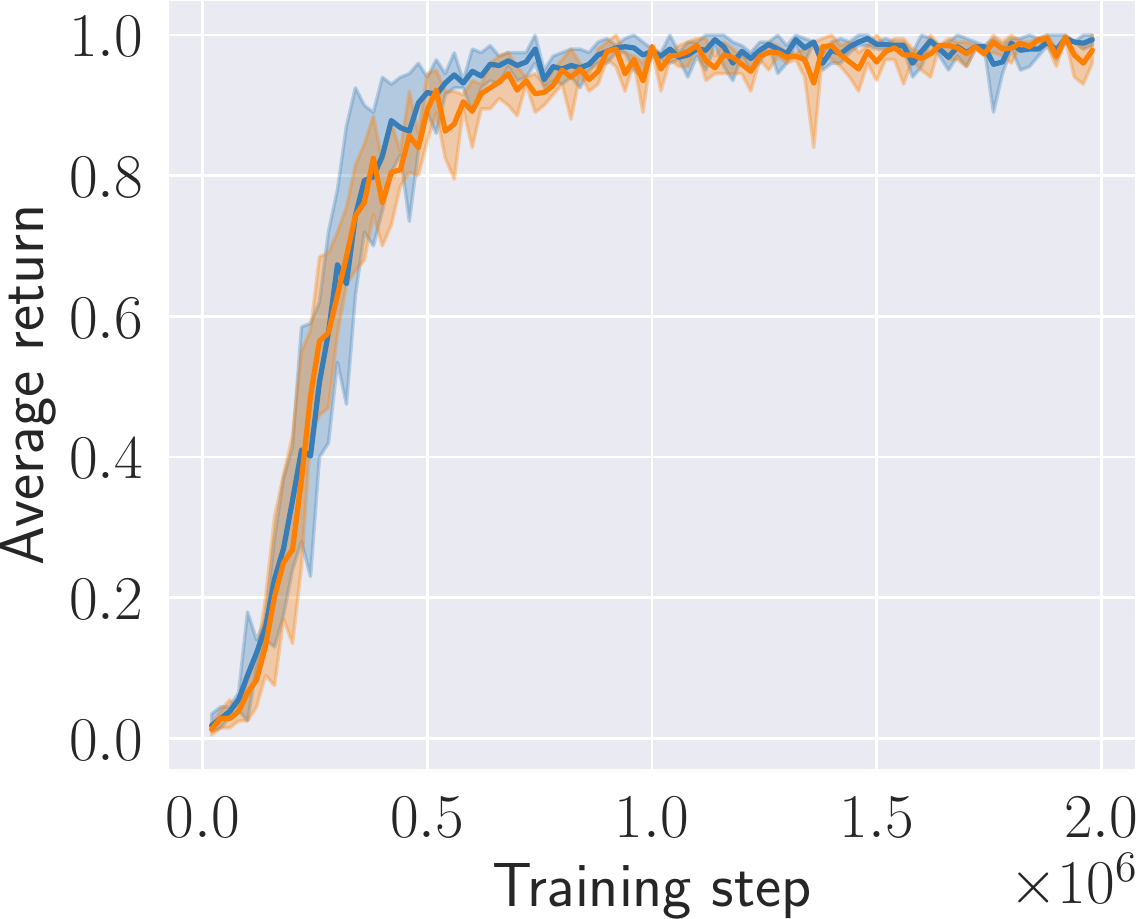}
        \caption{IQL.}
    \end{subfigure}
    \begin{subfigure}[b]{0.24\textwidth}
        \centering
        \includegraphics[width=0.97\linewidth]{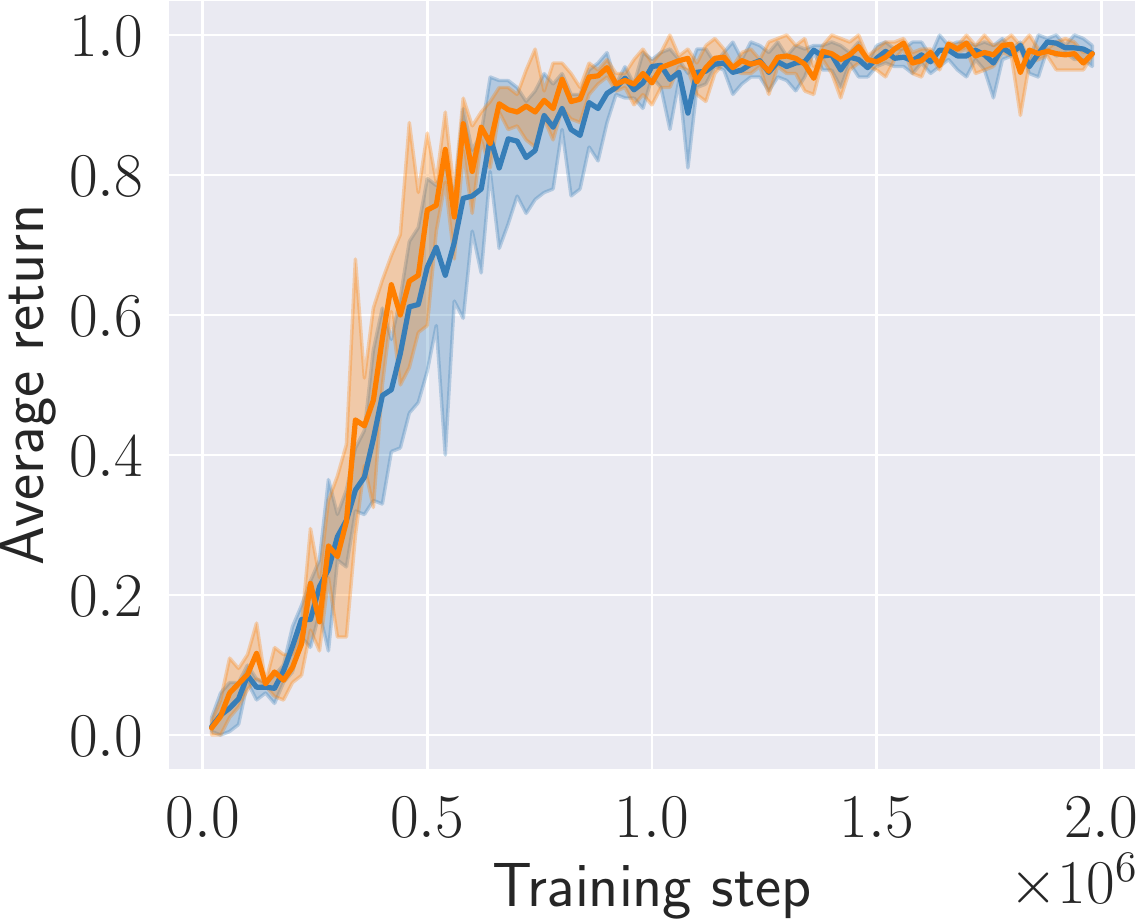}
        \caption{QMIX.}
    \end{subfigure}
    \begin{subfigure}[b]{0.24\textwidth}
        \centering
        \includegraphics[width=0.97\linewidth]{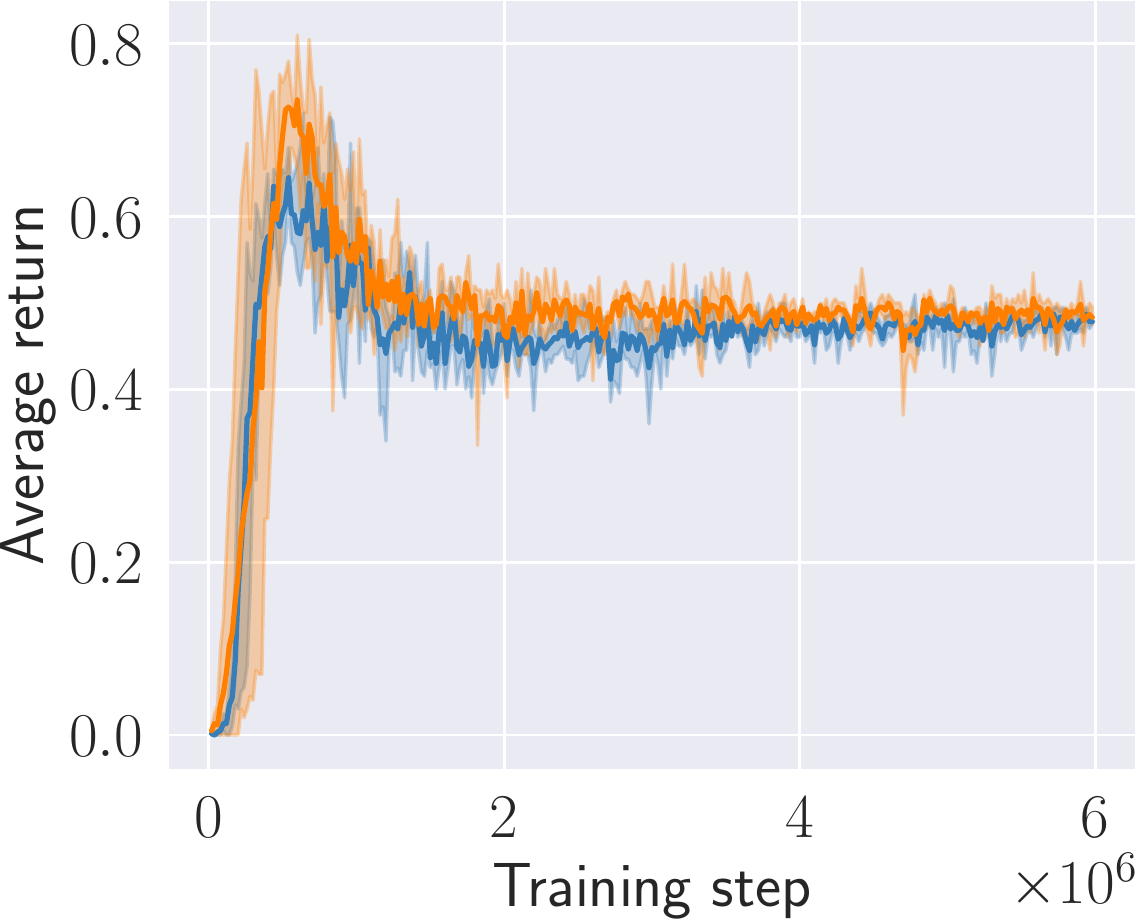}
        \caption{IPPO.}
    \end{subfigure}
    \begin{subfigure}[b]{0.24\textwidth}
        \centering
        \includegraphics[width=0.97\linewidth]{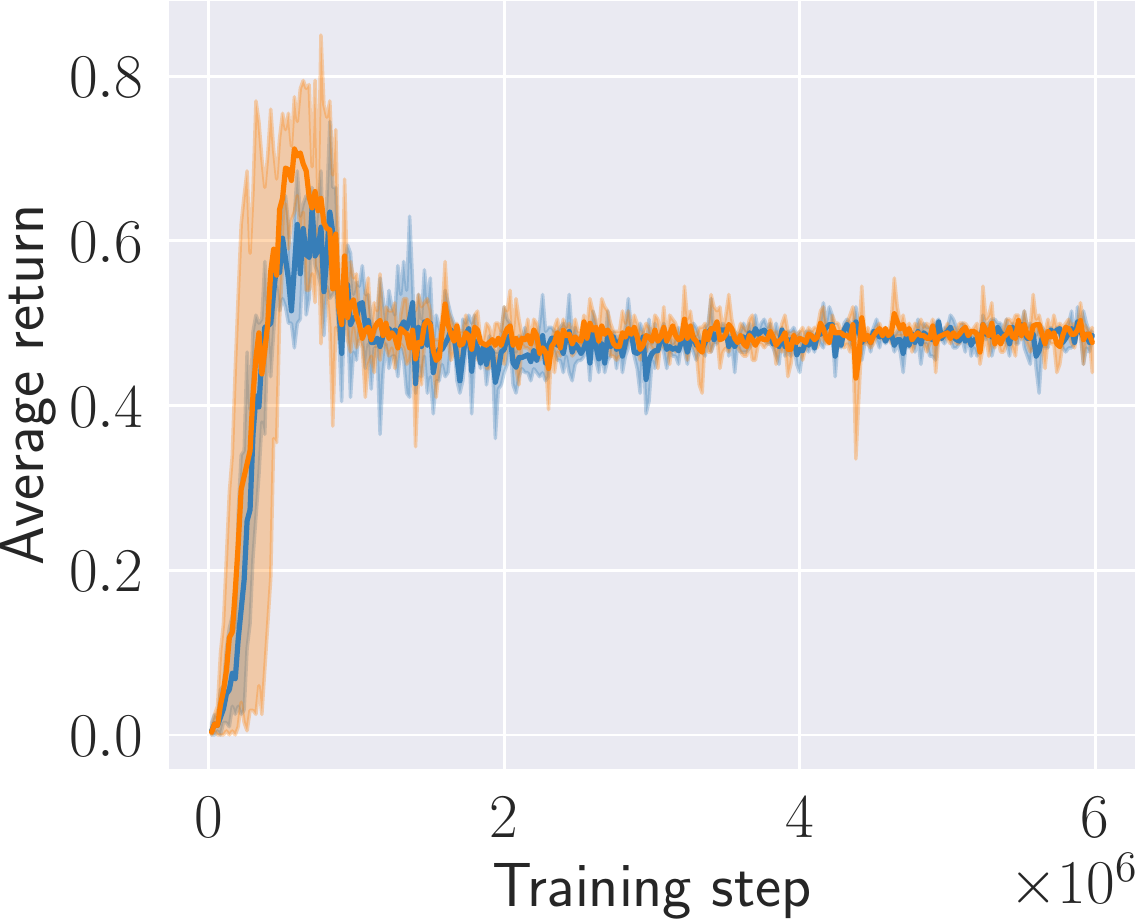}
        \caption{MAPPO.}
    \end{subfigure}
    \caption{(Foraging-2s-8x8-2p-2f-coop-v2, Modified) Mean episodic returns for $p \sim \mathcal{U}(0,1)$ during training.}
    \label{fig:appendix:no_gap_envs:lbf_modified}
\end{figure}

To illustrate the aforementioned, we compare the performance of the Obs. baseline (fully decentralized) and the Oracle baseline (fully centralized, without considering communication failures) under three different environments. In Table~\ref{tab:appendix:no_gap_envs:simple_spread} and Fig.~\ref{fig:appendix:no_gap_envs:simple_spread}, we display the performance values for the standard SimpleSpread-v0 environment from the Multi-Agent Particle Environment \cite{lowe2017multi}. As can be seen, the performance of the fully centralized approach (Oracle baseline) is similar to that of the fully decentralized approach (Obs. baseline). This happens because, for this environment, each agent gets to observe the relative positions of all other agents, as well as the positions of the landmarks, making the environment fully observable. Therefore, there are no gains in allowing information sharing between the agents. In Table~\ref{tab:appendix:no_gap_envs:lbf_original} and Fig.~\ref{fig:appendix:no_gap_envs:lbf_original}, we display the experimental results for the Foraging-2s-15x15-2p-2f-coop-v2 from the Level-Based Foraging environment \cite{lowe2017multi}. As can be seen, once again, the performances of the Obs. and Oracle baselines are very similar. This is the case because, in the Level-Based Foraging environment, the observation of each agent contains the relative positions of the agents and foods that are inside the field-of-view of the agent (as common to previous works, we consider the flag \textit{grid\_observation=False}). Since the relative positions of the agents/foods are considered, observation-sharing between the agents is of little help to improve agents' observability. Finally, in Table~\ref{tab:appendix:no_gap_envs:lbf_modified} and Fig.~\ref{fig:appendix:no_gap_envs:lbf_modified}, we display the results obtained for the modified LBF environment, under which the agents' observations are modified to include the absolute position of the agents (more details in the next section). This environment is similar to the one considered in the main text but comprises a smaller grid size. As can be seen, the performance of both the Obs. and Oracle baselines are similar, across algorithms. We hypothesize that this is the case because, even though information sharing can help to improve agents' observability, the small grid size already contributes to alleviating the partial observability of the agents, thus making the task easier for the fully decentralized agents.

Given the results just presented, it is clear that we do not expect the cooperative team of agents to always benefit from information-sharing among the agents. Some environments are simply already fully observable from the perspective of each agent, while for others, observation-sharing may not contribute to improved performance. With that in mind, we decided to propose new multi-agent environments, described in the next section, which extend standard benchmarks in order to emphasize the negative impact of partial observability in MARL and, under which, the problem of hybrid execution is relevant. We believe the proposed environments are representative of a diverse number of multi-agent tasks.

\begin{table}[h]
\centering
\caption{Mean episodic returns for the SimpleSpread-v0 environment ($p \sim \mathcal{U}(0,1)$). Higher is better.}
\label{tab:appendix:no_gap_envs:simple_spread}
\centering
\begin{tabular}{c c c}\toprule
\multicolumn{1}{c }{\textbf{}} & \multicolumn{2}{c }{\textbf{SimpleSpread-v0}} \\  
\cmidrule(lr){2-3}
\multicolumn{1}{ l }{\textbf{Algorithm}} & \textbf{Obs.} & \textbf{Oracle}  \\
\cmidrule{1-3}
\multicolumn{1}{ l }{IQL} & -393.9 \tiny{(-4.5,+7.8)} & -384.8 \tiny{(-7.4,+3.8)} \\ \cmidrule{1-3}
\multicolumn{1}{ l }{QMIX} & -380.2 \tiny{(-7.0,+6.8)} & -373.8 \tiny{(-4.5,+4.5)} \\ \cmidrule{1-3}
\multicolumn{1}{ l }{IPPO} & -400.9 \tiny{(-2.9,+1.9)} & -404.8 \tiny{(-4.2,+3.6)} \\ \cmidrule{1-3}
\multicolumn{1}{ l }{MAPPO} & -401.8 \tiny{(-1.2,+1.4)} & -404.8 \tiny{(-4.2,+3.6)} \\
\bottomrule
\end{tabular}
\end{table}

\clearpage

\begin{table}
\centering
\caption{Mean episodic returns for the Foraging-2s-15x15-2p-2f-coop-v2 (Original) environment ($p \sim \mathcal{U}(0,1)$). Higher is better.}
\label{tab:appendix:no_gap_envs:lbf_original}
\centering
\begin{tabular}{c c c}\toprule
\multicolumn{1}{c }{\textbf{}} & \multicolumn{2}{c }{\textbf{Foraging-2s-15x15-2p-2f-coop-v2 (Original)}} \\  
\cmidrule(lr){2-3}
\multicolumn{1}{ l }{\textbf{Algorithm}} & \textbf{Obs.} & \textbf{Oracle}  \\
\cmidrule{1-3}
\multicolumn{1}{ l }{IQL} & 0.156 \tiny{(-0.012,+0.012)} & 0.173 \tiny{(-0.019,+0.021)} \\ \cmidrule{1-3}
\multicolumn{1}{ l }{QMIX} & 0.308 \tiny{(-0.039,+0.058)}  & 0.335 \tiny{(-0.033,+0.026)} \\ \cmidrule{1-3}
\multicolumn{1}{ l }{IPPO} & 0.227 \tiny{(-0.014,+0.019)}  & 0.197 \tiny{(-0.045,+0.046)} \\ \cmidrule{1-3}
\multicolumn{1}{ l }{MAPPO} & 0.211 \tiny{(-0.016,+0.014)}  & 0.197 \tiny{(-0.046,+0.046)} \\
\bottomrule
\end{tabular}
\end{table}

\begin{table}
\centering
\caption{Mean episodic returns for the Foraging-2s-8x8-2p-2f-coop-v2 (Modified) environment ($p \sim \mathcal{U}(0,1)$). Higher is better.}
\label{tab:appendix:no_gap_envs:lbf_modified}
\centering
\begin{tabular}{c c c}\toprule
\multicolumn{1}{c }{\textbf{}} & \multicolumn{2}{c }{\textbf{Foraging-2s-8x8-2p-2f-coop-v2 (Modified)}} \\  
\cmidrule(lr){2-3}
\multicolumn{1}{ l }{\textbf{Algorithm}} & \textbf{Obs.} & \textbf{Oracle}  \\
\cmidrule{1-3}
\multicolumn{1}{ l }{IQL} & 0.986 \tiny{(-0.002,+0.001)} & 0.982 \tiny{(-0.002,+0.002)}\\ \cmidrule{1-3}
\multicolumn{1}{ l }{QMIX} & 0.978 \tiny{(-0.007,+0.007)} & 0.971 \tiny{(-0.012,+0.008)}  \\ \cmidrule{1-3}
\multicolumn{1}{ l }{IPPO} & 0.477 \tiny{(-0.009,+0.004)} & 0.488 \tiny{(-0.003,+0.004)} \\ \cmidrule{1-3}
\multicolumn{1}{ l }{MAPPO} & 0.486 \tiny{(-0.003,+0.004)}  & 0.489 \tiny{(-0.006,+0.008)} \\
\bottomrule
\end{tabular}
\end{table}

\clearpage
\subsubsection{Environments description}
\label{appendix:experimental_evaluation:scenarios:description}
\paragraph{SimpleSpeakerListener (SL)} Taken from the Multi-Agent Particle Environment \cite{lowe2017multi}.

\paragraph{HearSee (HS)} The environment consists of two heterogeneous agents and a single landmark in a 2D map. At the start of each episode both the position of the agents and of the landmark is randomly generated. The goal of the agents is to cooperate in order for both of them to cover the landmark: agents are (globally) rewarded considering how far the closest agent is to each landmark (sum of the minimum distances). In this scenario, one of the agents (``Hear'' agent) is provided with the absolute position of the landmark in its observation. However, it does not have access to its own position. The other agent (``See'' agent) is able to access the position and velocities of both agents in its observation, yet does not have access to the position of the landmark. Only through communicating with the central proxy, can the agents have access to both their positions and the position of the landmark in order to complete the task.

\paragraph{SpreadXY-2 (SXY-2)} The environment consists of two heterogeneous agents and two designated landmarks in a 2D map. At the start of each episode both the position of the agents and of the landmarks is randomly generated. The goal of the agents is to cover all the landmarks while avoiding collisions: agents are (globally) rewarded considering how far the closest agent is to each landmark (sum of the minimum distances) and are (locally) penalized if they collide with other agents. Differently from SSB, one of the agents has access to the X position and velocity of both agents, while the other agent has access to the Y position and velocity of both agents. Both agents observe as well the absolute position of all landmarks. Through communication with the central proxy, the agents can access the complete position and velocities of the other agents and cover the landmarks.

\paragraph{SpreadXY-4 (SXY-4)} This environment is similar to \textbf{SpreadXY-2} but comprises two teams of two agents each. Within each team, one of the two agents has access to the X position and velocity of both agents, and the other has access to the Y position and velocity of both agents. Agents must cover the four landmarks while avoiding collisions between each other.

\paragraph{Spreadblindfold (SBF)} The environment consists of three agents and three designated landmarks in a 2D map. At the start of each episode both the position of the agents and of the landmarks is randomly generated. The goal of the agents is to cover all the landmarks while avoiding collisions: agents are (globally) rewarded considering how far the closest agent is to each landmark (sum of the minimum distances) and are (locally) penalized if they collide with other agents. Differently from the original Simple Spread environment, the agent's observation only includes the position and velocity of the agent itself and the relative position of all landmarks. Through communication with the central proxy, the agents can access the position and velocities of the other agents.

\paragraph{Foraging-2s-15x15-2p-2f-coop-v2 (LBF)} Taken from the Level-Based Foraging environment \cite{papoudakis2021benchmarking}, but with the agents' observations modified to include the absolute position of the agents (if inside the field-of-view) instead of their relative positions. The positions of the food inside the field-of-view of each agent appear as relative positions in the observation vector as in the original environment. We consider parameter \textit{grid\_observation=False}.

We refer to \citet{lowe2017multi} and \cite{papoudakis2021benchmarking} for a visual depiction of the environments.

\subsection{Experimental Methodology, Implementation and Hyperparameters}
\label{appendix:experimental_evaluation:exprimental_methodology}
We employ the same RL controller networks across all evaluations. The RL networks include recurrent layers to mitigate the effects of partial observability. We consider four different MARL algorithms: IQL, QMIX, IPPO, and MAPPO. We perform 3 training runs for each experimental setting and 100 evaluation rollouts for each training run. We report, both in tables and plots, the 95\% bootstrapped confidence interval alongside the corresponding scalar mean value. We assume that $p=1$ at $t=0$ for all algorithms. We display our training hyperparameters for the RL controllers and the predictive model in Tables \ref{appendix:mpe_hyperparameters}, \ref{appendix:lbf_hyperparameters} and \ref{appendix:pred_model_hyperparams}. We developed our code in a Python environment using the EPyMARL framework \citep{papoudakis2021benchmarking} and PyTorch \citep{paszke_2019}. The computational code is available at \url{https://github.com/PPSantos/hybrid-marl}

\begin{table*}[h]
\centering
\caption{Hyperparameters for the RL controllers (MPE environments).}
\label{appendix:mpe_hyperparameters}
\begin{subtable}{0.45\textwidth}
\centering
\caption{IQL}
\begin{tabular}{@{}lc@{}}
\toprule
hidden dimension       & $256$                 \\
learning rate          & $0.0005$              \\
reward standardisation & True                  \\
network type           & GRU                   \\
evaluation epsilon     & $0.0$                 \\
epsilon anneal         & $500,000$             \\
target update          & $200$                 \\ \bottomrule
\end{tabular}
\end{subtable}
\hfill
\begin{subtable}{0.45\textwidth}
\centering
\caption{QMIX}
\begin{tabular}{@{}lcc@{}}
\toprule
hidden dimension       & $256$                 \\
learning rate          & $0.0005$              \\
reward standardisation & True                  \\
network type           & GRU                   \\
evaluation epsilon     & $0.0$                 \\
epsilon anneal         & $50,000$              \\
target update          & $200$                 \\ \bottomrule
\end{tabular}
\end{subtable}
\begin{subtable}{0.45\textwidth}
\centering
\caption{IPPO}
\begin{tabular}{@{}lc@{}}
\toprule
hidden dimension       & $256$                 \\
learning rate          & $0.0003$              \\
reward standardisation & True                  \\
network type           & GRU                   \\
entropy coefficient    & 0.01                      \\
target update          & 200                       \\
n-step                 & 5                      \\ \bottomrule
\end{tabular}
\end{subtable}
\hfill
\begin{subtable}{0.45\textwidth}
\centering
\caption{MAPPO}
\begin{tabular}{@{}lcc@{}}
\toprule
hidden dimension       & $256$                 \\
learning rate          & $0.0003$              \\
reward standardisation & True                  \\
network type           & GRU                   \\
entropy coefficient    & 0.01                      \\
target update          & 200                       \\
n-step                  & 5                     \\ \bottomrule
\end{tabular}
\end{subtable}
\end{table*}

\begin{table*}[h]
\centering
\caption{Hyperparameters for the RL controllers (LBF environments).}
\label{appendix:lbf_hyperparameters}
\begin{subtable}{0.45\textwidth}
\centering
\caption{IQL}
\begin{tabular}{@{}lc@{}}
\toprule
hidden dimension       & $256$                 \\
learning rate          & $0.0003$              \\
reward standardisation & True                  \\
network type           & GRU                   \\
evaluation epsilon     & $0.0$                 \\
epsilon anneal         & $100,000$             \\
target update          & $200$                 \\ \bottomrule
\end{tabular}
\end{subtable}
\hfill
\begin{subtable}{0.45\textwidth}
\centering
\caption{QMIX}
\begin{tabular}{@{}lcc@{}}
\toprule
hidden dimension       & $256$                 \\
learning rate          & $0.0001$              \\
reward standardisation & True                  \\
network type           & GRU                   \\
evaluation epsilon     & $0.0$                 \\
epsilon anneal         & $100,000$              \\
target update          & $200$                 \\ \bottomrule
\end{tabular}
\end{subtable}
\begin{subtable}{0.45\textwidth}
\centering
\caption{IPPO}
\begin{tabular}{@{}lc@{}}
\toprule
hidden dimension       & $256$                 \\
learning rate          & $0.0001$              \\
reward standardisation & False                  \\
network type           & GRU                   \\
entropy coefficient    & 0.001                      \\
target update          & 200                       \\
n-step                 & 5                      \\ \bottomrule
\end{tabular}
\end{subtable}
\hfill
\begin{subtable}{0.45\textwidth}
\centering
\caption{MAPPO}
\begin{tabular}{@{}lcc@{}}
\toprule
hidden dimension       & $256$                 \\
learning rate          & $0.0001$              \\
reward standardisation & False                  \\
network type           & GRU                   \\
entropy coefficient    & 0.001                      \\
target update          & 200                       \\
n-step                 & 5                     \\ \bottomrule
\end{tabular}
\end{subtable}
\end{table*}

\begin{table*}[h]
\centering
\caption{Hyperparameters for the predictive model across all environments and algorithms.}
\label{appendix:pred_model_hyperparams}
\begin{tabular}{@{}lcc@{}}
\toprule
hidden dimension       & $128$                \\
learning rate          & $0.001$              \\
grad clip              & 1.0                  \\
buffer size            & 5 000                  \\
batch size             & 32                  \\
\bottomrule
\end{tabular}
\end{table*}

\clearpage
\subsection{Experimental Results}
\label{appendix:experimental_evaluation:experimental_results}
In this section, we display the complete experimental results. We present our main results in Sec.~\ref{appendix:experimental_evaluation:experimental_results:main_results}. In Sec.~\ref{appendix:experimental_evaluation:experimental_results:communication_matrix}, we display the results of MARO under different communication protocols. In Sec. \ref{appendix:experimental_evaluation:experimental_results:trajectory_prediction}, we display a set of figures that illustrates the predictions made by the predictive model. 


\subsubsection{Main Experimental Results}
\label{appendix:experimental_evaluation:experimental_results:main_results}
In this section, we present the complete experimental results of all approaches across all environments and algorithms. The results herein presented correspond to the full results for Sec.~\ref{sec:evaluation:results} of the main text. In this section, we display the mean episodic returns: (i) for a specific communication level $p$, when the communication level is explicitly referred; or (ii) for our default communication setting $p_{\textrm{default}}$, under which $p_{i,j} = p_{j,i} = p$ with $p \sim \mathcal{U}(0,1)$, with communication matrices sampled at the beginning of each episode. The Oracle baseline is always evaluated with $p=1$.

\clearpage

\begin{table}
\centering
\noindent
\caption{(SpeakerListener) Mean episodic returns for $p_{\textrm{default}}$ at execution time.}
\vspace{0.1cm}
\resizebox{\linewidth}{!}{%
\begin{tabular}{c c c c c c c c }\toprule
\multicolumn{1}{c }{\textbf{}} & \multicolumn{7}{c }{\textbf{SpeakerListener ($p_\textrm{default}$)}} \\  
\cmidrule(lr){2-8}
\multicolumn{1}{ l }{\textbf{Algorithm}} & \textbf{Obs.} & \textbf{Oracle} & \textbf{Masked j. obs.} & \textbf{MD} & \textbf{MD w/ masks} & \textbf{MARO} & \textbf{MARO w/ drop.} \\
\cmidrule{1-8}
\multicolumn{1}{ l }{IQL} & -40.0 \tiny{(-0.4,+0.4)} & -24.2 \tiny{(-0.1,+0.2)} & -45.3 \tiny{(-1.2,+1.9)} & -25.4 \tiny{(-0.6,+1.1)} & -25.5 \tiny{(-0.6,+1.1)} & -25.3 \tiny{(-0.6,+1.0)} & -25.3 \tiny{(-0.5,+1.0)} \\ \cmidrule{1-8}
\multicolumn{1}{ l }{QMIX} & -24.9 \tiny{(-0.1,+0.0)} & -23.9 \tiny{(-0.2,+0.2)} & -40.5 \tiny{(-0.8,+0.8)} & -25.2 \tiny{(-0.6,+1.1)} & -25.2 \tiny{(-0.6,+1.1)} & -25.1 \tiny{(-0.6,+1.2)} & -24.8 \tiny{(-0.8,+0.8)} \\ \cmidrule{1-8}
\multicolumn{1}{ l }{IPPO} & -33.3 \tiny{(-11.6,+5.9)} & -25.1 \tiny{(-0.3,+0.6)} & -39.5 \tiny{(-0.8,+0.7)} & -45.2 \tiny{(-5.0,+5.0)} & -47.0 \tiny{(-3.2,+4.0)} & -25.2 \tiny{(-0.1,+0.1)} & -28.9 \tiny{(-1.6,+1.1)} \\ \cmidrule{1-8}
\multicolumn{1}{ l }{MAPPO} & -59.6 \tiny{(-0.5,+0.6)} & -25.0 \tiny{(-0.2,+0.3)} & -39.3 \tiny{(-1.2,+1.0)} & -28.2 \tiny{(-0.3,+0.3)} & -27.5 \tiny{(-0.4,+0.5)} & -25.2 \tiny{(-0.1,+0.1)} & -28.1 \tiny{(-0.7,+0.9)} \\
\bottomrule
\end{tabular}
}
\end{table}

\begin{figure}
    \centering
    \begin{subfigure}[b]{0.24\textwidth}
        \centering
        \includegraphics[width=0.97\linewidth]{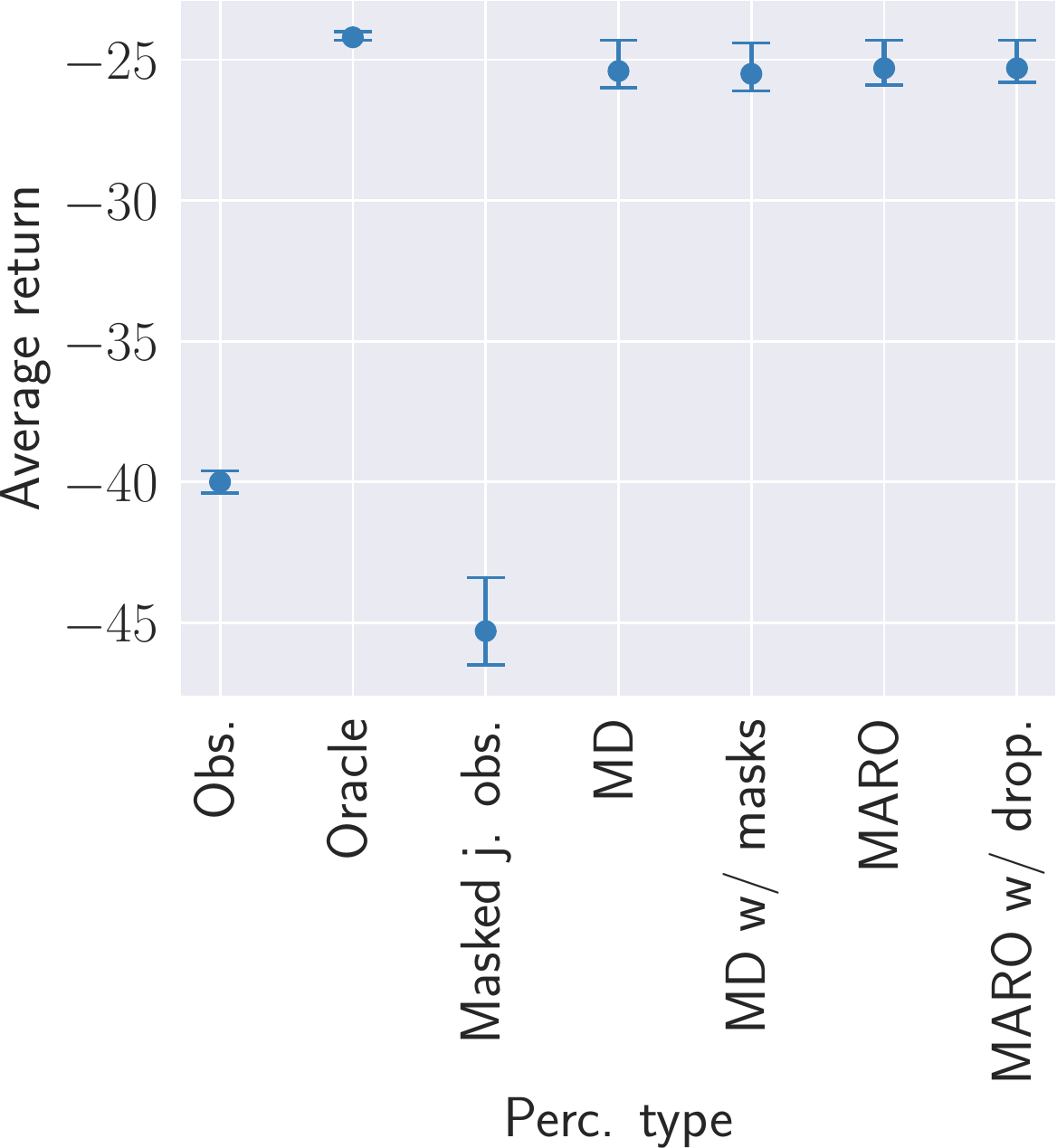}
        \caption{IQL.}
    \end{subfigure}
    \begin{subfigure}[b]{0.24\textwidth}
        \centering
        \includegraphics[width=0.97\linewidth]{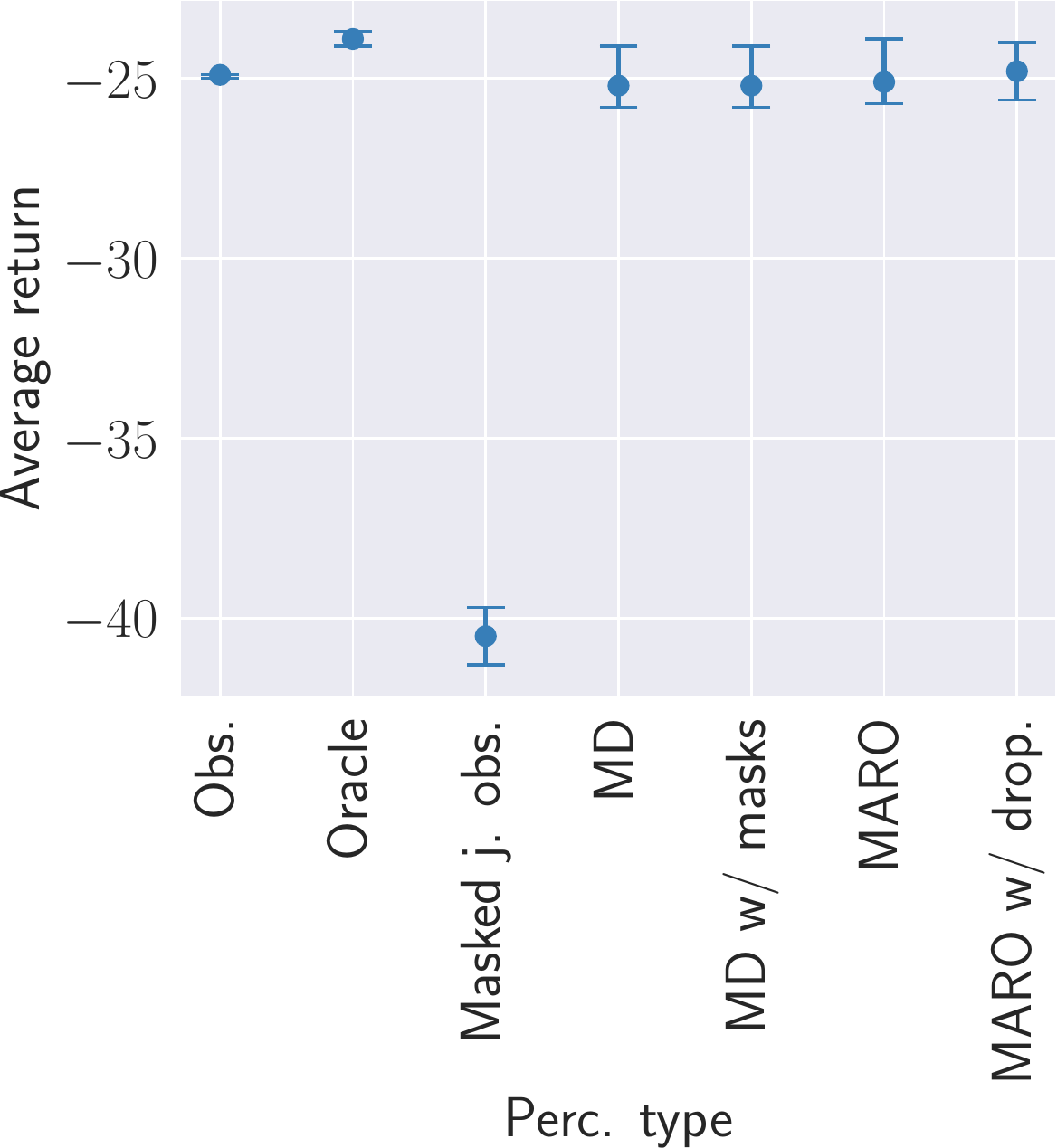}
        \caption{QMIX.}
    \end{subfigure}
    \begin{subfigure}[b]{0.24\textwidth}
        \centering
        \includegraphics[width=0.97\linewidth]{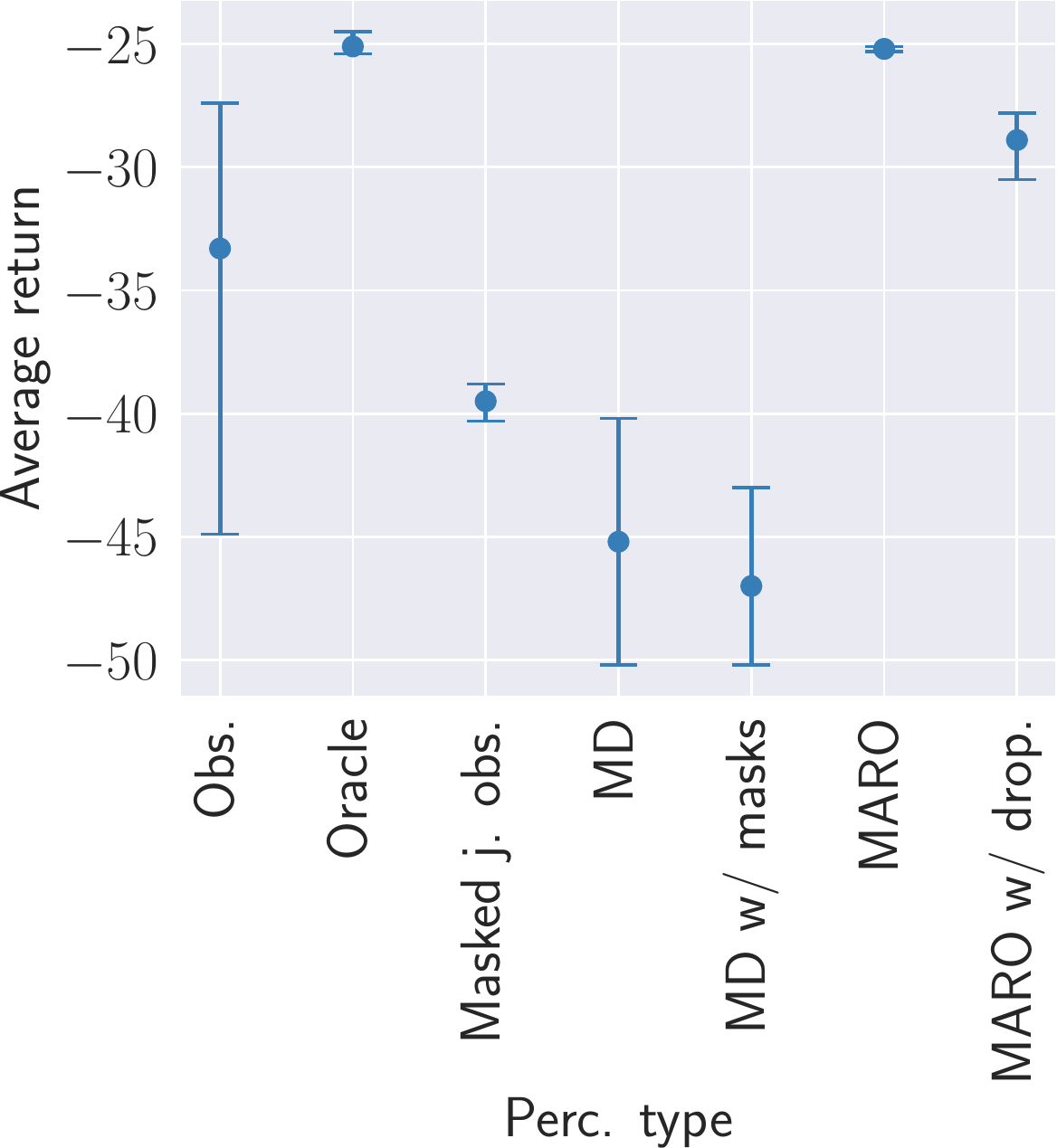}
        \caption{IPPO.}
    \end{subfigure}
    \begin{subfigure}[b]{0.24\textwidth}
        \centering
        \includegraphics[width=0.97\linewidth]{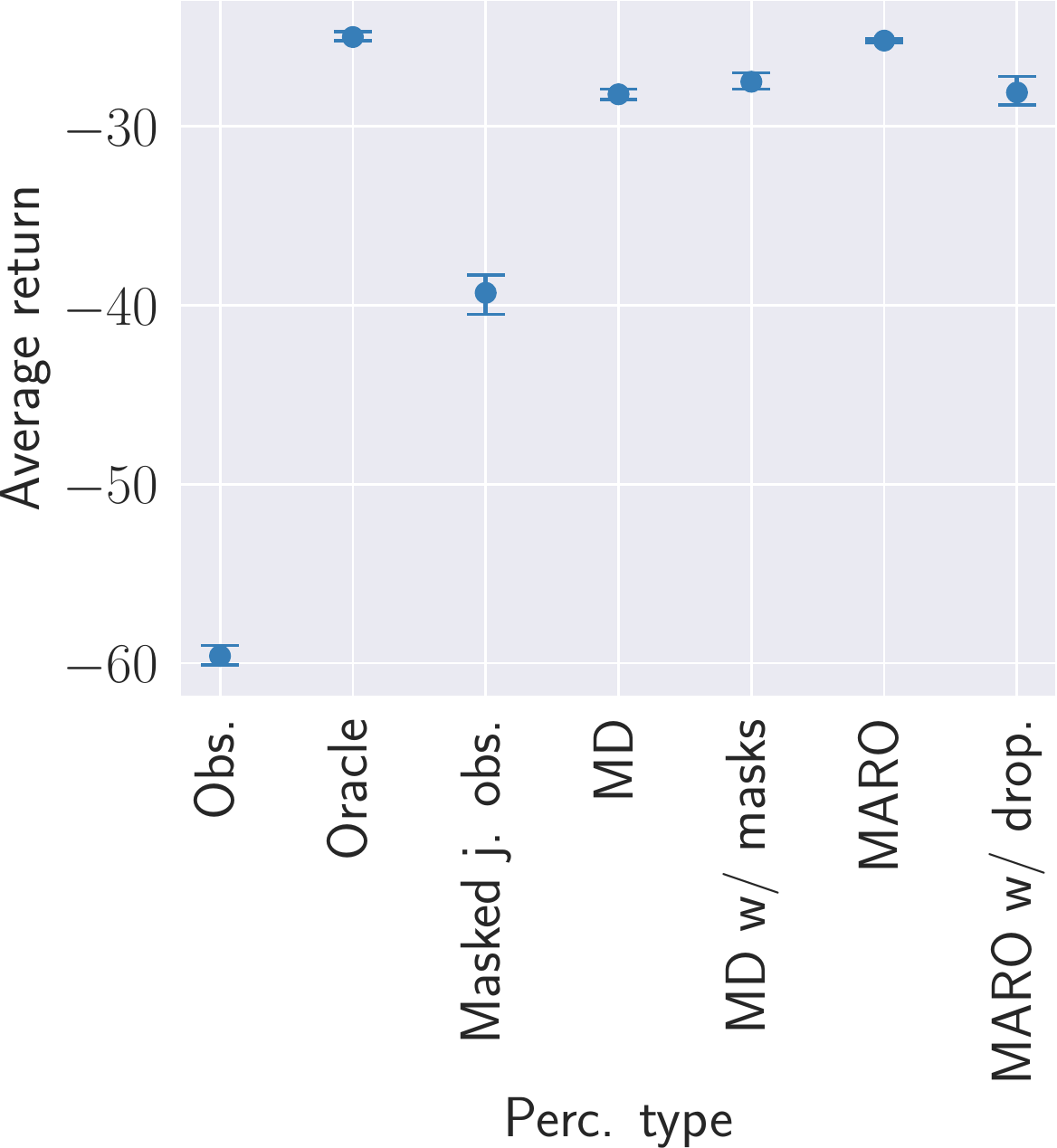}
        \caption{MAPPO.}
    \end{subfigure}
    \caption{(SpeakerListener) Mean episodic returns for $p_\textrm{default}$ at execution time.}
\end{figure}

\begin{figure}
    \centering
    \begin{subfigure}[b]{0.24\textwidth}
        \centering
        \includegraphics[width=0.97\linewidth]{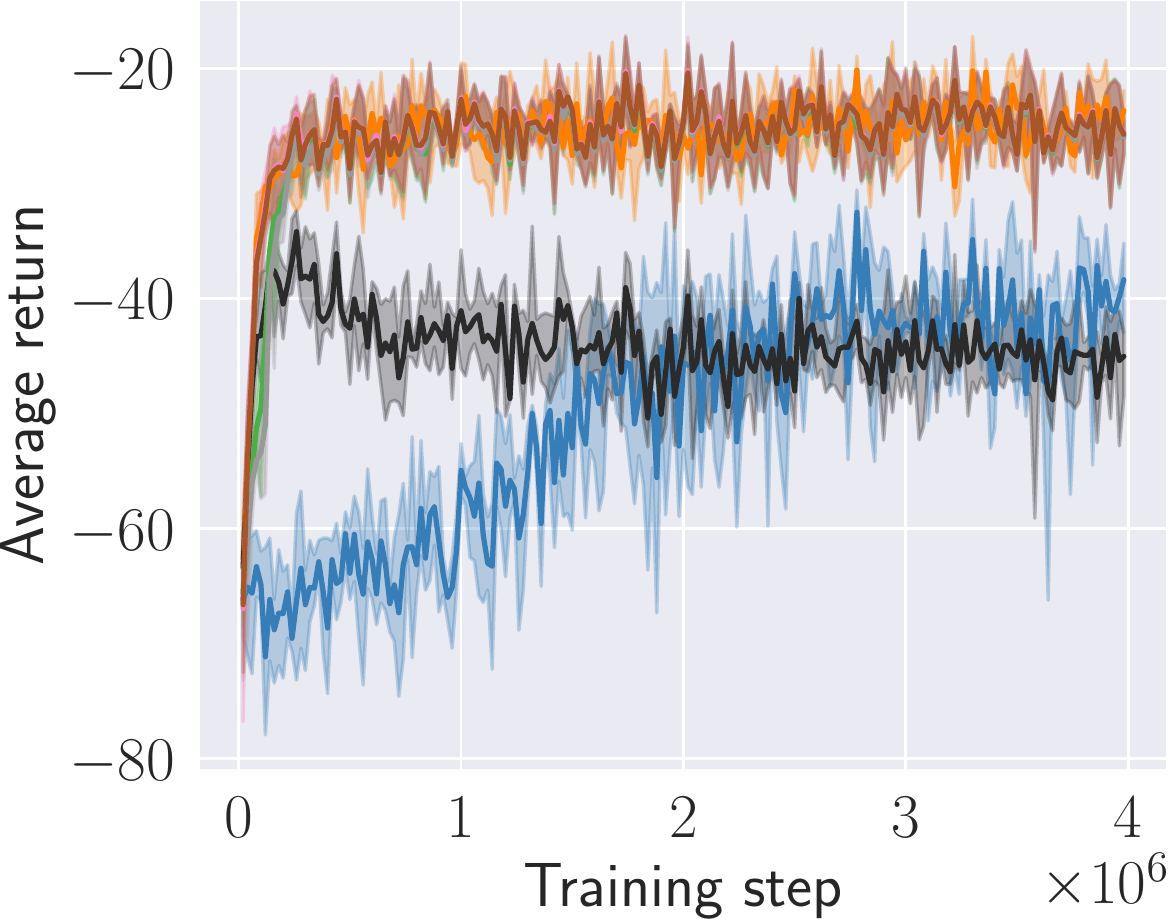}
        \caption{IQL.}
    \end{subfigure}
    \begin{subfigure}[b]{0.24\textwidth}
        \centering
        \includegraphics[width=0.97\linewidth]{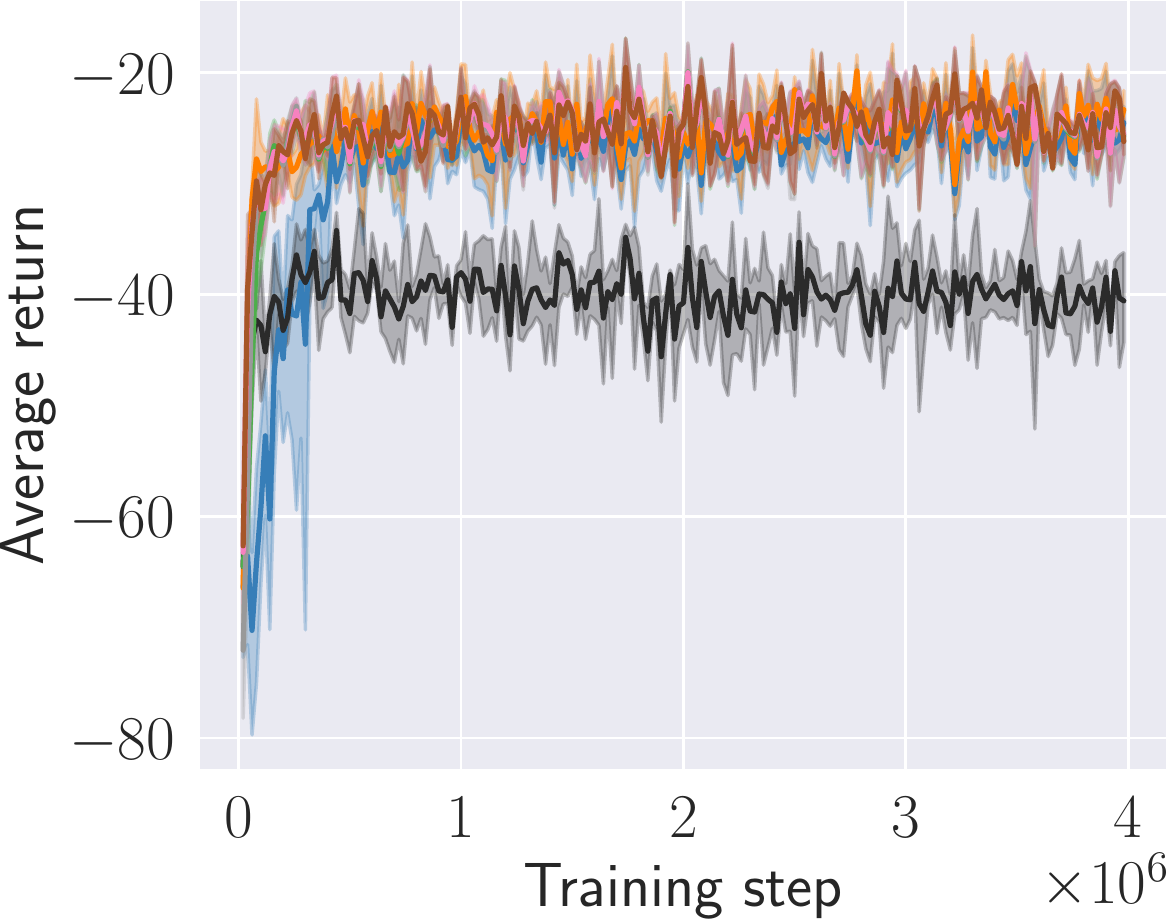}
        \caption{QMIX.}
    \end{subfigure}
    \begin{subfigure}[b]{0.24\textwidth}
        \centering
        \includegraphics[width=0.97\linewidth]{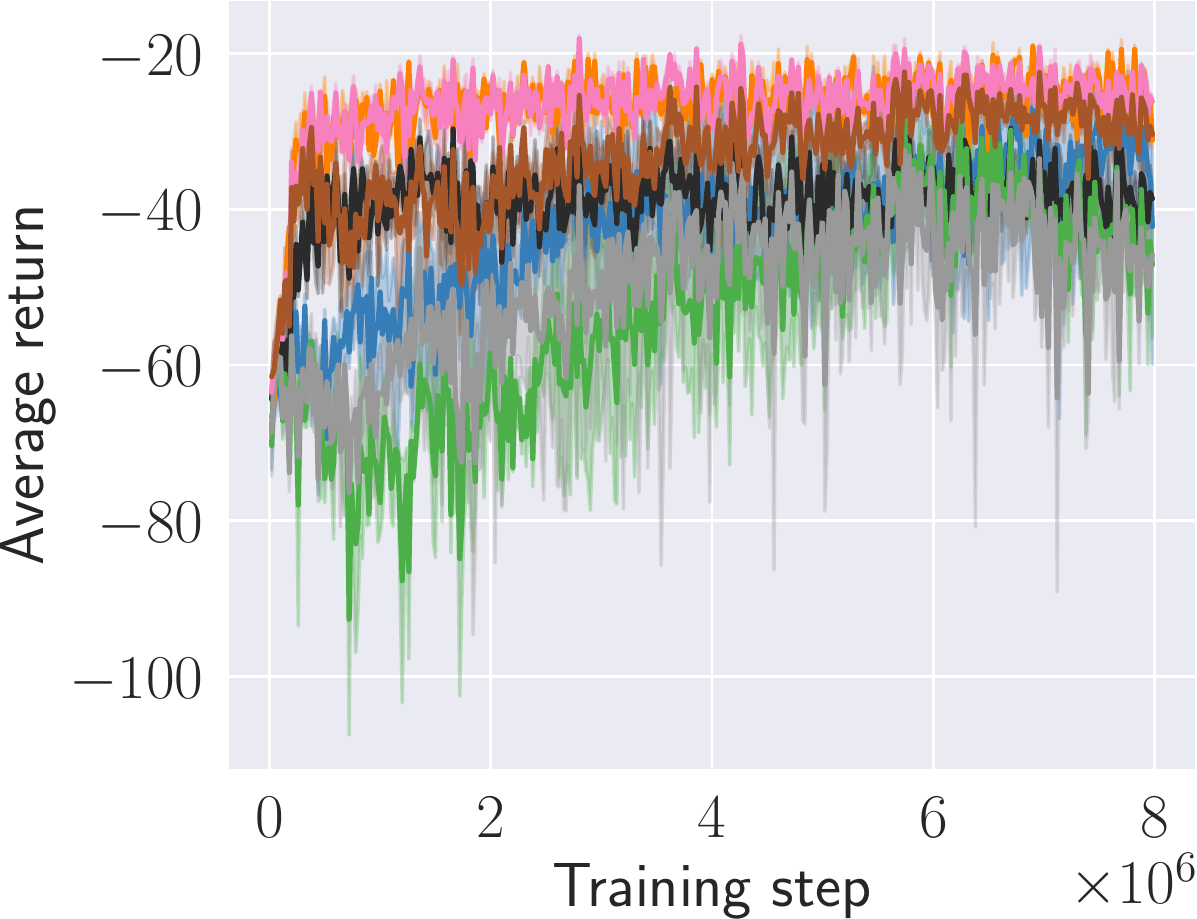}
        \caption{IPPO.}
    \end{subfigure}
    \begin{subfigure}[b]{0.24\textwidth}
        \centering
        \includegraphics[width=0.97\linewidth]{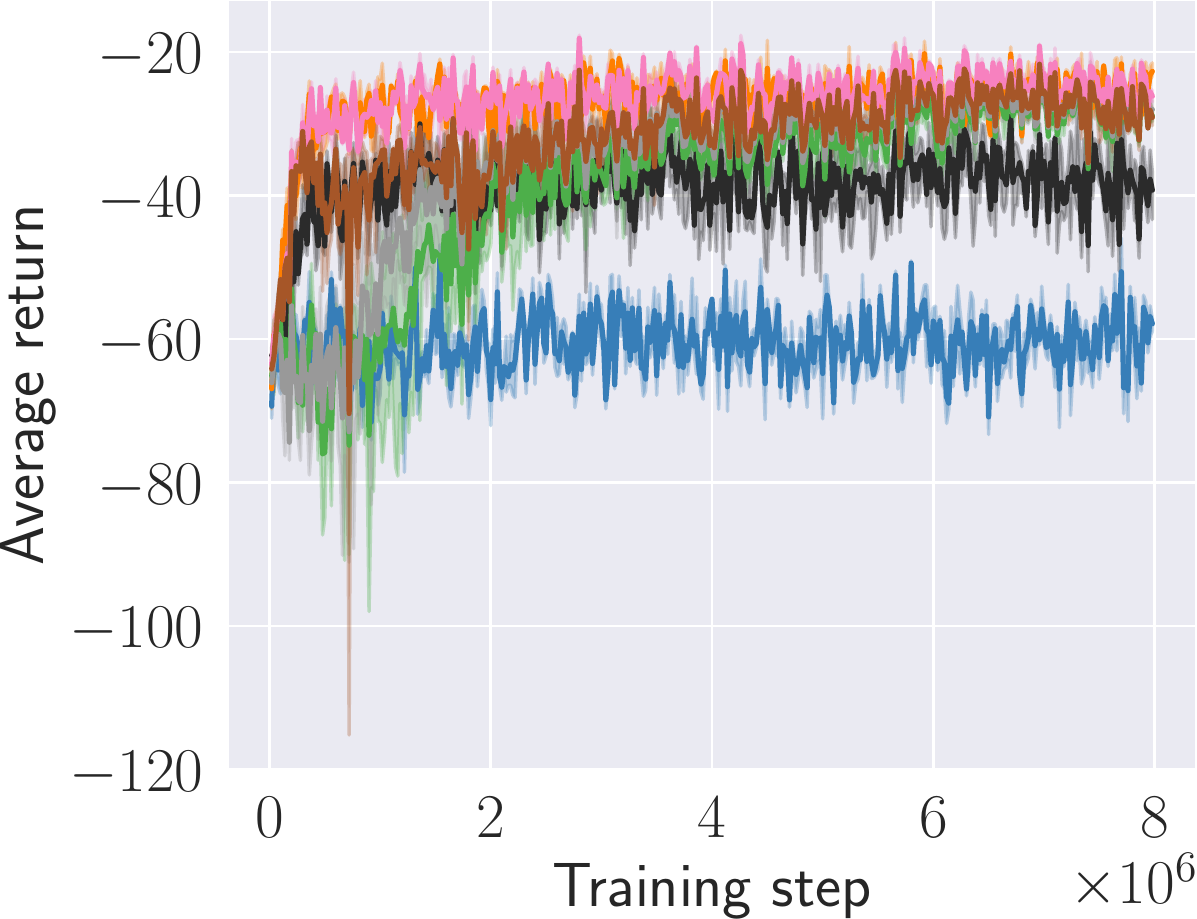}
        \caption{MAPPO.}
    \end{subfigure}
    \caption{(SpeakerListener) Mean episodic returns for $p_\textrm{default}$ during training.}
\end{figure}

\begin{figure}
    \centering
    \begin{subfigure}[b]{0.24\textwidth}
        \centering
        \includegraphics[width=0.97\linewidth]{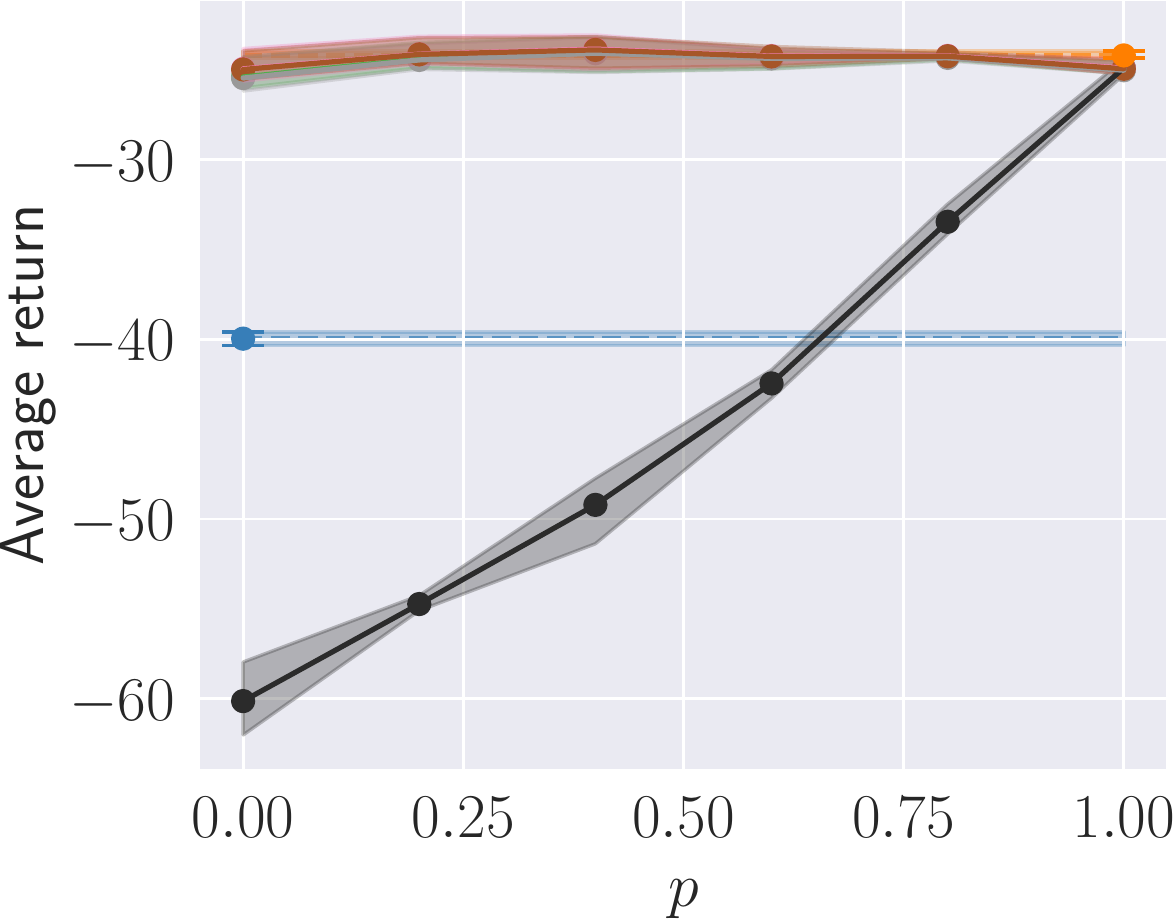}
        \caption{IQL.}
    \end{subfigure}
    \begin{subfigure}[b]{0.24\textwidth}
        \centering
        \includegraphics[width=0.97\linewidth]{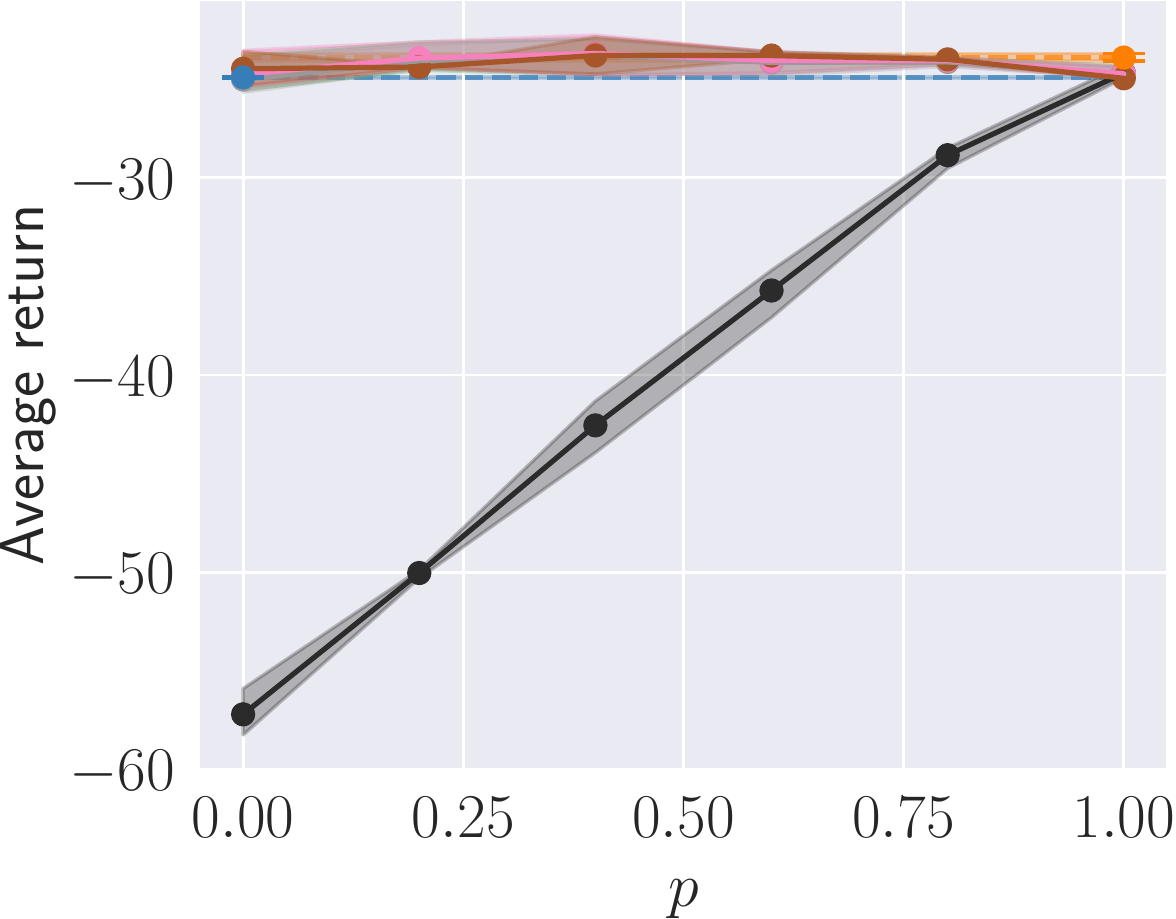}
        \caption{QMIX.}
    \end{subfigure}
    \begin{subfigure}[b]{0.24\textwidth}
        \centering
        \includegraphics[width=0.97\linewidth]{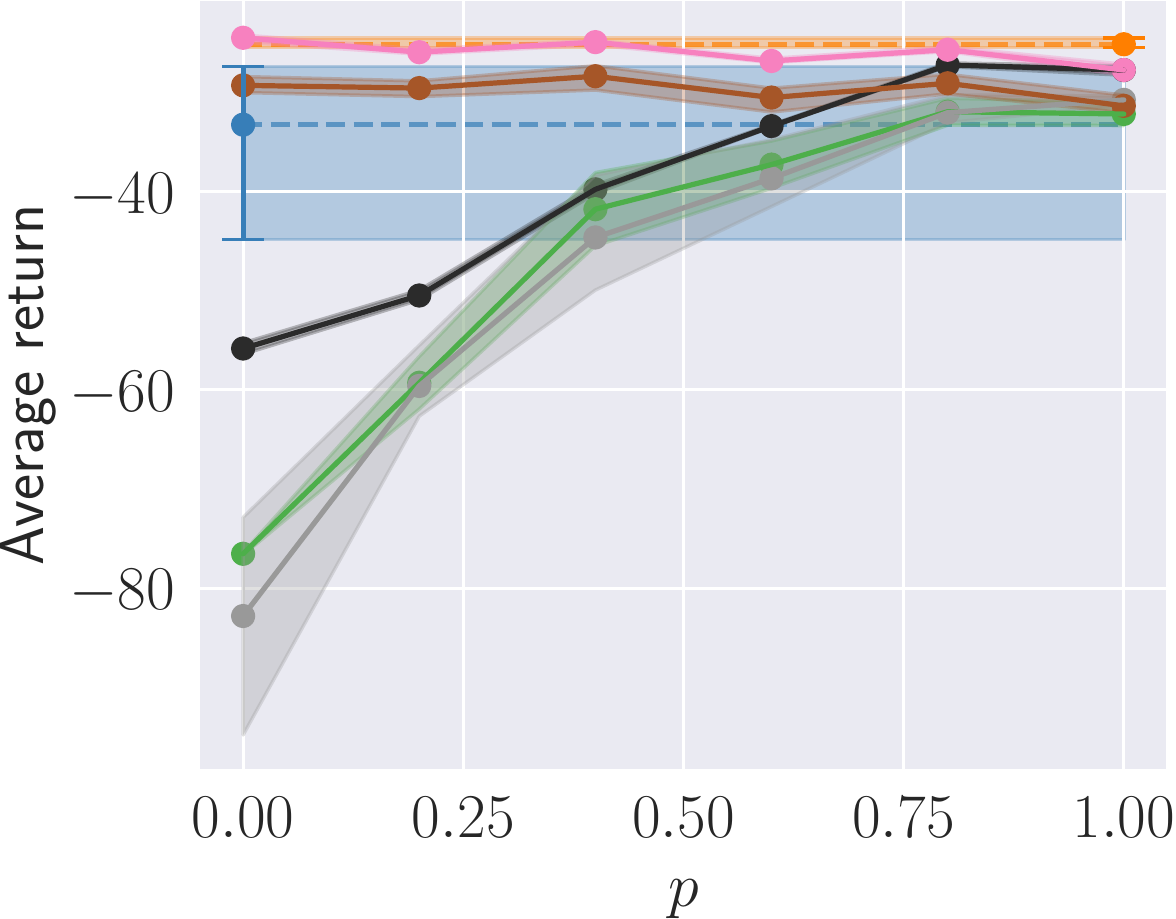}
        \caption{IPPO.}
    \end{subfigure}
    \begin{subfigure}[b]{0.24\textwidth}
        \centering
        \includegraphics[width=0.97\linewidth]{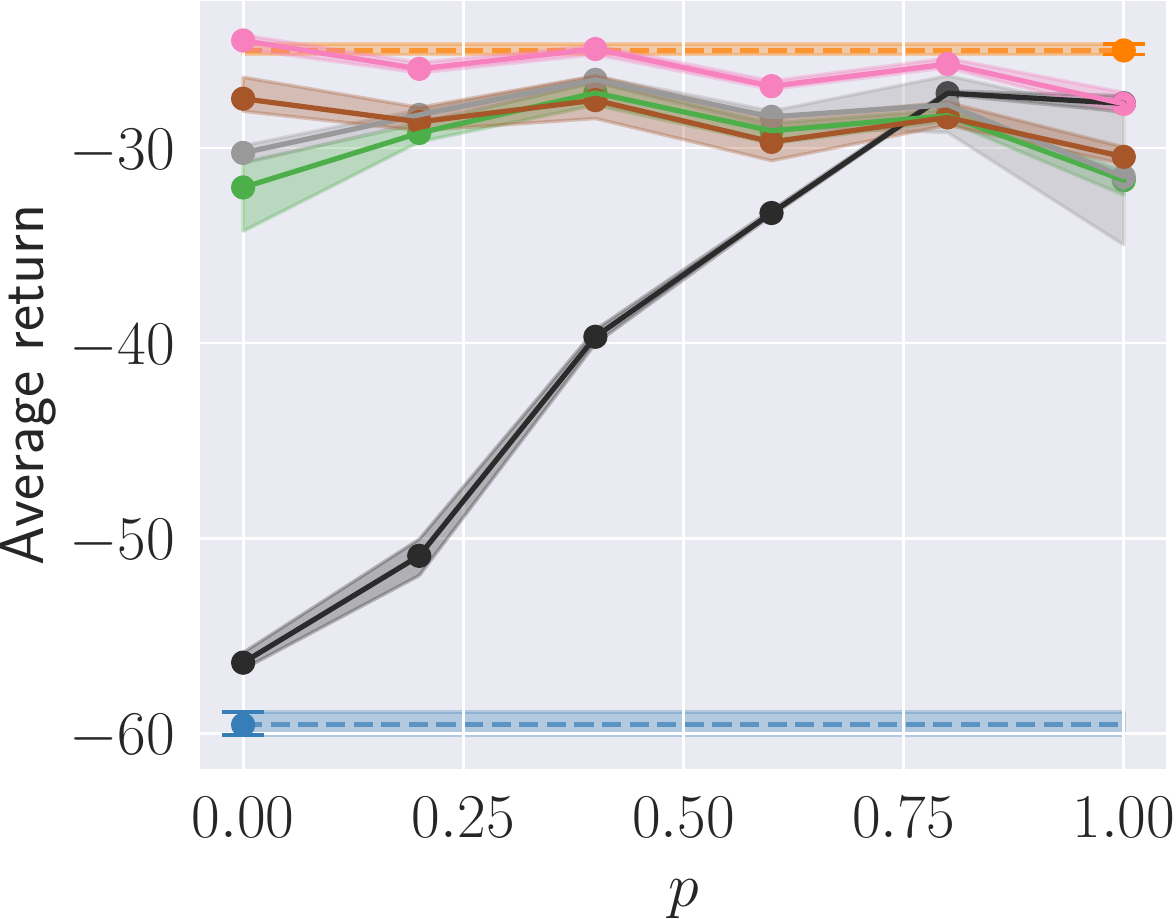}
        \caption{MAPPO.}
    \end{subfigure}
    \caption{(SpeakerListener) Mean episodic returns for different $p$ values at execution time.}
\end{figure}

\begin{figure}
    \centering
    \includegraphics[height=0.7cm]{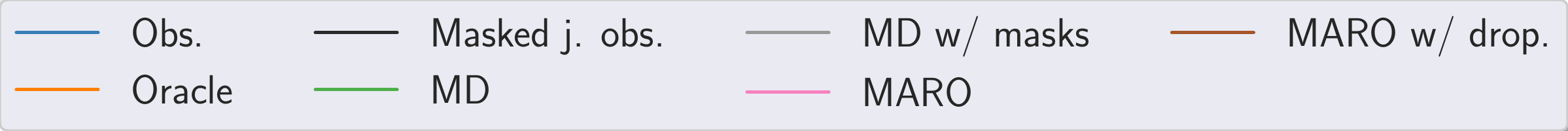}
    \caption{Legend.}
\end{figure}

\clearpage

\begin{table}
\centering
\noindent
\caption{(HearSee) Mean episodic returns for $p_{\textrm{default}}$ at execution time.}
\vspace{0.1cm}
\resizebox{\linewidth}{!}{%
\begin{tabular}{c c c c c c c c }\toprule
\multicolumn{1}{c }{\textbf{}} & \multicolumn{7}{c }{\textbf{HearSee ($p_\textrm{default}$)}} \\  
\cmidrule(lr){2-8}
\multicolumn{1}{ l }{\textbf{Algorithm}} & \textbf{Obs.} & \textbf{Oracle} & \textbf{Masked j. obs.} & \textbf{MD} & \textbf{MD w/ masks} & \textbf{MARO} & \textbf{MARO w/ drop.} \\
\cmidrule{1-8}
\multicolumn{1}{ l }{IQL} & -114.5 \tiny{(-2.0,+1.6)} & -24.5 \tiny{(-0.6,+0.9)} & -64.0 \tiny{(-2.1,+2.5)} & -34.8 \tiny{(-1.6,+1.7)} & -34.1 \tiny{(-1.3,+2.5)} & -29.6 \tiny{(-1.0,+0.7)} & -29.9 \tiny{(-1.1,+0.6)} \\ \cmidrule{1-8}
\multicolumn{1}{ l }{QMIX} & -62.2 \tiny{(-1.9,+1.4)} & -23.6 \tiny{(-0.7,+0.7)} & -67.4 \tiny{(-4.6,+3.0)} & -29.2 \tiny{(-1.3,+0.9)} & -29.1 \tiny{(-0.9,+0.8)} & -28.8 \tiny{(-1.2,+1.9)} & -26.0 \tiny{(-1.0,+0.8)} \\ \cmidrule{1-8}
\multicolumn{1}{ l }{IPPO} & -114.1 \tiny{(-37.7,+27.7)} & -25.7 \tiny{(-0.2,+0.2)} & -81.8 \tiny{(-5.8,+7.0)} & -102.9 \tiny{(-20.9,+22.5)} & -101.9 \tiny{(-19.4,+20.1)} & -31.4 \tiny{(-1.4,+1.2)} & -29.8 \tiny{(-1.1,+1.6)} \\ \cmidrule{1-8}
\multicolumn{1}{ l }{MAPPO} & -70.5 \tiny{(-18.3,+10.1)} & -25.7 \tiny{(-0.2,+0.2)} & -82.2 \tiny{(-6.4,+3.3)} & -32.4 \tiny{(-0.0,+0.0)} & -30.9 \tiny{(-3.5,+2.6)} & -31.4 \tiny{(-1.4,+1.2)} & -28.7 \tiny{(-1.4,+1.2)} \\
\bottomrule
\end{tabular}
}
\end{table}

\begin{figure}
    \centering
    \begin{subfigure}[b]{0.24\textwidth}
        \centering
        \includegraphics[width=0.97\linewidth]{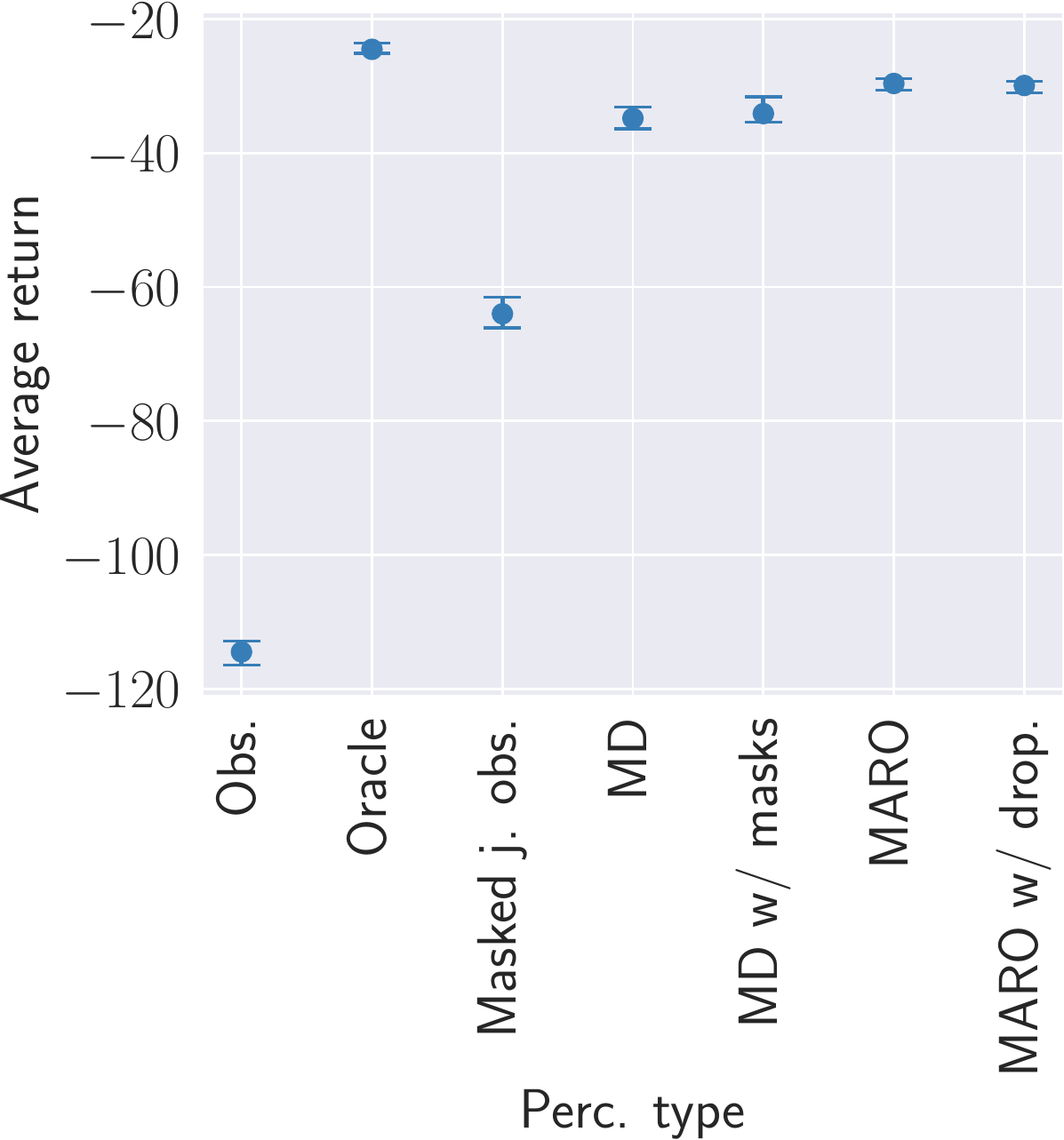}
        \caption{IQL.}
    \end{subfigure}
    \begin{subfigure}[b]{0.24\textwidth}
        \centering
        \includegraphics[width=0.97\linewidth]{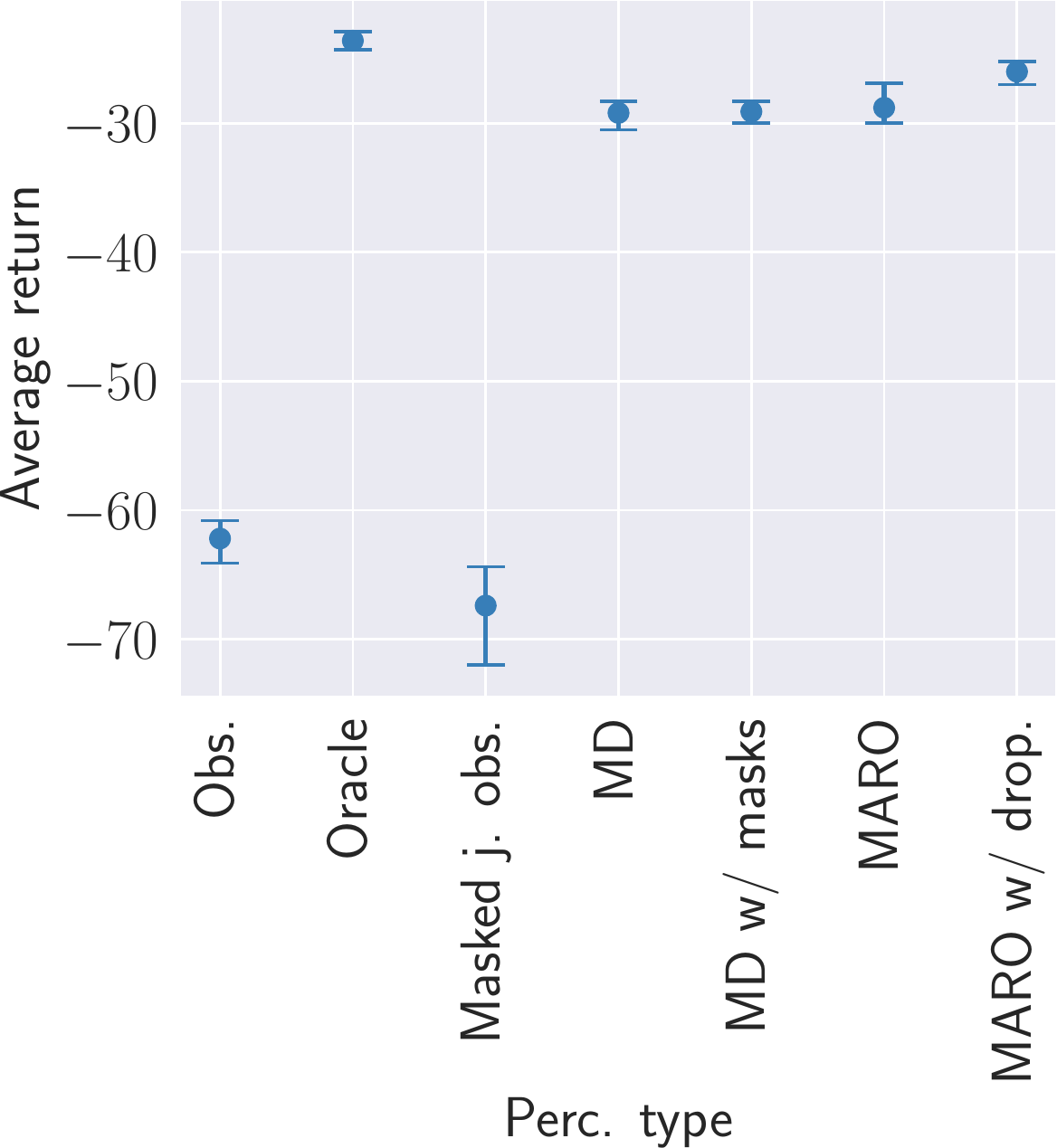}
        \caption{QMIX.}
    \end{subfigure}
    \begin{subfigure}[b]{0.24\textwidth}
        \centering
        \includegraphics[width=0.97\linewidth]{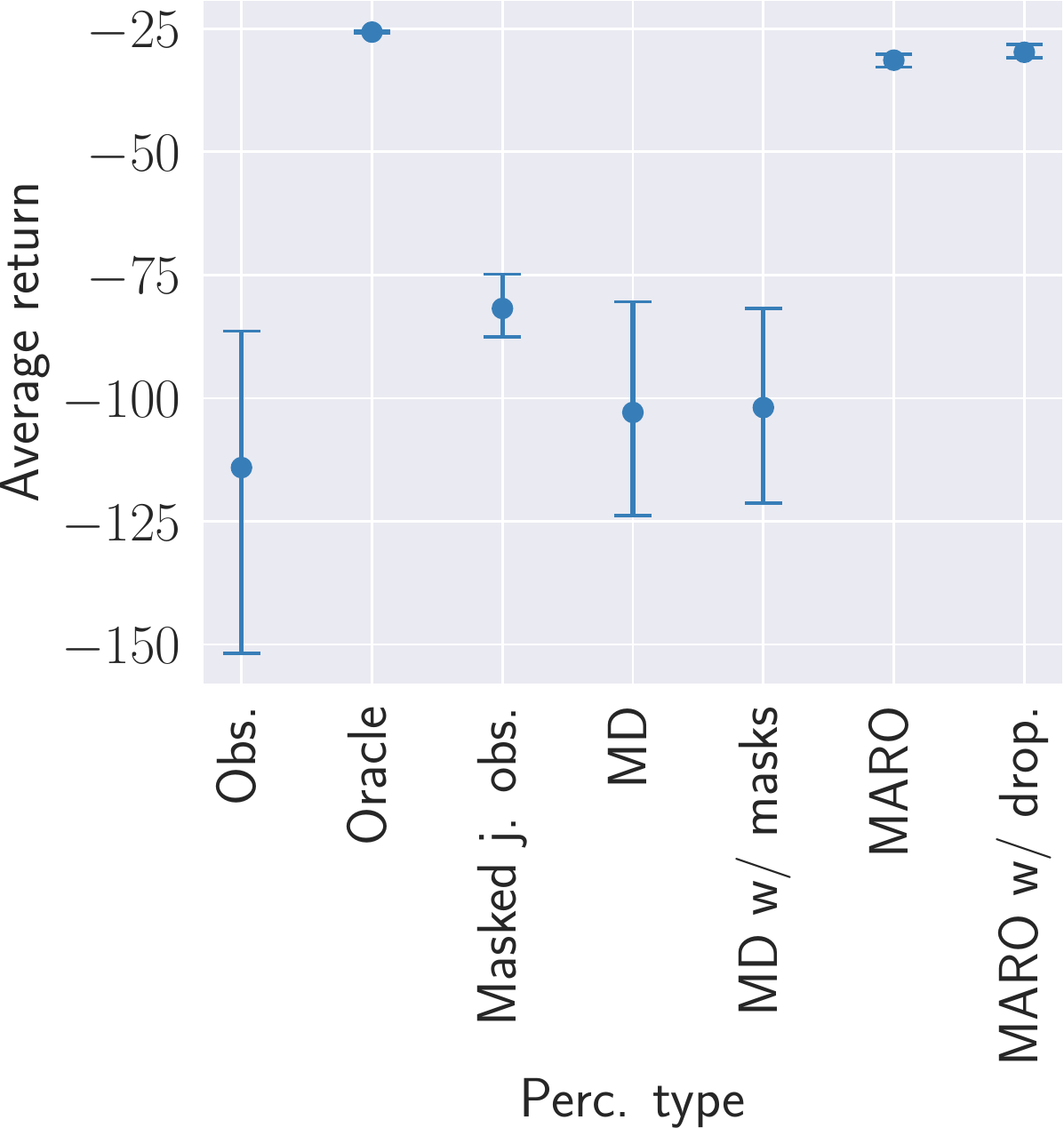}
        \caption{IPPO.}
    \end{subfigure}
    \begin{subfigure}[b]{0.24\textwidth}
        \centering
        \includegraphics[width=0.97\linewidth]{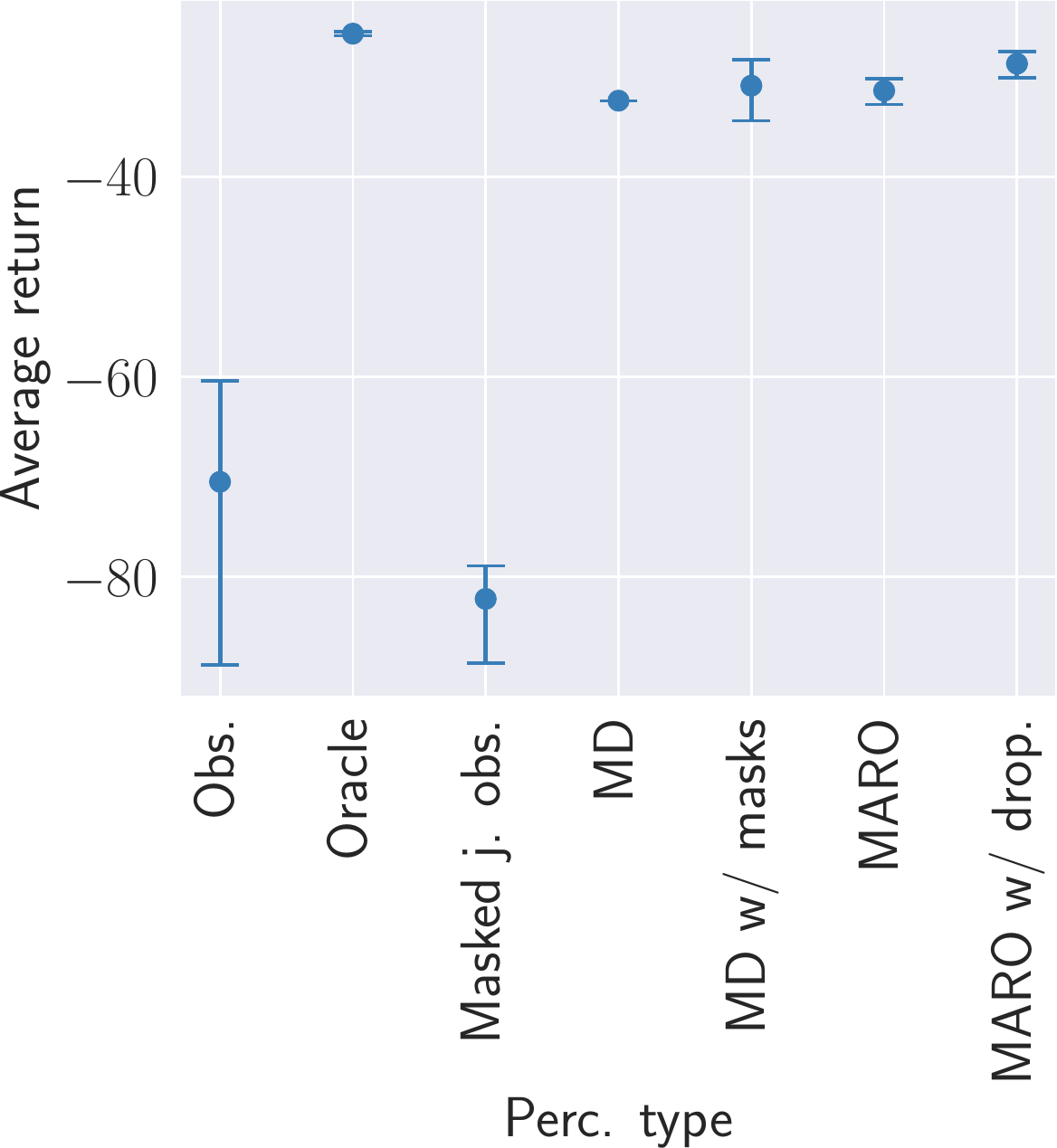}
        \caption{MAPPO.}
    \end{subfigure}
    \caption{(HearSee) Mean episodic returns for $p_\textrm{default}$ at execution time.}
\end{figure}

\begin{figure}
    \centering
    \begin{subfigure}[b]{0.24\textwidth}
        \centering
        \includegraphics[width=0.97\linewidth]{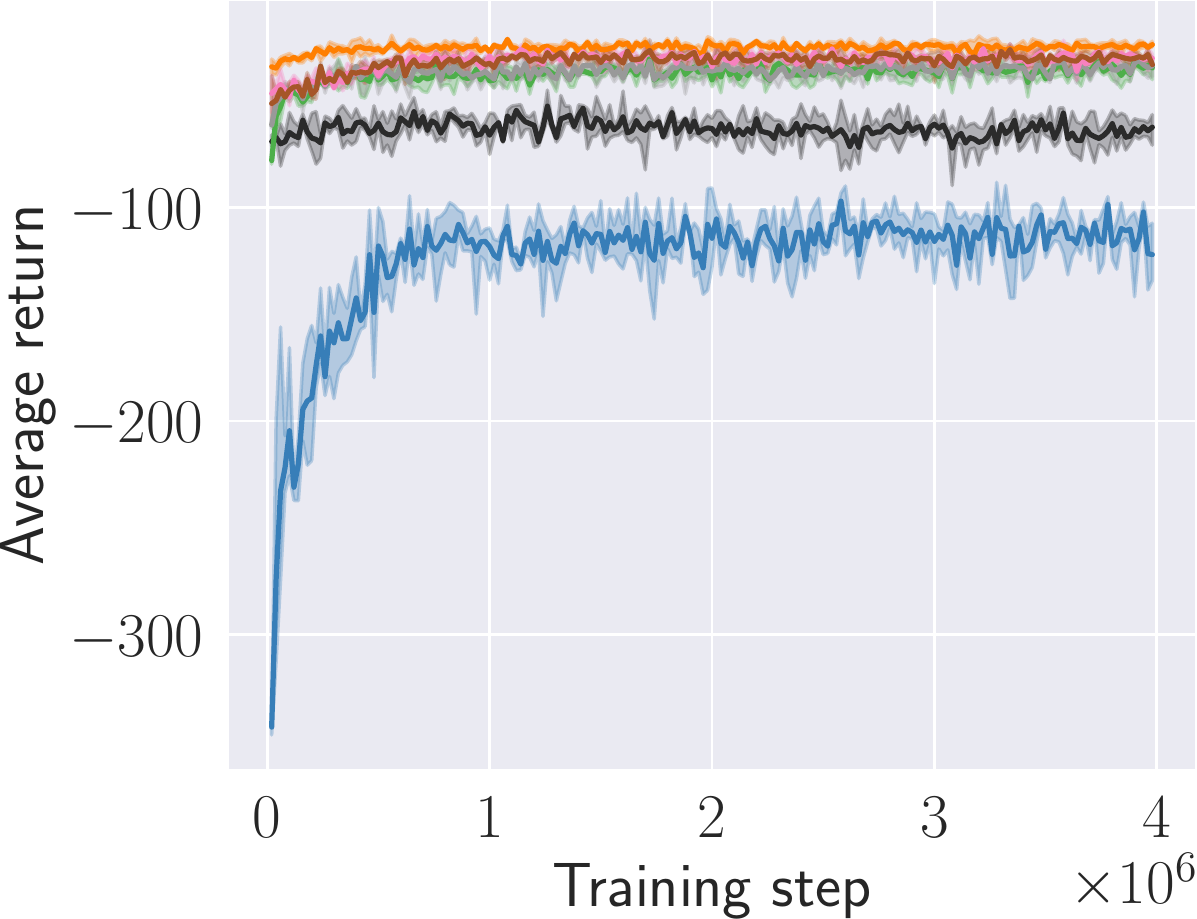}
        \caption{IQL.}
    \end{subfigure}
    \begin{subfigure}[b]{0.24\textwidth}
        \centering
        \includegraphics[width=0.97\linewidth]{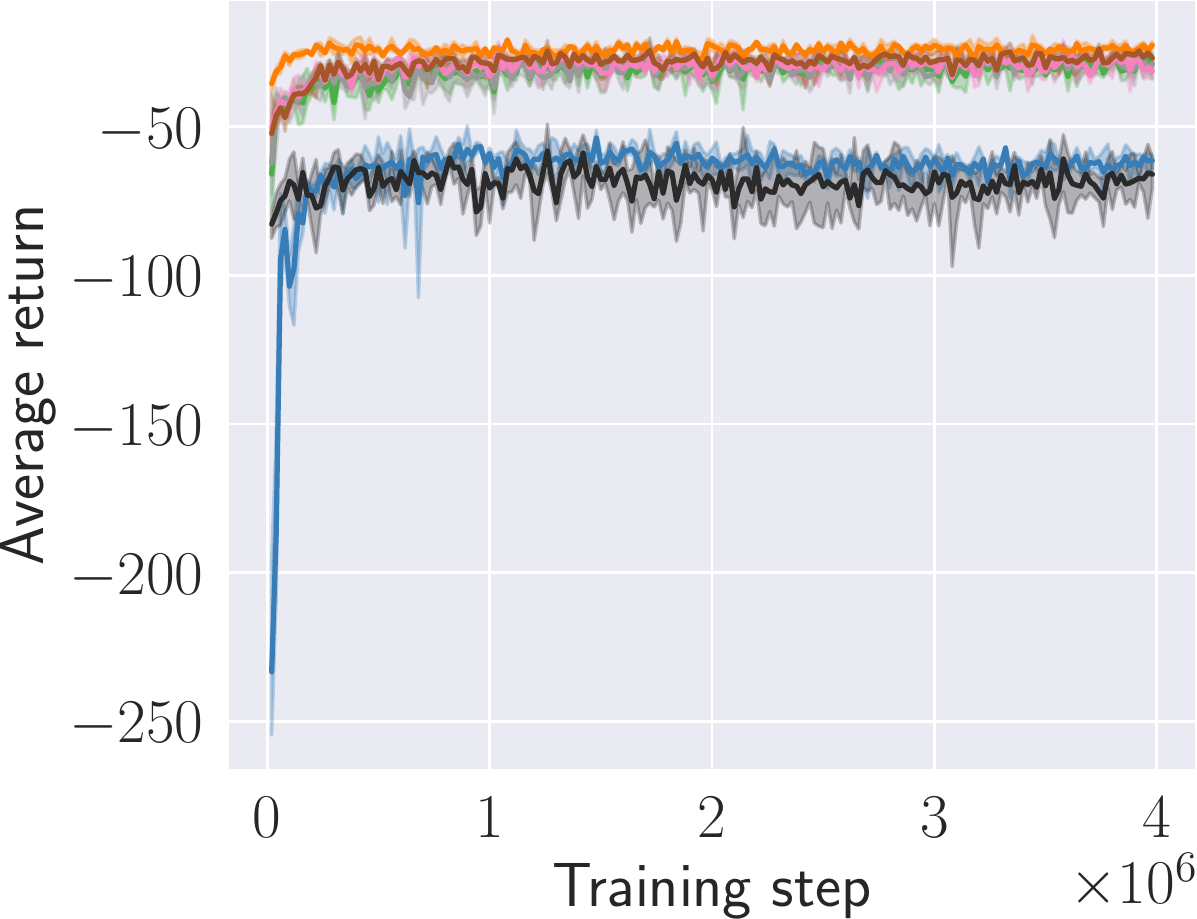}
        \caption{QMIX.}
    \end{subfigure}
    \begin{subfigure}[b]{0.24\textwidth}
        \centering
        \includegraphics[width=0.97\linewidth]{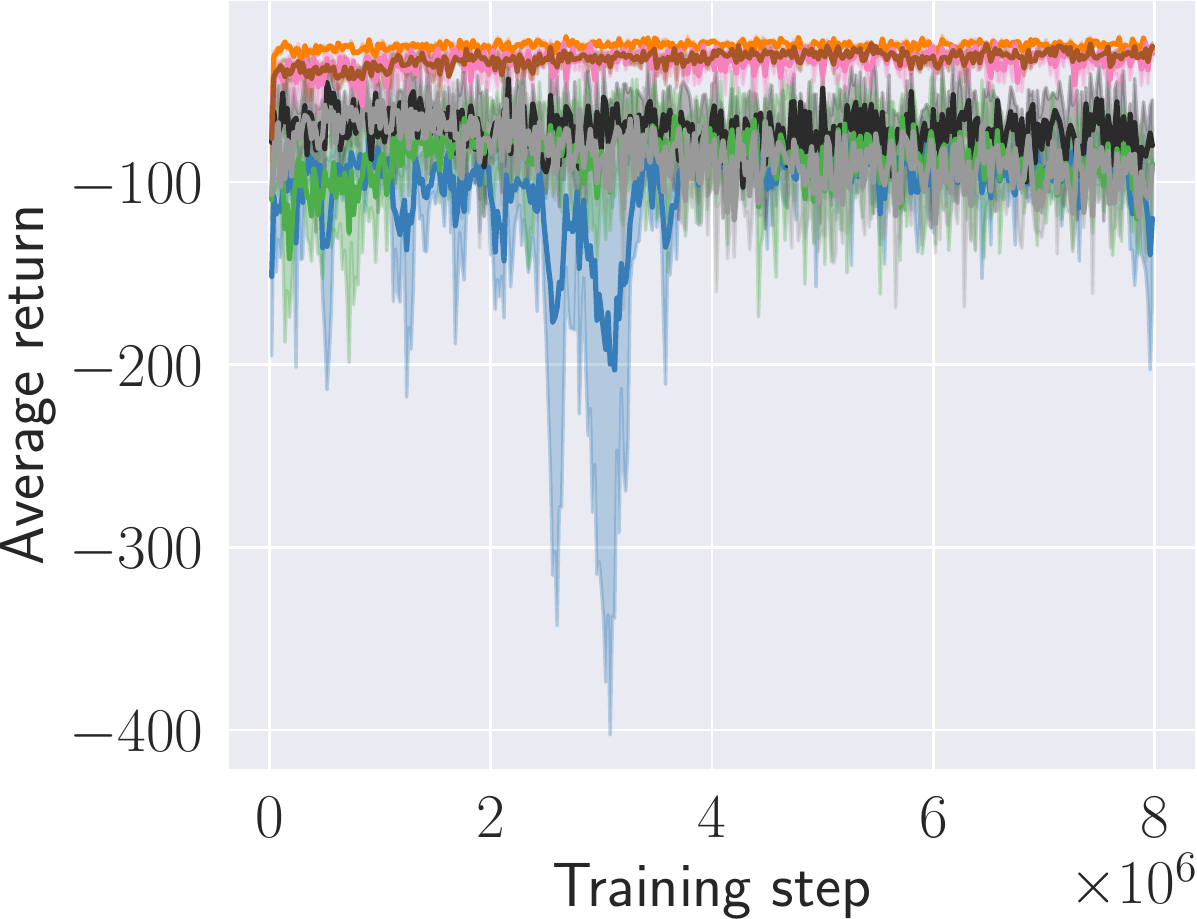}
        \caption{IPPO.}
    \end{subfigure}
    \begin{subfigure}[b]{0.24\textwidth}
        \centering
        \includegraphics[width=0.97\linewidth]{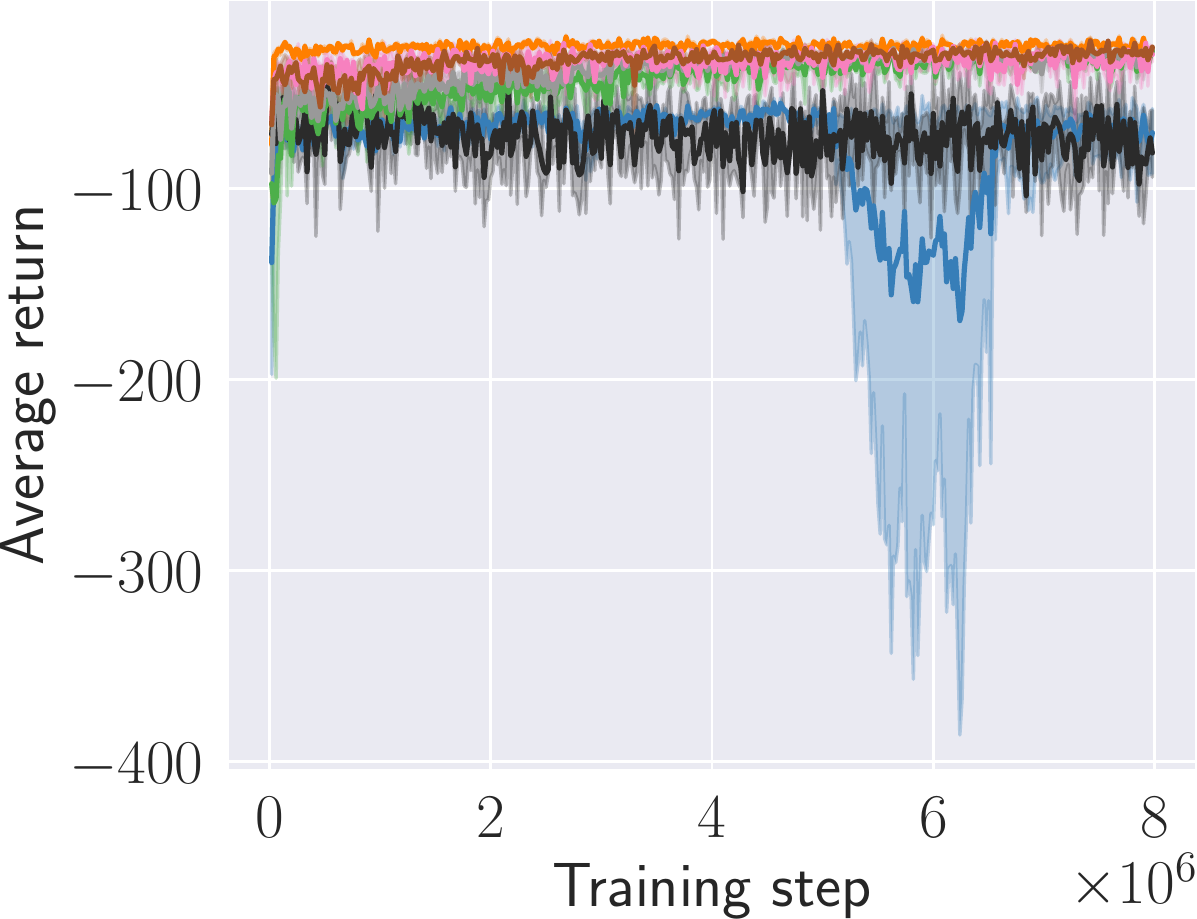}
        \caption{MAPPO.}
    \end{subfigure}
    \caption{(HearSee) Mean episodic returns for $p_\textrm{default}$ during training.}
\end{figure}

\begin{figure}
    \centering
    \begin{subfigure}[b]{0.24\textwidth}
        \centering
        \includegraphics[width=0.97\linewidth]{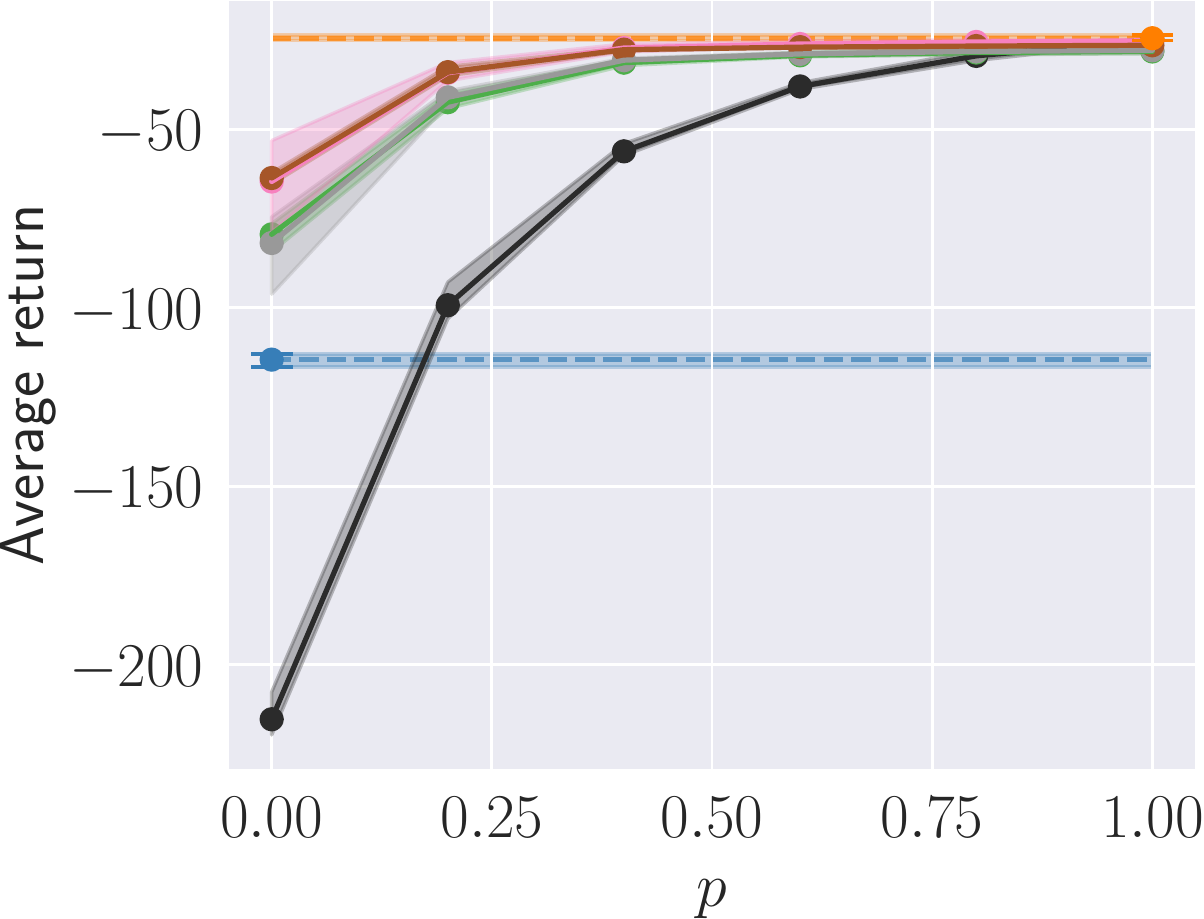}
        \caption{IQL.}
    \end{subfigure}
    \begin{subfigure}[b]{0.24\textwidth}
        \centering
        \includegraphics[width=0.97\linewidth]{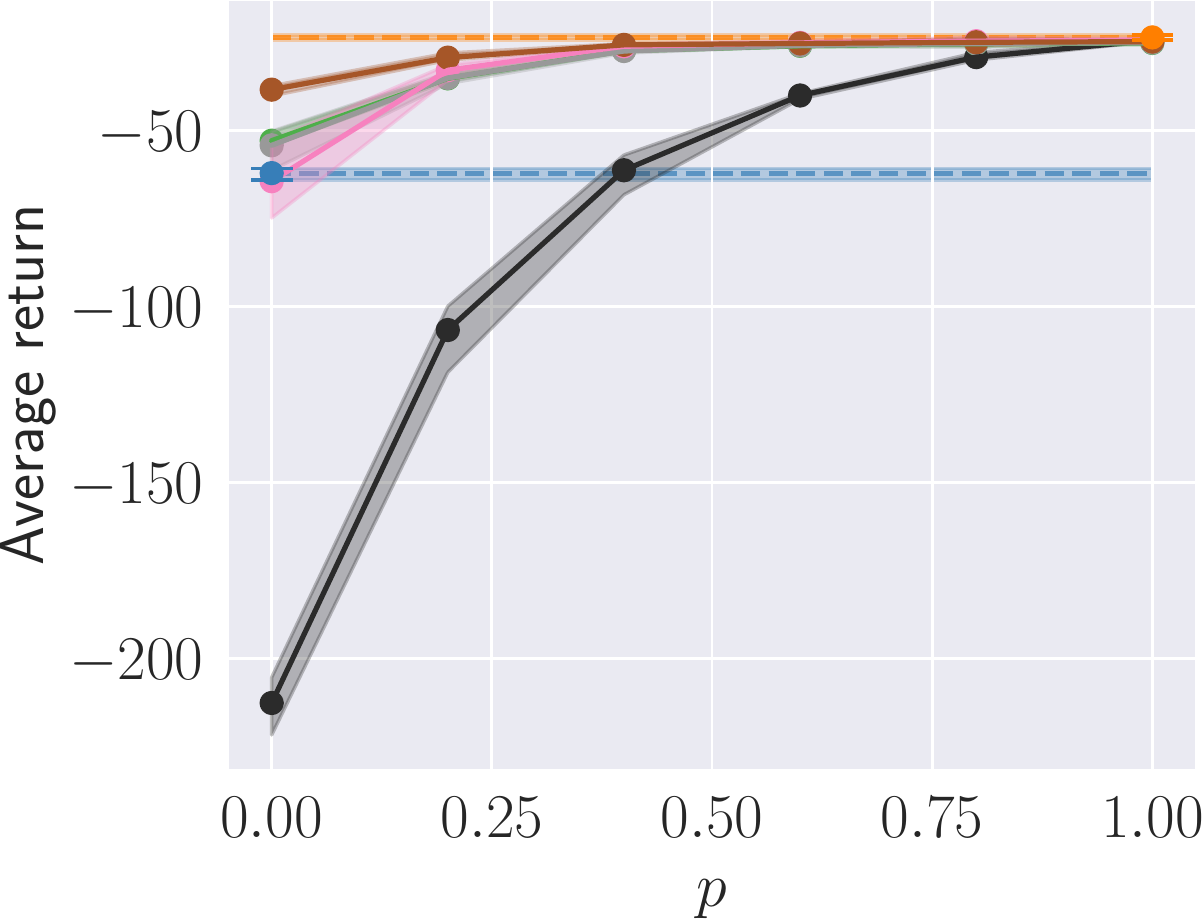}
        \caption{QMIX.}
    \end{subfigure}
    \begin{subfigure}[b]{0.24\textwidth}
        \centering
        \includegraphics[width=0.97\linewidth]{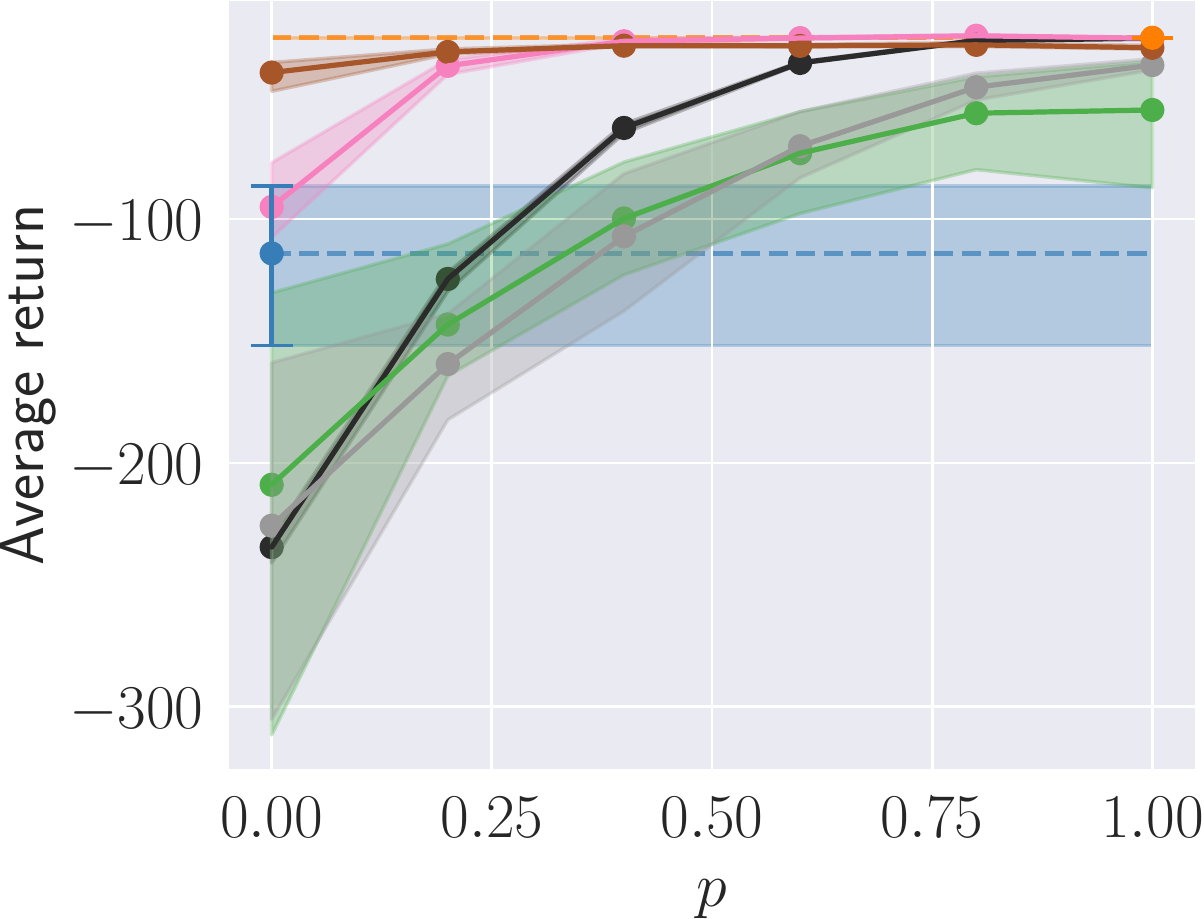}
        \caption{IPPO.}
    \end{subfigure}
    \begin{subfigure}[b]{0.24\textwidth}
        \centering
        \includegraphics[width=0.97\linewidth]{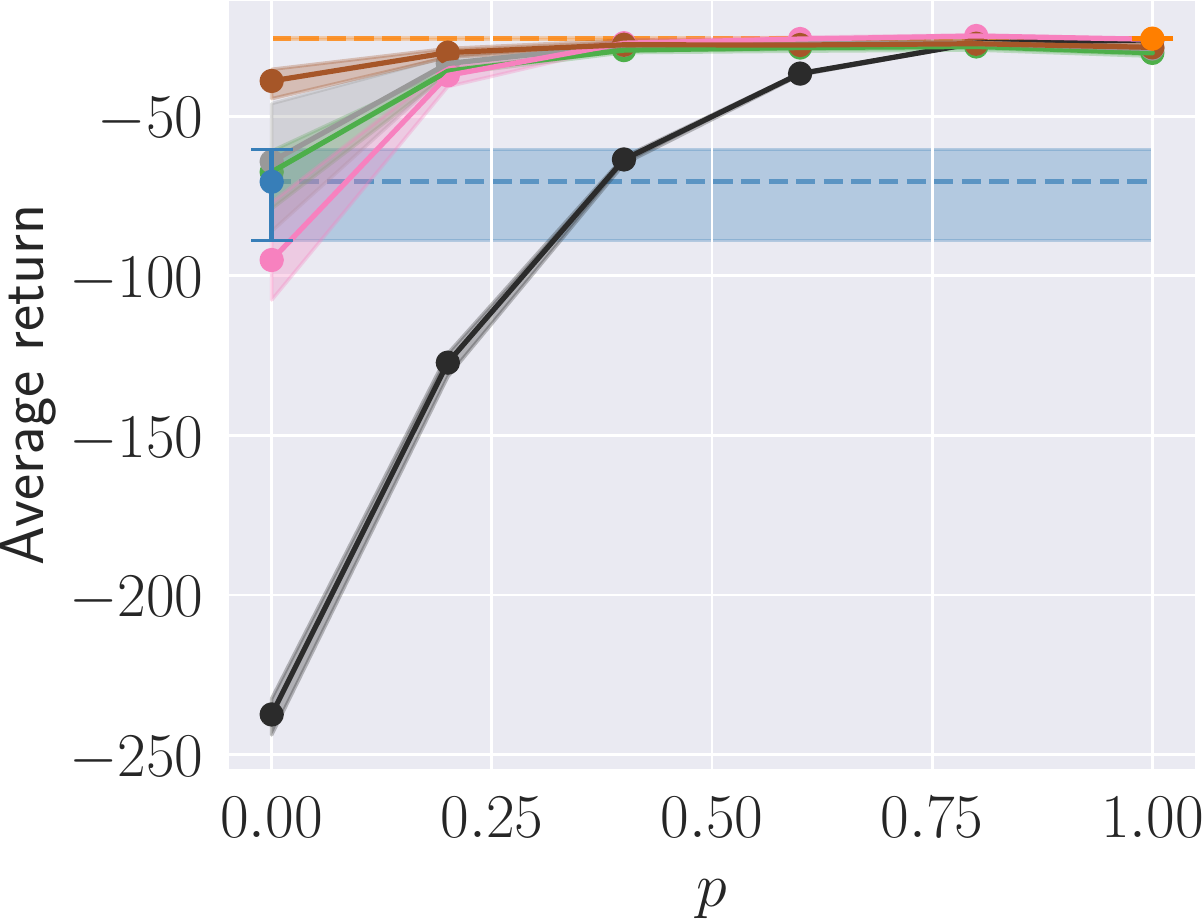}
        \caption{MAPPO.}
    \end{subfigure}
    \caption{(HearSee) Mean episodic returns for different $p$ values at execution time.}
\end{figure}

\begin{figure}
    \centering
    \includegraphics[height=0.7cm]{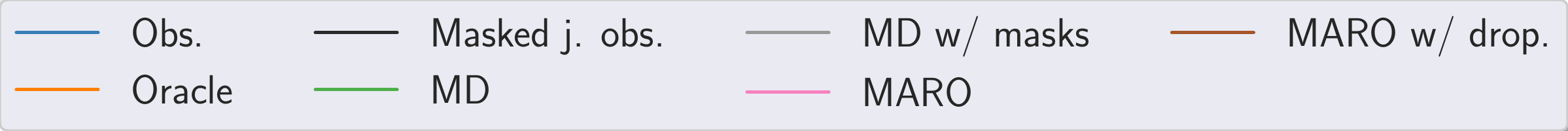}
    \caption{Legend.}
\end{figure}

\clearpage

\begin{table}
\centering
\noindent
\caption{(SpreadXY-2) Mean episodic returns for $p_{\textrm{default}}$ at execution time.}
\vspace{0.1cm}
\resizebox{\linewidth}{!}{%
\begin{tabular}{c c c c c c c c }\toprule
\multicolumn{1}{c }{\textbf{}} & \multicolumn{7}{c }{\textbf{SpreadXY-2 ($p_\textrm{default}$)}} \\  
\cmidrule(lr){2-8}
\multicolumn{1}{ l }{\textbf{Algorithm}} & \textbf{Obs.} & \textbf{Oracle} & \textbf{Masked j. obs.} & \textbf{MD} & \textbf{MD w/ masks} & \textbf{MARO} & \textbf{MARO w/ drop.} \\
\cmidrule{1-8}
\multicolumn{1}{ l }{IQL} & -199.6 \tiny{(-0.9,+0.9)} & -139.4 \tiny{(-0.5,+0.6)} & -202.9 \tiny{(-2.3,+3.7)} & -165.2 \tiny{(-0.5,+0.7)} & -160.7 \tiny{(-1.7,+1.6)} & -148.0 \tiny{(-0.4,+0.5)} & -158.8 \tiny{(-1.2,+1.2)} \\ \cmidrule{1-8}
\multicolumn{1}{ l }{QMIX} & -177.8 \tiny{(-7.6,+4.1)} & -138.6 \tiny{(-0.4,+0.4)} & -201.2 \tiny{(-2.3,+2.3)} & -157.2 \tiny{(-1.4,+0.7)} & -154.5 \tiny{(-0.5,+1.0)} & -145.7 \tiny{(-0.9,+0.5)} & -152.7 \tiny{(-0.5,+0.7)} \\ \cmidrule{1-8}
\multicolumn{1}{ l }{IPPO} & -235.6 \tiny{(-0.6,+0.6)} & -160.8 \tiny{(-2.8,+1.4)} & -214.8 \tiny{(-5.4,+4.6)} & -184.9 \tiny{(-5.8,+3.6)} & -175.9 \tiny{(-3.0,+2.7)} & -160.7 \tiny{(-1.3,+1.0)} & -166.6 \tiny{(-1.0,+1.0)} \\ \cmidrule{1-8}
\multicolumn{1}{ l }{MAPPO} & -212.6 \tiny{(-13.7,+24.5)} & -160.8 \tiny{(-1.8,+3.1)} & -221.5 \tiny{(-2.3,+3.2)} & -181.1 \tiny{(-3.9,+2.2)} & -163.5 \tiny{(-2.3,+2.7)} & -161.2 \tiny{(-0.9,+0.8)} & -159.6 \tiny{(-0.3,+0.3)} \\
\bottomrule
\end{tabular}
}
\end{table}

\begin{figure}
    \centering
    \begin{subfigure}[b]{0.24\textwidth}
        \centering
        \includegraphics[width=0.97\linewidth]{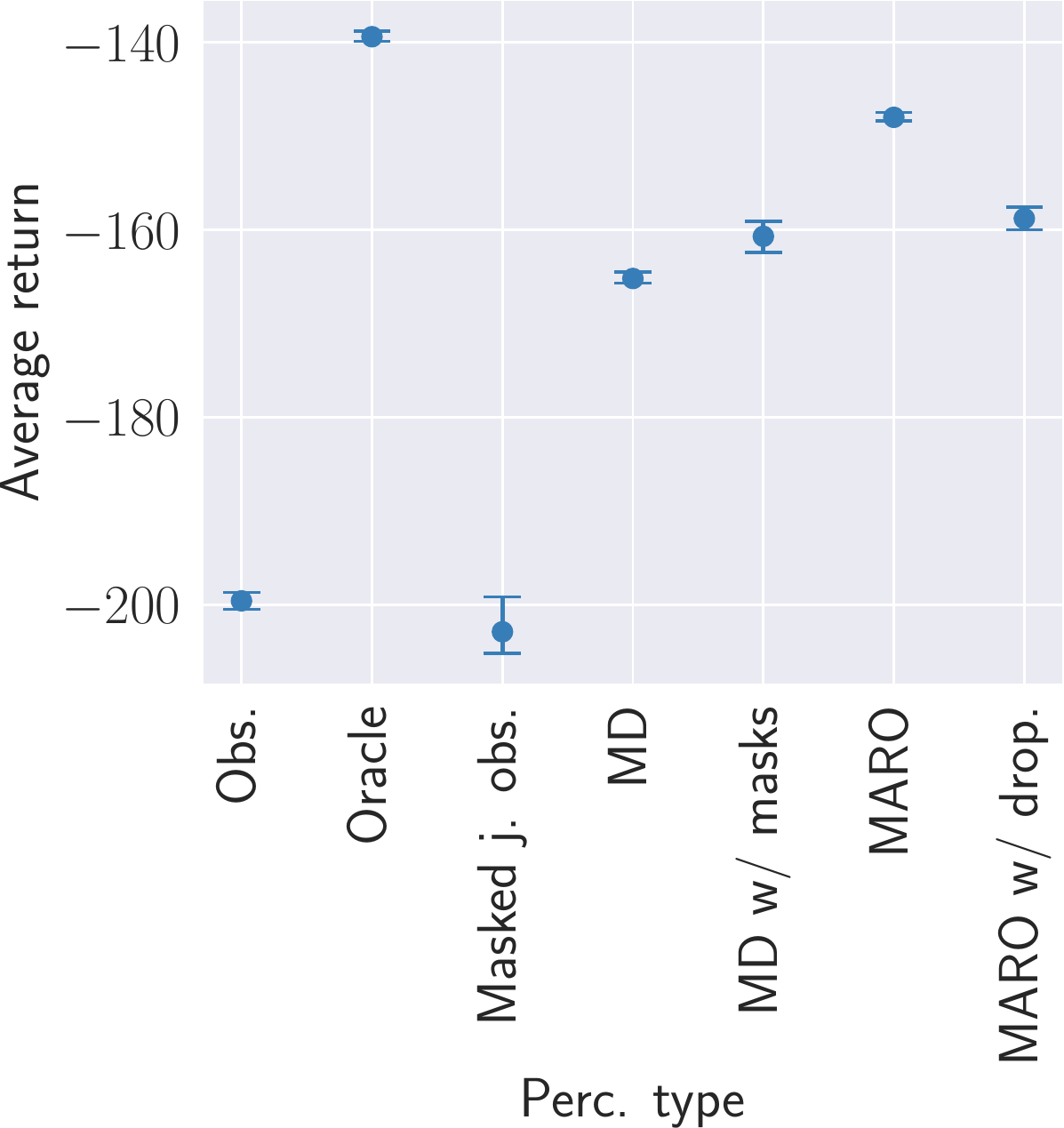}
        \caption{IQL.}
    \end{subfigure}
    \begin{subfigure}[b]{0.24\textwidth}
        \centering
        \includegraphics[width=0.97\linewidth]{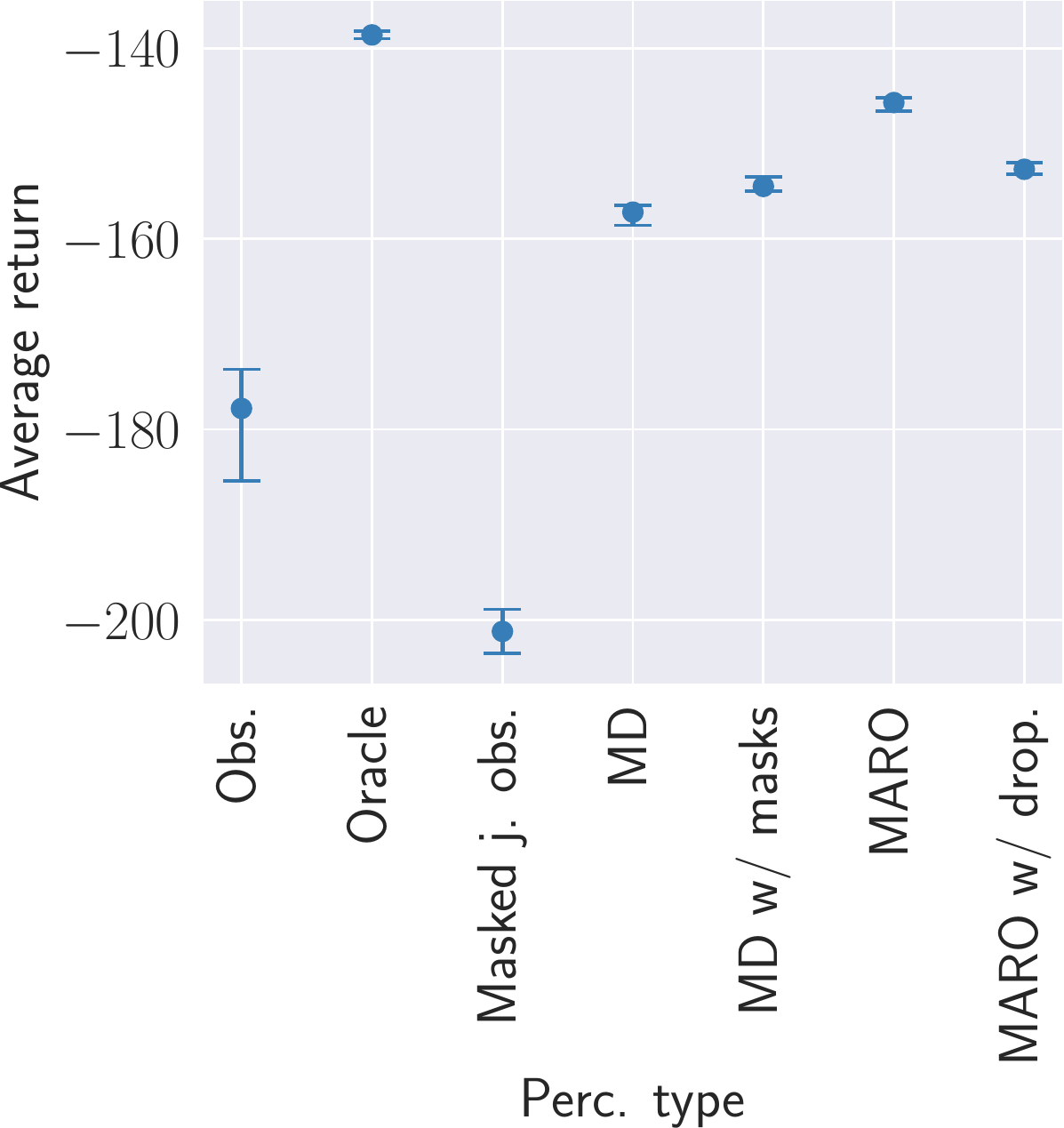}
        \caption{QMIX.}
    \end{subfigure}
    \begin{subfigure}[b]{0.24\textwidth}
        \centering
        \includegraphics[width=0.97\linewidth]{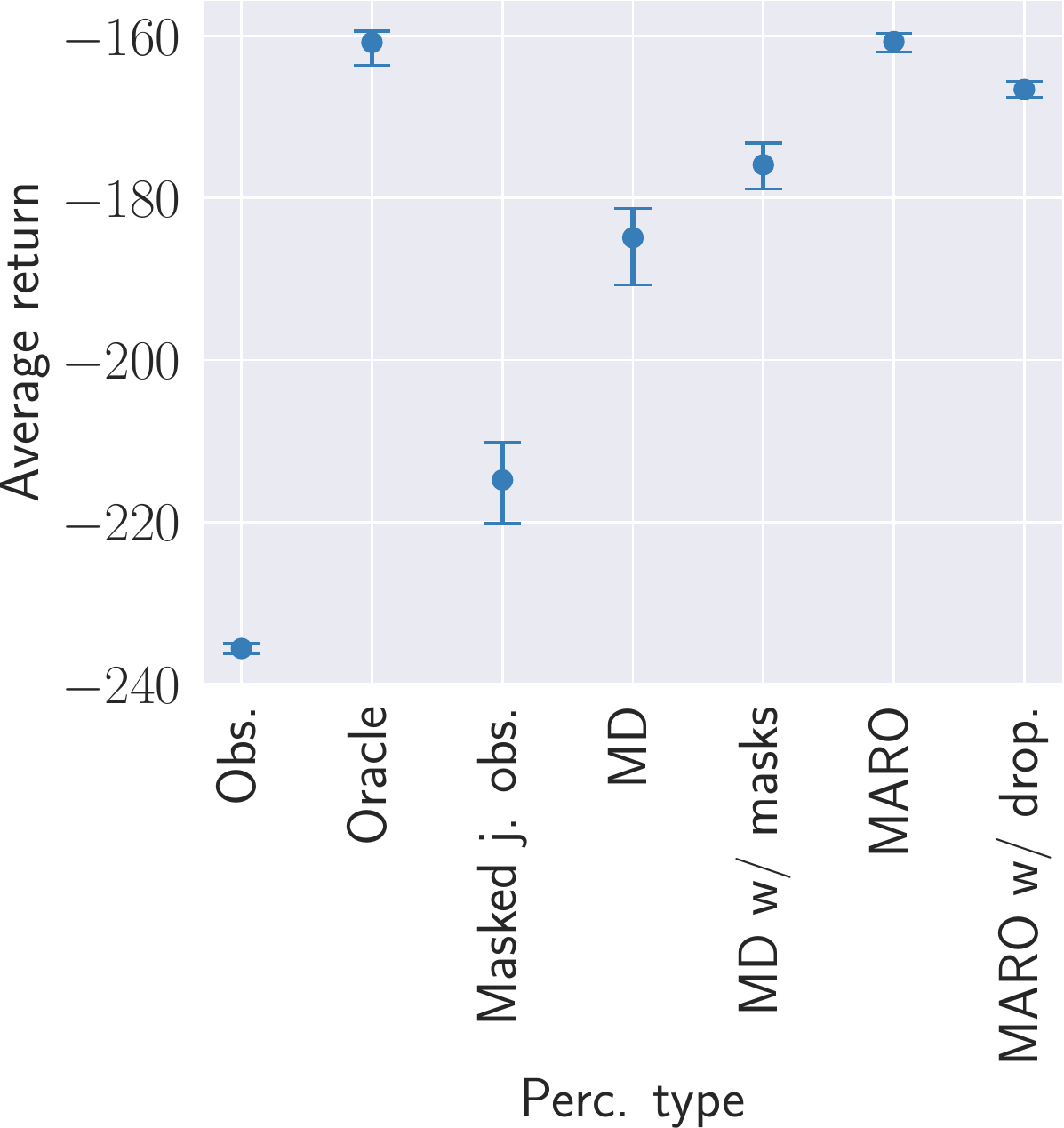}
        \caption{IPPO.}
    \end{subfigure}
    \begin{subfigure}[b]{0.24\textwidth}
        \centering
        \includegraphics[width=0.97\linewidth]{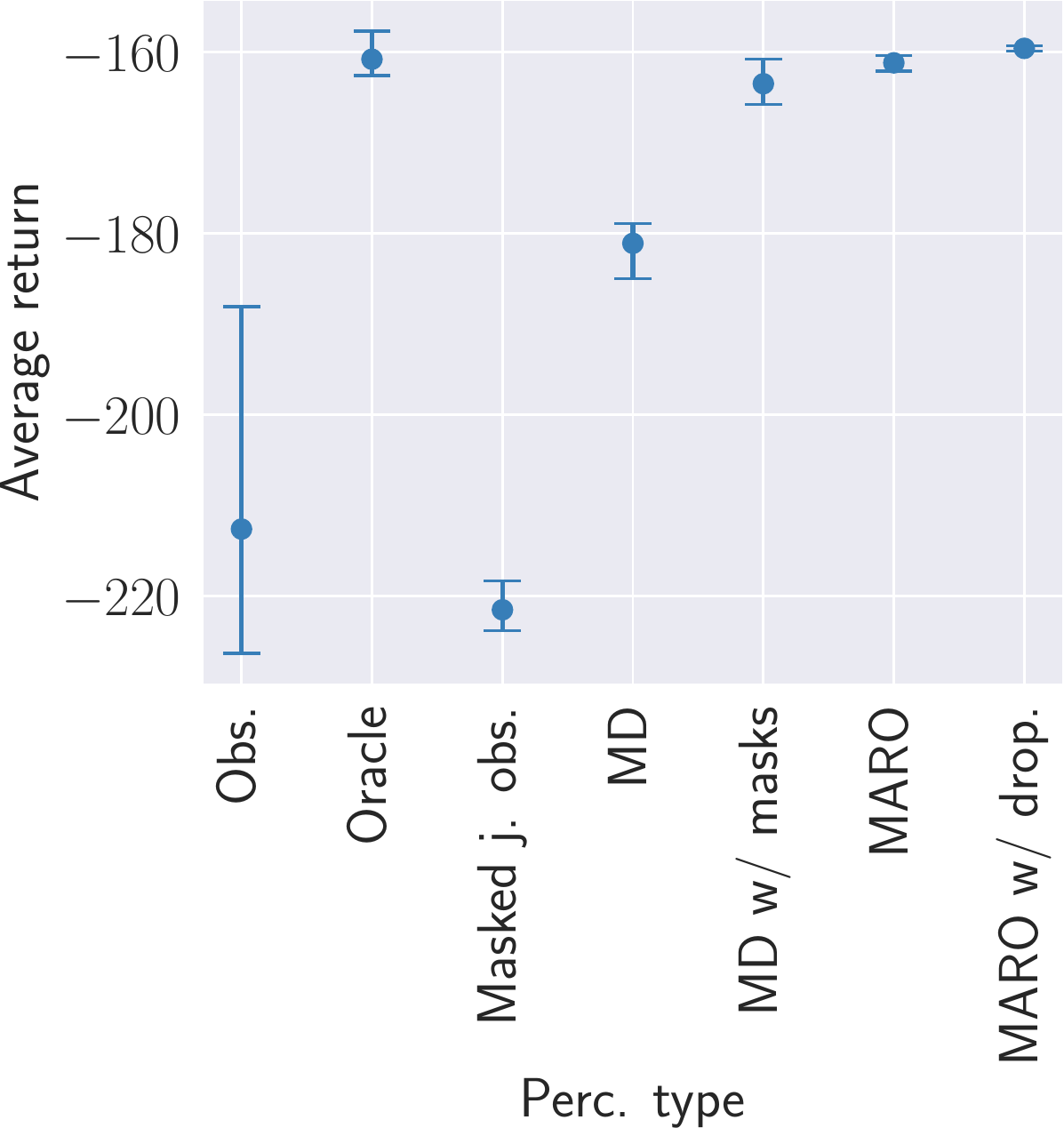}
        \caption{MAPPO.}
    \end{subfigure}
    \caption{(SpreadXY-2) Mean episodic returns for $p_\textrm{default}$ at execution time.}
\end{figure}

\begin{figure}
    \centering
    \begin{subfigure}[b]{0.24\textwidth}
        \centering
        \includegraphics[width=0.97\linewidth]{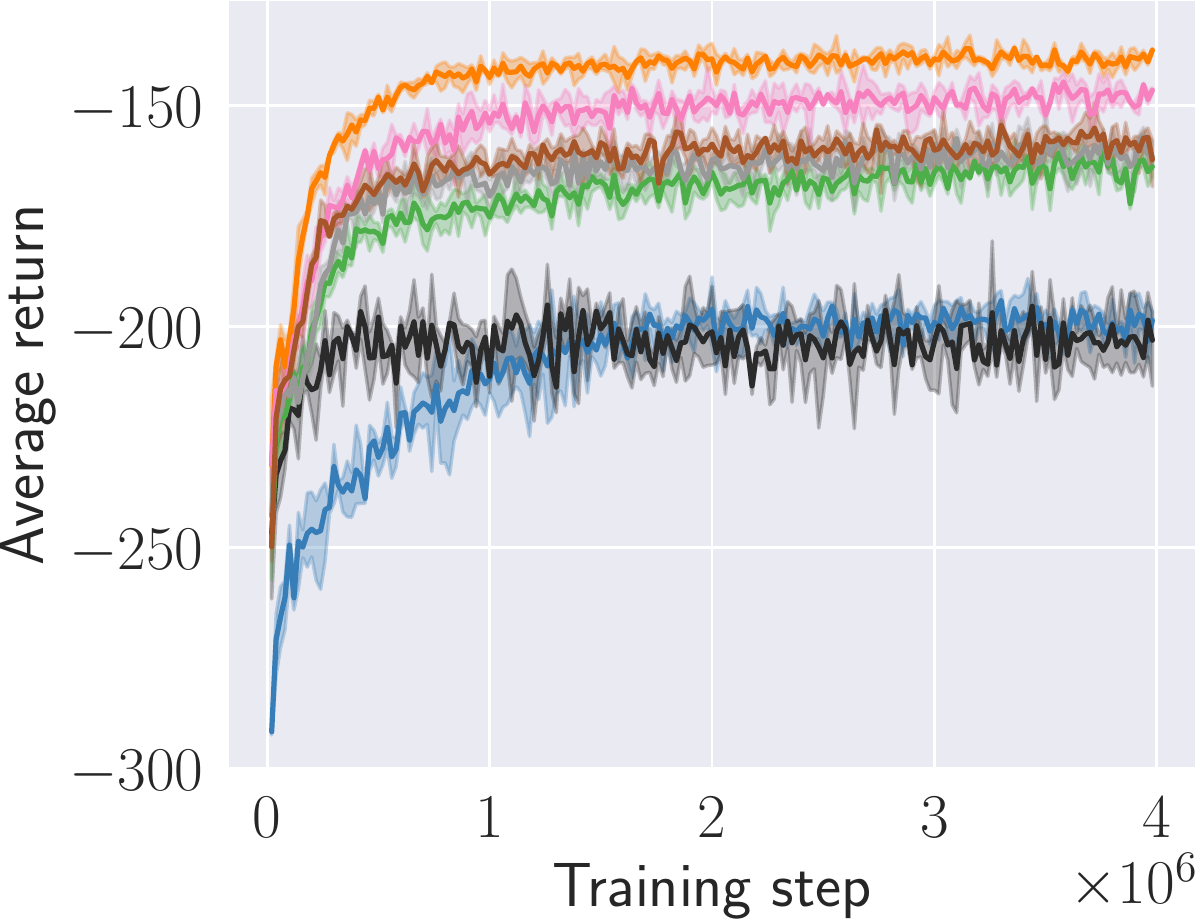}
        \caption{IQL.}
    \end{subfigure}
    \begin{subfigure}[b]{0.24\textwidth}
        \centering
        \includegraphics[width=0.97\linewidth]{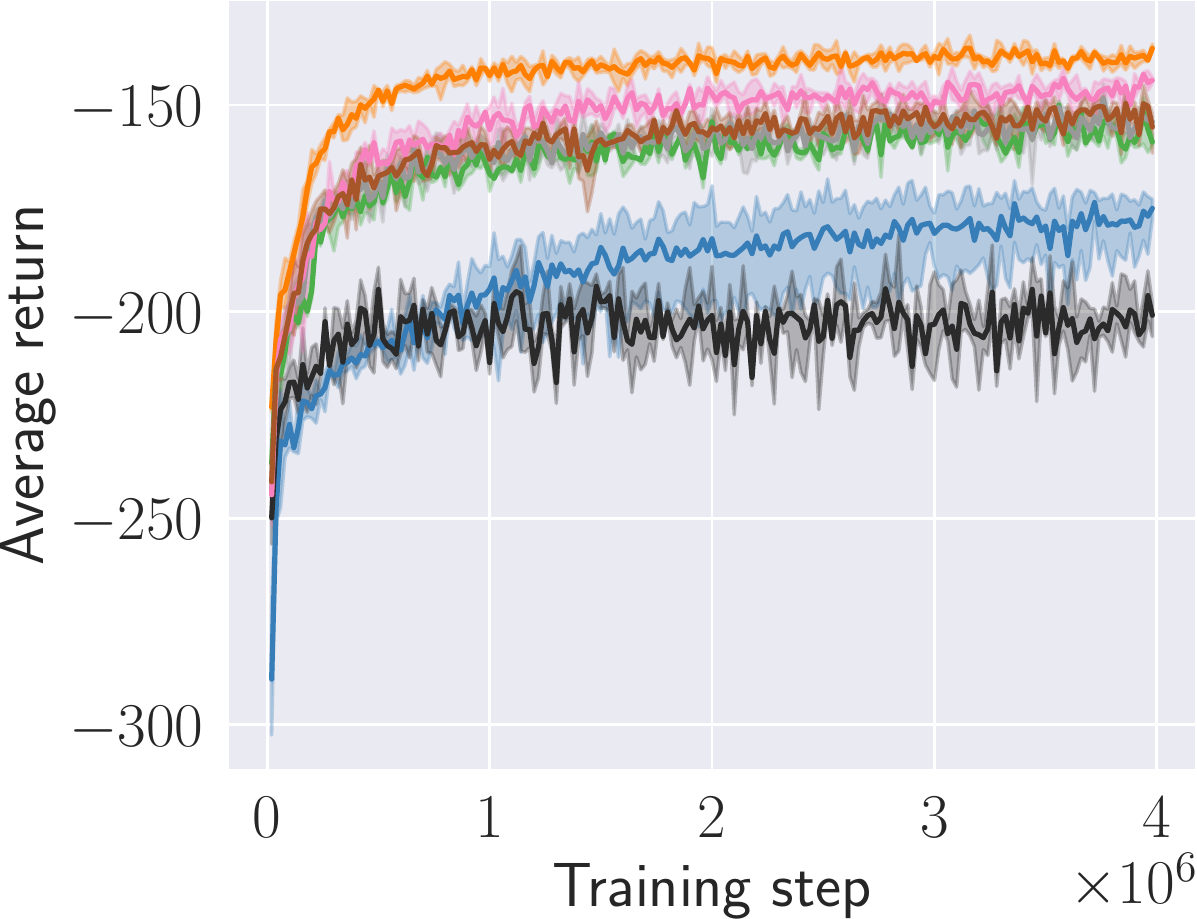}
        \caption{QMIX.}
    \end{subfigure}
    \begin{subfigure}[b]{0.24\textwidth}
        \centering
        \includegraphics[width=0.97\linewidth]{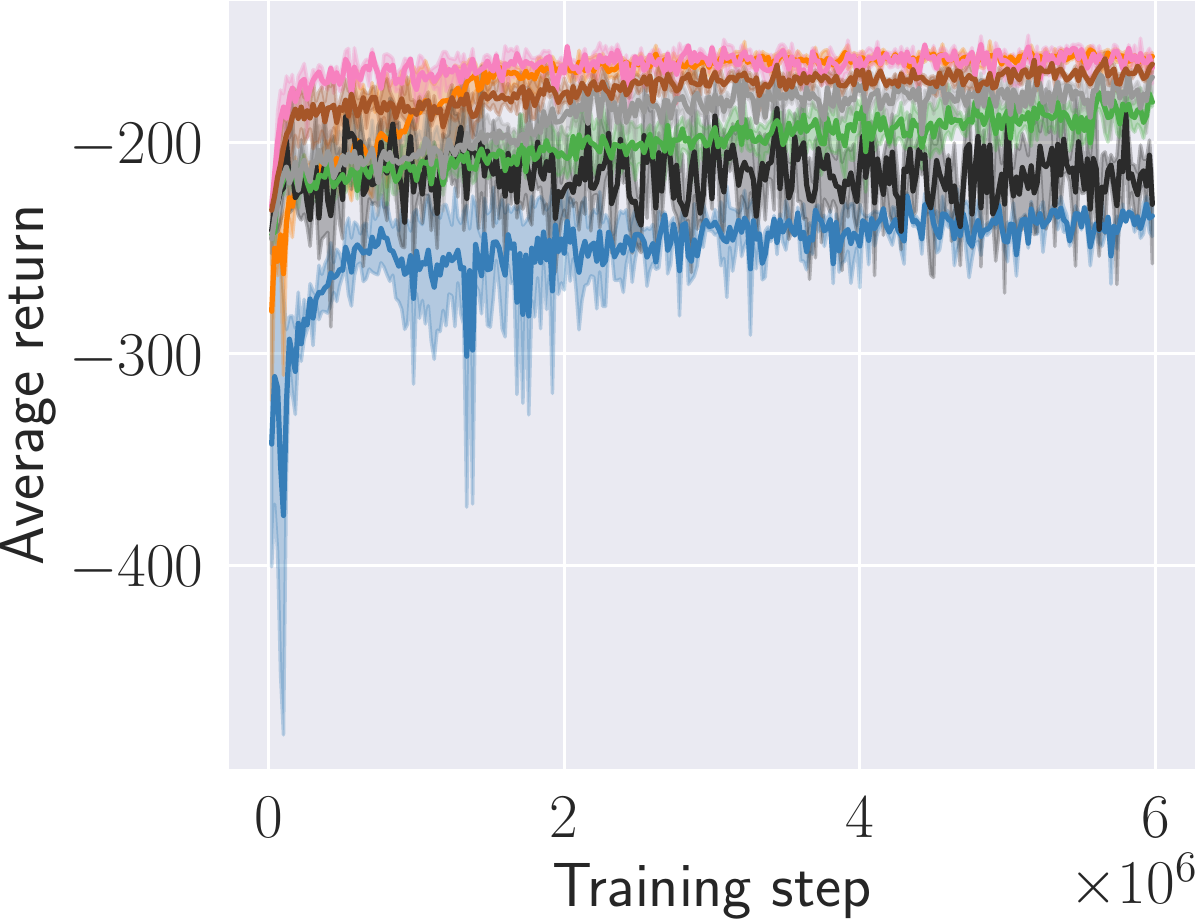}
        \caption{IPPO.}
    \end{subfigure}
    \begin{subfigure}[b]{0.24\textwidth}
        \centering
        \includegraphics[width=0.97\linewidth]{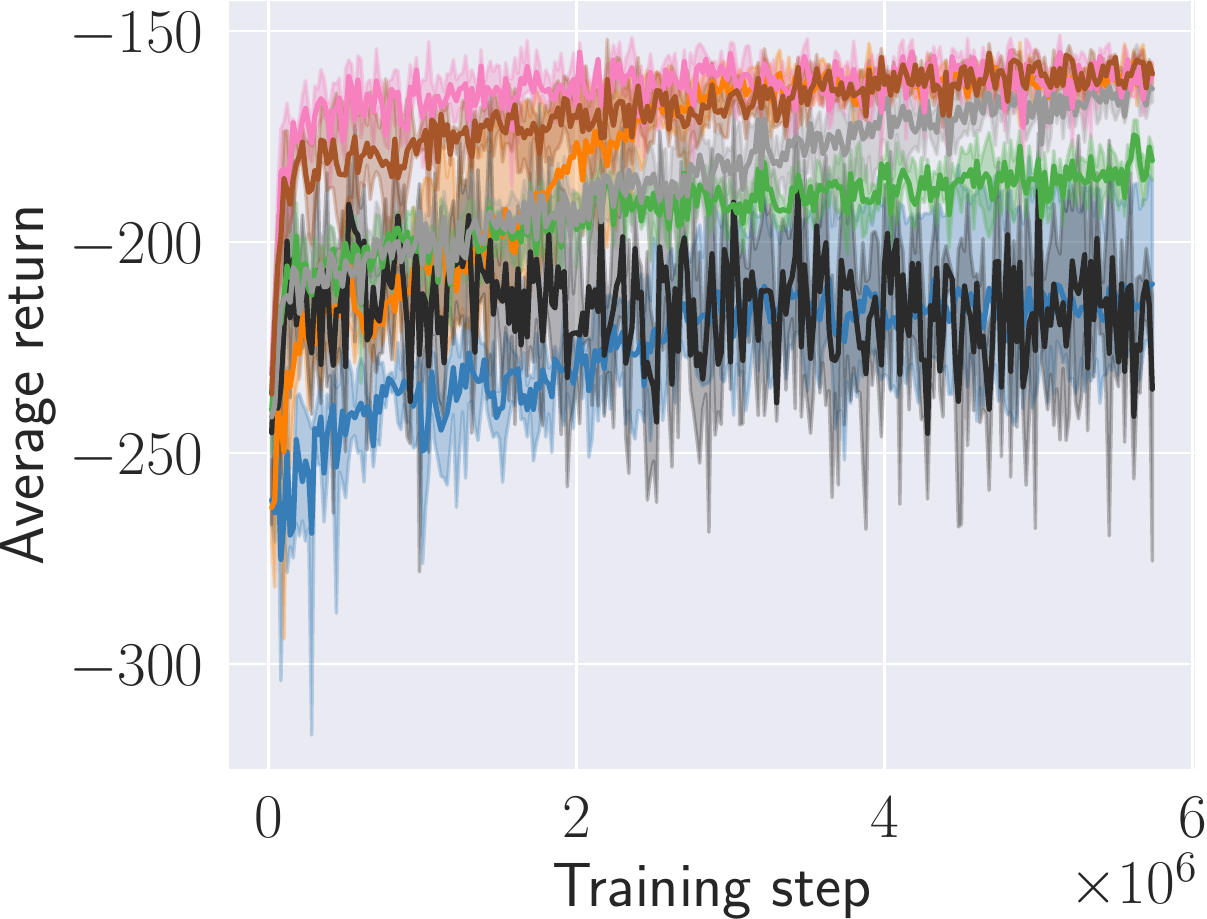}
        \caption{MAPPO.}
    \end{subfigure}
    \caption{(SpreadXY-2) Mean episodic returns for $p_\textrm{default}$ during training.}
\end{figure}

\begin{figure}
    \centering
    \begin{subfigure}[b]{0.24\textwidth}
        \centering
        \includegraphics[width=0.97\linewidth]{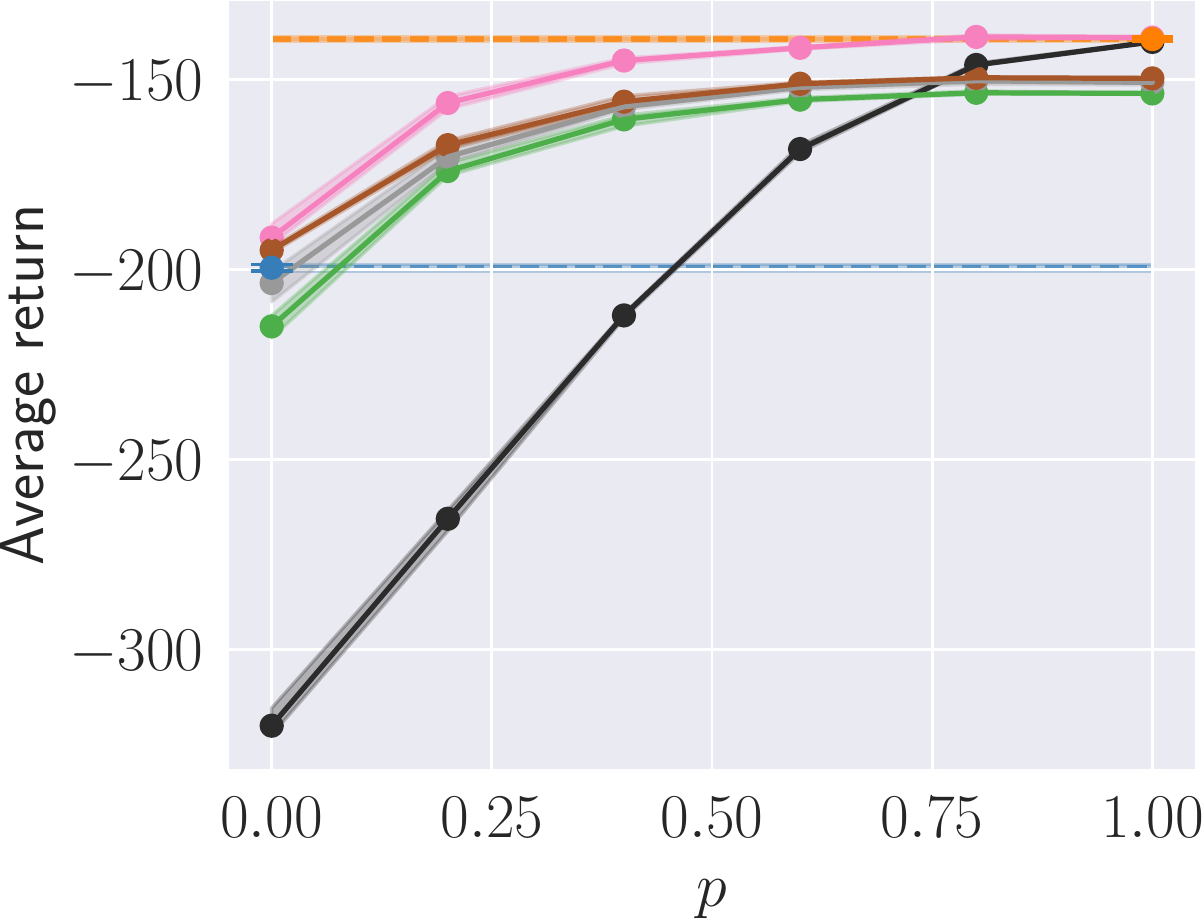}
        \caption{IQL.}
    \end{subfigure}
    \begin{subfigure}[b]{0.24\textwidth}
        \centering
        \includegraphics[width=0.97\linewidth]{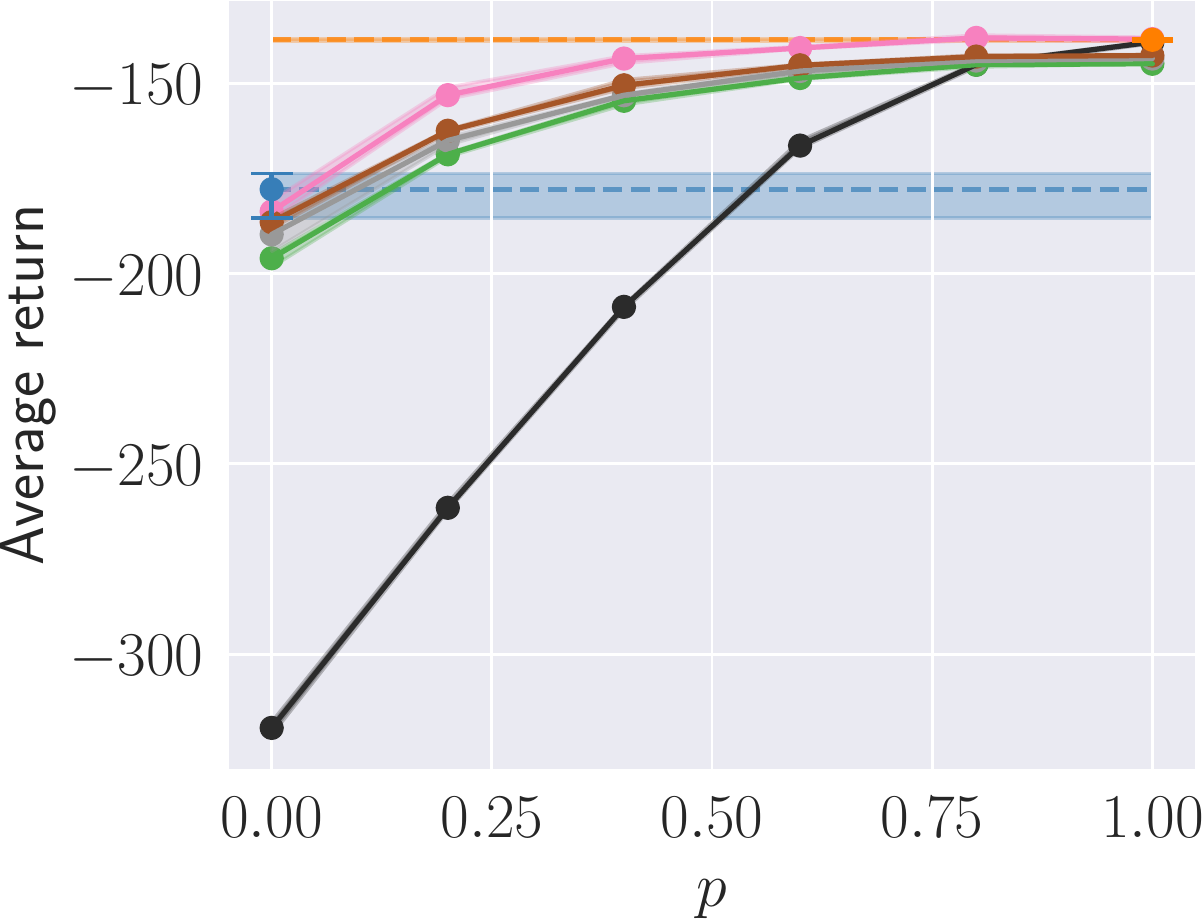}
        \caption{QMIX.}
    \end{subfigure}
    \begin{subfigure}[b]{0.24\textwidth}
        \centering
        \includegraphics[width=0.97\linewidth]{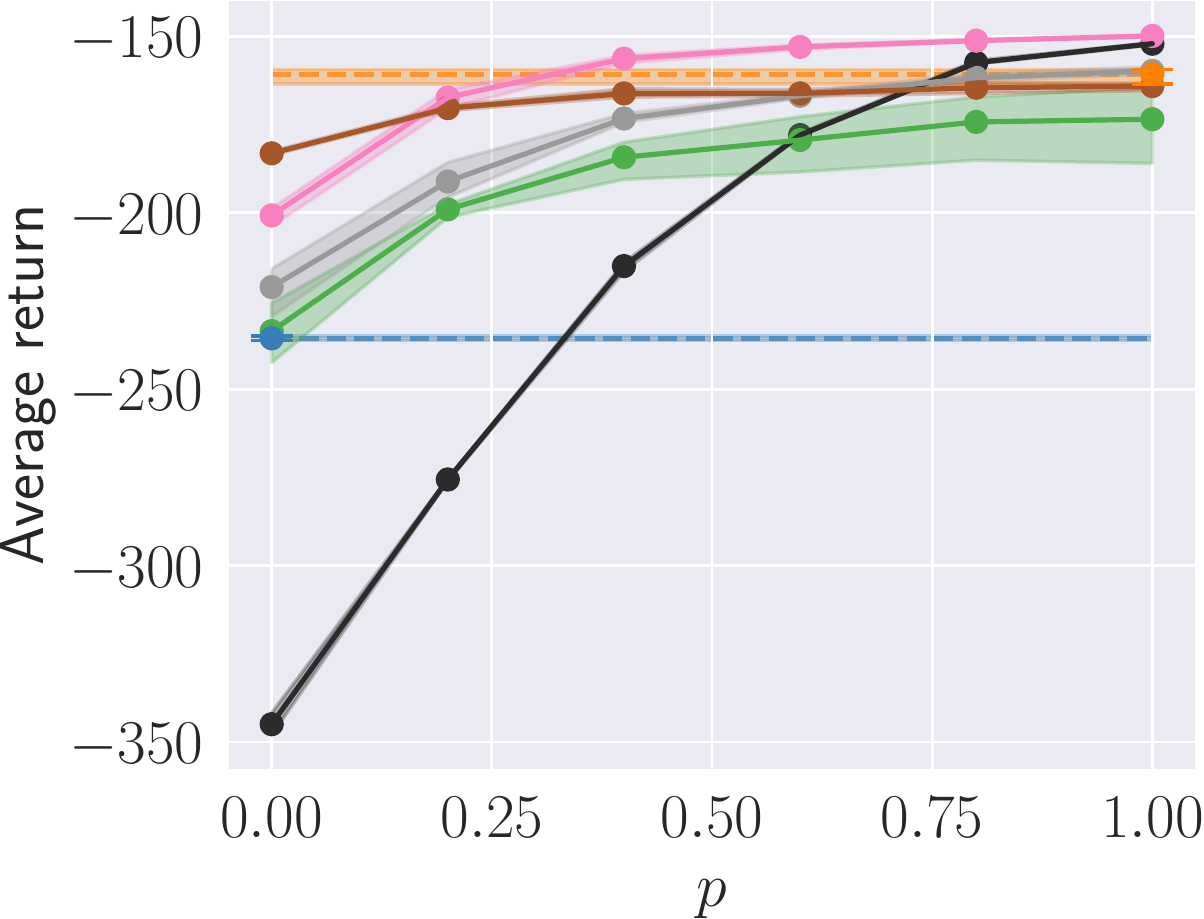}
        \caption{IPPO.}
    \end{subfigure}
    \begin{subfigure}[b]{0.24\textwidth}
        \centering
        \includegraphics[width=0.97\linewidth]{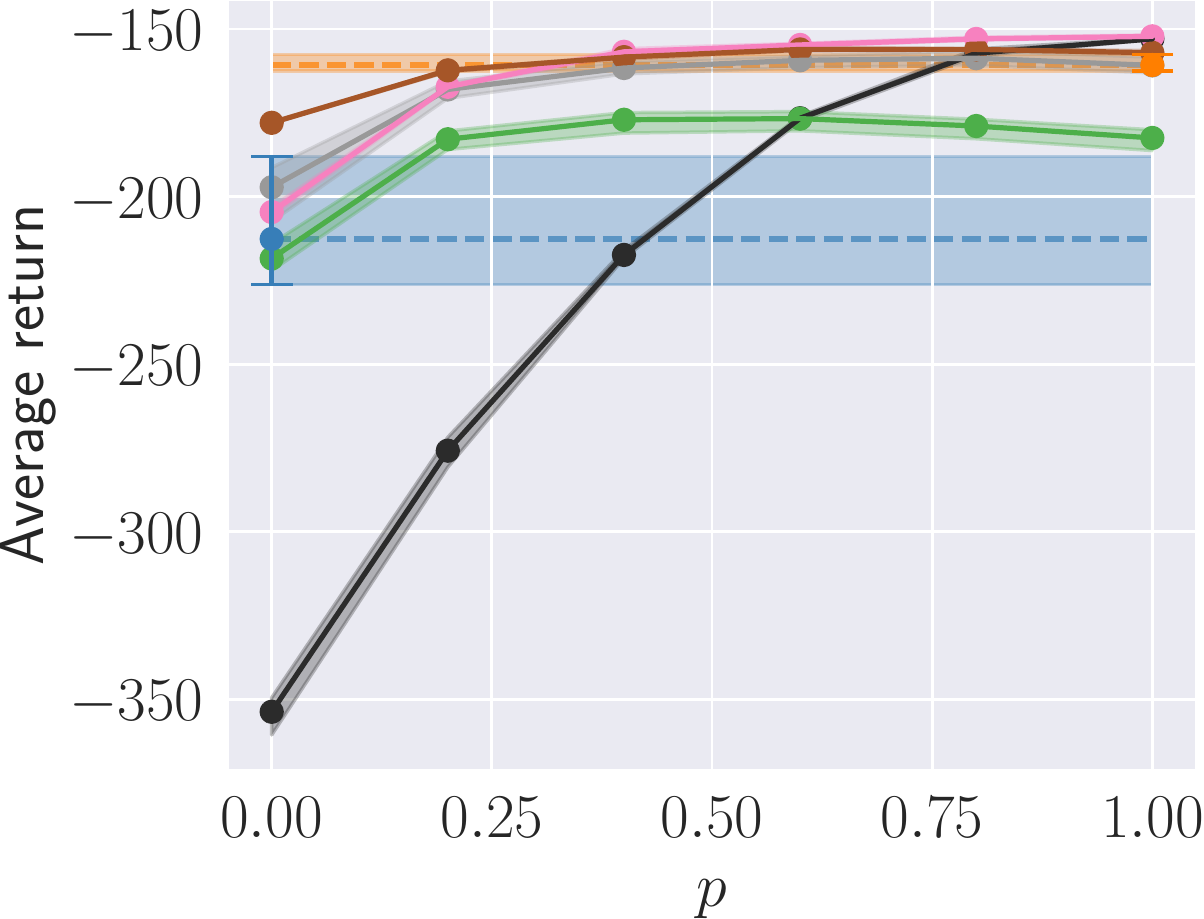}
        \caption{MAPPO.}
    \end{subfigure}
    \caption{(SpreadXY-2) Mean episodic returns for different $p$ values at execution time.}
\end{figure}

\begin{figure}
    \centering
    \includegraphics[height=0.7cm]{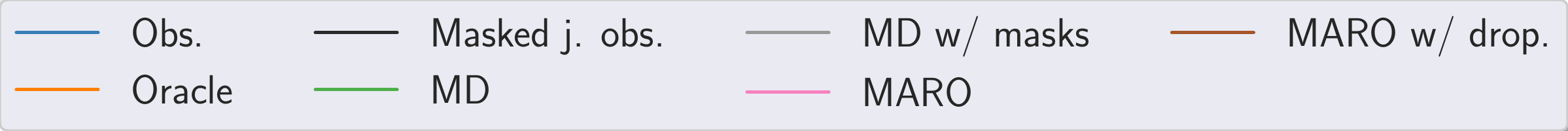}
    \caption{Legend.}
\end{figure}

\clearpage

\begin{table}
\centering
\noindent
\caption{(SpreadXY-4) Mean episodic returns for $p_{\textrm{default}}$ at execution time.}
\vspace{0.1cm}
\resizebox{\linewidth}{!}{%
\begin{tabular}{c c c c c c c c }\toprule
\multicolumn{1}{c }{\textbf{}} & \multicolumn{7}{c }{\textbf{SpreadXY-4 ($p_\textrm{default}$)}} \\  
\cmidrule(lr){2-8}
\multicolumn{1}{ l }{\textbf{Algorithm}} & \textbf{Obs.} & \textbf{Oracle} & \textbf{Masked j. obs.} & \textbf{MD} & \textbf{MD w/ masks} & \textbf{MARO} & \textbf{MARO w/ drop.} \\
\cmidrule{1-8}
\multicolumn{1}{ l }{IQL} & -1225.5 \tiny{(-4.2,+4.9)} & -902.3 \tiny{(-58.3,+58.3)} & -1161.2 \tiny{(-7.6,+9.4)} & -1157.0 \tiny{(-1.2,+1.0)} & -1164.5 \tiny{(-10.8,+13.3)} & -988.3 \tiny{(-21.7,+37.4)} & -1050.1 \tiny{(-35.7,+54.1)} \\ \cmidrule{1-8}
\multicolumn{1}{ l }{QMIX} & -1132.6 \tiny{(-6.6,+5.9)} & -796.9 \tiny{(-9.0,+12.7)} & -1146.4 \tiny{(-12.7,+22.1)} & -1024.6 \tiny{(-39.5,+54.9)} & -1014.0 \tiny{(-40.4,+40.6)} & -850.0 \tiny{(-22.5,+17.0)} & -945.6 \tiny{(-47.3,+65.9)} \\ \cmidrule{1-8}
\multicolumn{1}{ l }{IPPO} & -1133.2 \tiny{(-7.1,+8.4)} & -781.6 \tiny{(-18.0,+10.5)} & -1162.1 \tiny{(-33.8,+33.8)} & -1124.7 \tiny{(-27.6,+16.9)} & -1177.9 \tiny{(-29.4,+38.5)} & -920.6 \tiny{(-50.1,+92.7)} & -874.6 \tiny{(-7.7,+6.5)} \\ \cmidrule{1-8}
\multicolumn{1}{ l }{MAPPO} & -1116.9 \tiny{(-43.2,+79.0)} & -832.8 \tiny{(-123.3,+65.9)} & -1196.5 \tiny{(-29.8,+20.3)} & -1112.1 \tiny{(-28.9,+26.4)} & -1149.7 \tiny{(-12.5,+16.3)} & -827.4 \tiny{(-7.8,+5.8)} & -820.6 \tiny{(-3.6,+3.6)} \\
\bottomrule
\end{tabular}
}
\end{table}

\begin{figure}
    \centering
    \begin{subfigure}[b]{0.24\textwidth}
        \centering
        \includegraphics[width=0.97\linewidth]{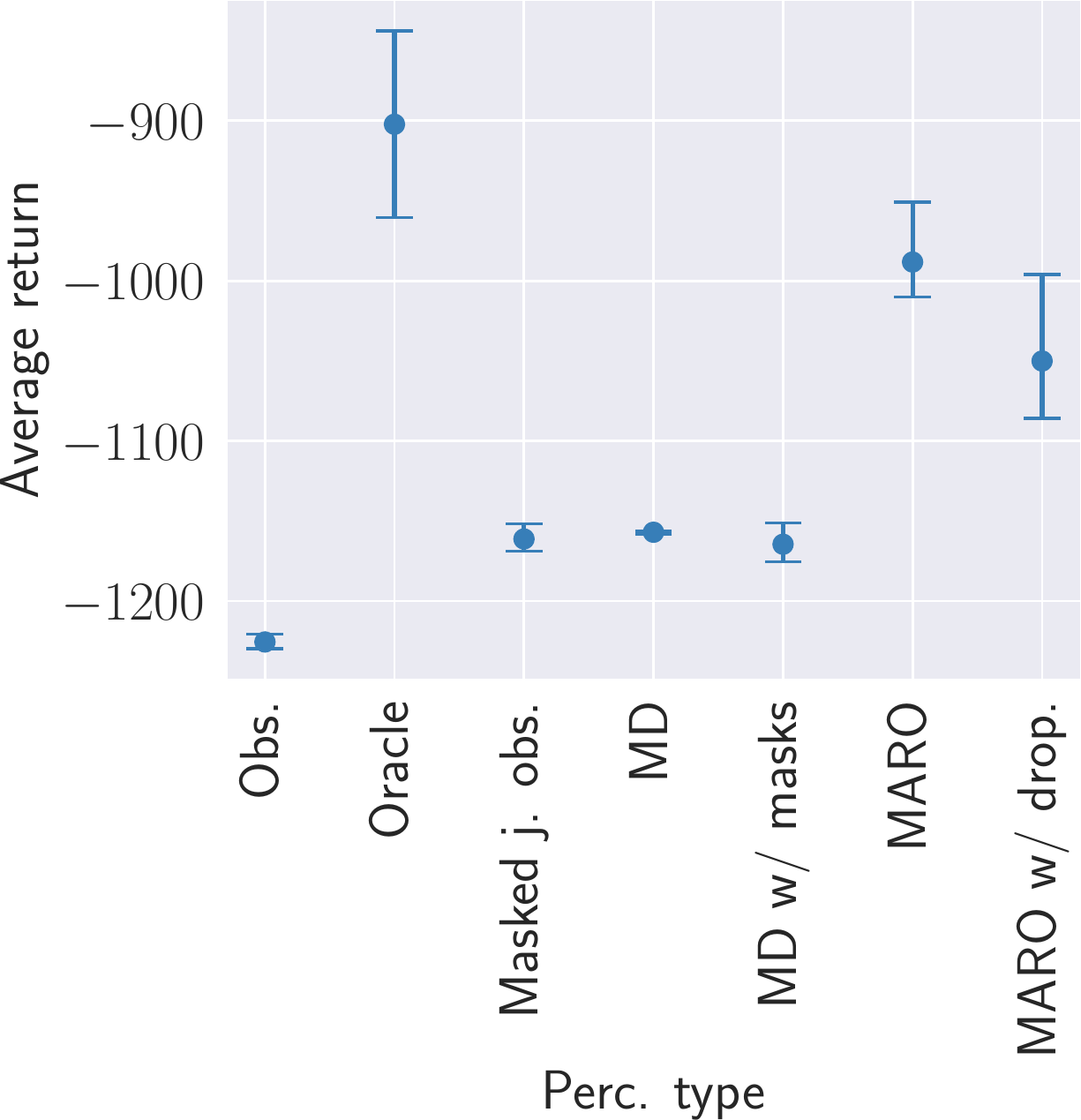}
        \caption{IQL.}
    \end{subfigure}
    \begin{subfigure}[b]{0.24\textwidth}
        \centering
        \includegraphics[width=0.97\linewidth]{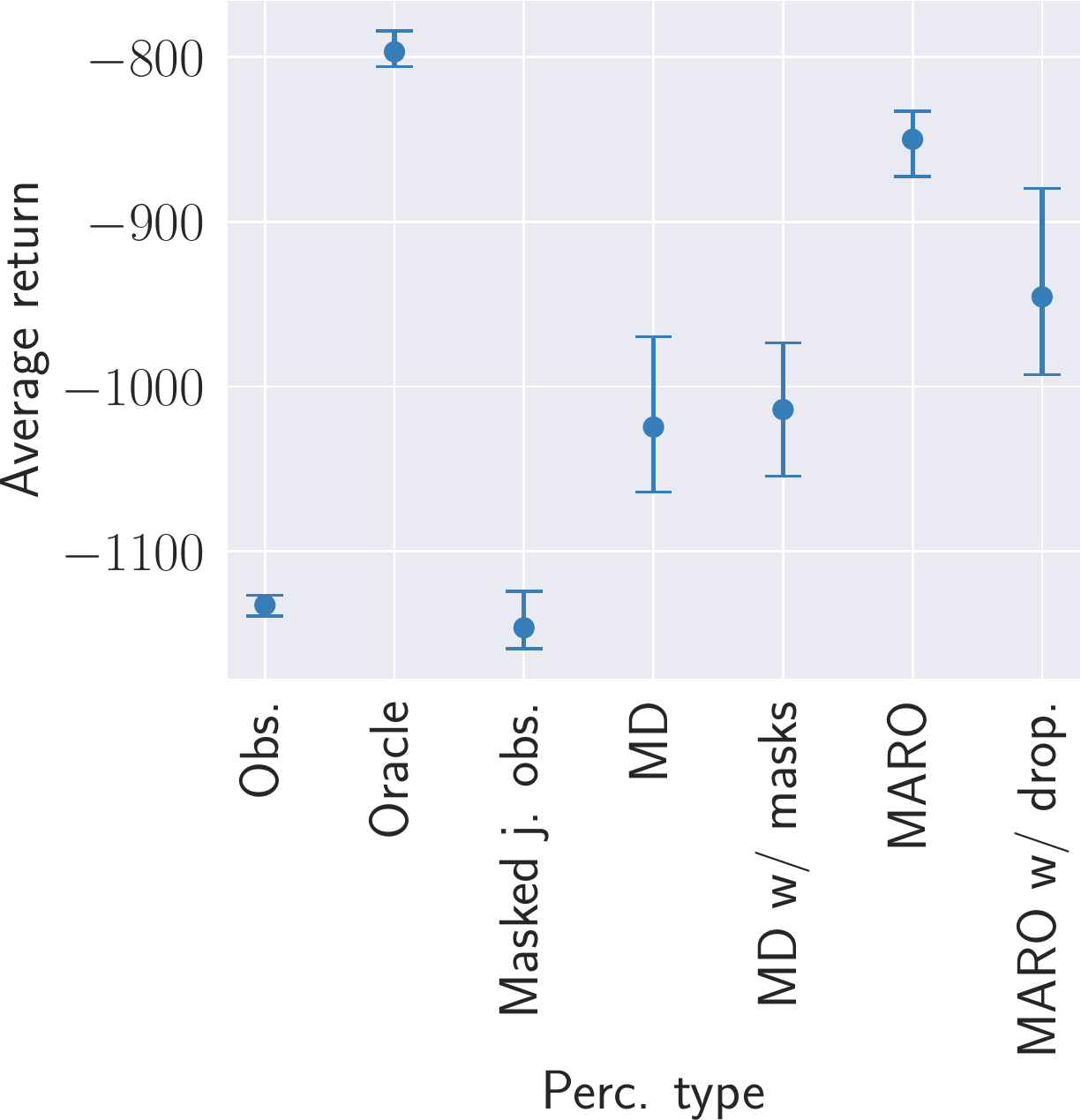}
        \caption{QMIX.}
    \end{subfigure}
    \begin{subfigure}[b]{0.24\textwidth}
        \centering
        \includegraphics[width=0.97\linewidth]{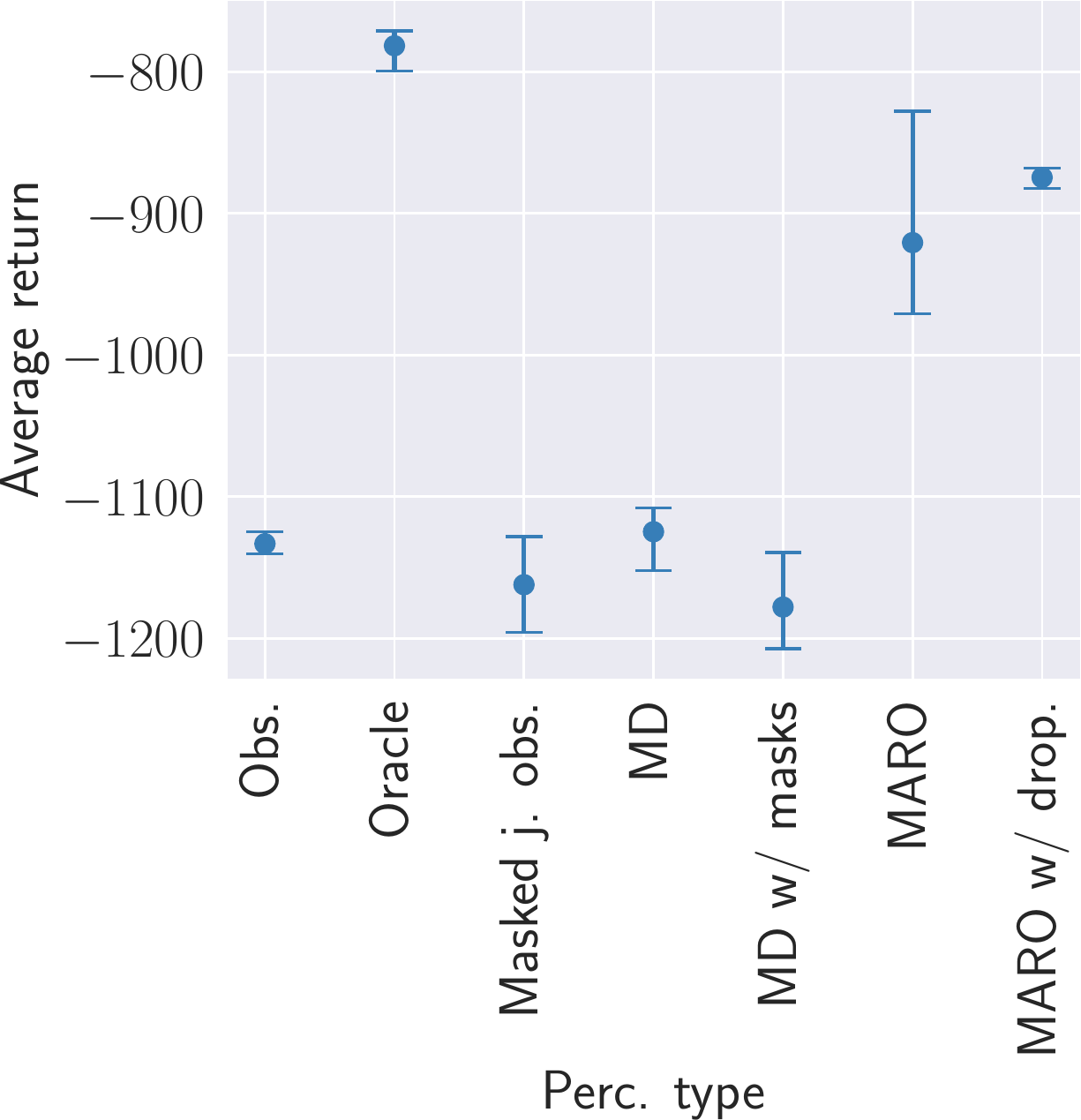}
        \caption{IPPO.}
    \end{subfigure}
    \begin{subfigure}[b]{0.24\textwidth}
        \centering
        \includegraphics[width=0.97\linewidth]{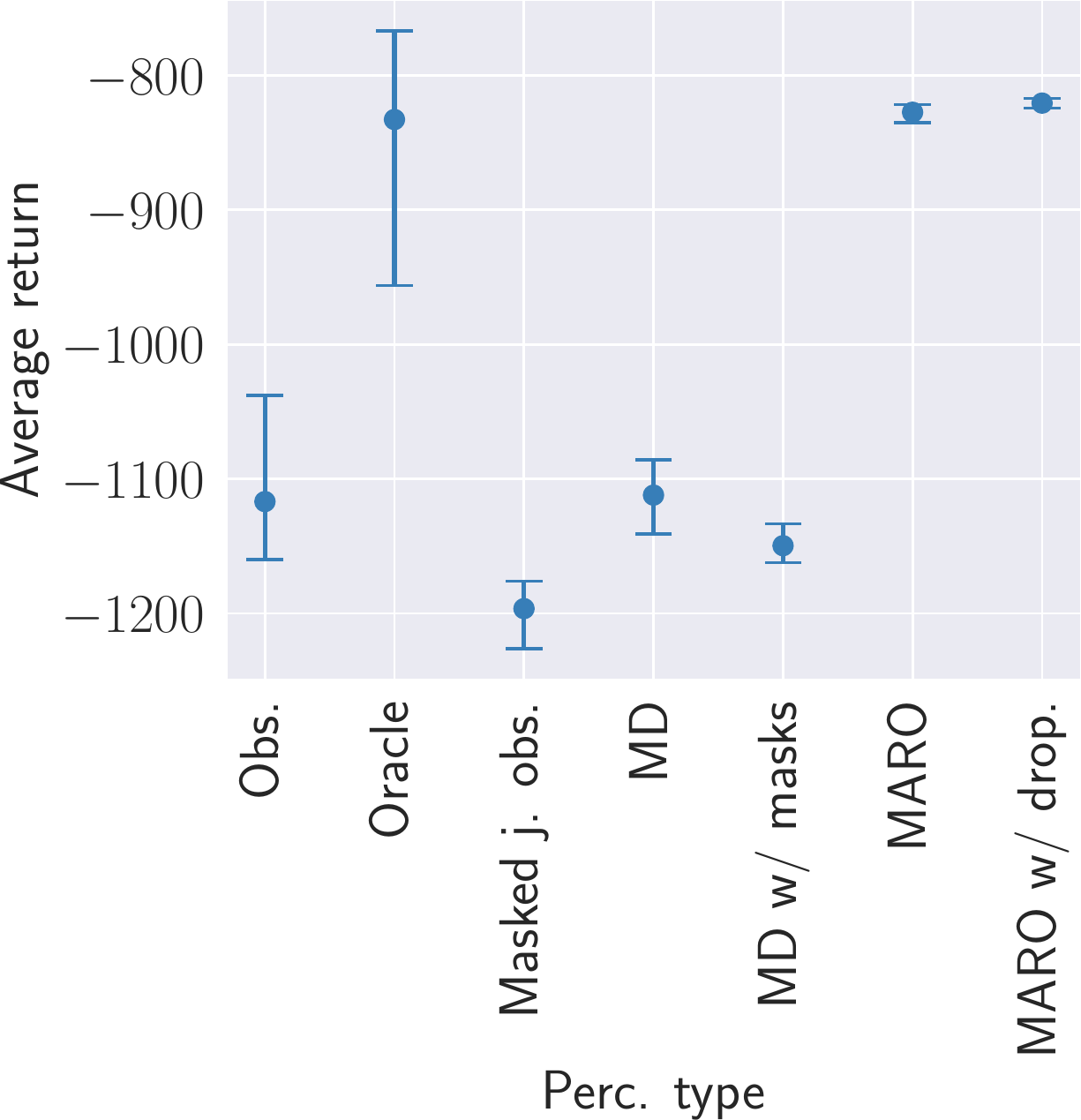}
        \caption{MAPPO.}
    \end{subfigure}
    \caption{(SpreadXY-4) Mean episodic returns for $p_\textrm{default}$ at execution time.}
\end{figure}

\begin{figure}
    \centering
    \begin{subfigure}[b]{0.24\textwidth}
        \centering
        \includegraphics[width=0.97\linewidth]{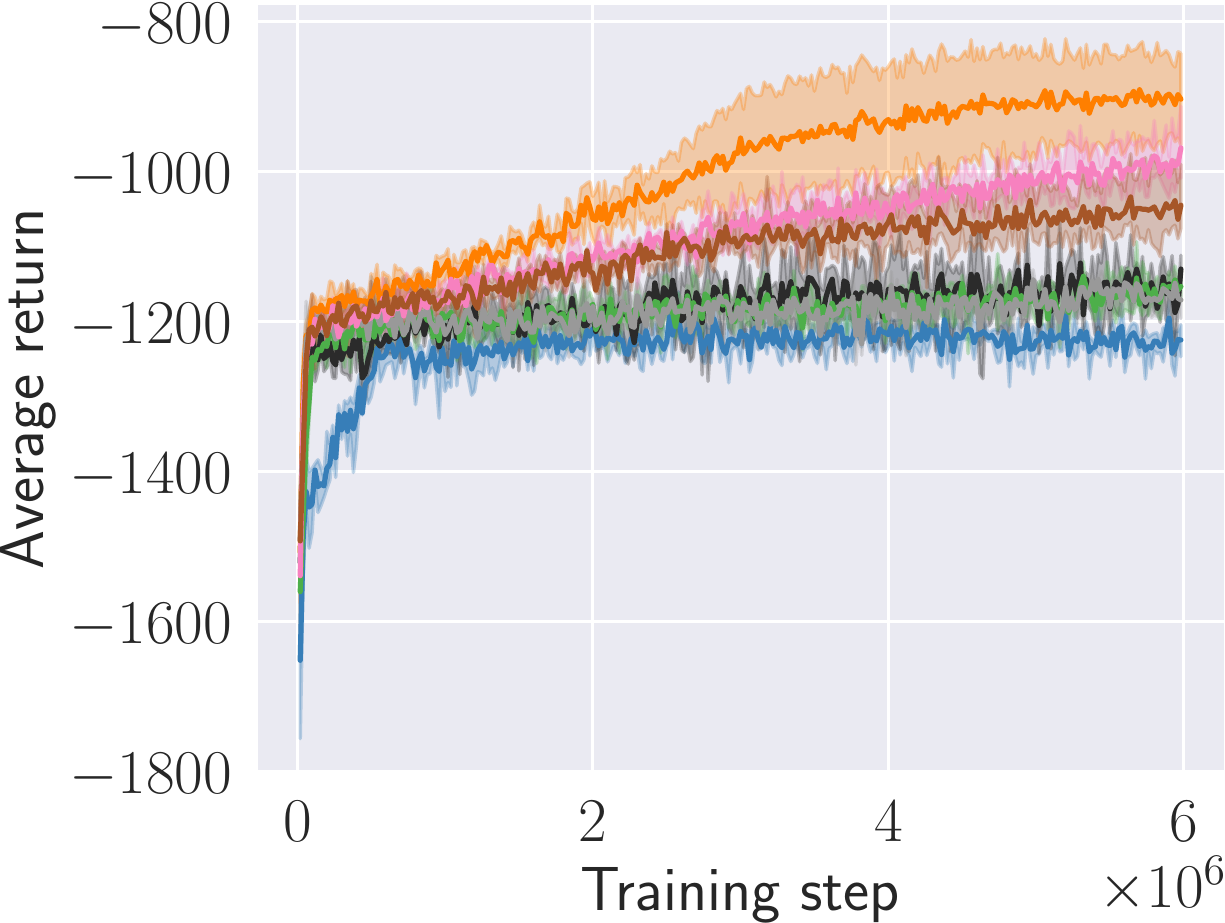}
        \caption{IQL.}
    \end{subfigure}
    \begin{subfigure}[b]{0.24\textwidth}
        \centering
        \includegraphics[width=0.97\linewidth]{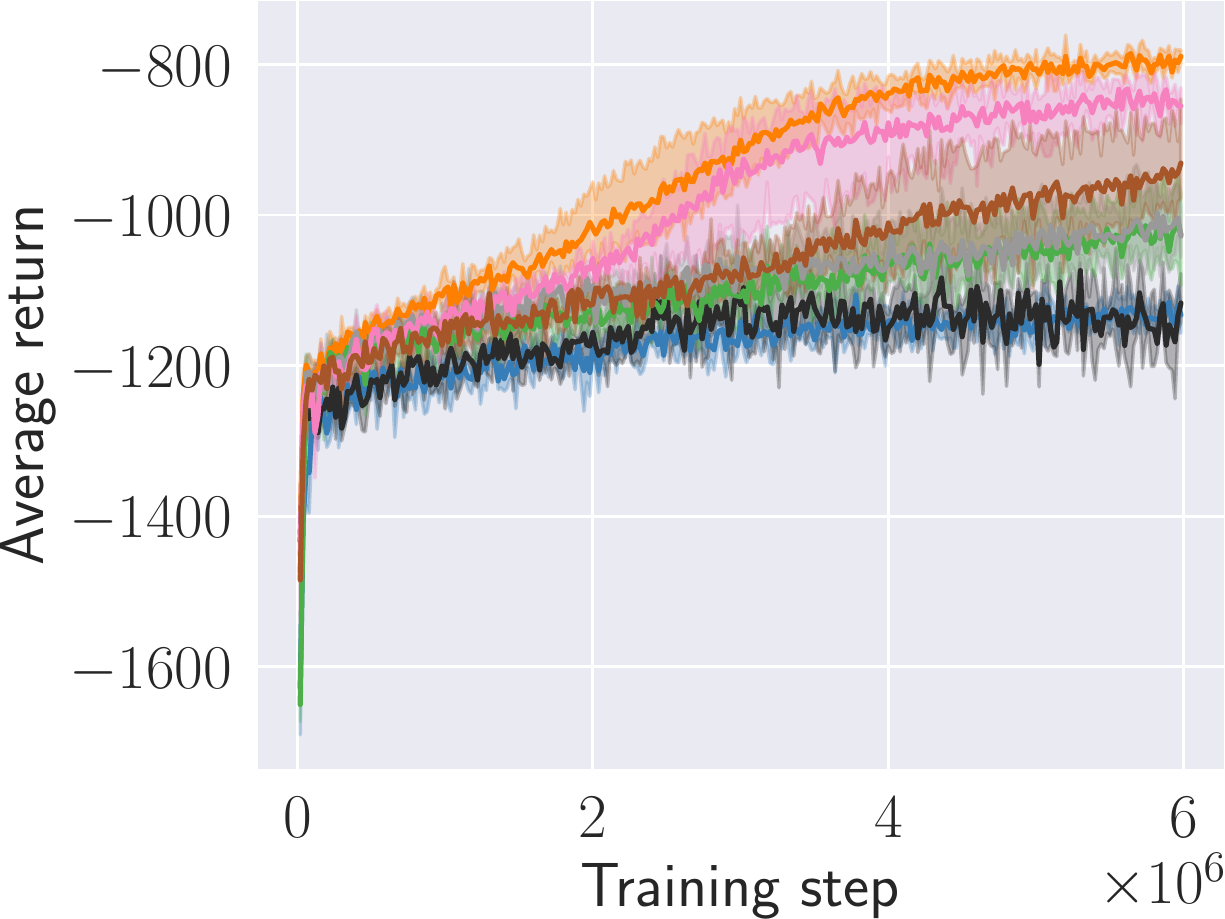}
        \caption{QMIX.}
    \end{subfigure}
    \begin{subfigure}[b]{0.24\textwidth}
        \centering
        \includegraphics[width=0.97\linewidth]{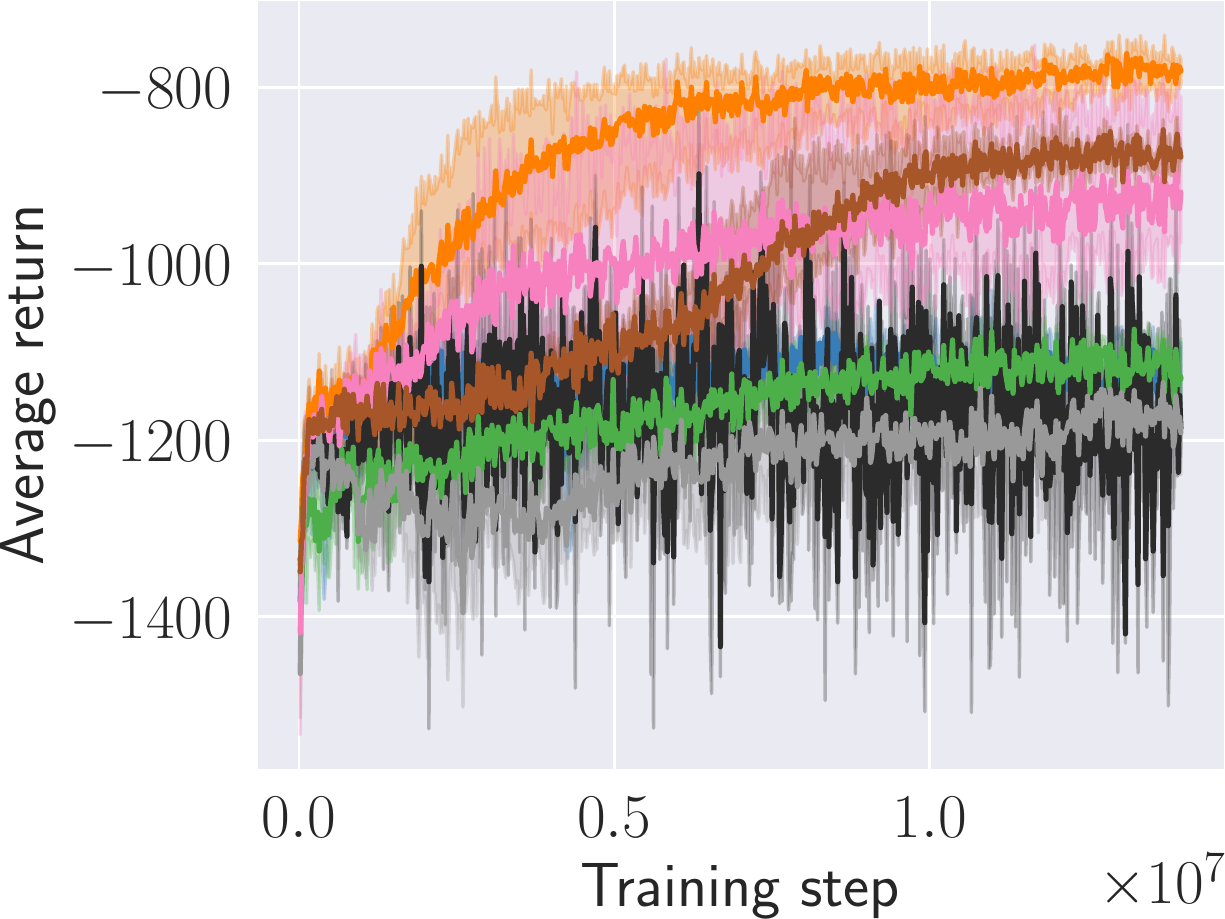}
        \caption{IPPO.}
    \end{subfigure}
    \begin{subfigure}[b]{0.24\textwidth}
        \centering
        \includegraphics[width=0.97\linewidth]{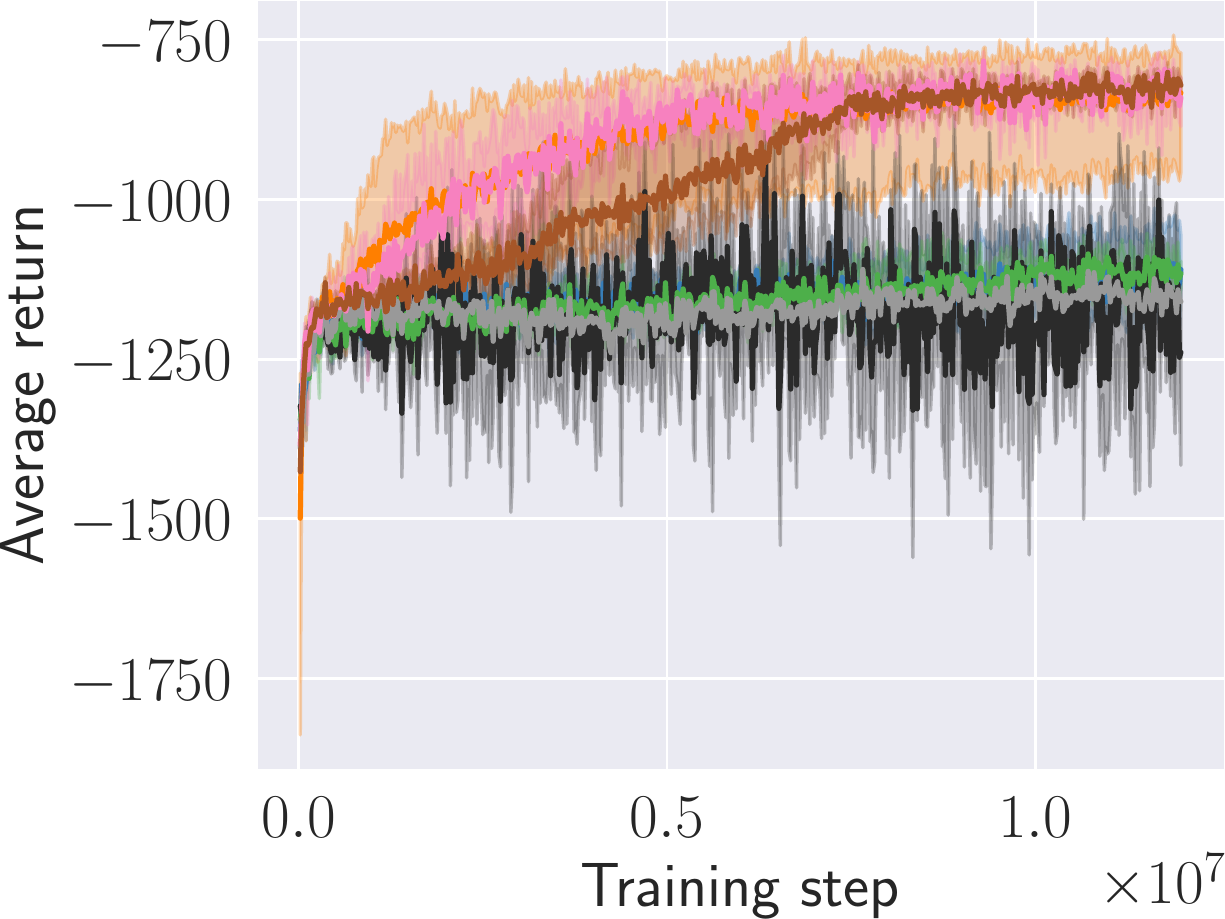}
        \caption{MAPPO.}
    \end{subfigure}
    \caption{(SpreadXY-4) Mean episodic returns for $p_\textrm{default}$ during training.}
\end{figure}

\begin{figure}
    \centering
    \begin{subfigure}[b]{0.24\textwidth}
        \centering
        \includegraphics[width=0.97\linewidth]{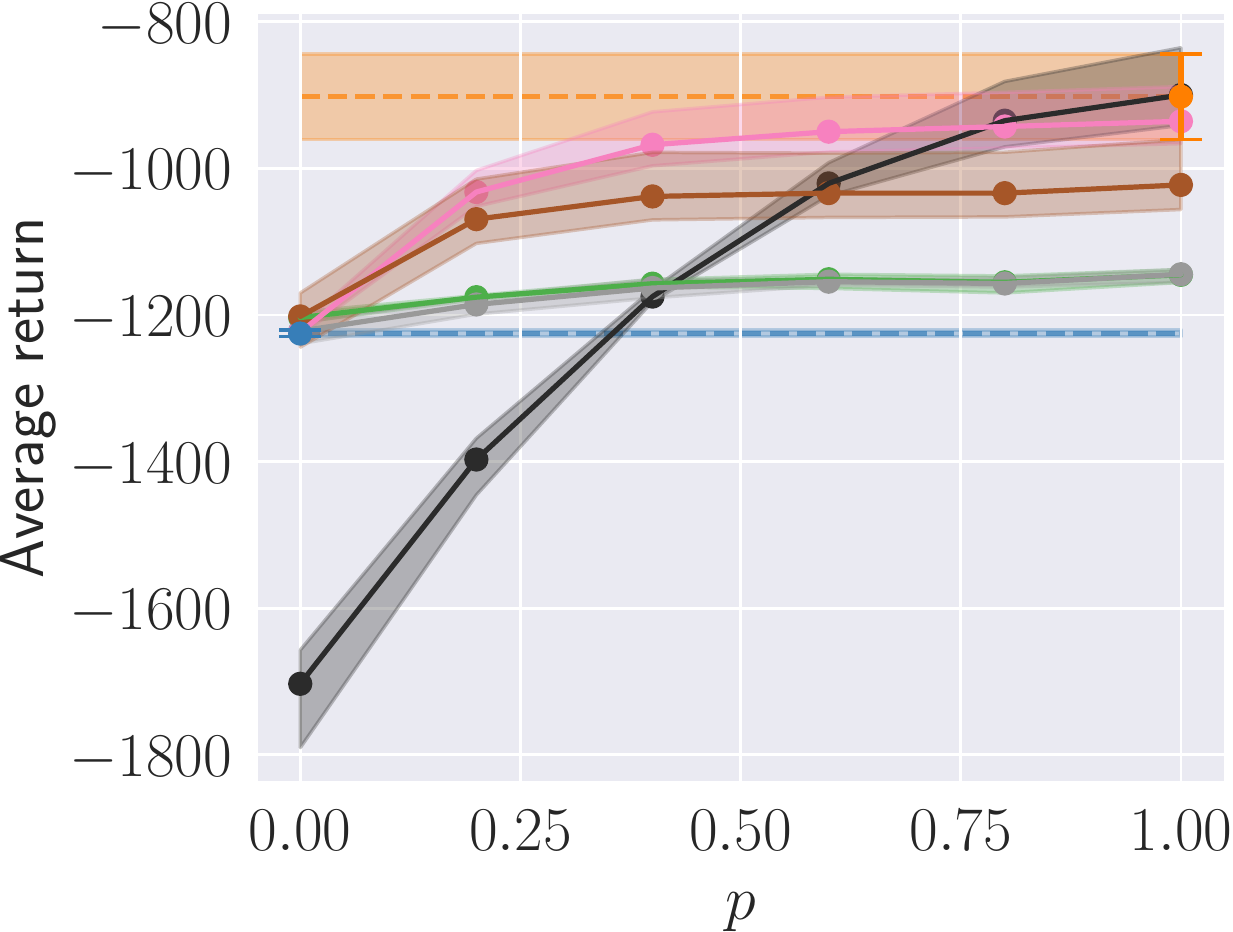}
        \caption{IQL.}
    \end{subfigure}
    \begin{subfigure}[b]{0.24\textwidth}
        \centering
        \includegraphics[width=0.97\linewidth]{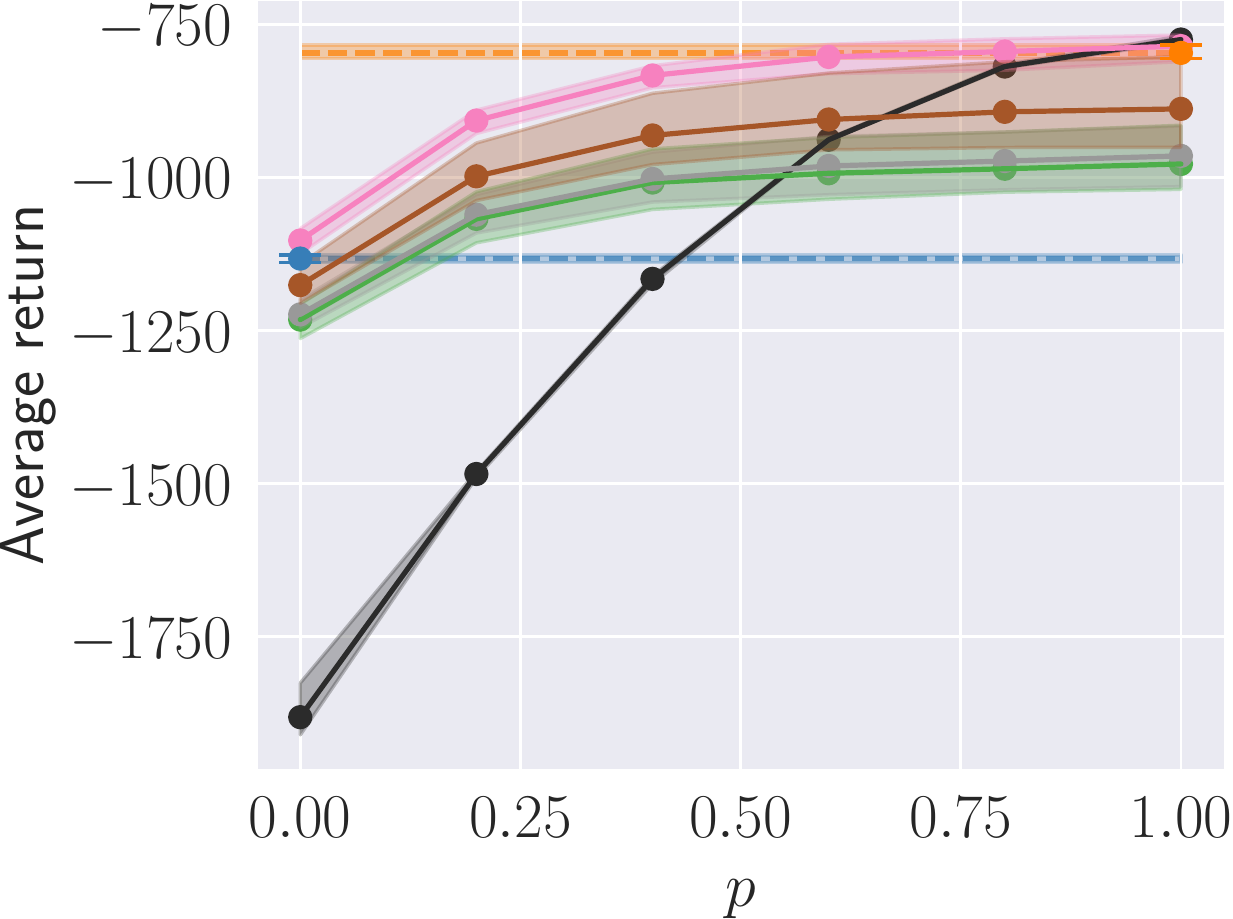}
        \caption{QMIX.}
    \end{subfigure}
    \begin{subfigure}[b]{0.24\textwidth}
        \centering
        \includegraphics[width=0.97\linewidth]{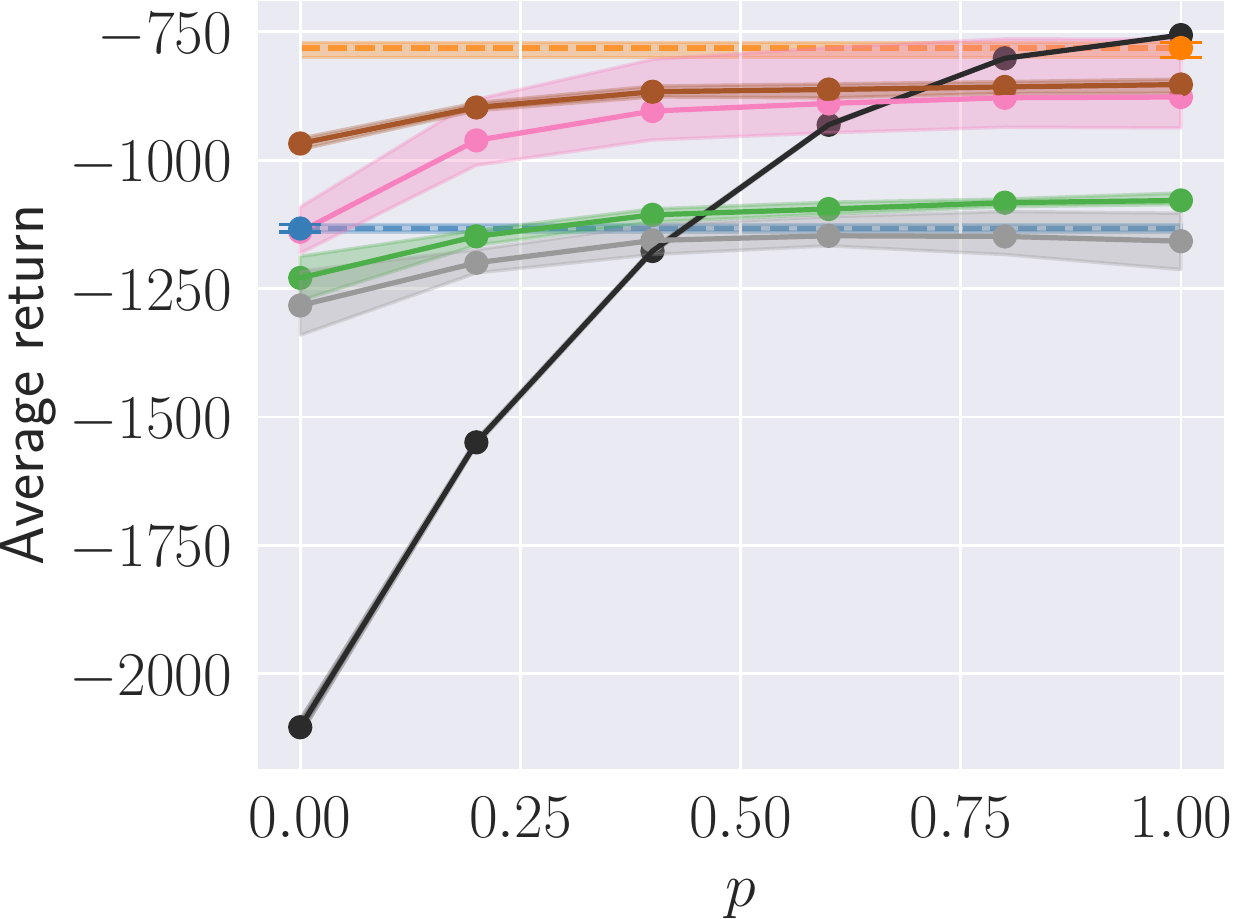}
        \caption{IPPO.}
    \end{subfigure}
    \begin{subfigure}[b]{0.24\textwidth}
        \centering
        \includegraphics[width=0.97\linewidth]{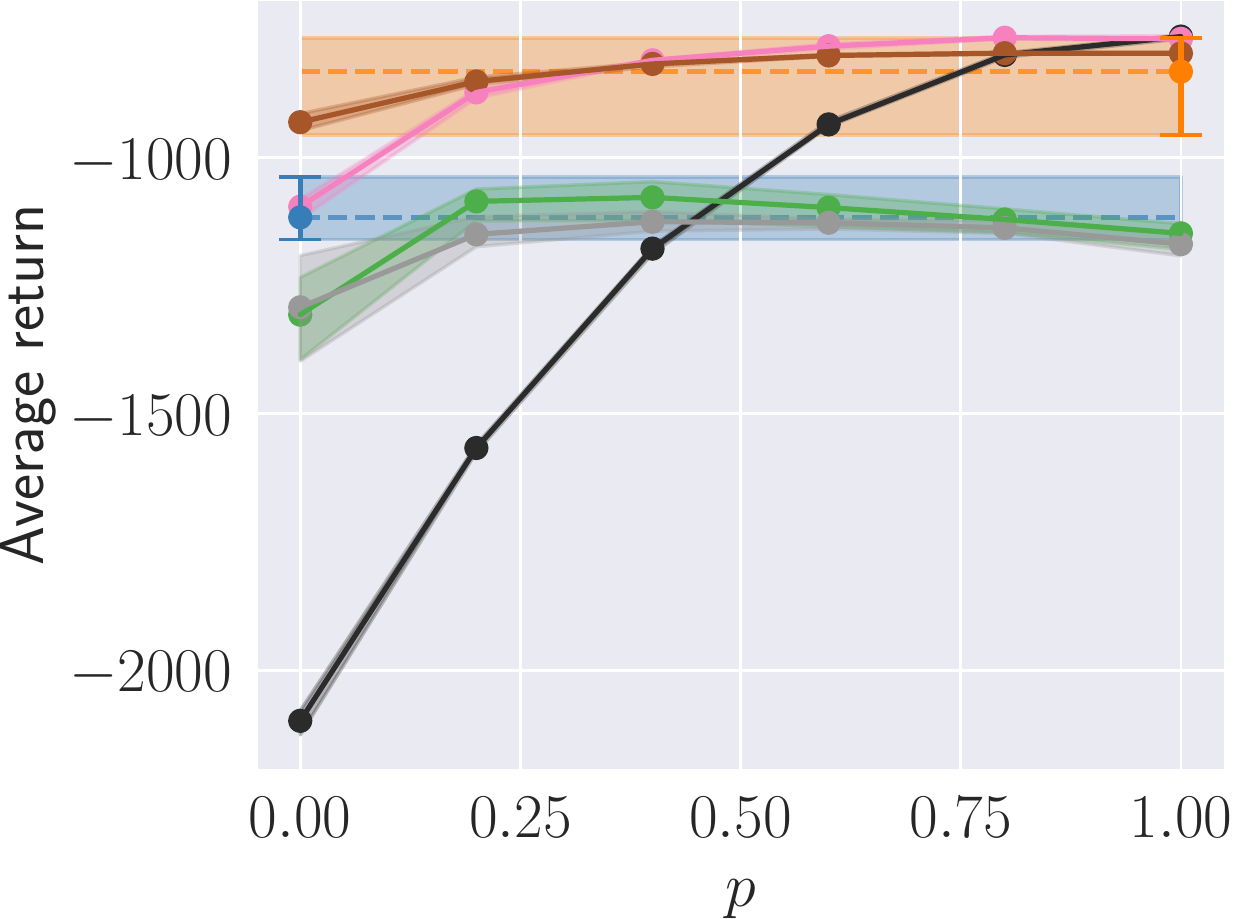}
        \caption{MAPPO.}
    \end{subfigure}
    \caption{(SpreadXY-4) Mean episodic returns for different $p$ values at execution time.}
\end{figure}

\begin{figure}
    \centering
    \includegraphics[height=0.7cm]{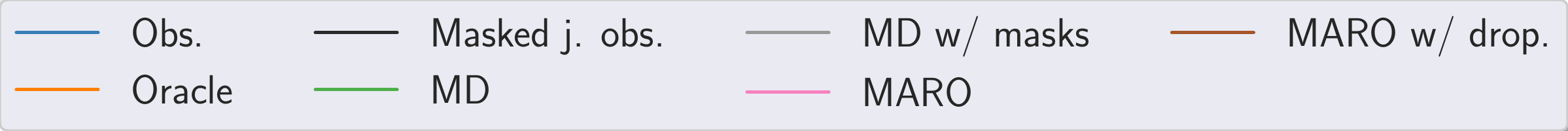}
    \caption{Legend.}
\end{figure}

\clearpage

\begin{table}
\centering
\noindent
\caption{(SpreadBlindFold) Mean episodic returns for $p_{\textrm{default}}$ at execution time.}
\vspace{0.1cm}
\resizebox{\linewidth}{!}{%
\begin{tabular}{c c c c c c c c }\toprule
\multicolumn{1}{c }{\textbf{}} & \multicolumn{7}{c }{\textbf{SpreadBlindFold ($p_\textrm{default}$)}} \\  
\cmidrule(lr){2-8}
\multicolumn{1}{ l }{\textbf{Algorithm}} & \textbf{Obs.} & \textbf{Oracle} & \textbf{Masked j. obs.} & \textbf{MD} & \textbf{MD w/ masks} & \textbf{MARO} & \textbf{MARO w/ drop.} \\
\cmidrule{1-8}
\multicolumn{1}{ l }{IQL} & -425.1 \tiny{(-1.1,+1.4)} & -395.4 \tiny{(-2.8,+2.5)} & -415.5 \tiny{(-7.5,+4.1)} & -401.2 \tiny{(-6.5,+8.5)} & -400.5 \tiny{(-5.5,+8.2)} & -399.3 \tiny{(-5.2,+6.4)} & -389.5 \tiny{(-1.3,+1.6)} \\ \cmidrule{1-8}
\multicolumn{1}{ l }{QMIX} & -416.1 \tiny{(-10.0,+7.7)} & -376.4 \tiny{(-4.7,+4.5)} & -407.3 \tiny{(-1.9,+1.5)} & -401.4 \tiny{(-4.9,+3.5)} & -398.3 \tiny{(-3.0,+2.3)} & -382.3 \tiny{(-5.2,+5.7)} & -373.7 \tiny{(-3.4,+1.8)} \\ \cmidrule{1-8}
\multicolumn{1}{ l }{IPPO} & -436.3 \tiny{(-74.5,+38.4)} & -407.5 \tiny{(-2.8,+1.6)} & -403.1 \tiny{(-2.1,+2.7)} & -446.6 \tiny{(-5.6,+6.4)} & -472.7 \tiny{(-9.7,+12.5)} & -401.7 \tiny{(-0.5,+0.6)} & -417.1 \tiny{(-7.4,+11.0)} \\ \cmidrule{1-8}
\multicolumn{1}{ l }{MAPPO} & -420.3 \tiny{(-0.4,+0.4)} & -404.5 \tiny{(-2.8,+2.0)} & -403.3 \tiny{(-0.8,+1.2)} & -407.3 \tiny{(-1.8,+2.8)} & -412.5 \tiny{(-1.9,+3.3)} & -399.9 \tiny{(-1.2,+0.9)} & -401.4 \tiny{(-2.3,+2.1)} \\
\bottomrule
\end{tabular}
}
\end{table}

\begin{figure}
    \centering
    \begin{subfigure}[b]{0.24\textwidth}
        \centering
        \includegraphics[width=0.97\linewidth]{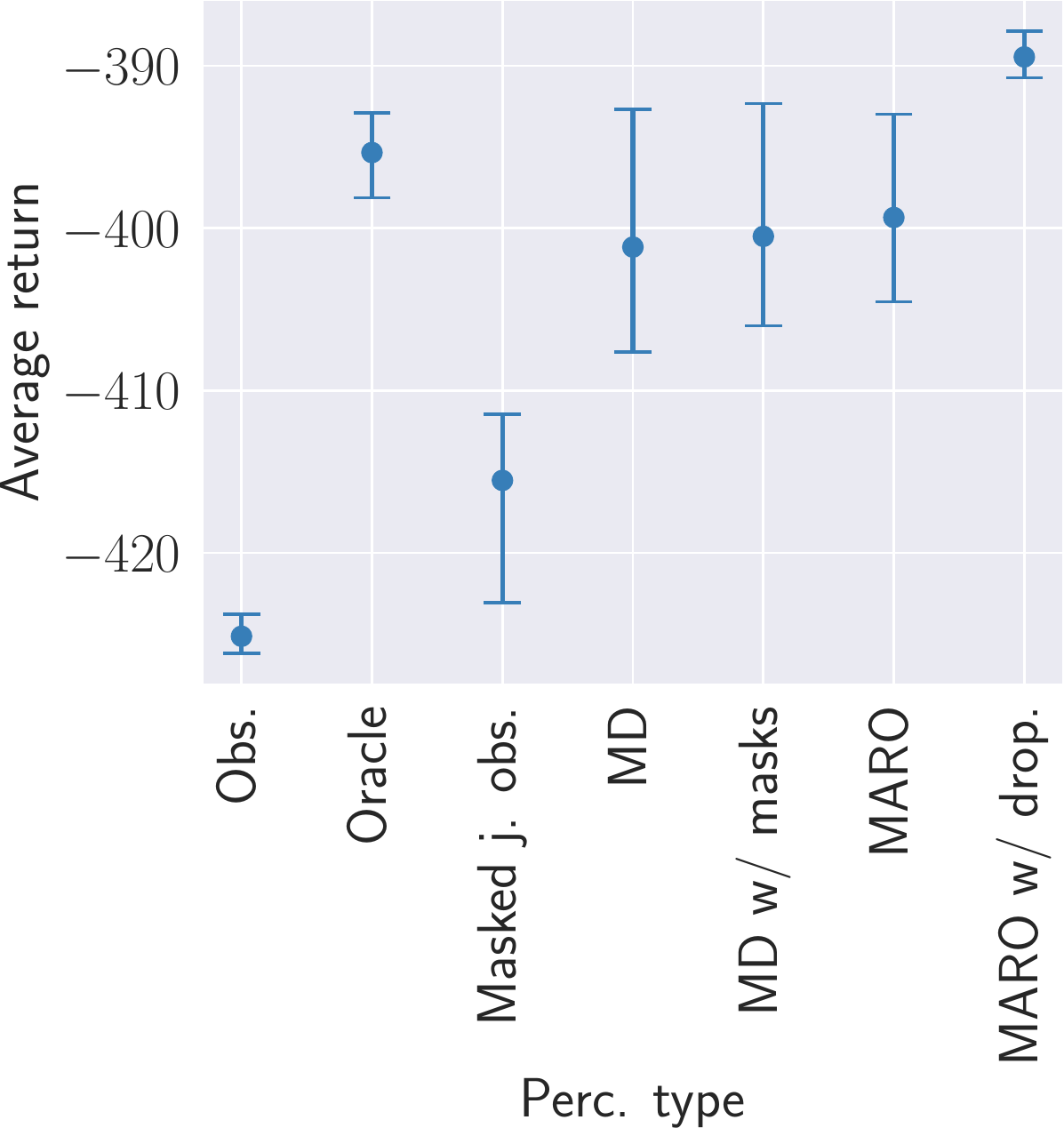}
        \caption{IQL.}
    \end{subfigure}
    \begin{subfigure}[b]{0.24\textwidth}
        \centering
        \includegraphics[width=0.97\linewidth]{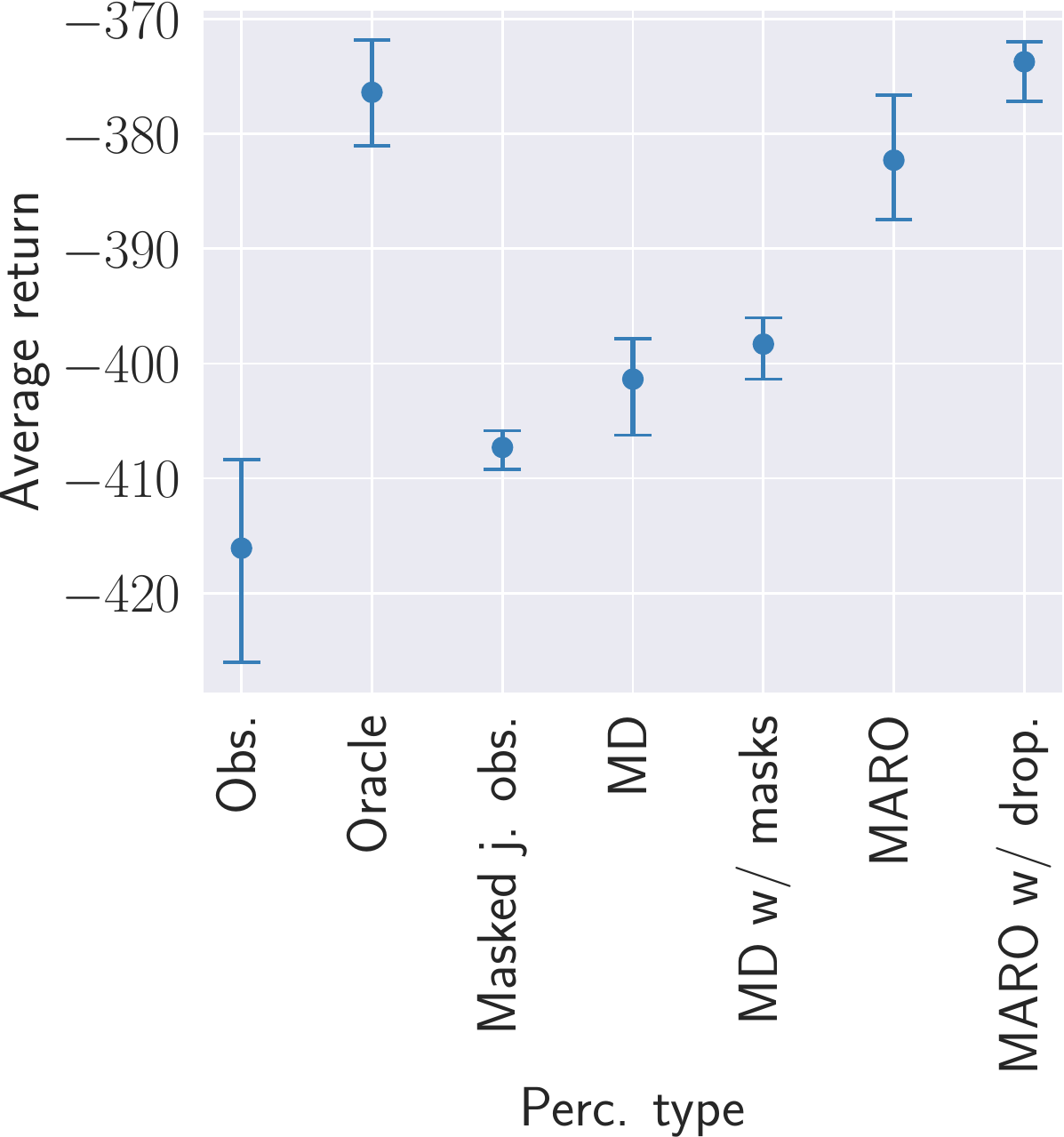}
        \caption{QMIX.}
    \end{subfigure}
    \begin{subfigure}[b]{0.24\textwidth}
        \centering
        \includegraphics[width=0.97\linewidth]{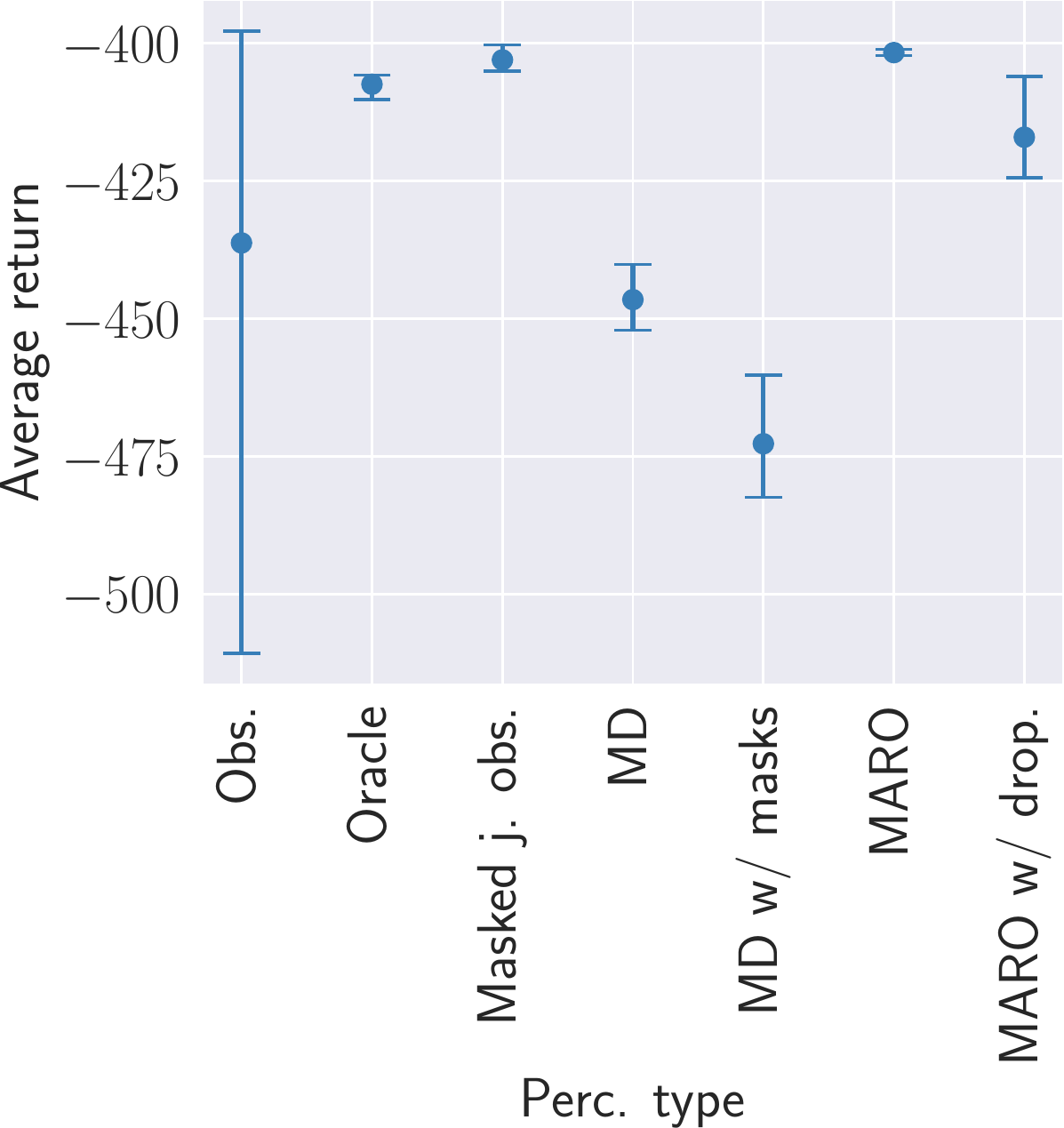}
        \caption{IPPO.}
    \end{subfigure}
    \begin{subfigure}[b]{0.24\textwidth}
        \centering
        \includegraphics[width=0.97\linewidth]{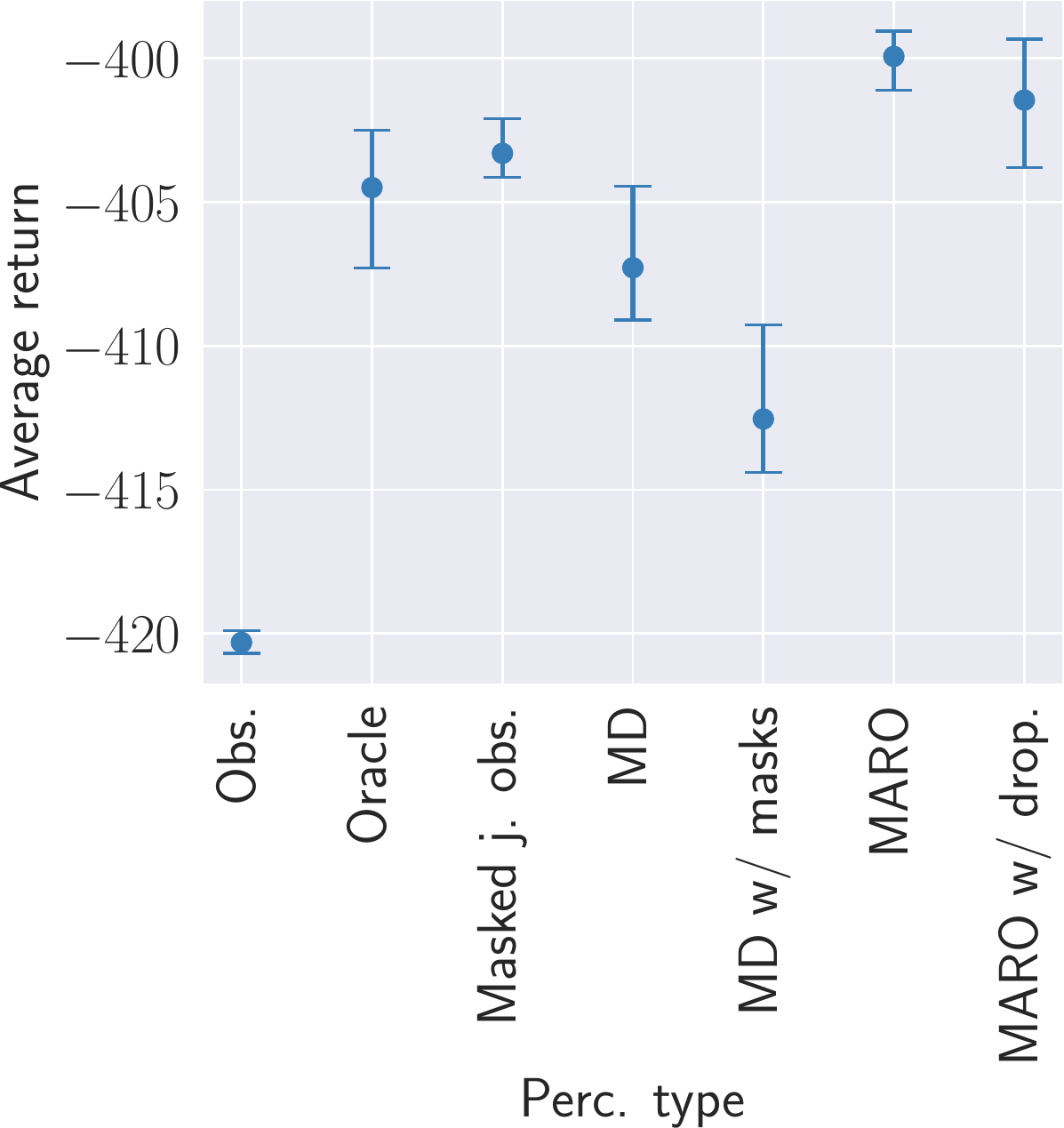}
        \caption{MAPPO.}
    \end{subfigure}
    \caption{(SpreadBlindFold) Mean episodic returns for $p_\textrm{default}$ at execution time.}
\end{figure}

\begin{figure}
    \centering
    \begin{subfigure}[b]{0.24\textwidth}
        \centering
        \includegraphics[width=0.97\linewidth]{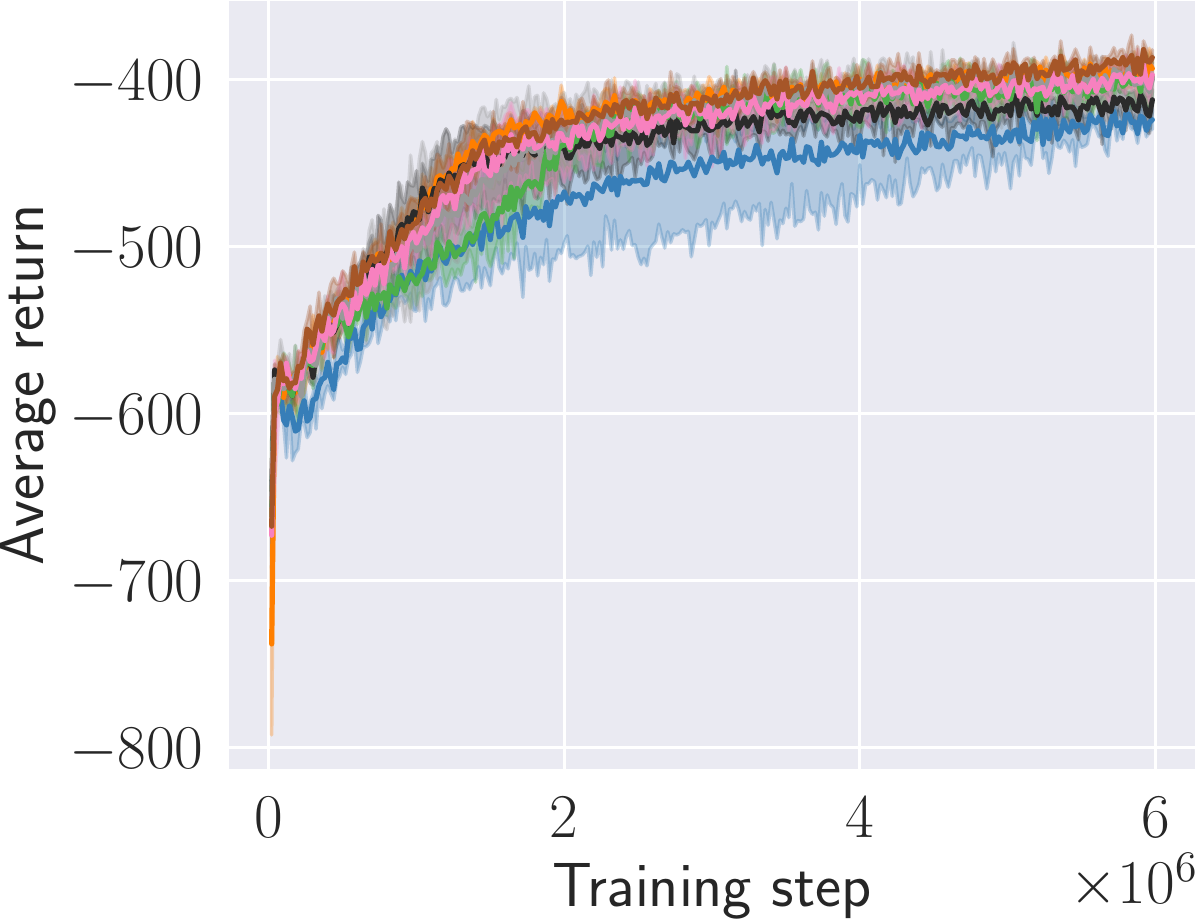}
        \caption{IQL.}
    \end{subfigure}
    \begin{subfigure}[b]{0.24\textwidth}
        \centering
        \includegraphics[width=0.97\linewidth]{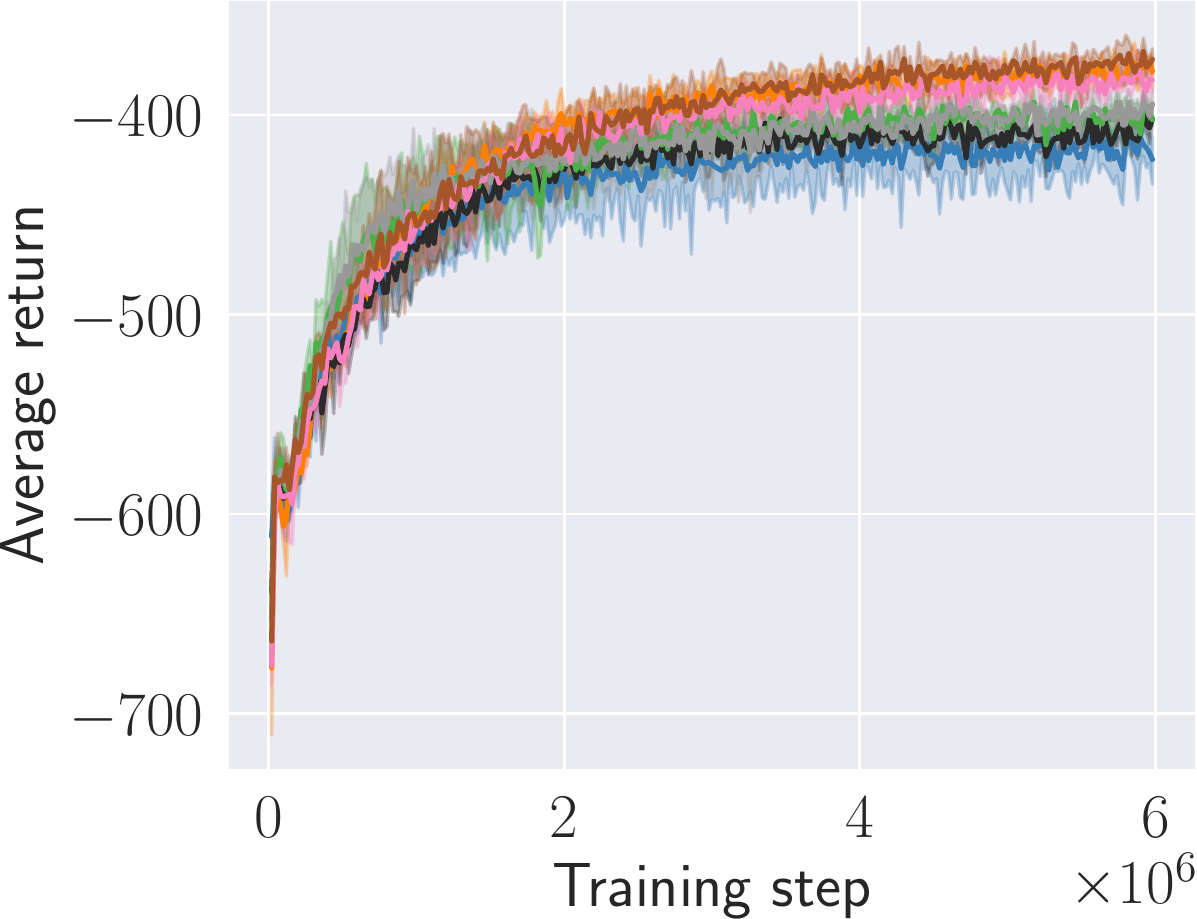}
        \caption{QMIX.}
    \end{subfigure}
    \begin{subfigure}[b]{0.24\textwidth}
        \centering
        \includegraphics[width=0.97\linewidth]{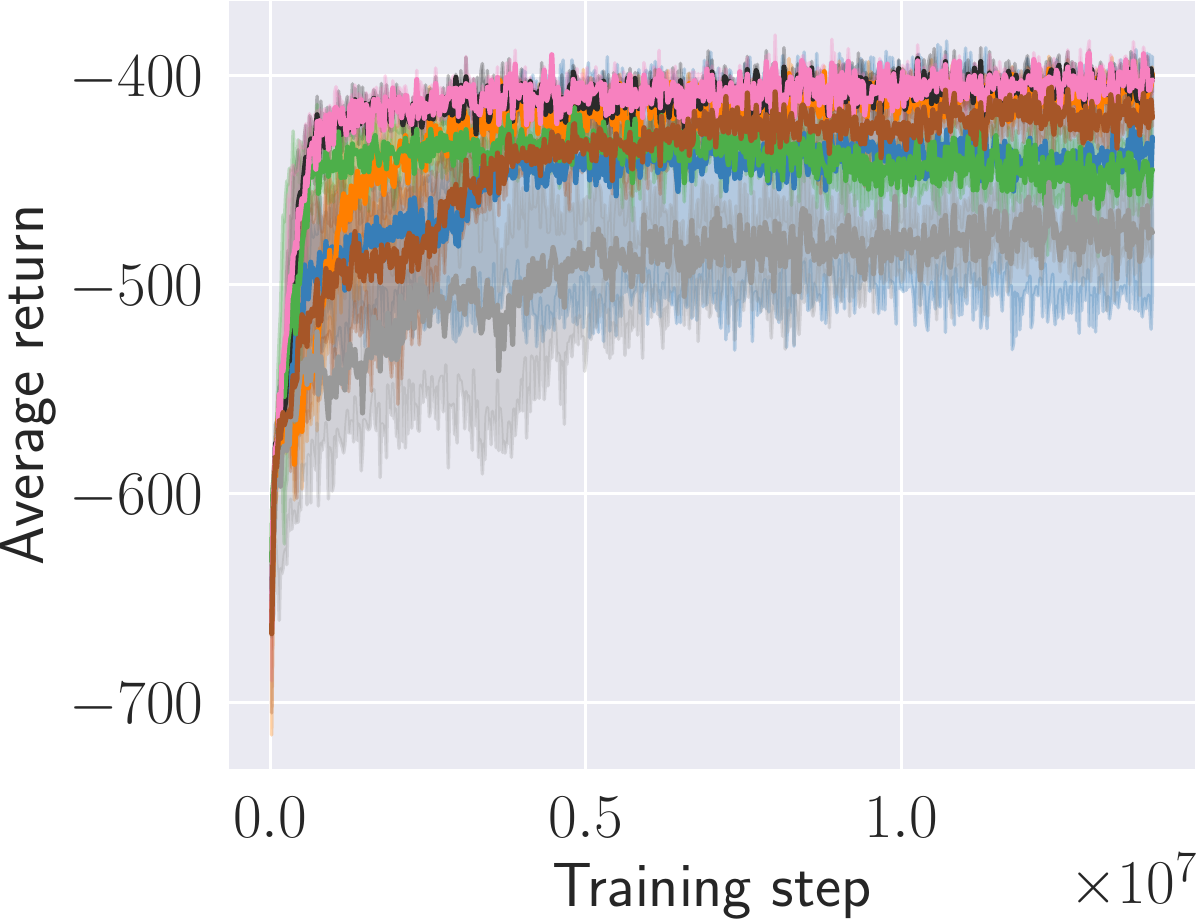}
        \caption{IPPO.}
    \end{subfigure}
    \begin{subfigure}[b]{0.24\textwidth}
        \centering
        \includegraphics[width=0.97\linewidth]{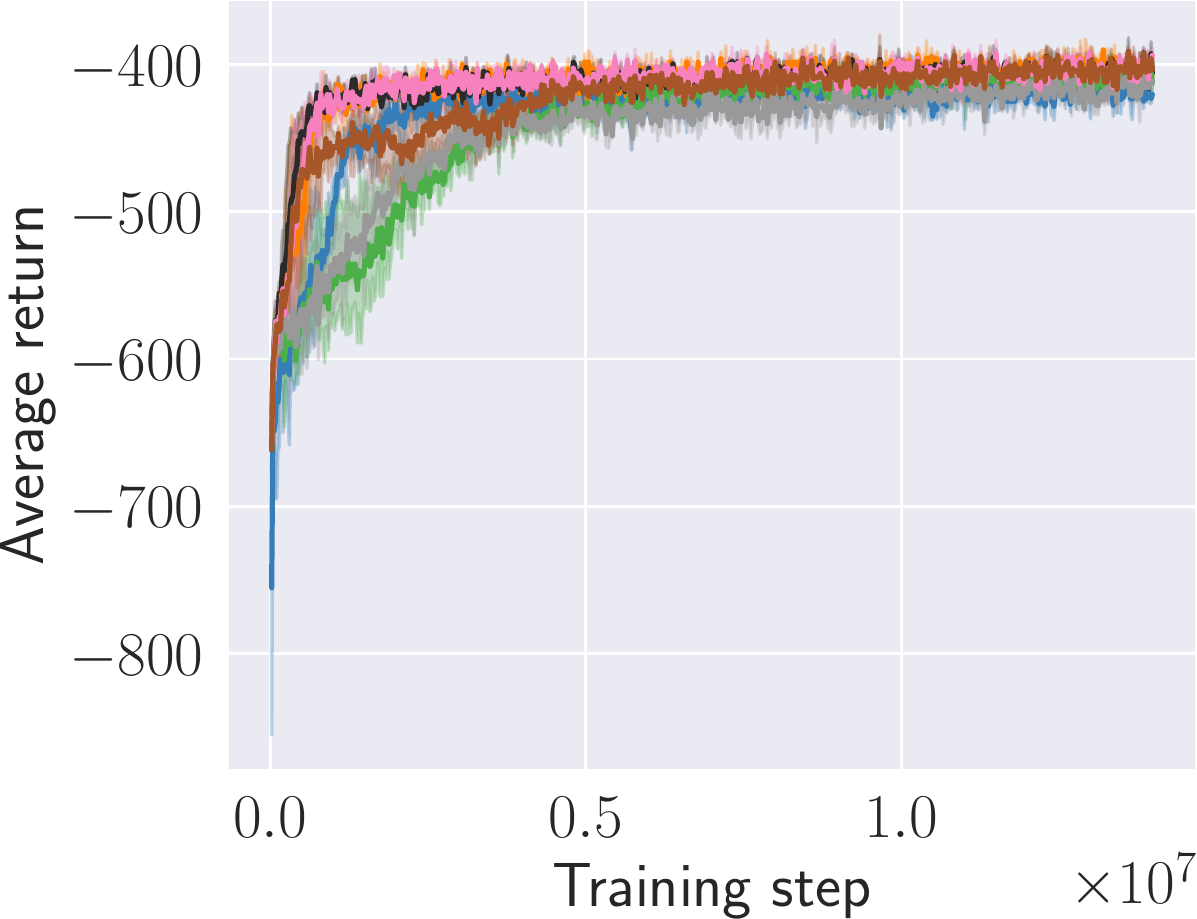}
        \caption{MAPPO.}
    \end{subfigure}
    \caption{(SpreadBlindFold) Mean episodic returns for $p_\textrm{default}$ during training.}
\end{figure}

\begin{figure}
    \centering
    \begin{subfigure}[b]{0.24\textwidth}
        \centering
        \includegraphics[width=0.97\linewidth]{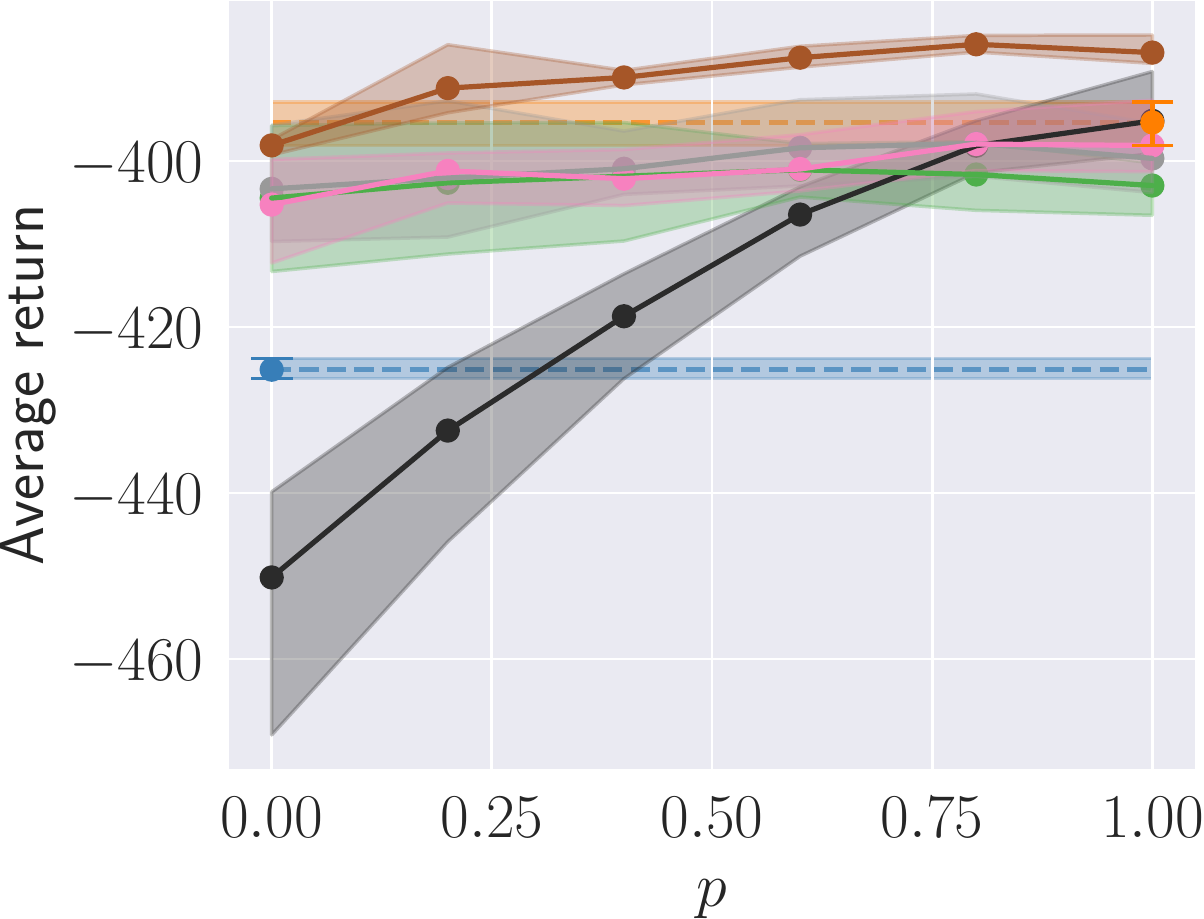}
        \caption{IQL.}
    \end{subfigure}
    \begin{subfigure}[b]{0.24\textwidth}
        \centering
        \includegraphics[width=0.97\linewidth]{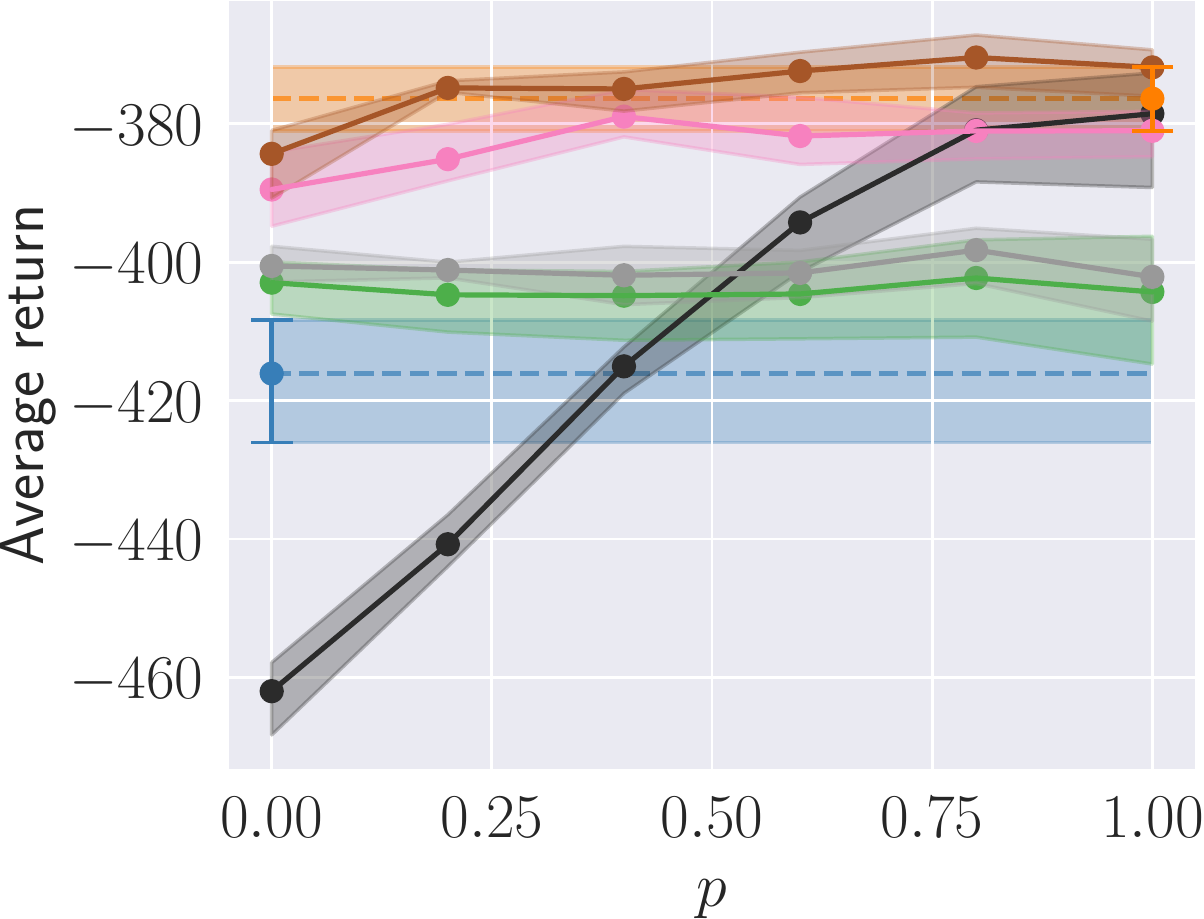}
        \caption{QMIX.}
    \end{subfigure}
    \begin{subfigure}[b]{0.24\textwidth}
        \centering
        \includegraphics[width=0.97\linewidth]{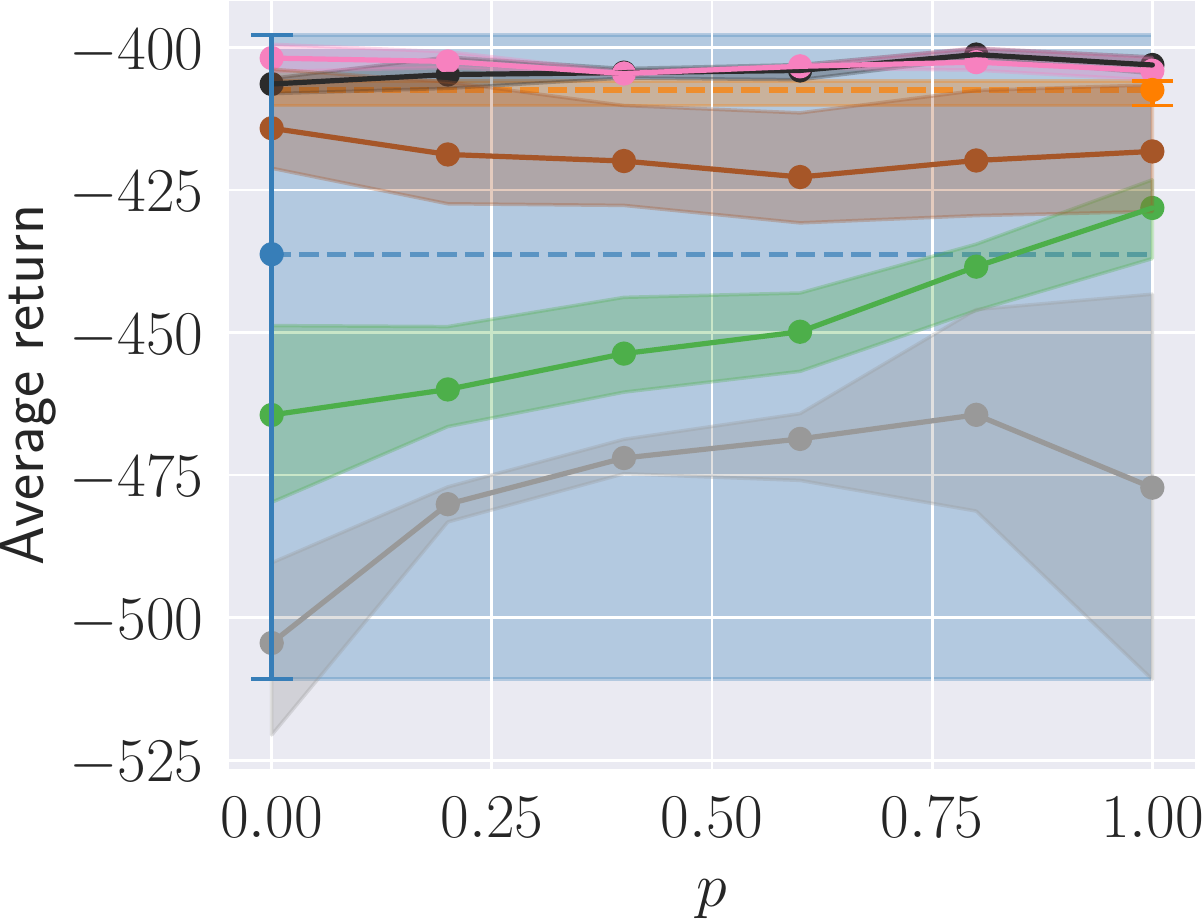}
        \caption{IPPO.}
    \end{subfigure}
    \begin{subfigure}[b]{0.24\textwidth}
        \centering
        \includegraphics[width=0.97\linewidth]{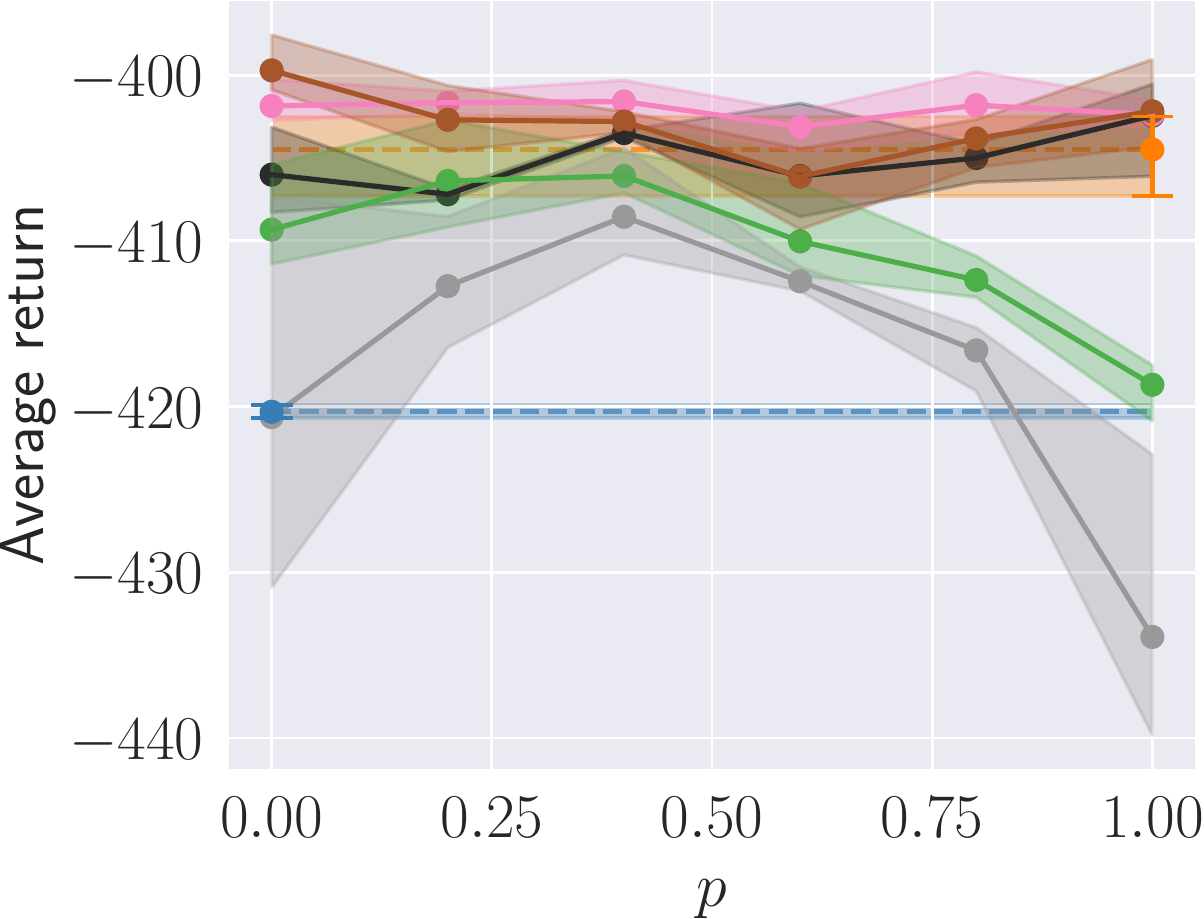}
        \caption{MAPPO.}
    \end{subfigure}
    \caption{(SpreadBlindFold) Mean episodic returns for different $p$ values at execution time.}
\end{figure}

\begin{figure}
    \centering
    \includegraphics[height=0.7cm]{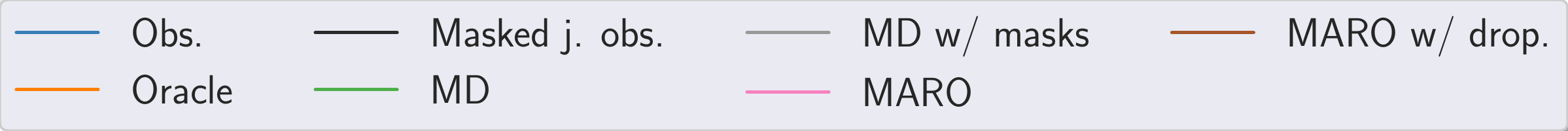}
    \caption{Legend.}
\end{figure}

\clearpage

\begin{table}
\centering
\noindent
\caption{(Foraging-2s-15x15-2p-2f-coop-v2) Mean episodic returns for $p_{\textrm{default}}$ at execution time.}
\vspace{0.1cm}
\resizebox{\linewidth}{!}{%
\begin{tabular}{c c c c c c c c }\toprule
\multicolumn{1}{c }{\textbf{}} & \multicolumn{7}{c }{\textbf{Foraging-2s-15x15-2p-2f-coop-v2 ($p_\textrm{default}$)}} \\  
\cmidrule(lr){2-8}
\multicolumn{1}{ l }{\textbf{Algorithm}} & \textbf{Obs.} & \textbf{Oracle} & \textbf{Masked j. obs.} & \textbf{MD} & \textbf{MD w/ masks} & \textbf{MARO} & \textbf{MARO w/ drop.} \\
\cmidrule{1-8}
\multicolumn{1}{ l }{IQL} & 0.38 \tiny{(-0.03,+0.01)} & 0.48 \tiny{(-0.03,+0.02)} & 0.19 \tiny{(-0.02,+0.01)} & 0.53 \tiny{(-0.03,+0.02)} & 0.52 \tiny{(-0.02,+0.02)} & 0.35 \tiny{(-0.01,+0.02)} & 0.52 \tiny{(-0.01,+0.01)} \\ \cmidrule{1-8}
\multicolumn{1}{ l }{QMIX} & 0.55 \tiny{(-0.01,+0.0)} & 0.68 \tiny{(-0.02,+0.03)} & 0.25 \tiny{(-0.05,+0.03)} & 0.58 \tiny{(-0.01,+0.02)} & 0.59 \tiny{(-0.02,+0.02)} & 0.44 \tiny{(-0.02,+0.02)} & 0.60 \tiny{(-0.01,+0.02)} \\ \cmidrule{1-8}
\multicolumn{1}{ l }{IPPO} & 0.31 \tiny{(-0.0,+0.0)} & 0.44 \tiny{(-0.0,+0.0)} & 0.30 \tiny{(-0.01,+0.01)} & 0.03 \tiny{(-0.02,+0.05)} & 0.01 \tiny{(-0.01,+0.02)} & 0.37 \tiny{(-0.01,+0.02)} & 0.01 \tiny{(-0.0,+0.0)} \\ \cmidrule{1-8}
\multicolumn{1}{ l }{MAPPO} & 0.36 \tiny{(-0.0,+0.0)} & 0.45 \tiny{(-0.01,+0.01)} & 0.31 \tiny{(-0.02,+0.03)} & 0.02 \tiny{(-0.01,+0.02)} & 0.02 \tiny{(-0.02,+0.03)} & 0.38 \tiny{(-0.03,+0.04)} & 0.09 \tiny{(-0.08,+0.06)} \\
\bottomrule
\end{tabular}
}
\end{table}

\begin{figure}
    \centering
    \begin{subfigure}[b]{0.24\textwidth}
        \centering
        \includegraphics[width=0.97\linewidth]{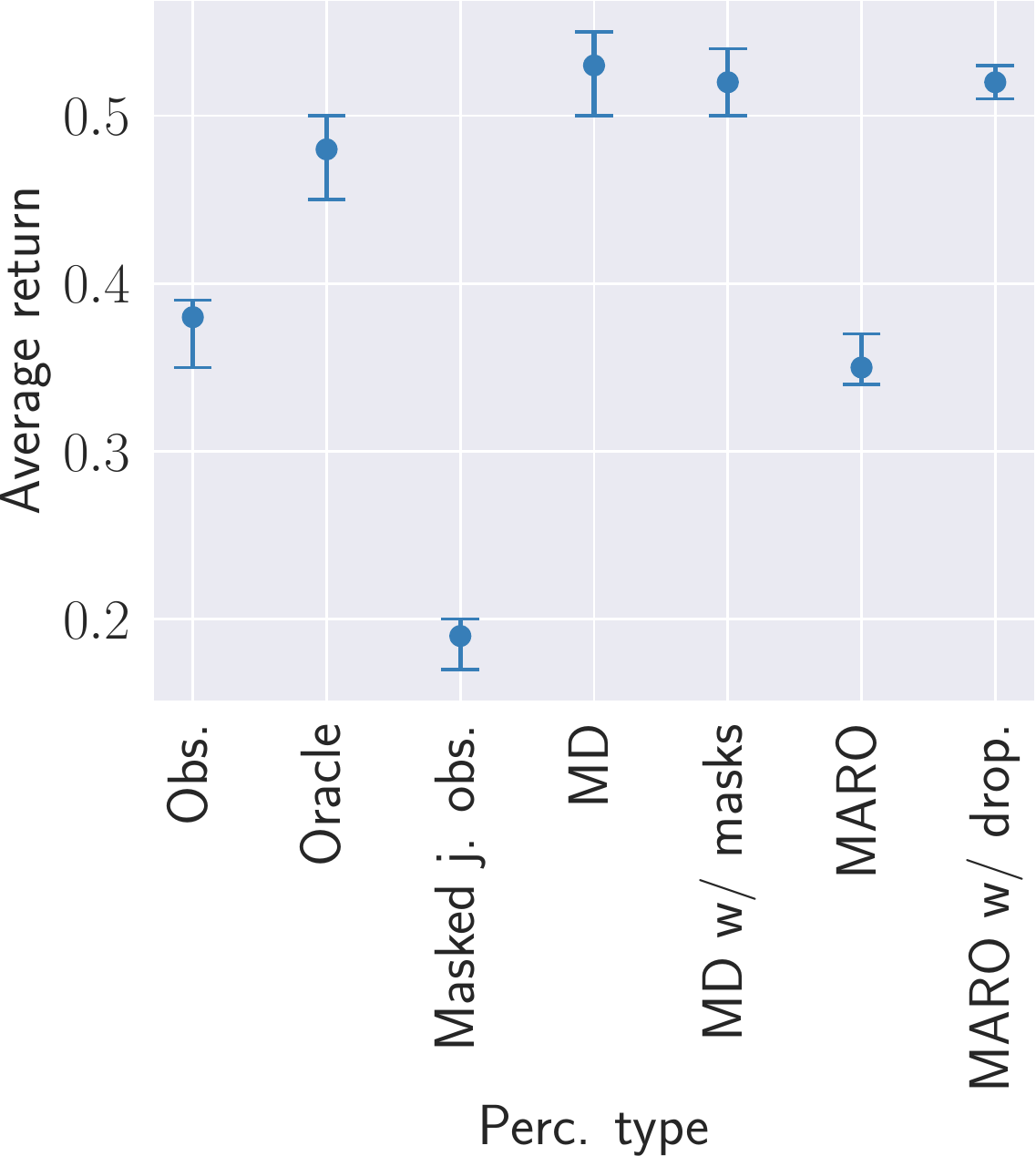}
        \caption{IQL.}
    \end{subfigure}
    \begin{subfigure}[b]{0.24\textwidth}
        \centering
        \includegraphics[width=0.97\linewidth]{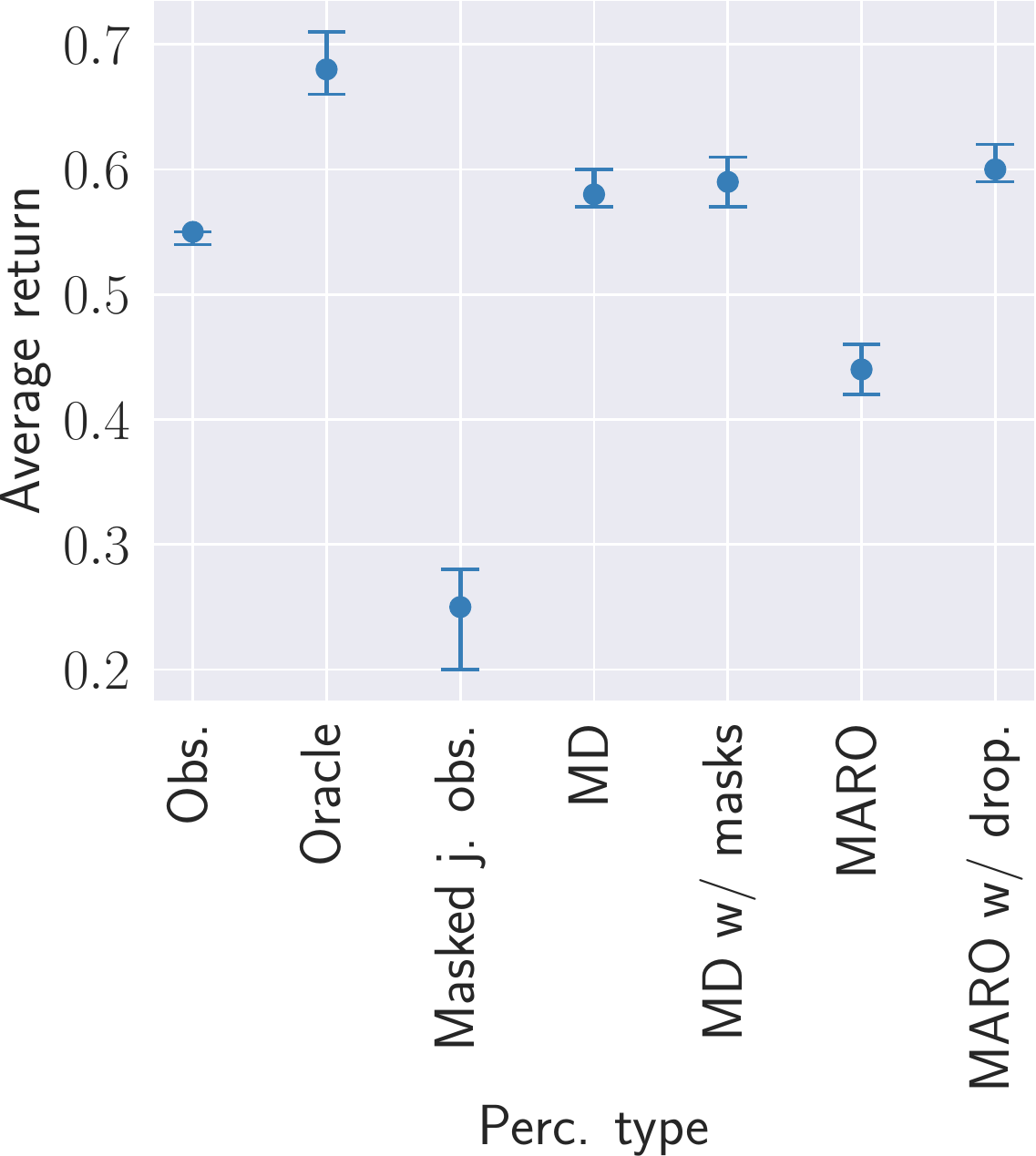}
        \caption{QMIX.}
    \end{subfigure}
    \begin{subfigure}[b]{0.24\textwidth}
        \centering
        \includegraphics[width=0.97\linewidth]{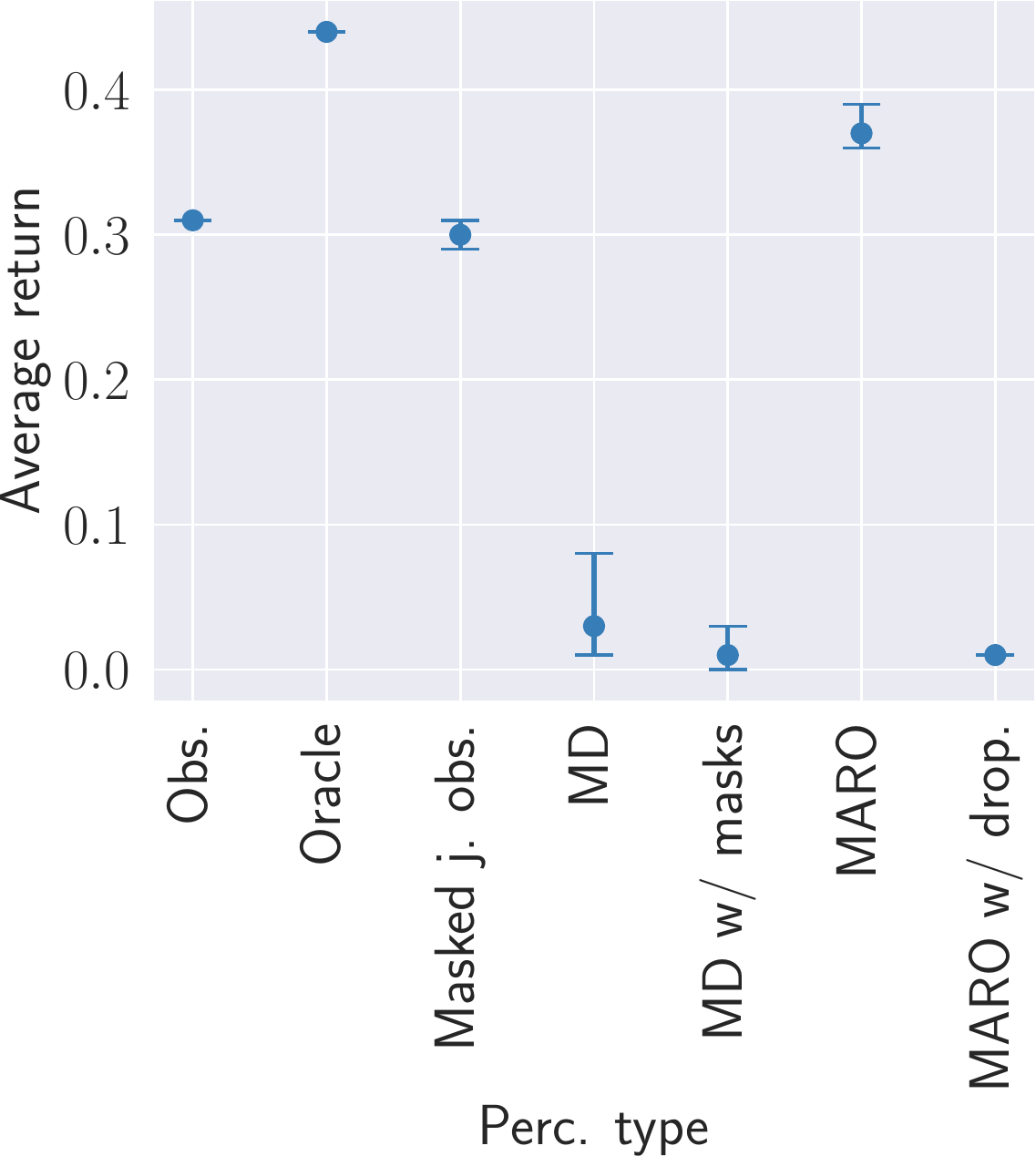}
        \caption{IPPO.}
    \end{subfigure}
    \begin{subfigure}[b]{0.24\textwidth}
        \centering
        \includegraphics[width=0.97\linewidth]{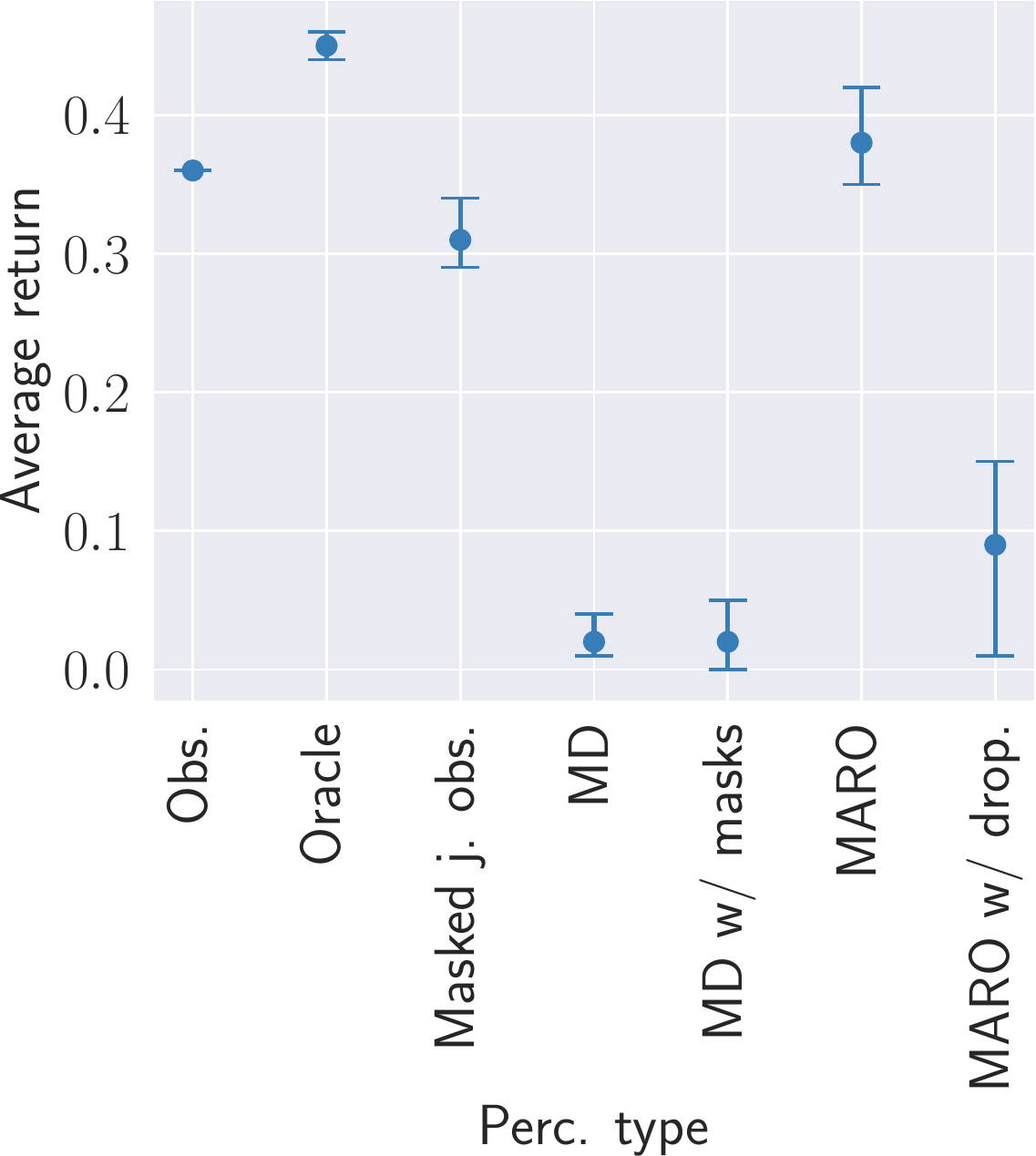}
        \caption{MAPPO.}
    \end{subfigure}
    \caption{(Foraging-2s-15x15-2p-2f-coop-v2) Mean episodic returns for $p_\textrm{default}$ at execution time.}
\end{figure}

\begin{figure}
    \centering
    \begin{subfigure}[b]{0.24\textwidth}
        \centering
        \includegraphics[width=0.97\linewidth]{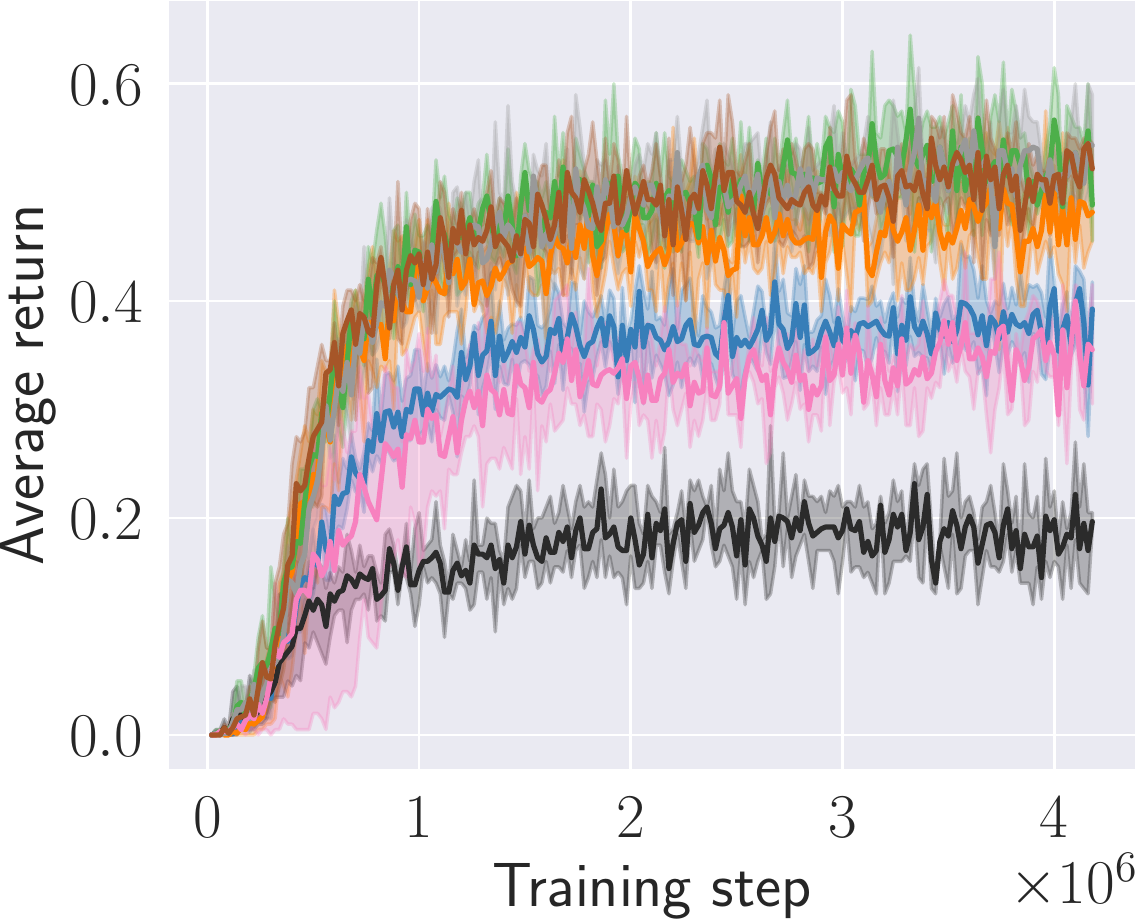}
        \caption{IQL.}
    \end{subfigure}
    \begin{subfigure}[b]{0.24\textwidth}
        \centering
        \includegraphics[width=0.97\linewidth]{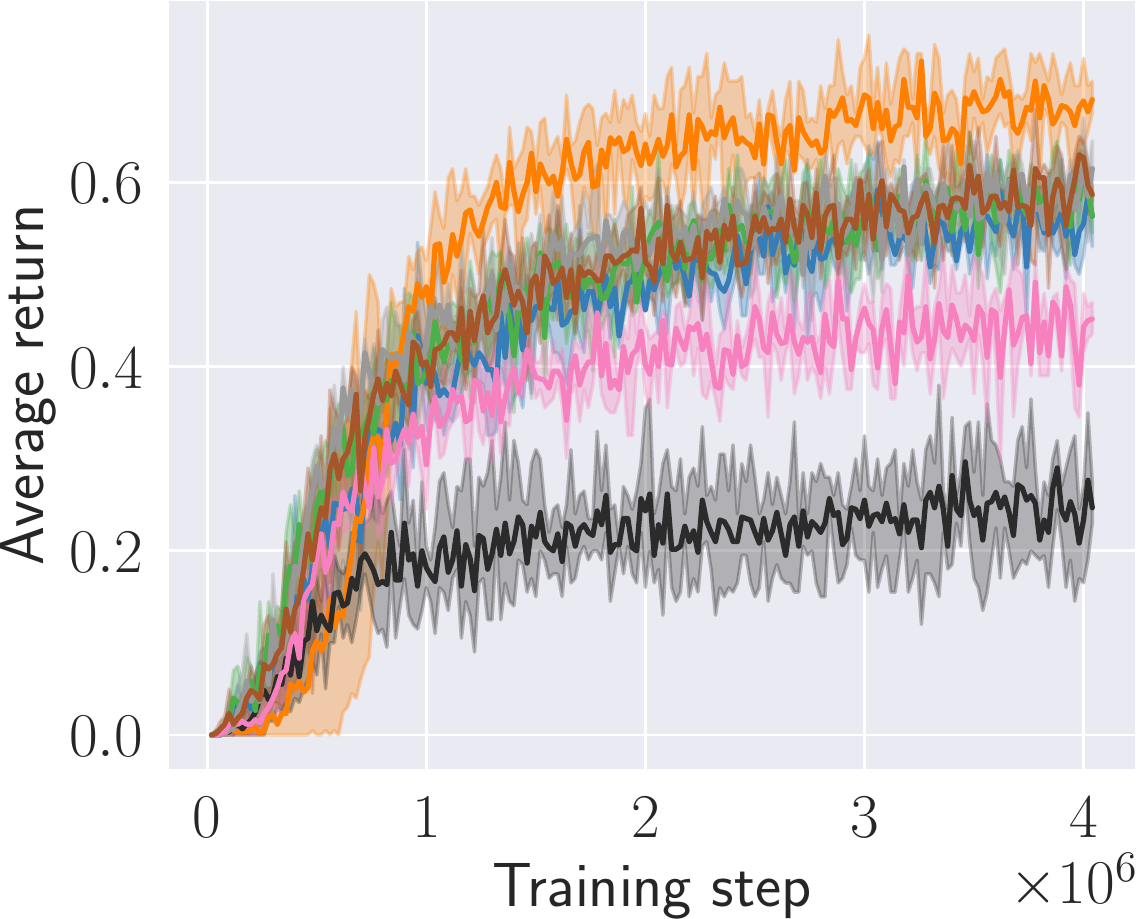}
        \caption{QMIX.}
    \end{subfigure}
    \begin{subfigure}[b]{0.24\textwidth}
        \centering
        \includegraphics[width=0.97\linewidth]{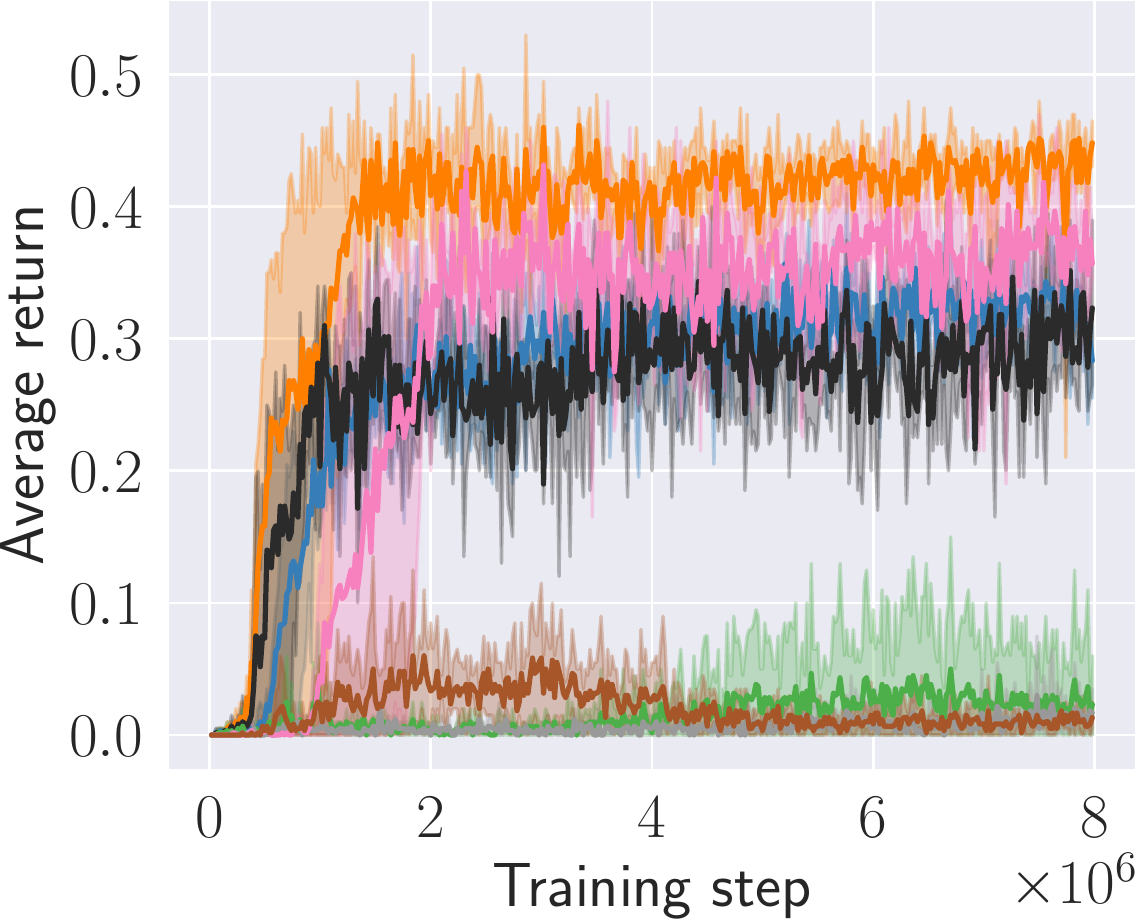}
        \caption{IPPO.}
    \end{subfigure}
    \begin{subfigure}[b]{0.24\textwidth}
        \centering
        \includegraphics[width=0.97\linewidth]{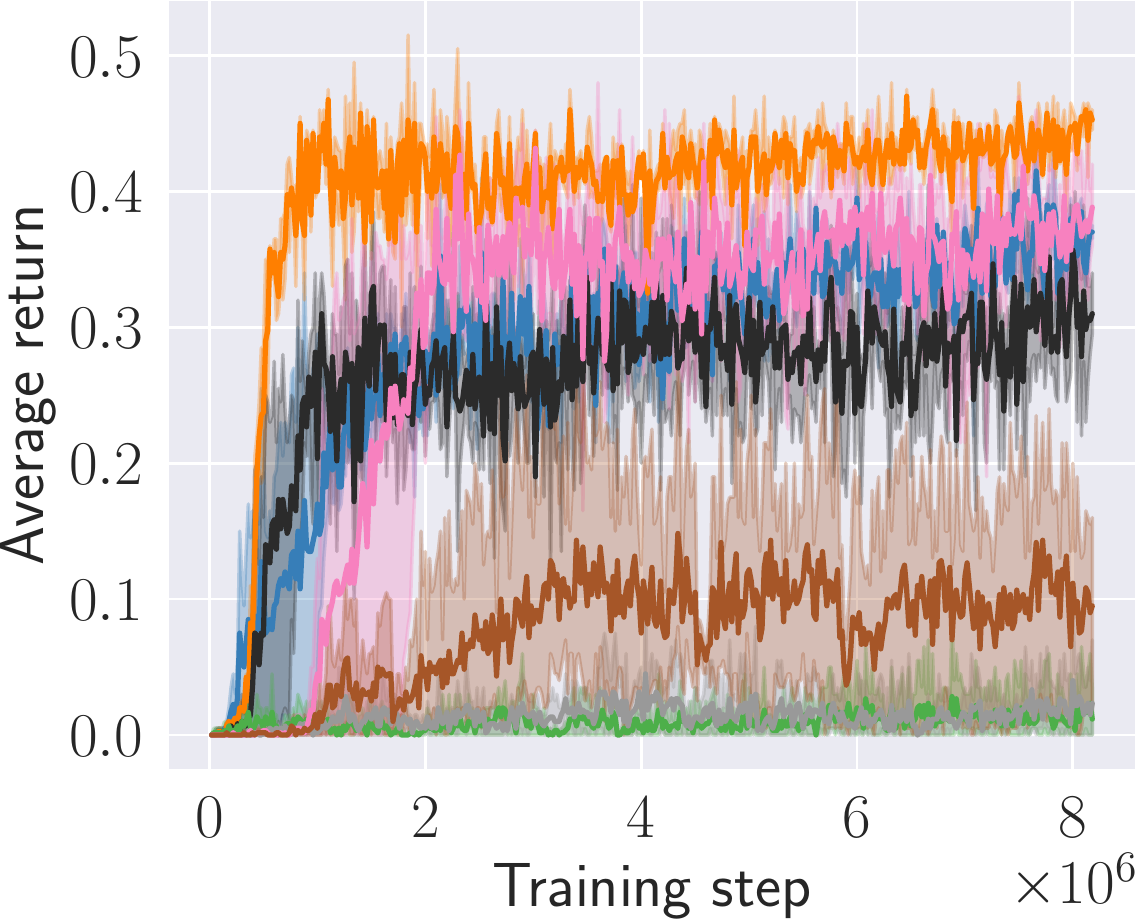}
        \caption{MAPPO.}
    \end{subfigure}
    \caption{(Foraging-2s-15x15-2p-2f-coop-v2) Mean episodic returns for $p_\textrm{default}$ during training.}
\end{figure}

\begin{figure}
    \centering
    \begin{subfigure}[b]{0.24\textwidth}
        \centering
        \includegraphics[width=0.97\linewidth]{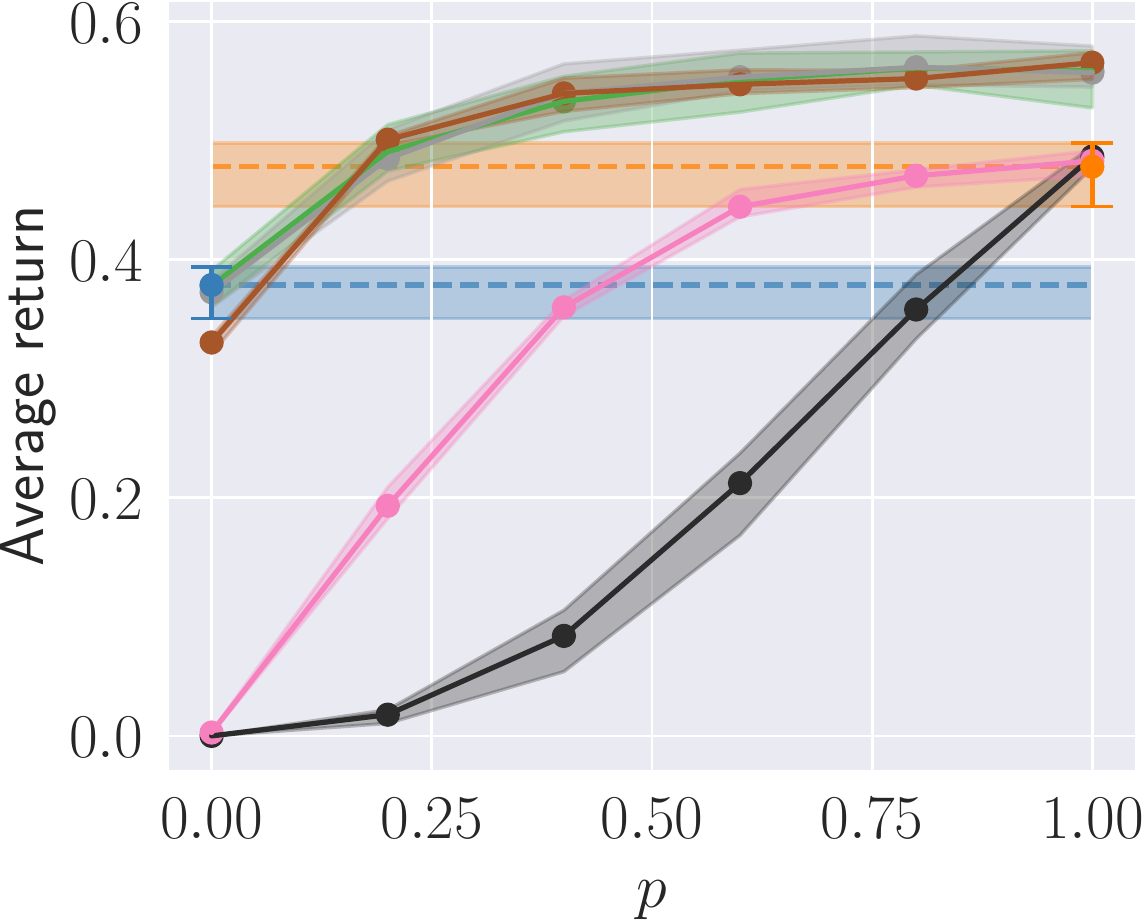}
        \caption{IQL.}
    \end{subfigure}
    \begin{subfigure}[b]{0.24\textwidth}
        \centering
        \includegraphics[width=0.97\linewidth]{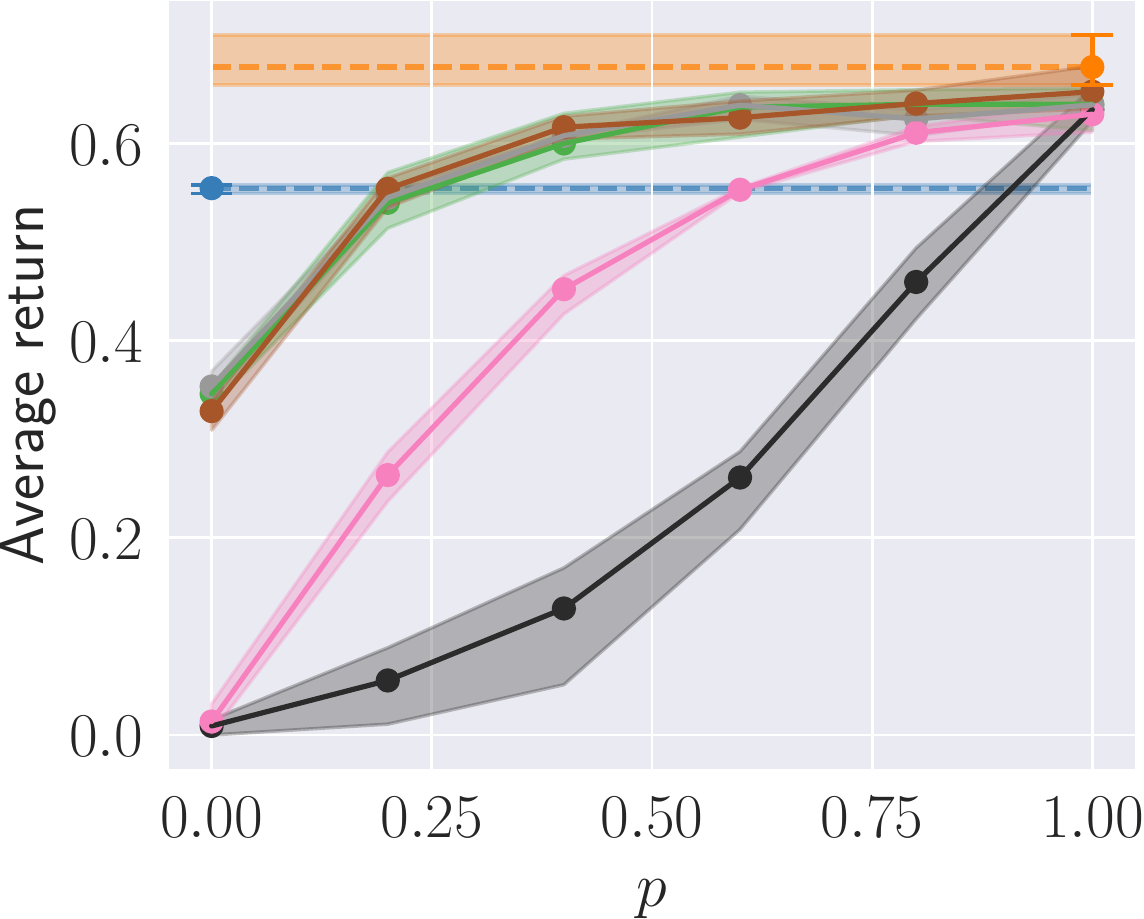}
        \caption{QMIX.}
    \end{subfigure}
    \begin{subfigure}[b]{0.24\textwidth}
        \centering
        \includegraphics[width=0.97\linewidth]{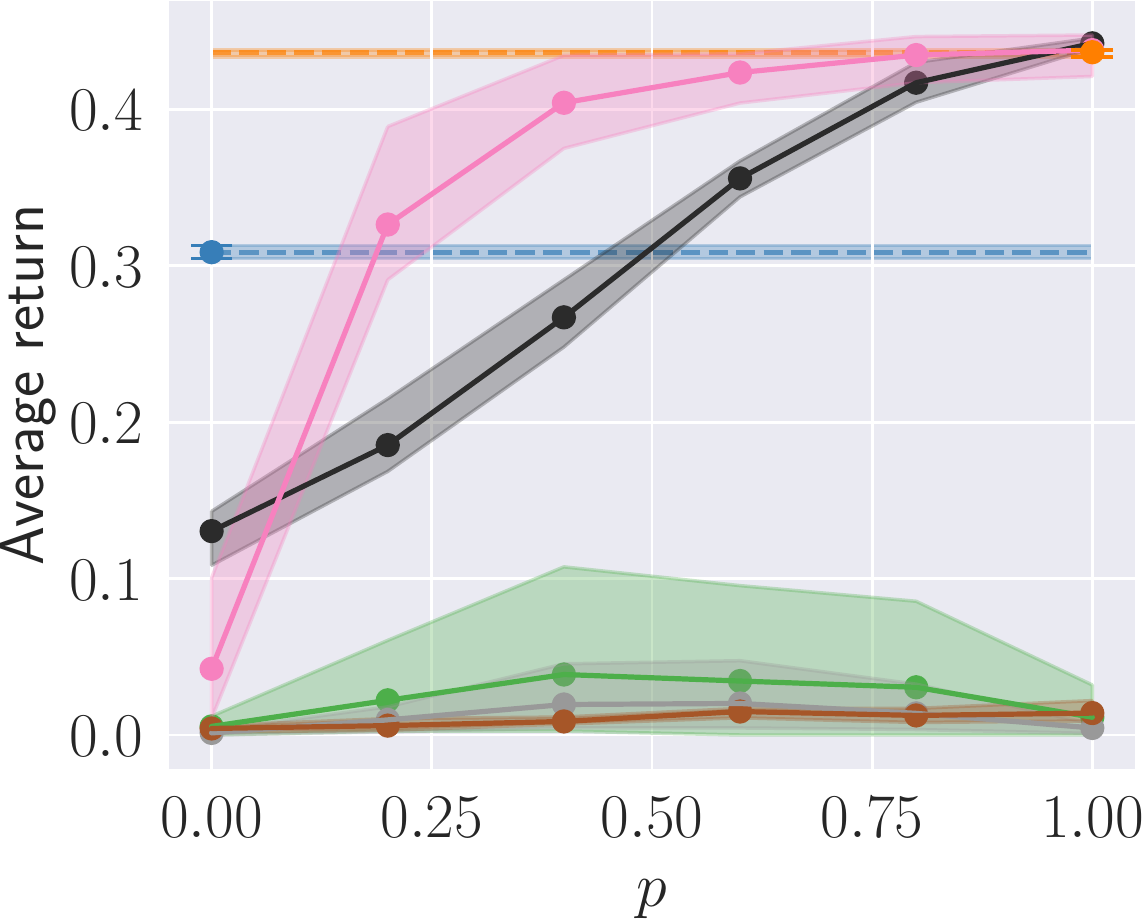}
        \caption{IPPO.}
    \end{subfigure}
    \begin{subfigure}[b]{0.24\textwidth}
        \centering
        \includegraphics[width=0.97\linewidth]{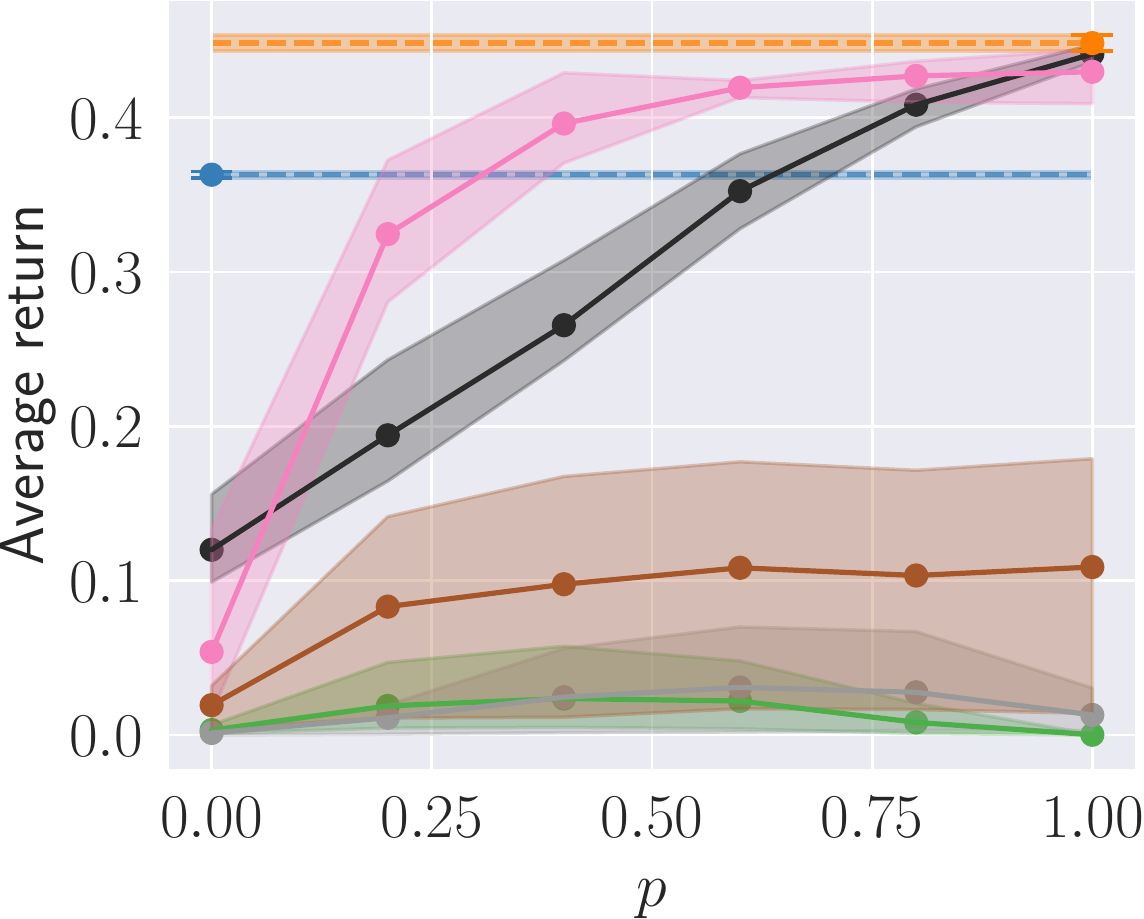}
        \caption{MAPPO.}
    \end{subfigure}
    \caption{(Foraging-2s-15x15-2p-2f-coop-v2) Mean episodic returns for different $p$ values at execution time.}
\end{figure}

\begin{figure}
    \centering
    \includegraphics[height=0.7cm]{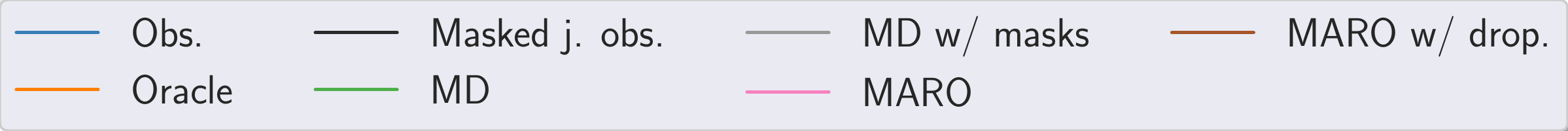}
    \caption{Legend.}
\end{figure}


\clearpage
\subsubsection{Sampling Schemes of the Communication Matrix}
\label{appendix:experimental_evaluation:experimental_results:communication_matrix}
In this section, we present the complete experimental results for the different sampling schemes of the communication matrix. The results herein presented correspond to the full results for Sec.~\ref{sec:evaluation:results:comm_protocols} of the main text. More precisely, we display the mean episodic returns for: (i) the communication setting $p_{\textrm{asymmetric}}$, which features communication matrices $C$ such that $p_{i, j} \neq p_{j, i} \sim \mathcal{U}(0,1)$, sampled at the beginning of each episode; and (ii) the communication setting $p_{\textrm{dynamic}}$, similar to (i) but with $C$'s sampled every $5$ time steps. The Oracle baseline is always evaluated with $p=1$.

\clearpage


\begin{table}
\centering
\noindent
\caption{(SpeakerListener) Mean episodic returns for $p_{\textrm{asymmetric}}$ at execution time.}
\vspace{0.1cm}
\resizebox{\linewidth}{!}{%
\begin{tabular}{c c c c c c c c }\toprule
\multicolumn{1}{c }{\textbf{}} & \multicolumn{7}{c }{\textbf{SpeakerListener ($p_\textrm{asymmetric}$)}} \\  
\cmidrule(lr){2-8}
\multicolumn{1}{ l }{\textbf{Algorithm}} & \textbf{Obs.} & \textbf{Oracle} & \textbf{Masked j. obs.} & \textbf{MD} & \textbf{MD w/ masks} & \textbf{MARO} & \textbf{MARO w/ drop.} \\
\cmidrule{1-8}
\multicolumn{1}{ l }{IQL} & -40.0 \tiny{(-0.4,+0.4)} & -24.2 \tiny{(-0.1,+0.2)} & -45.6 \tiny{(-1.4,+1.9)} & -25.5 \tiny{(-1.4,+1.1)} & -25.6 \tiny{(-1.4,+1.0)} & -25.4 \tiny{(-1.3,+1.0)} & -25.4 \tiny{(-1.3,+0.9)} \\ \cmidrule{1-8}
\multicolumn{1}{ l }{QMIX} & -24.9 \tiny{(-0.1,+0.0)} & -23.9 \tiny{(-0.2,+0.2)} & -40.4 \tiny{(-0.4,+0.2)} & -25.3 \tiny{(-1.4,+1.0)} & -25.3 \tiny{(-1.5,+1.1)} & -25.2 \tiny{(-1.5,+1.1)} & -24.5 \tiny{(-0.4,+0.4)} \\ \cmidrule{1-8}
\multicolumn{1}{ l }{IPPO} & -33.3 \tiny{(-11.6,+5.9)} & -25.1 \tiny{(-0.3,+0.6)} & -36.9 \tiny{(-0.1,+0.1)} & -43.1 \tiny{(-3.1,+3.1)} & -44.8 \tiny{(-2.0,+2.6)} & -25.5 \tiny{(-0.2,+0.3)} & -29.0 \tiny{(-1.0,+0.8)} \\ \cmidrule{1-8}
\multicolumn{1}{ l }{MAPPO} & -59.6 \tiny{(-0.5,+0.6)} & -25.0 \tiny{(-0.2,+0.3)} & -36.7 \tiny{(-0.1,+0.2)} & -28.8 \tiny{(-0.1,+0.1)} & -27.8 \tiny{(-0.8,+0.8)} & -25.5 \tiny{(-0.2,+0.3)} & -28.2 \tiny{(-0.3,+0.6)} \\
\bottomrule
\end{tabular}
}
\end{table}

\begin{table}
\centering
\noindent
\caption{(SpeakerListener) Mean episodic returns for $p_{\textrm{dynamic}}$ at execution time.}
\vspace{0.1cm}
\resizebox{\linewidth}{!}{%
\begin{tabular}{c c c c c c c c }\toprule
\multicolumn{1}{c }{\textbf{}} & \multicolumn{7}{c }{\textbf{SpeakerListener ($p_\textrm{dynamic}$)}} \\  
\cmidrule(lr){2-8}
\multicolumn{1}{ l }{\textbf{Algorithm}} & \textbf{Obs.} & \textbf{Oracle} & \textbf{Masked j. obs.} & \textbf{MD} & \textbf{MD w/ masks} & \textbf{MARO} & \textbf{MARO w/ drop.} \\
\cmidrule{1-8}
\multicolumn{1}{ l }{IQL} & -40.0 \tiny{(-0.4,+0.4)} & -24.2 \tiny{(-0.1,+0.2)} & -44.8 \tiny{(-1.6,+0.9)} & -24.7 \tiny{(-0.4,+0.3)} & -24.8 \tiny{(-0.4,+0.3)} & -24.6 \tiny{(-0.4,+0.4)} & -24.6 \tiny{(-0.3,+0.4)} \\ \cmidrule{1-8}
\multicolumn{1}{ l }{QMIX} & -24.9 \tiny{(-0.1,+0.0)} & -23.9 \tiny{(-0.2,+0.2)} & -39.7 \tiny{(-1.1,+1.2)} & -24.5 \tiny{(-0.4,+0.3)} & -24.5 \tiny{(-0.3,+0.2)} & -24.4 \tiny{(-0.4,+0.2)} & -24.5 \tiny{(-0.2,+0.2)} \\ \cmidrule{1-8}
\multicolumn{1}{ l }{IPPO} & -33.3 \tiny{(-11.6,+5.9)} & -25.1 \tiny{(-0.3,+0.6)} & -35.7 \tiny{(-0.1,+0.1)} & -40.1 \tiny{(-1.6,+1.6)} & -40.5 \tiny{(-2.2,+3.6)} & -25.1 \tiny{(-0.2,+0.2)} & -28.7 \tiny{(-1.0,+1.0)} \\ \cmidrule{1-8}
\multicolumn{1}{ l }{MAPPO} & -59.6 \tiny{(-0.5,+0.6)} & -25.0 \tiny{(-0.2,+0.3)} & -35.6 \tiny{(-0.6,+0.3)} & -27.8 \tiny{(-0.2,+0.4)} & -27.1 \tiny{(-0.7,+0.8)} & -25.1 \tiny{(-0.2,+0.2)} & -27.8 \tiny{(-0.5,+0.7)} \\
\bottomrule
\end{tabular}
}
\end{table}

\begin{figure}
    \centering
    \begin{subfigure}[b]{0.24\textwidth}
        \centering
        \includegraphics[width=0.97\linewidth]{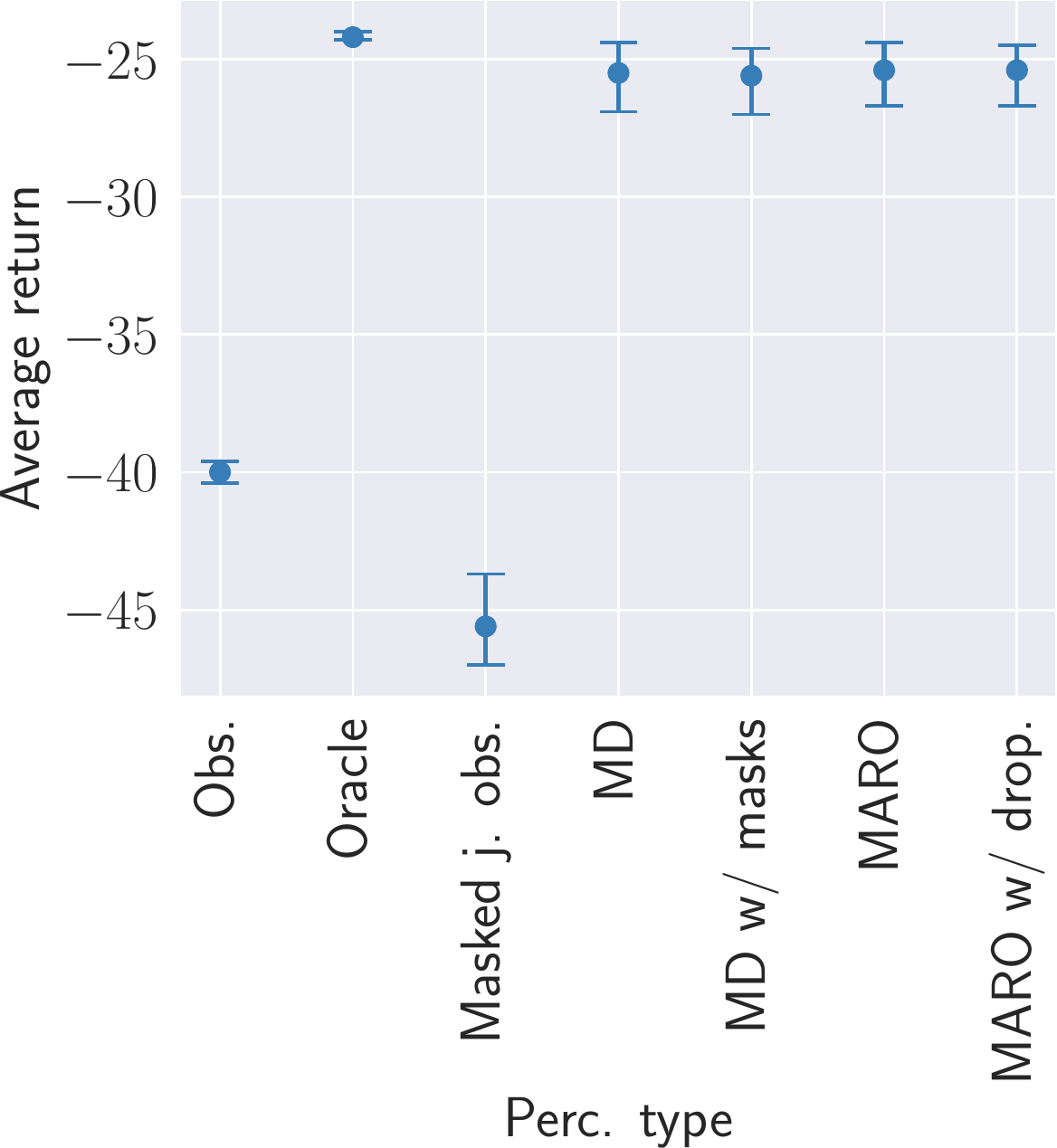}
        \caption{IQL.}
    \end{subfigure}
    \begin{subfigure}[b]{0.24\textwidth}
        \centering
        \includegraphics[width=0.97\linewidth]{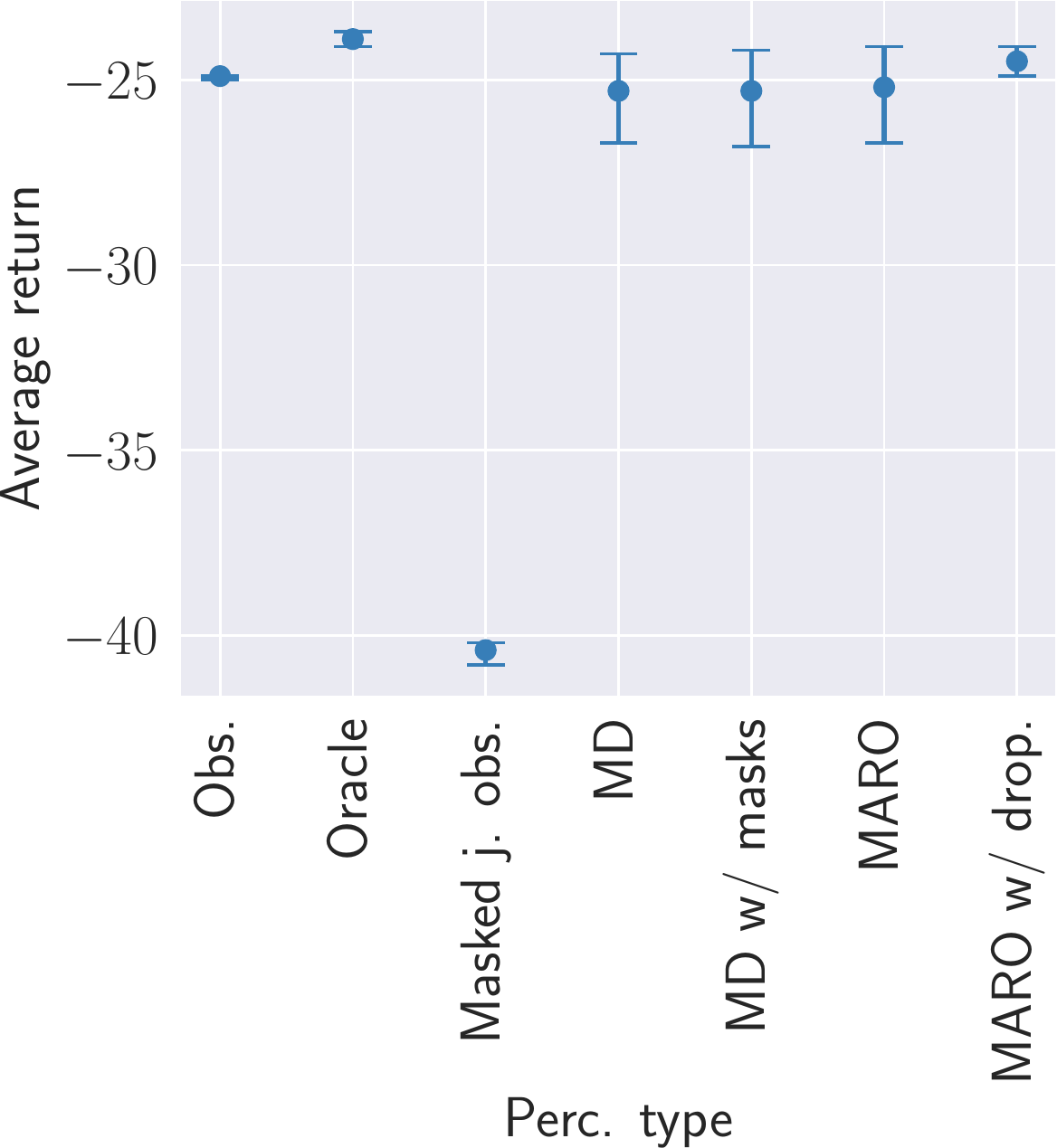}
        \caption{QMIX.}
    \end{subfigure}
    \begin{subfigure}[b]{0.24\textwidth}
        \centering
        \includegraphics[width=0.97\linewidth]{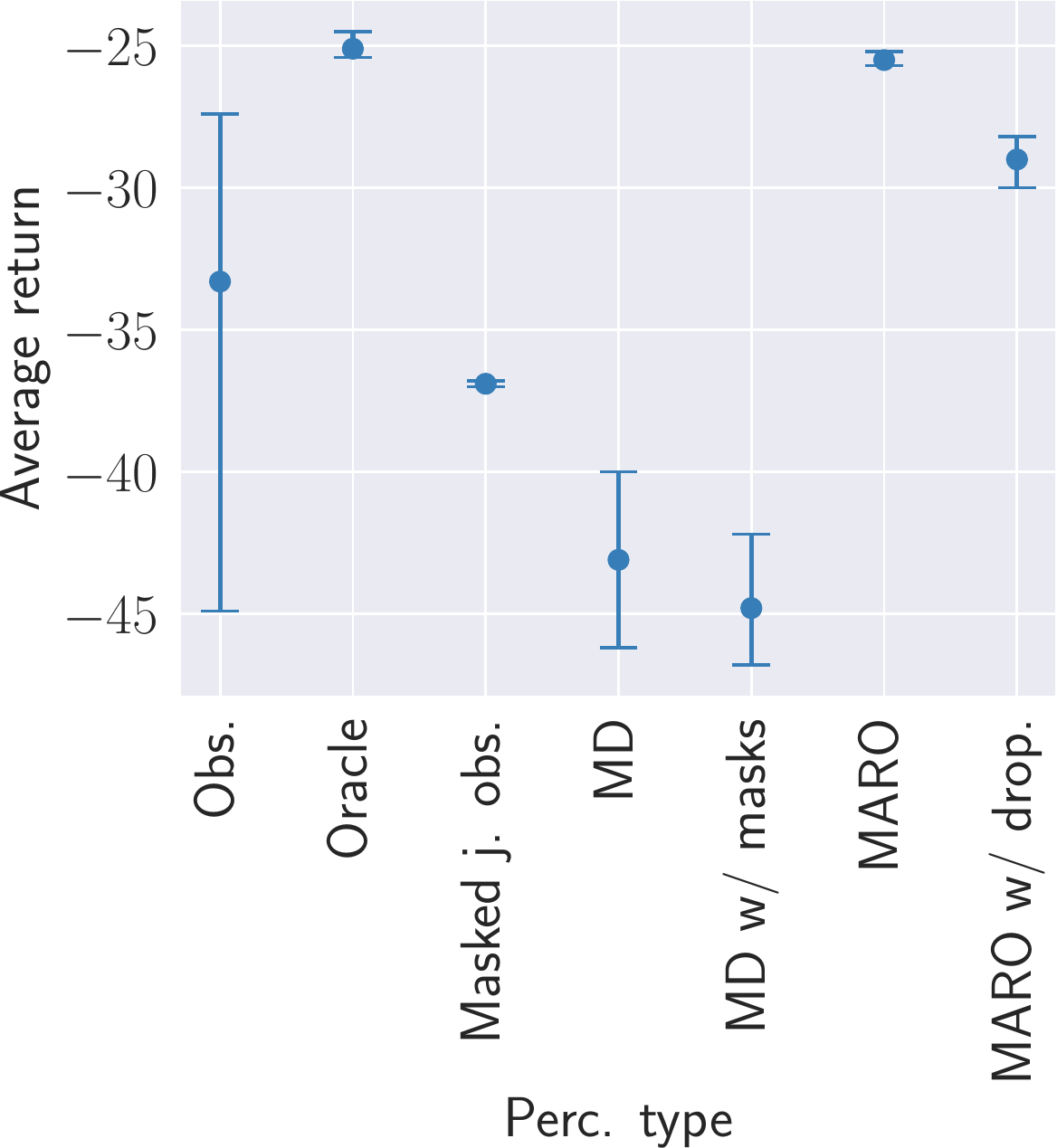}
        \caption{IPPO.}
    \end{subfigure}
    \begin{subfigure}[b]{0.24\textwidth}
        \centering
        \includegraphics[width=0.97\linewidth]{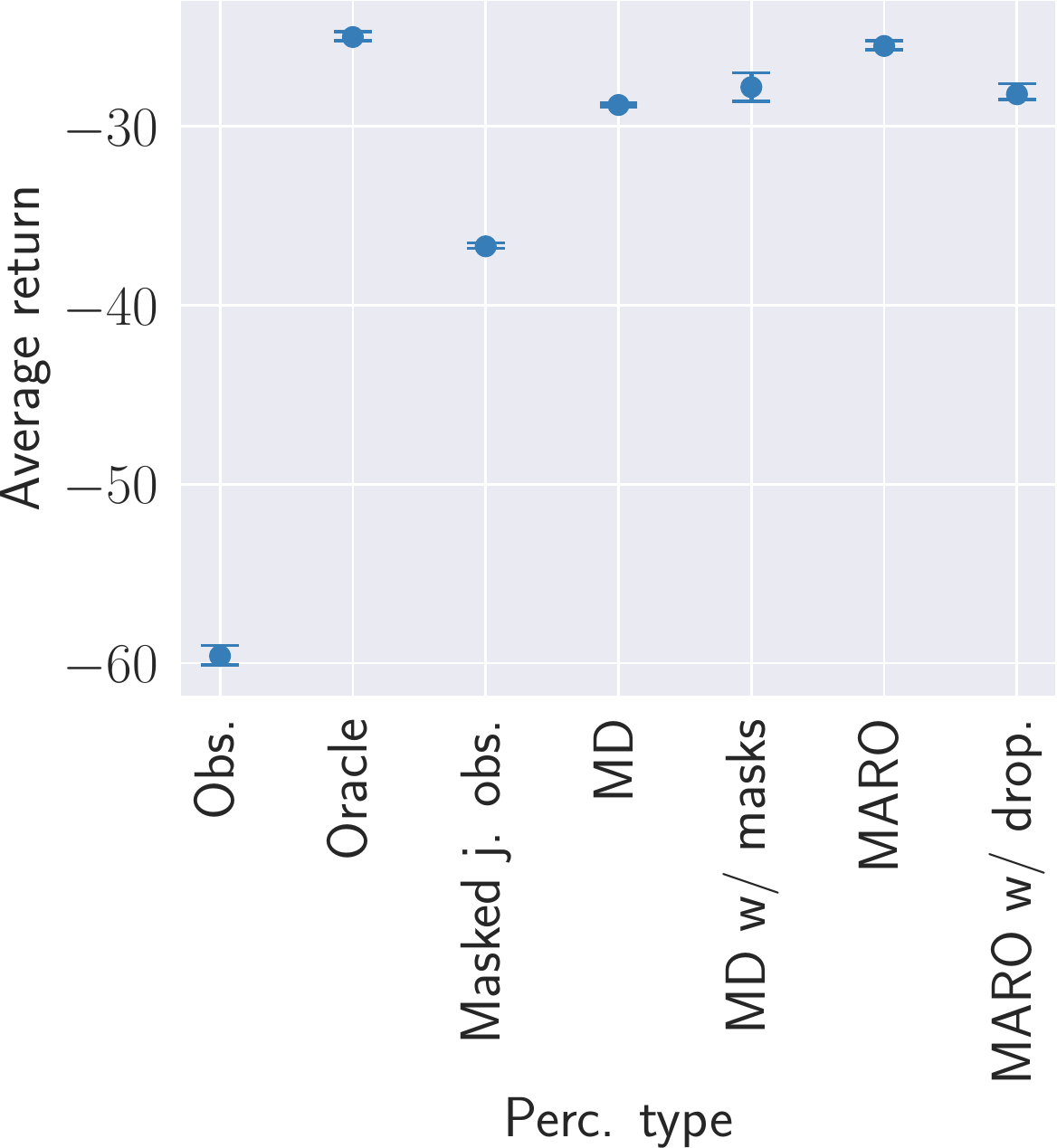}
        \caption{MAPPO.}
    \end{subfigure}
    \caption{(SpeakerListener) Mean episodic returns for $p_\textrm{asymmetric}$ during training.}
\end{figure}

\begin{figure}
    \centering
    \begin{subfigure}[b]{0.24\textwidth}
        \centering
        \includegraphics[width=0.97\linewidth]{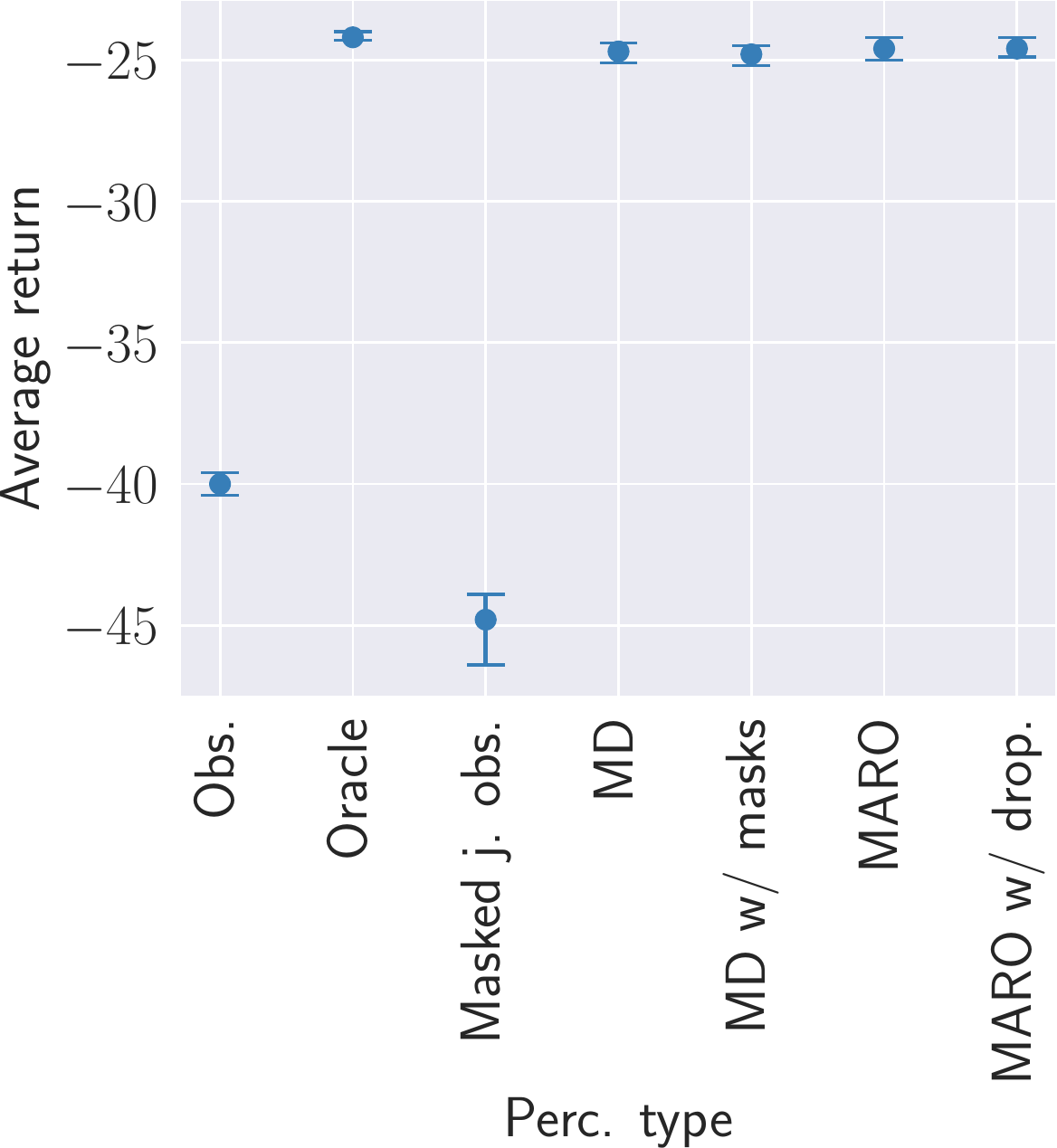}
        \caption{IQL.}
    \end{subfigure}
    \begin{subfigure}[b]{0.24\textwidth}
        \centering
        \includegraphics[width=0.97\linewidth]{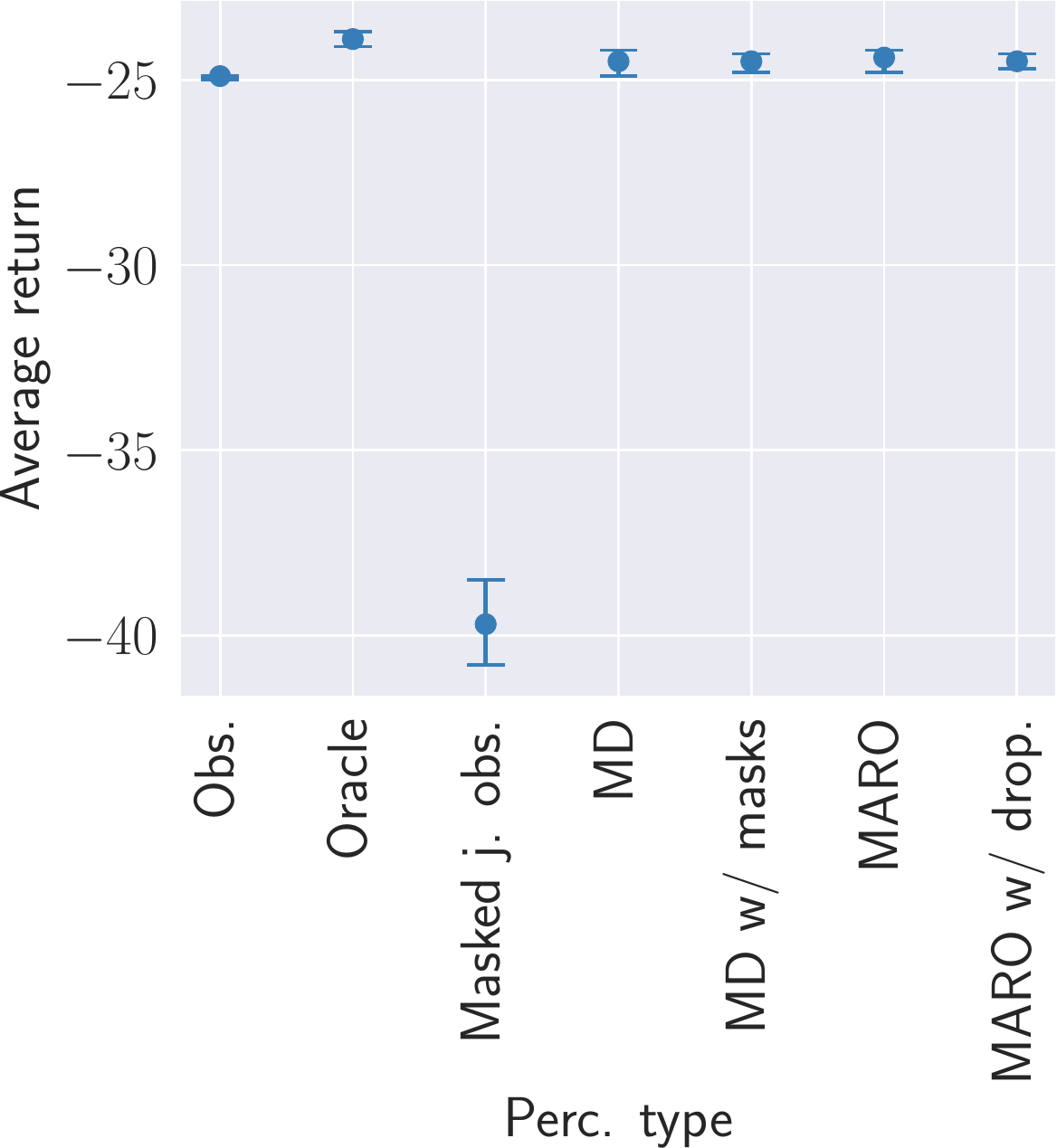}
        \caption{QMIX.}
    \end{subfigure}
    \begin{subfigure}[b]{0.24\textwidth}
        \centering
        \includegraphics[width=0.97\linewidth]{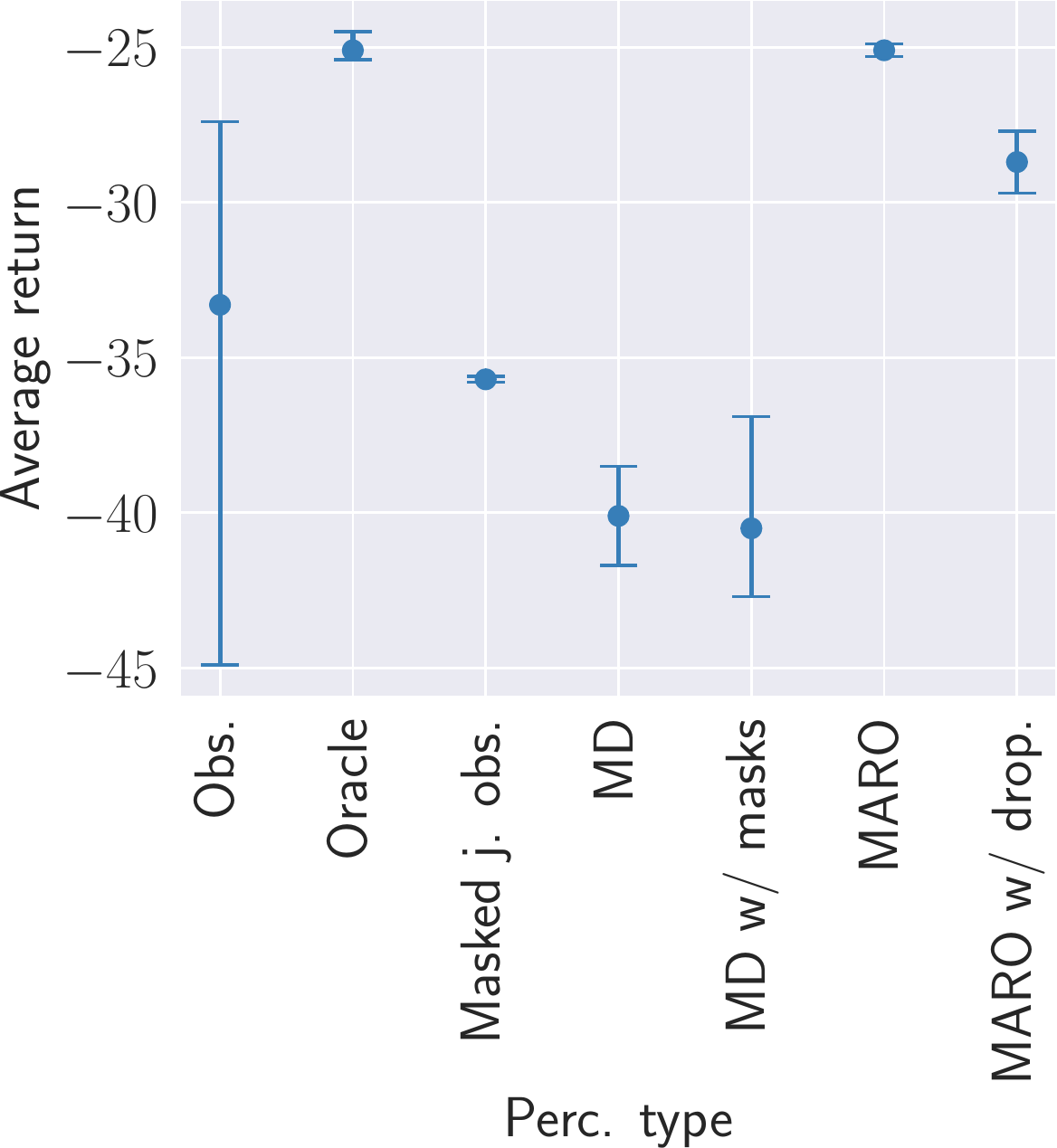}
        \caption{IPPO.}
    \end{subfigure}
    \begin{subfigure}[b]{0.24\textwidth}
        \centering
        \includegraphics[width=0.97\linewidth]{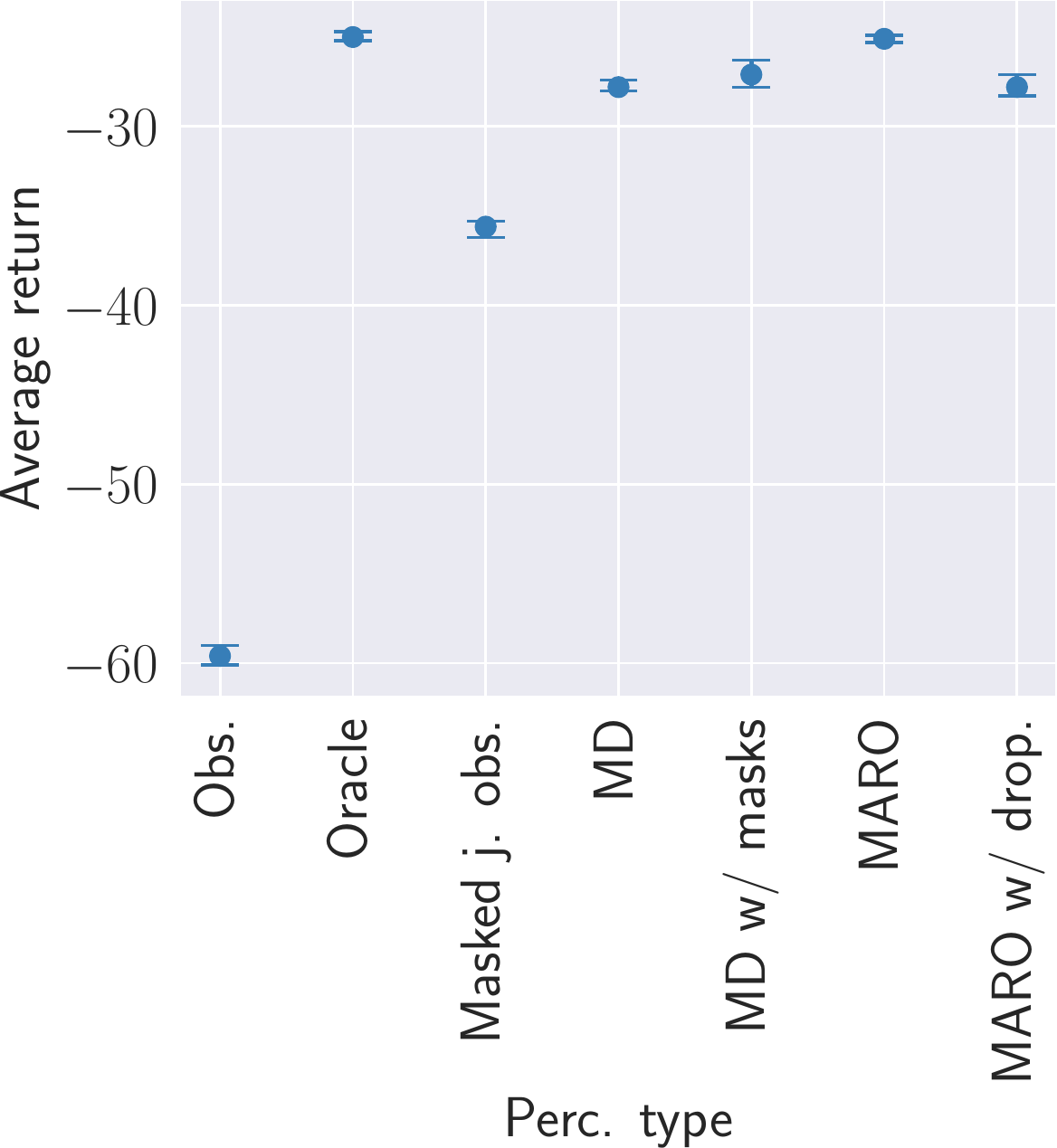}
        \caption{MAPPO.}
    \end{subfigure}
    \caption{(SpeakerListener) Mean episodic returns for $p_\textrm{dynamic}$ during training.}
\end{figure}

\clearpage

\begin{table}
\centering
\noindent
\caption{(HearSee) Mean episodic returns for $p_{\textrm{asymmetric}}$ at execution time.}
\vspace{0.1cm}
\resizebox{\linewidth}{!}{%
\begin{tabular}{c c c c c c c c }\toprule
\multicolumn{1}{c }{\textbf{}} & \multicolumn{7}{c }{\textbf{HearSee ($p_\textrm{asymmetric}$)}} \\  
\cmidrule(lr){2-8}
\multicolumn{1}{ l }{\textbf{Algorithm}} & \textbf{Obs.} & \textbf{Oracle} & \textbf{Masked j. obs.} & \textbf{MD} & \textbf{MD w/ masks} & \textbf{MARO} & \textbf{MARO w/ drop.} \\
\cmidrule{1-8}
\multicolumn{1}{ l }{IQL} & -114.5 \tiny{(-2.0,+1.6)} & -24.5 \tiny{(-0.6,+0.9)} & -65.6 \tiny{(-3.1,+4.5)} & -35.7 \tiny{(-1.1,+0.8)} & -34.9 \tiny{(-0.1,+0.1)} & -30.9 \tiny{(-1.8,+1.4)} & -30.8 \tiny{(-1.5,+0.9)} \\ \cmidrule{1-8}
\multicolumn{1}{ l }{QMIX} & -62.2 \tiny{(-1.9,+1.4)} & -23.6 \tiny{(-0.7,+0.7)} & -69.2 \tiny{(-1.6,+1.0)} & -30.5 \tiny{(-0.1,+0.2)} & -30.8 \tiny{(-0.1,+0.3)} & -29.9 \tiny{(-1.1,+1.2)} & -27.2 \tiny{(-0.1,+0.2)} \\ \cmidrule{1-8}
\multicolumn{1}{ l }{IPPO} & -114.1 \tiny{(-37.7,+27.7)} & -25.7 \tiny{(-0.2,+0.2)} & -77.5 \tiny{(-1.9,+2.0)} & -98.2 \tiny{(-25.1,+34.2)} & -100.6 \tiny{(-7.9,+11.5)} & -33.4 \tiny{(-2.0,+1.6)} & -30.2 \tiny{(-1.1,+1.5)} \\ \cmidrule{1-8}
\multicolumn{1}{ l }{MAPPO} & -70.5 \tiny{(-18.3,+10.1)} & -25.7 \tiny{(-0.2,+0.2)} & -79.1 \tiny{(-1.4,+1.1)} & -32.2 \tiny{(-1.1,+1.2)} & -30.4 \tiny{(-1.0,+1.0)} & -33.4 \tiny{(-2.0,+1.6)} & -29.0 \tiny{(-1.4,+1.3)} \\ 
\bottomrule
\end{tabular}
}
\end{table}

\begin{table}
\centering
\noindent
\caption{(HearSee) Mean episodic returns for $p_{\textrm{dynamic}}$ at execution time.}
\vspace{0.1cm}
\resizebox{\linewidth}{!}{%
\begin{tabular}{c c c c c c c c }\toprule
\multicolumn{1}{c }{\textbf{}} & \multicolumn{7}{c }{\textbf{HearSee ($p_\textrm{dynamic}$)}} \\  
\cmidrule(lr){2-8}
\multicolumn{1}{ l }{\textbf{Algorithm}} & \textbf{Obs.} & \textbf{Oracle} & \textbf{Masked j. obs.} & \textbf{MD} & \textbf{MD w/ masks} & \textbf{MARO} & \textbf{MARO w/ drop.} \\
\cmidrule{1-8}
\multicolumn{1}{ l }{IQL} & -114.5 \tiny{(-2.0,+1.6)} & -24.5 \tiny{(-0.6,+0.9)} & -52.1 \tiny{(-1.7,+3.0)} & -33.0 \tiny{(-0.5,+0.7)} & -32.1 \tiny{(-0.6,+0.9)} & -28.5 \tiny{(-0.5,+0.9)} & -28.8 \tiny{(-0.5,+0.5)} \\ \cmidrule{1-8}
\multicolumn{1}{ l }{QMIX} & -62.2 \tiny{(-1.9,+1.4)} & -23.6 \tiny{(-0.7,+0.7)} & -55.1 \tiny{(-2.6,+2.3)} & -29.1 \tiny{(-0.3,+0.2)} & -29.0 \tiny{(-0.2,+0.3)} & -27.5 \tiny{(-1.2,+1.0)} & -26.7 \tiny{(-0.4,+0.3)} \\ \cmidrule{1-8}
\multicolumn{1}{ l }{IPPO} & -114.1 \tiny{(-37.7,+27.7)} & -25.7 \tiny{(-0.2,+0.2)} & -52.5 \tiny{(-2.2,+3.4)} & -84.8 \tiny{(-27.8,+23.7)} & -82.1 \tiny{(-13.5,+9.8)} & -27.5 \tiny{(-1.2,+1.6)} & -27.7 \tiny{(-0.4,+0.7)} \\ \cmidrule{1-8}
\multicolumn{1}{ l }{MAPPO} & -70.5 \tiny{(-18.3,+10.1)} & -25.7 \tiny{(-0.2,+0.2)} & -53.3 \tiny{(-1.2,+1.3)} & -28.2 \tiny{(-0.6,+0.9)} & -26.8 \tiny{(-0.4,+0.6)} & -27.5 \tiny{(-1.2,+1.6)} & -26.6 \tiny{(-1.0,+1.6)} \\
\bottomrule
\end{tabular}
}
\end{table}

\begin{figure}
    \centering
    \begin{subfigure}[b]{0.24\textwidth}
        \centering
        \includegraphics[width=0.97\linewidth]{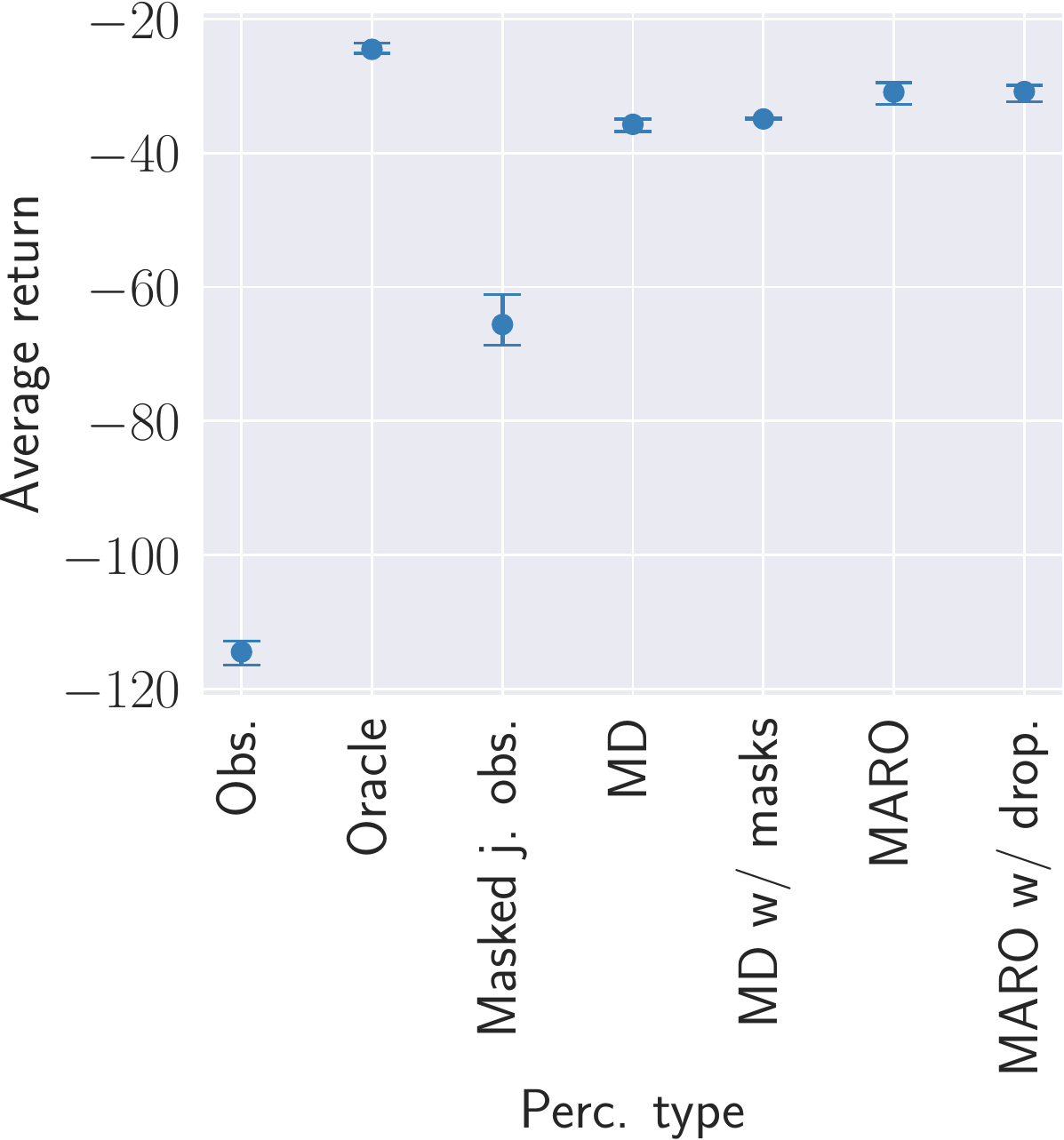}
        \caption{IQL.}
    \end{subfigure}
    \begin{subfigure}[b]{0.24\textwidth}
        \centering
        \includegraphics[width=0.97\linewidth]{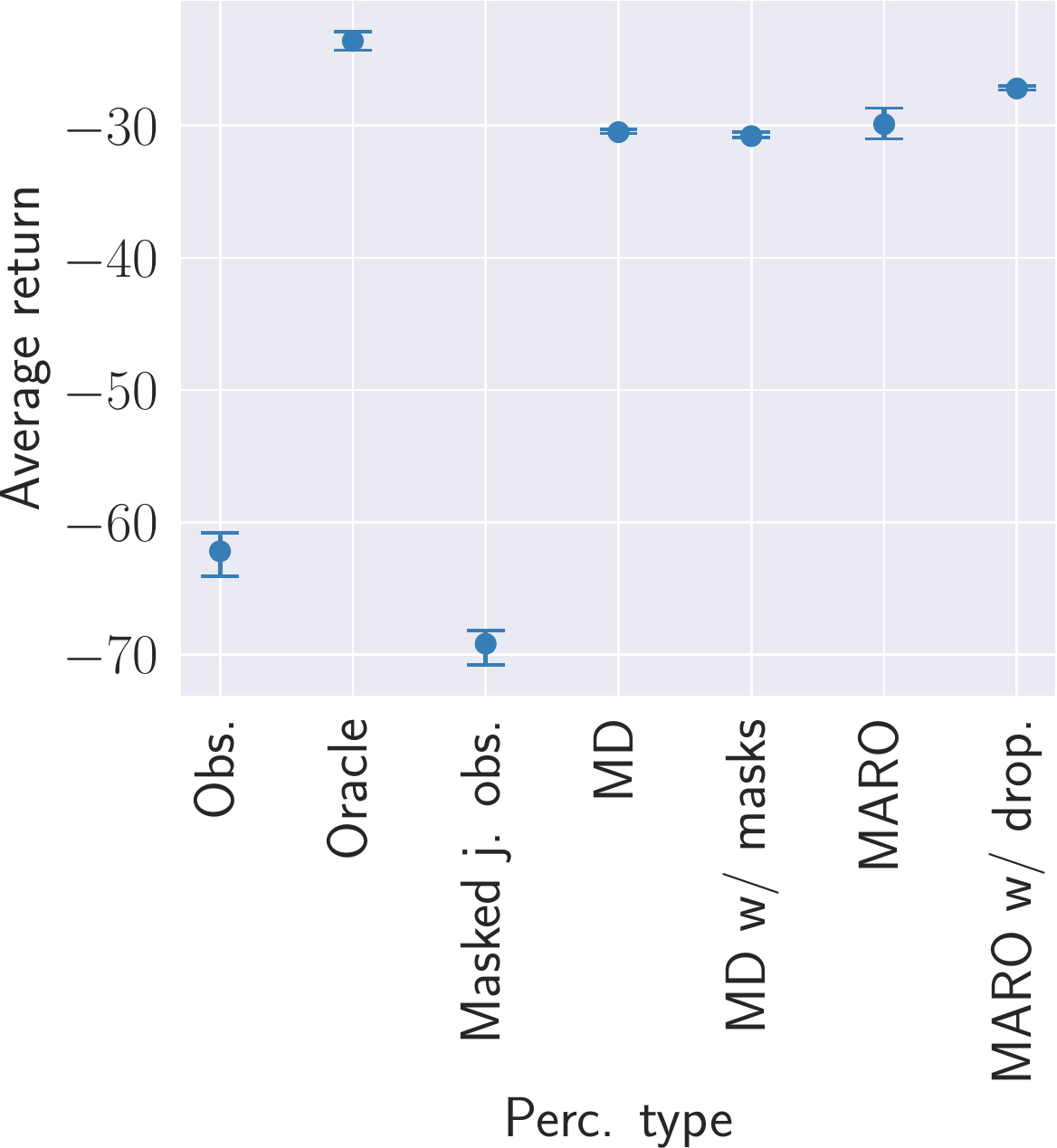}
        \caption{QMIX.}
    \end{subfigure}
    \begin{subfigure}[b]{0.24\textwidth}
        \centering
        \includegraphics[width=0.97\linewidth]{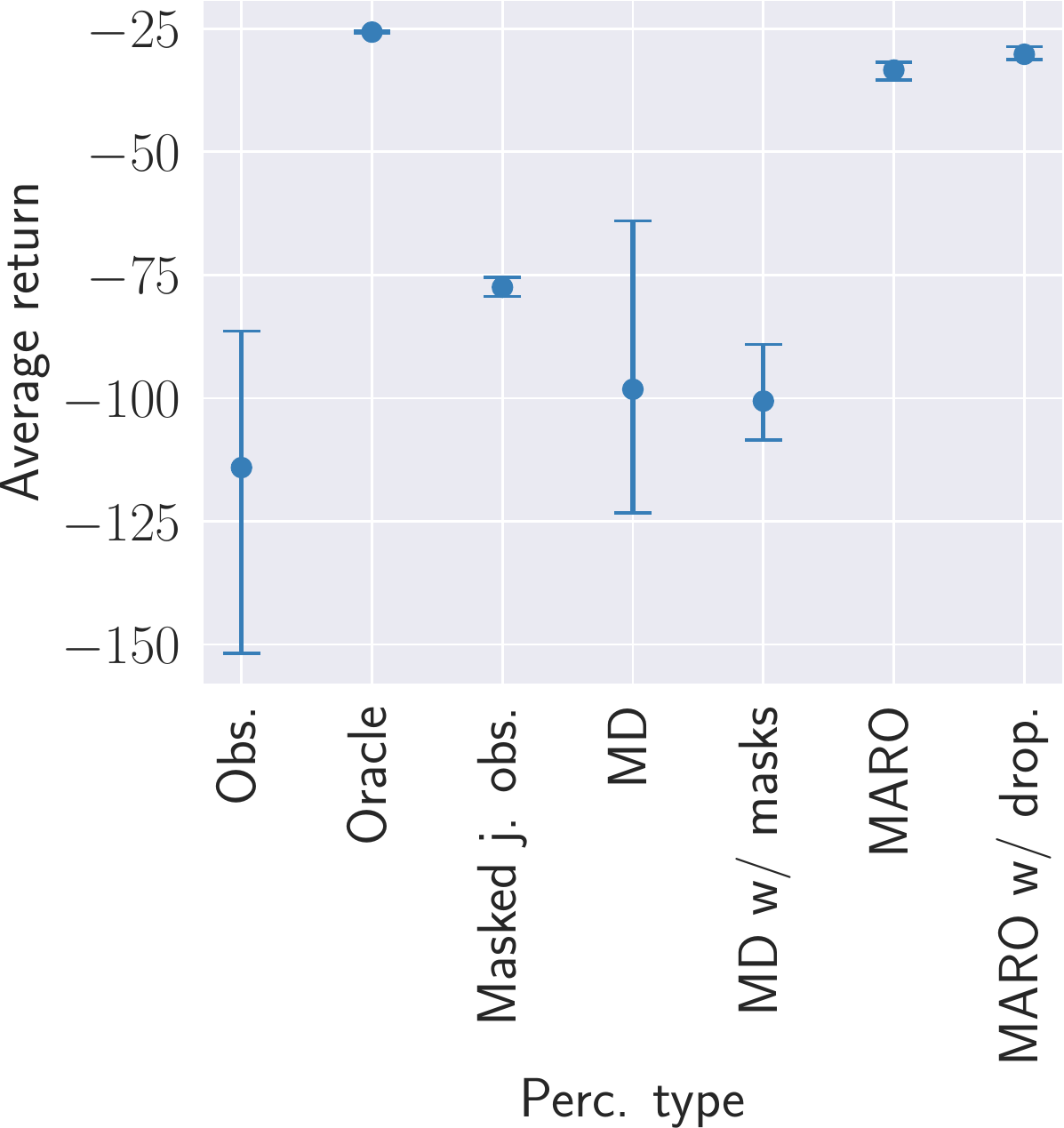}
        \caption{IPPO.}
    \end{subfigure}
    \begin{subfigure}[b]{0.24\textwidth}
        \centering
        \includegraphics[width=0.97\linewidth]{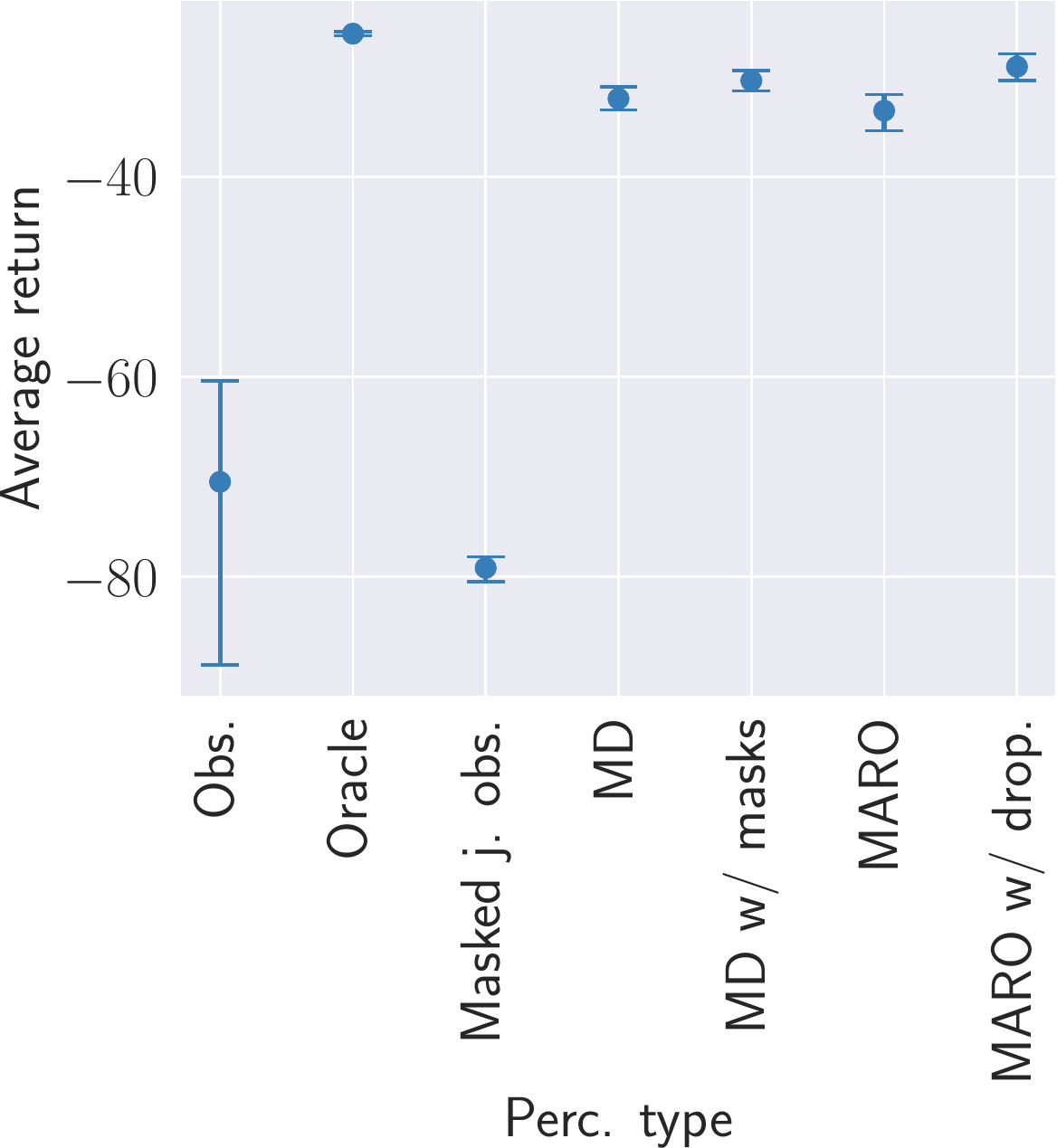}
        \caption{MAPPO.}
    \end{subfigure}
    \caption{(HearSee) Mean episodic returns for $p_\textrm{asymmetric}$ during training.}
\end{figure}

\begin{figure}
    \centering
    \begin{subfigure}[b]{0.24\textwidth}
        \centering
        \includegraphics[width=0.97\linewidth]{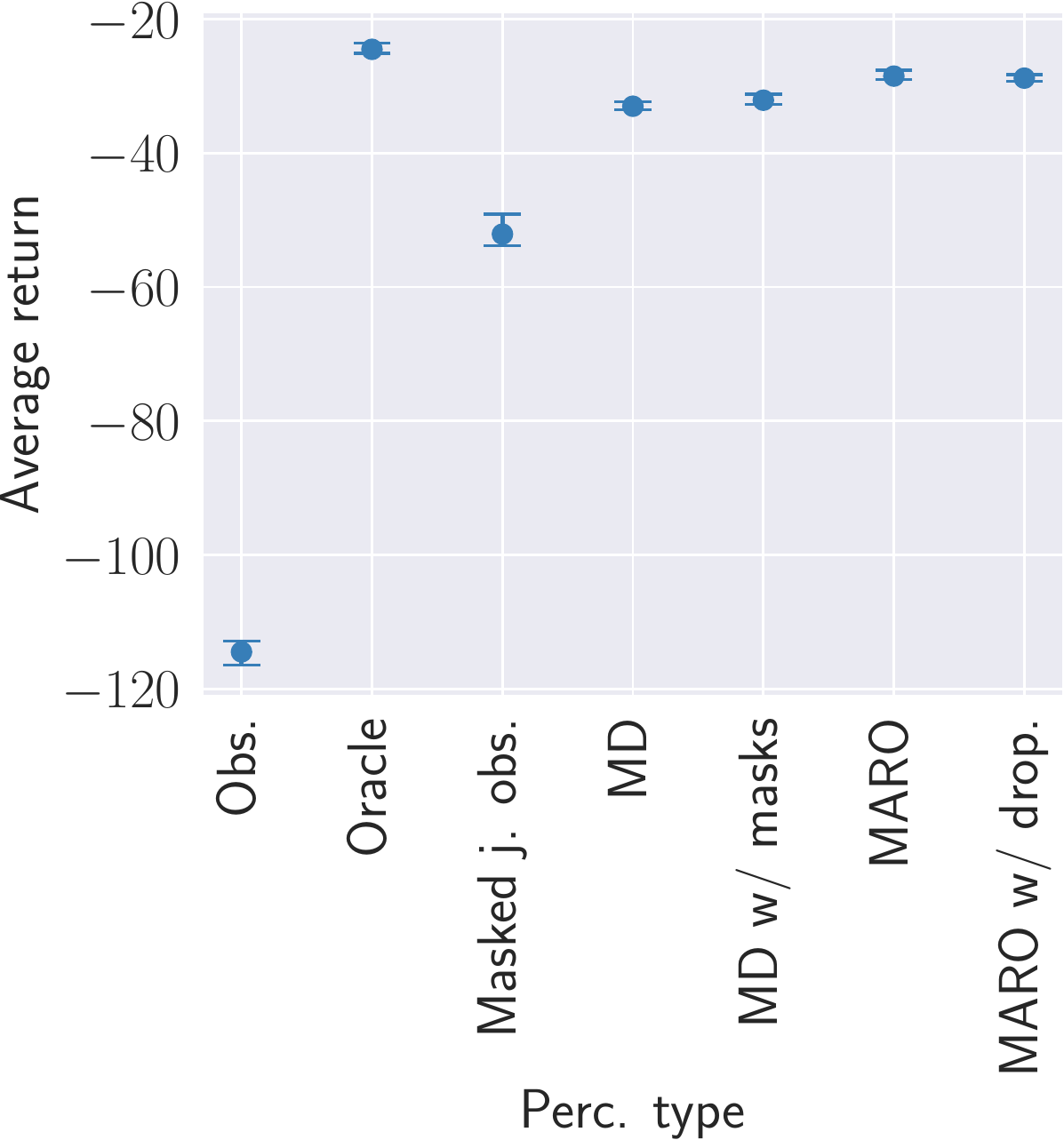}
        \caption{IQL.}
    \end{subfigure}
    \begin{subfigure}[b]{0.24\textwidth}
        \centering
        \includegraphics[width=0.97\linewidth]{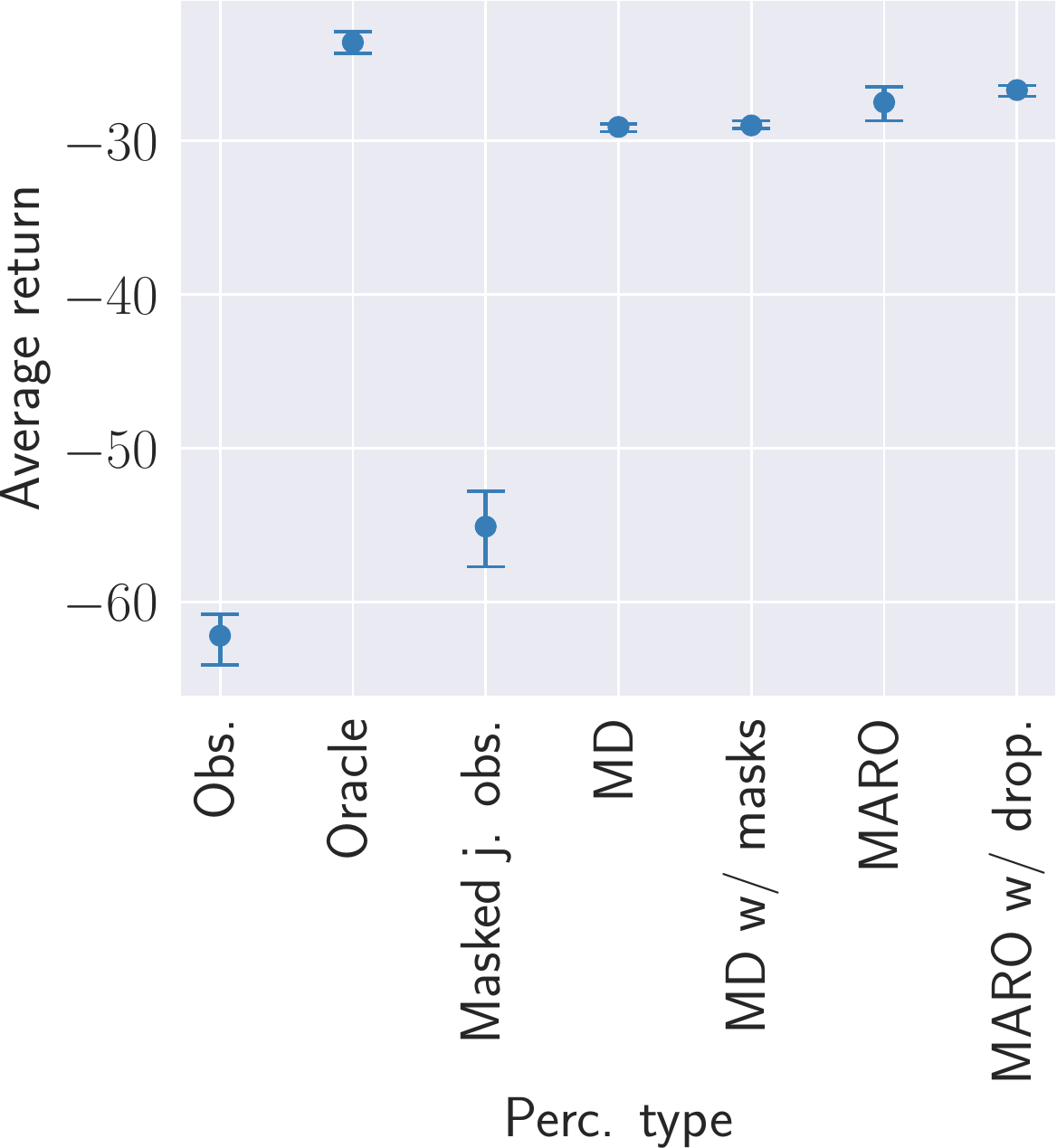}
        \caption{QMIX.}
    \end{subfigure}
    \begin{subfigure}[b]{0.24\textwidth}
        \centering
        \includegraphics[width=0.97\linewidth]{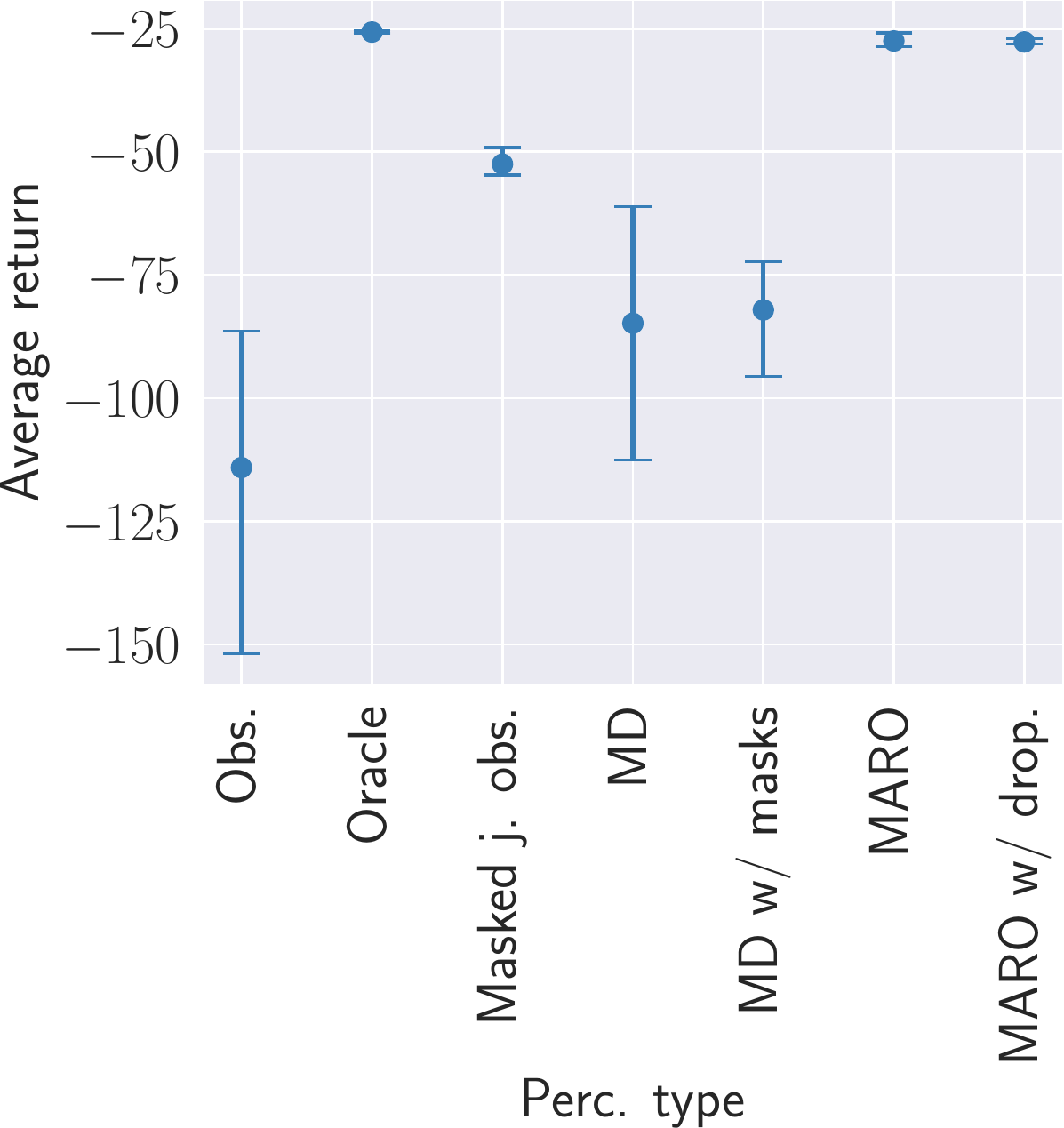}
        \caption{IPPO.}
    \end{subfigure}
    \begin{subfigure}[b]{0.24\textwidth}
        \centering
        \includegraphics[width=0.97\linewidth]{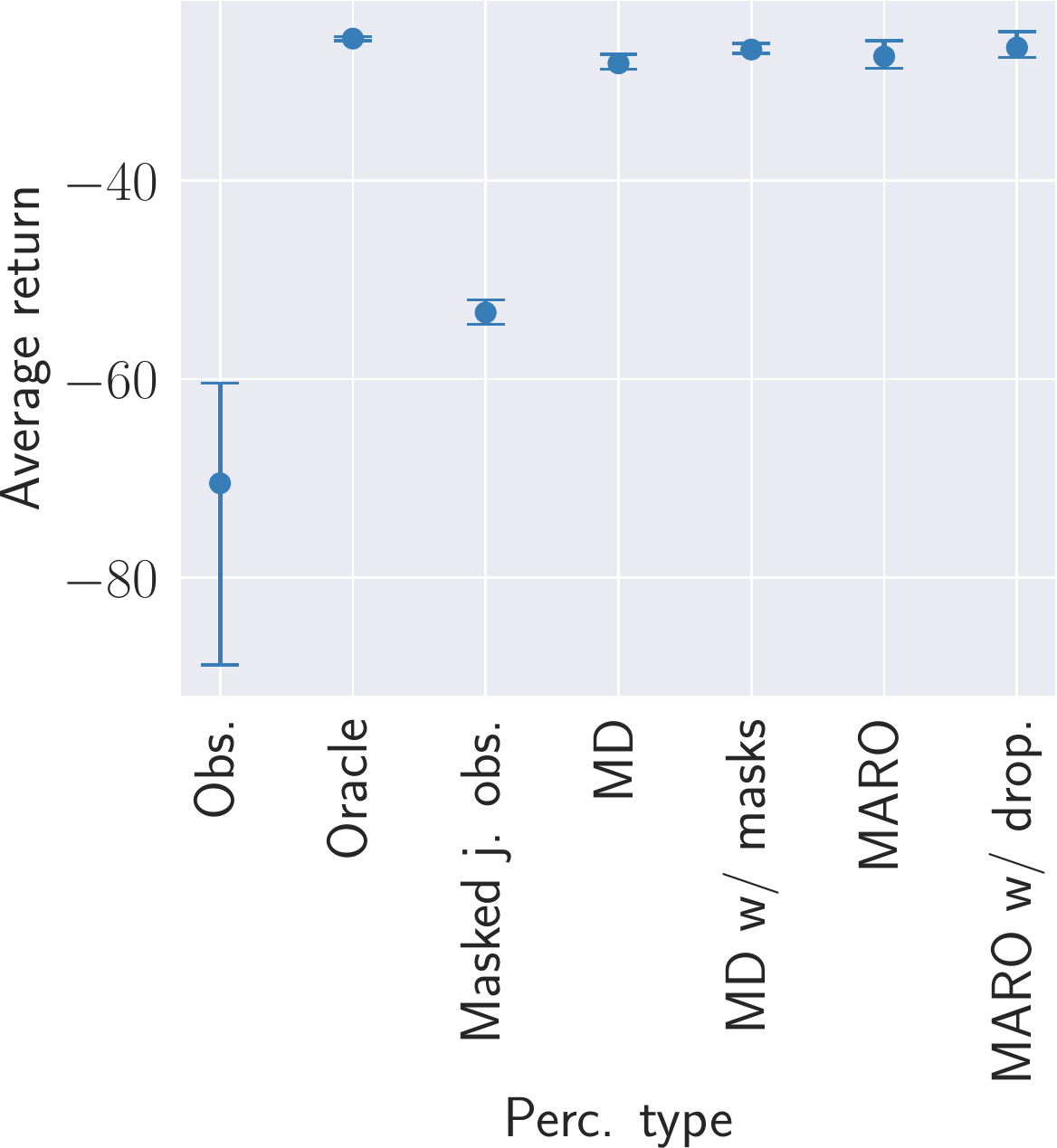}
        \caption{MAPPO.}
    \end{subfigure}
    \caption{(HearSee) Mean episodic returns for $p_\textrm{dynamic}$ during training.}
\end{figure}

\clearpage

\begin{table}
\centering
\noindent
\caption{(SpreadXY-2) Mean episodic returns for $p_{\textrm{asymmetric}}$ at execution time.}
\vspace{0.1cm}
\resizebox{\linewidth}{!}{%
\begin{tabular}{c c c c c c c c }\toprule
\multicolumn{1}{c }{\textbf{}} & \multicolumn{7}{c }{\textbf{SpreadXY-2 ($p_\textrm{asymmetric}$)}} \\  
\cmidrule(lr){2-8}
\multicolumn{1}{ l }{\textbf{Algorithm}} & \textbf{Obs.} & \textbf{Oracle} & \textbf{Masked j. obs.} & \textbf{MD} & \textbf{MD w/ masks} & \textbf{MARO} & \textbf{MARO w/ drop.} \\
\cmidrule{1-8}
\multicolumn{1}{ l }{IQL} & -199.6 \tiny{(-0.9,+0.9)} & -139.4 \tiny{(-0.5,+0.6)} & -200.2 \tiny{(-0.5,+0.4)} & -164.1 \tiny{(-1.1,+0.7)} & -160.5 \tiny{(-0.8,+0.6)} & -149.4 \tiny{(-0.2,+0.3)} & -158.3 \tiny{(-0.8,+0.6)} \\ \cmidrule{1-8}
\multicolumn{1}{ l }{QMIX} & -177.8 \tiny{(-7.6,+4.1)} & -138.6 \tiny{(-0.4,+0.4)} & -197.9 \tiny{(-1.0,+0.9)} & -157.8 \tiny{(-0.5,+0.2)} & -154.3 \tiny{(-1.3,+1.1)} & -147.3 \tiny{(-0.3,+0.3)} & -152.7 \tiny{(-0.8,+0.8)} \\ \cmidrule{1-8}
\multicolumn{1}{ l }{IPPO} & -235.6 \tiny{(-0.6,+0.6)} & -160.8 \tiny{(-2.8,+1.4)} & -205.4 \tiny{(-3.5,+2.4)} & -186.2 \tiny{(-8.7,+5.5)} & -175.0 \tiny{(-1.1,+1.3)} & -161.1 \tiny{(-2.0,+1.1)} & -166.8 \tiny{(-0.2,+0.3)} \\ \cmidrule{1-8}
\multicolumn{1}{ l }{MAPPO} & -212.6 \tiny{(-13.7,+24.5)} & -160.8 \tiny{(-1.8,+3.1)} & -209.4 \tiny{(-1.9,+1.1)} & -184.0 \tiny{(-3.2,+1.9)} & -165.7 \tiny{(-3.7,+2.4)} & -161.2 \tiny{(-0.4,+0.3)} & -160.6 \tiny{(-1.0,+0.8)} \\
\bottomrule
\end{tabular}
}
\end{table}

\begin{table}
\centering
\noindent
\caption{(SpreadXY-2) Mean episodic returns for $p_{\textrm{dynamic}}$ at execution time.}
\vspace{0.1cm}
\resizebox{\linewidth}{!}{%
\begin{tabular}{c c c c c c c c }\toprule
\multicolumn{1}{c }{\textbf{}} & \multicolumn{7}{c }{\textbf{SpreadXY-2 ($p_\textrm{dynamic}$)}} \\  
\cmidrule(lr){2-8}
\multicolumn{1}{ l }{\textbf{Algorithm}} & \textbf{Obs.} & \textbf{Oracle} & \textbf{Masked j. obs.} & \textbf{MD} & \textbf{MD w/ masks} & \textbf{MARO} & \textbf{MARO w/ drop.} \\
\cmidrule{1-8}
\multicolumn{1}{ l }{IQL} & -199.6 \tiny{(-0.9,+0.9)} & -139.4 \tiny{(-0.5,+0.6)} & -196.7 \tiny{(-2.3,+2.2)} & -161.2 \tiny{(-0.4,+0.3)} & -157.6 \tiny{(-0.8,+0.8)} & -145.3 \tiny{(-1.0,+1.5)} & -155.7 \tiny{(-0.4,+0.6)} \\ \cmidrule{1-8}
\multicolumn{1}{ l }{QMIX} & -177.8 \tiny{(-7.6,+4.1)} & -138.6 \tiny{(-0.4,+0.4)} & -193.7 \tiny{(-1.4,+2.1)} & -155.8 \tiny{(-0.8,+0.4)} & -152.6 \tiny{(-0.7,+1.2)} & -143.5 \tiny{(-1.1,+1.9)} & -150.2 \tiny{(-0.8,+0.8)} \\ \cmidrule{1-8}
\multicolumn{1}{ l }{IPPO} & -235.6 \tiny{(-0.6,+0.6)} & -160.8 \tiny{(-2.8,+1.4)} & -200.8 \tiny{(-2.2,+2.5)} & -183.5 \tiny{(-7.9,+4.7)} & -171.3 \tiny{(-1.4,+0.7)} & -157.3 \tiny{(-0.5,+0.3)} & -166.7 \tiny{(-1.2,+0.9)} \\ \cmidrule{1-8}
\multicolumn{1}{ l }{MAPPO} & -212.6 \tiny{(-13.7,+24.5)} & -160.8 \tiny{(-1.8,+3.1)} & -203.3 \tiny{(-2.4,+1.7)} & -179.6 \tiny{(-3.3,+2.3)} & -162.9 \tiny{(-2.5,+2.1)} & -157.7 \tiny{(-0.3,+0.3)} & -158.8 \tiny{(-0.6,+0.4)} \\
\bottomrule
\end{tabular}
}
\end{table}

\begin{figure}
    \centering
    \begin{subfigure}[b]{0.24\textwidth}
        \centering
        \includegraphics[width=0.97\linewidth]{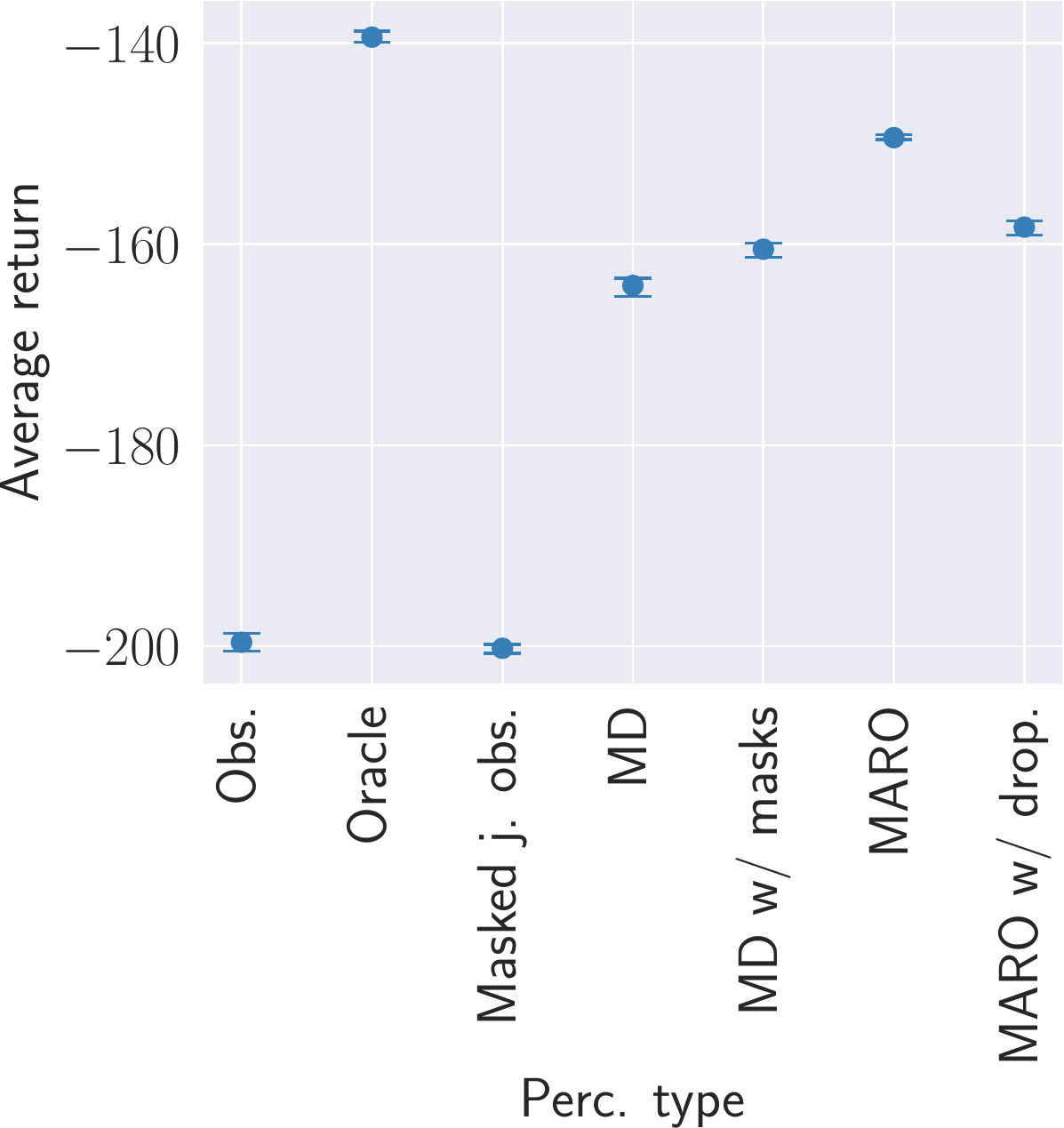}
        \caption{IQL.}
    \end{subfigure}
    \begin{subfigure}[b]{0.24\textwidth}
        \centering
        \includegraphics[width=0.97\linewidth]{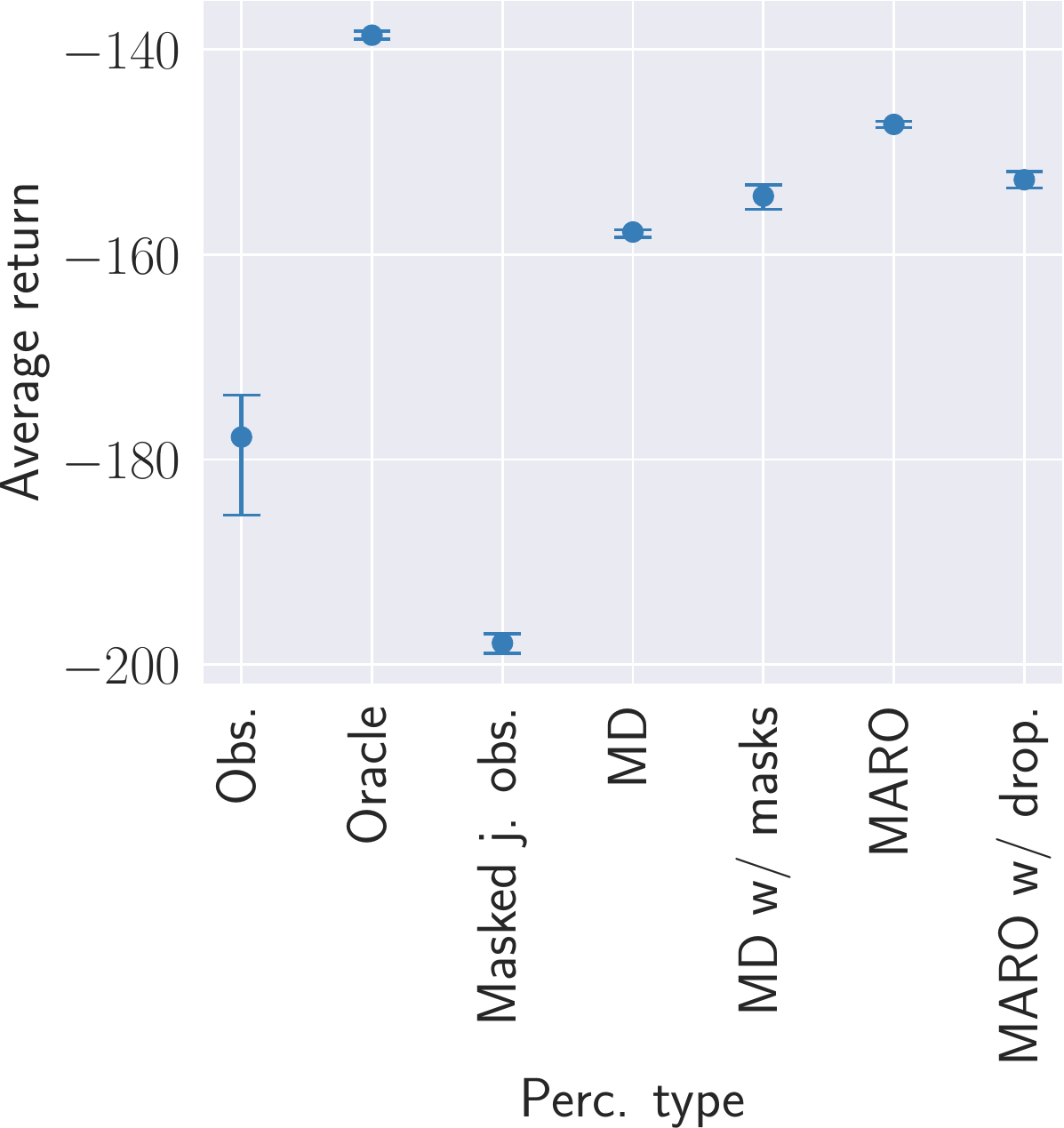}
        \caption{QMIX.}
    \end{subfigure}
    \begin{subfigure}[b]{0.24\textwidth}
        \centering
        \includegraphics[width=0.97\linewidth]{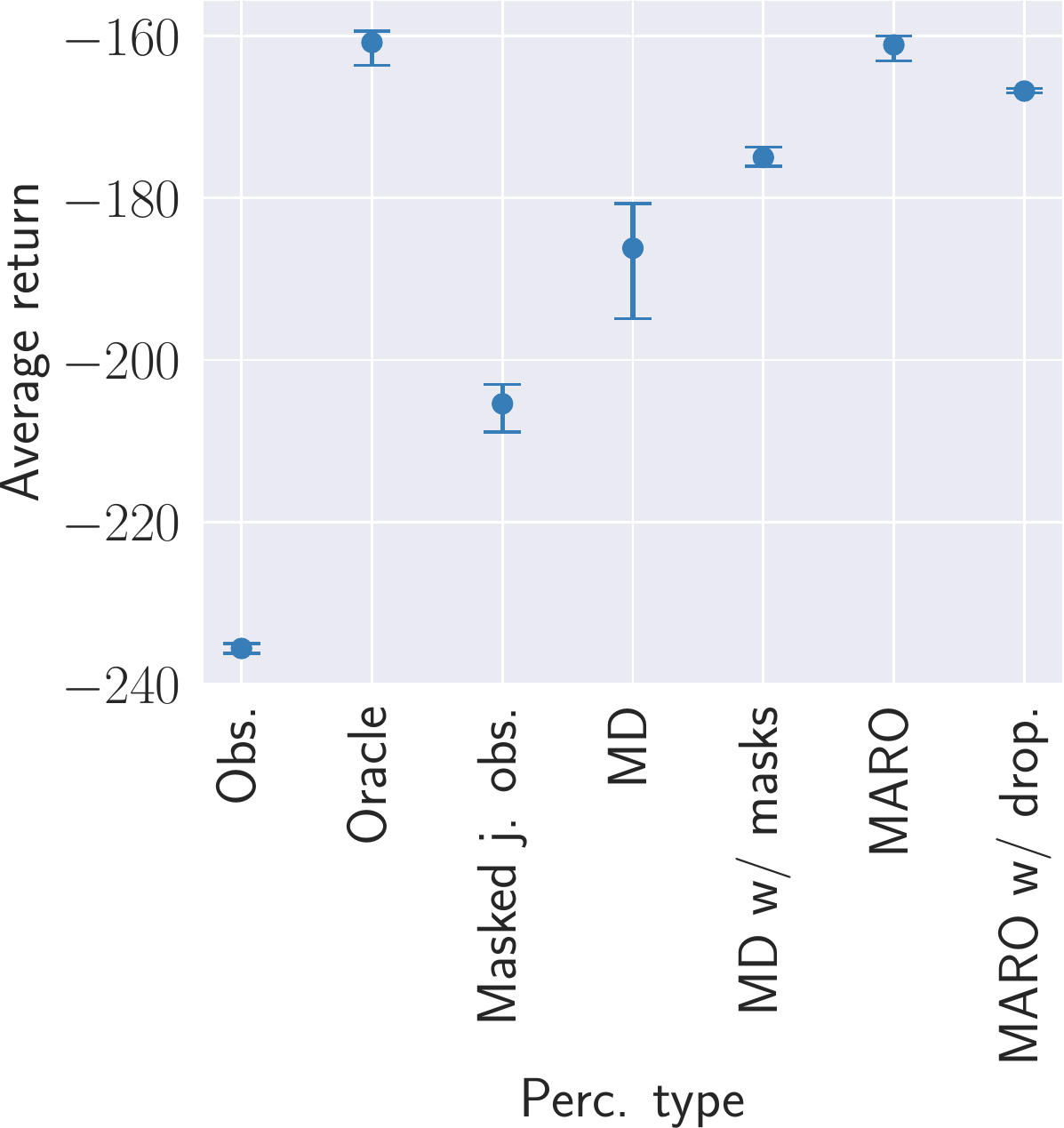}
        \caption{IPPO.}
    \end{subfigure}
    \begin{subfigure}[b]{0.24\textwidth}
        \centering
        \includegraphics[width=0.97\linewidth]{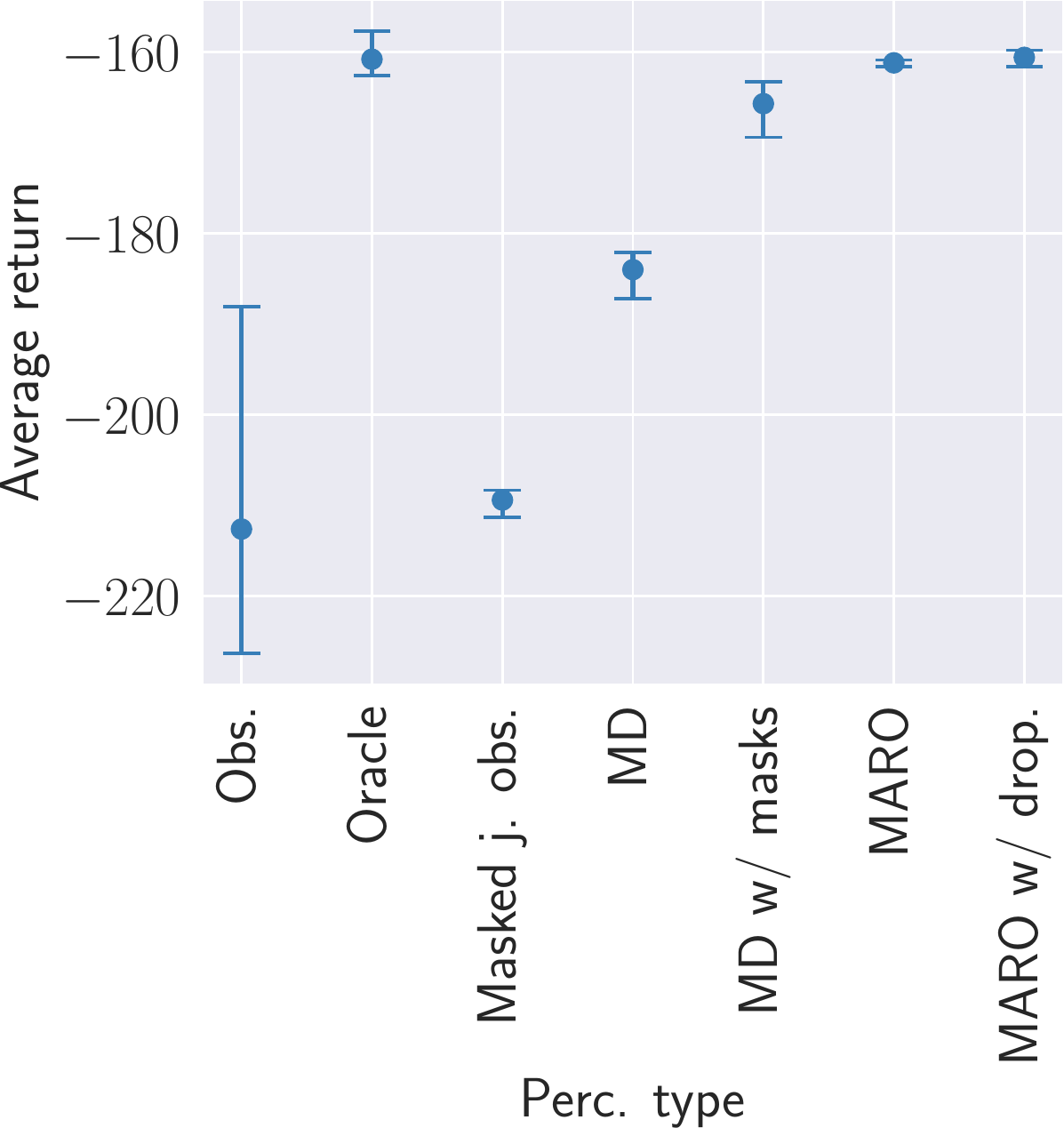}
        \caption{MAPPO.}
    \end{subfigure}
    \caption{(SpreadXY-2) Mean episodic returns for $p_\textrm{asymmetric}$ during training.}
\end{figure}

\begin{figure}
    \centering
    \begin{subfigure}[b]{0.24\textwidth}
        \centering
        \includegraphics[width=0.97\linewidth]{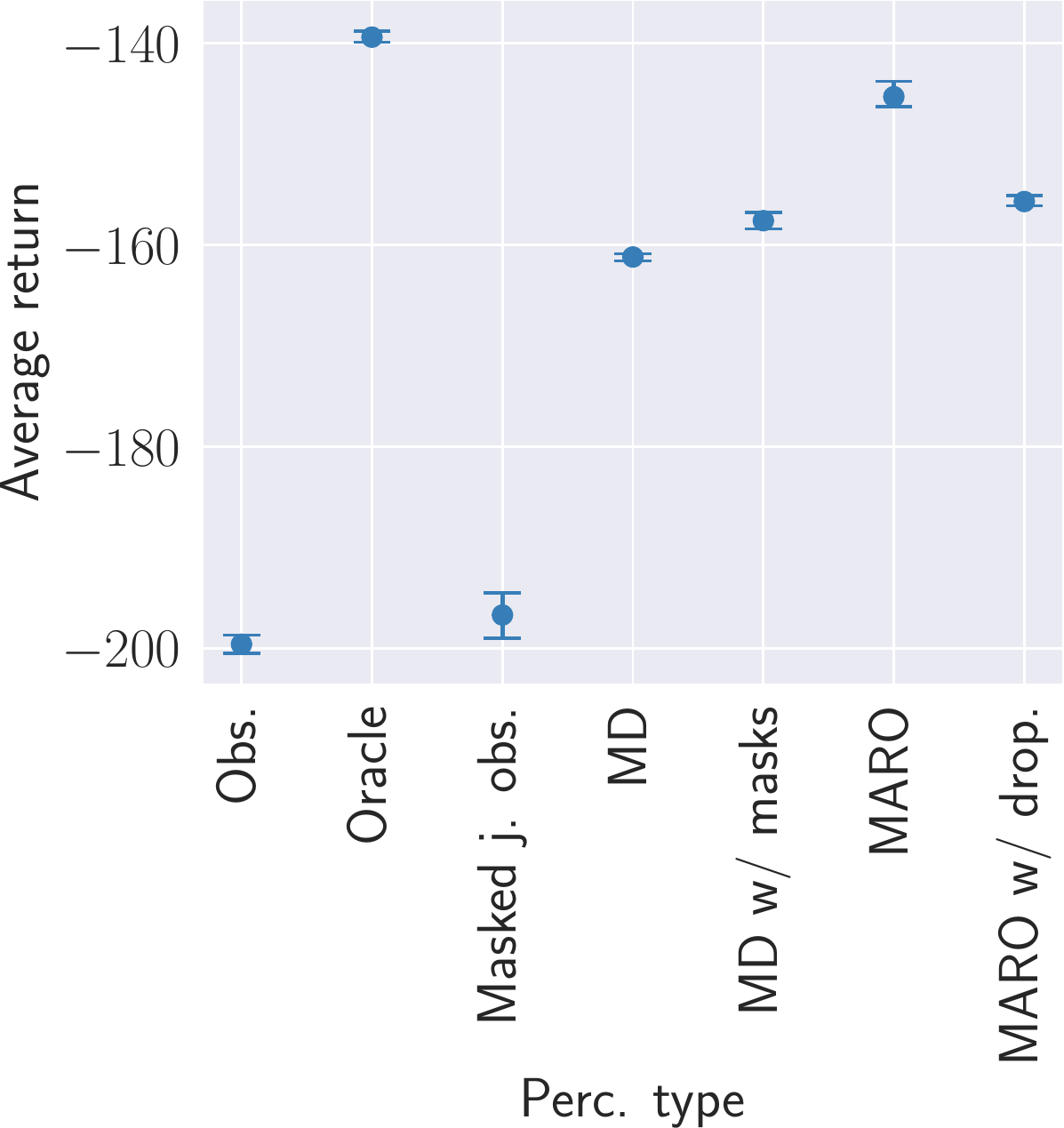}
        \caption{IQL.}
    \end{subfigure}
    \begin{subfigure}[b]{0.24\textwidth}
        \centering
        \includegraphics[width=0.97\linewidth]{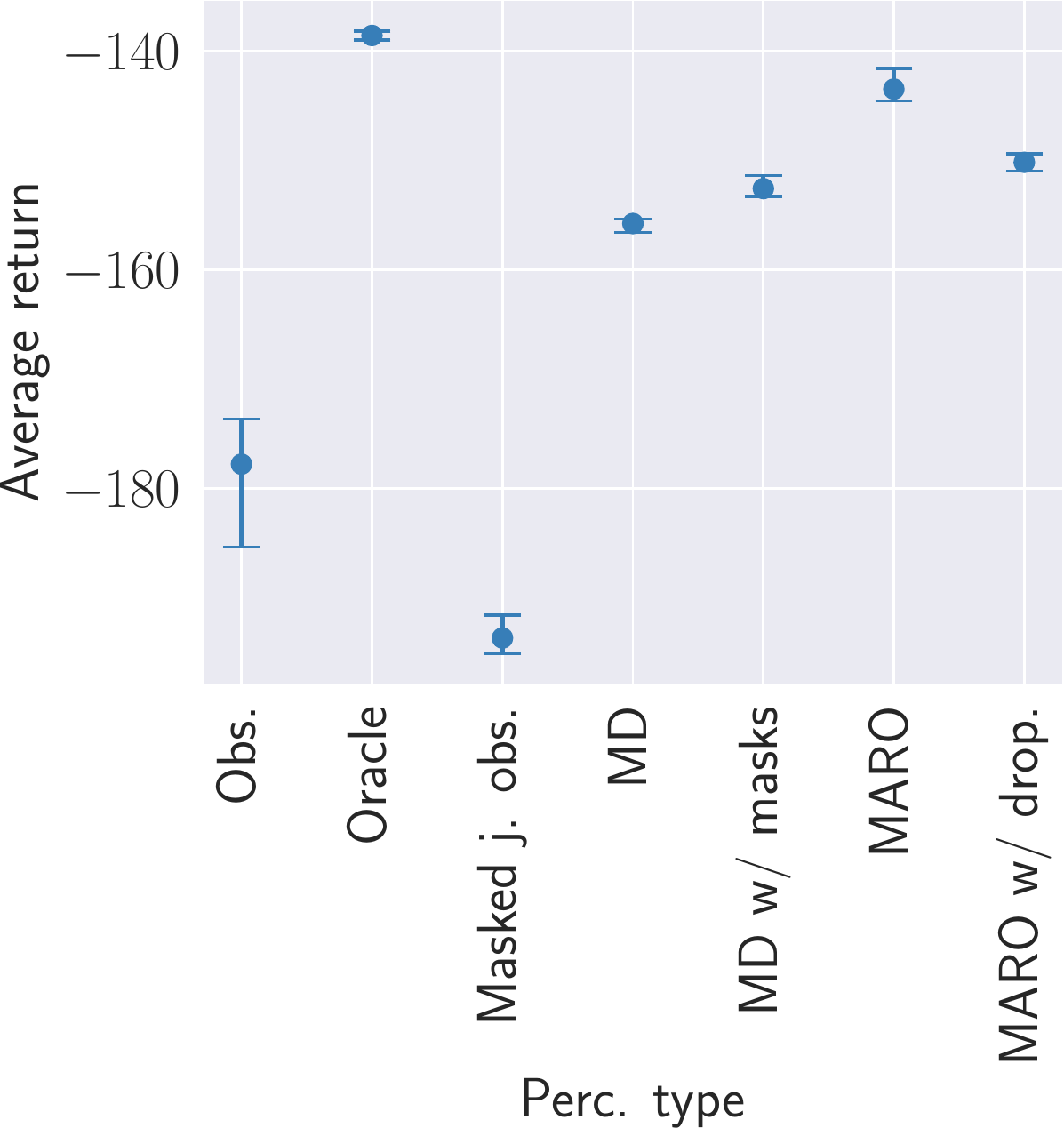}
        \caption{QMIX.}
    \end{subfigure}
    \begin{subfigure}[b]{0.24\textwidth}
        \centering
        \includegraphics[width=0.97\linewidth]{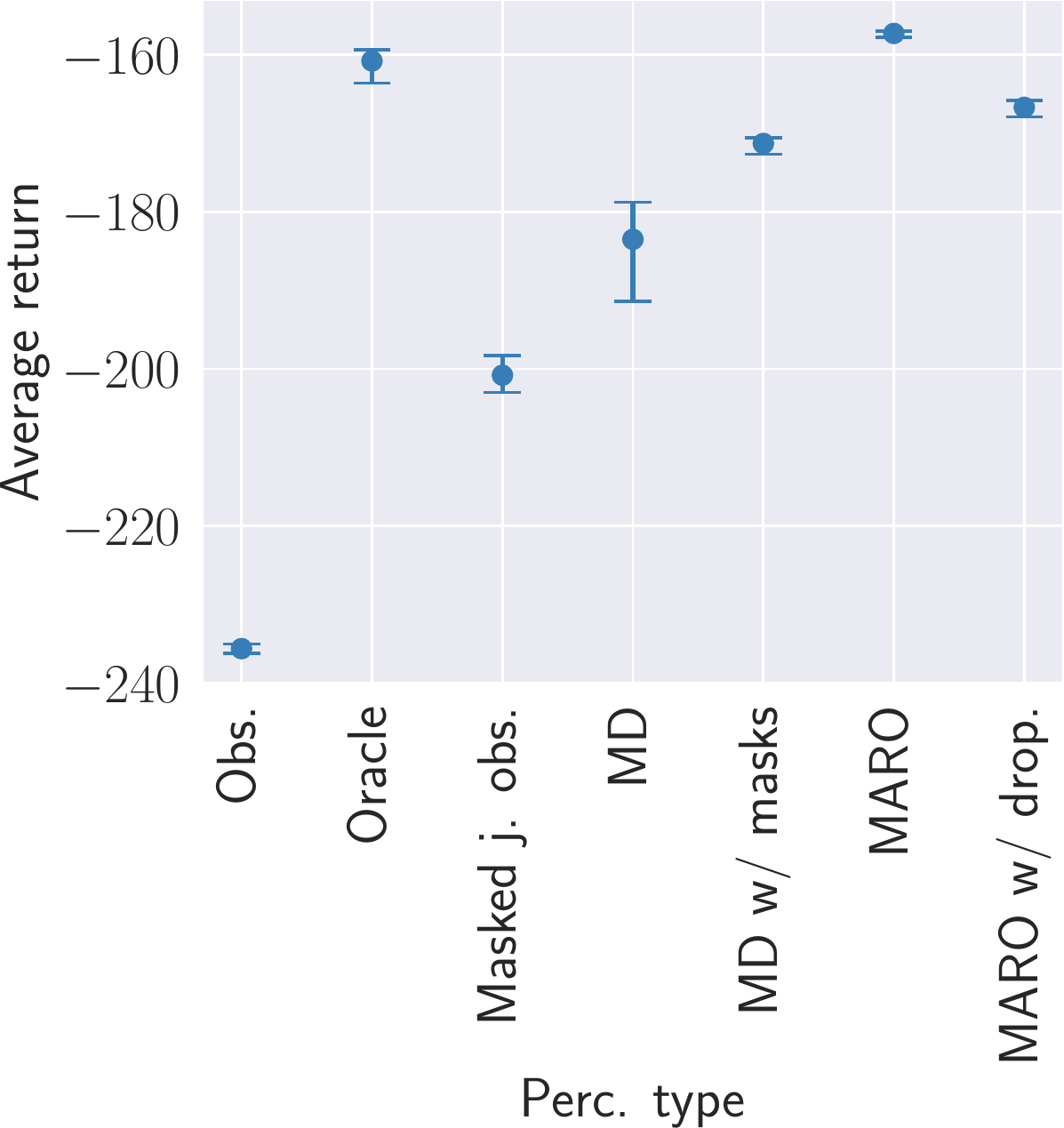}
        \caption{IPPO.}
    \end{subfigure}
    \begin{subfigure}[b]{0.24\textwidth}
        \centering
        \includegraphics[width=0.97\linewidth]{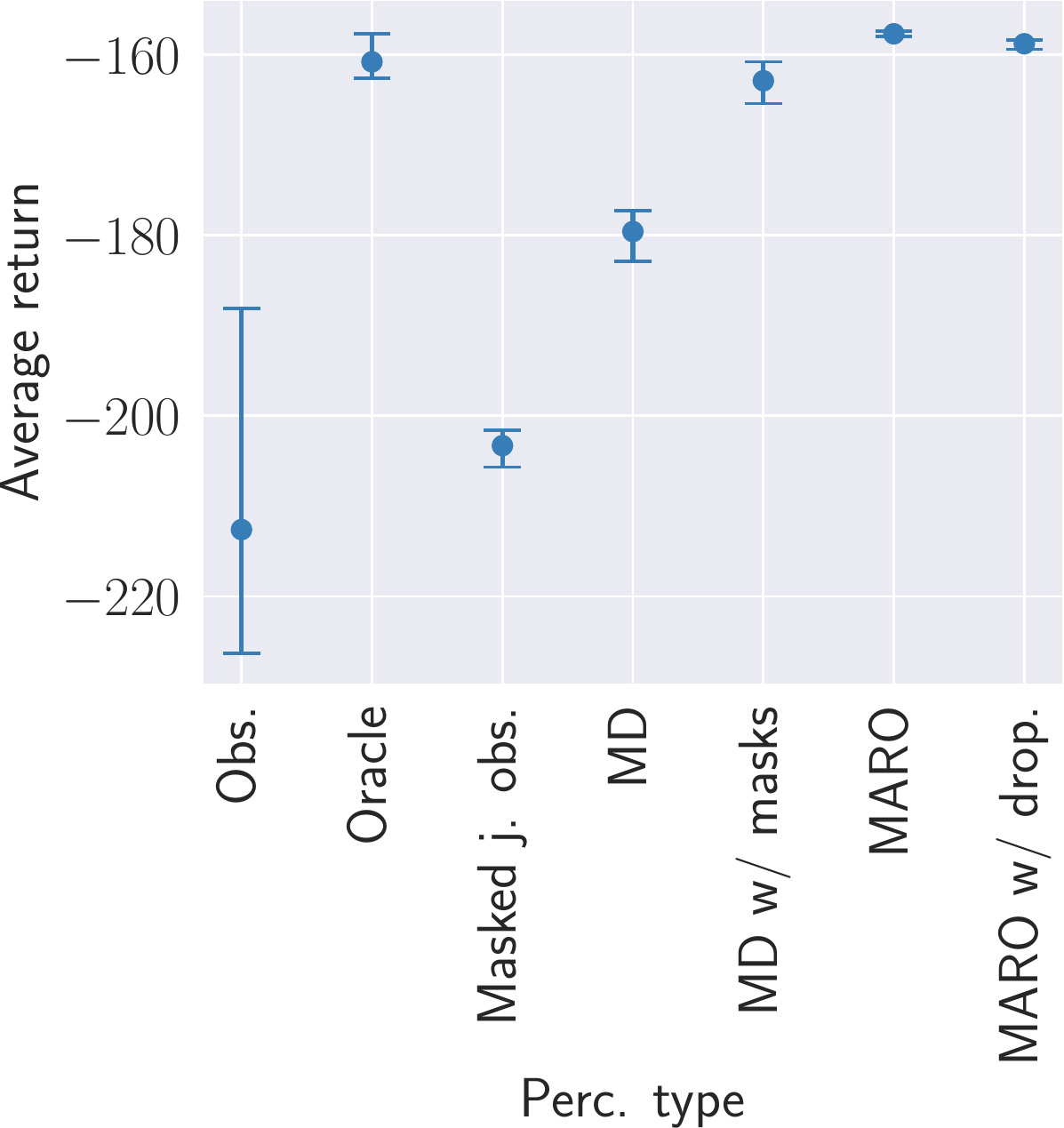}
        \caption{MAPPO.}
    \end{subfigure}
    \caption{(SpreadXY-2) Mean episodic returns for $p_\textrm{dynamic}$ during training.}
\end{figure}

\clearpage

\begin{table}
\centering
\noindent
\caption{(SpreadXY-4) Mean episodic returns for $p_{\textrm{asymmetric}}$ at execution time.}
\vspace{0.1cm}
\resizebox{\linewidth}{!}{%
\begin{tabular}{c c c c c c c c }\toprule
\multicolumn{1}{c }{\textbf{}} & \multicolumn{7}{c }{\textbf{SpreadXY-4 ($p_\textrm{asymmetric}$)}} \\  
\cmidrule(lr){2-8}
\multicolumn{1}{ l }{\textbf{Algorithm}} & \textbf{Obs.} & \textbf{Oracle} & \textbf{Masked j. obs.} & \textbf{MD} & \textbf{MD w/ masks} & \textbf{MARO} & \textbf{MARO w/ drop.} \\
\cmidrule{1-8}
\multicolumn{1}{ l }{IQL} & -1225.5 \tiny{(-4.2,+4.9)} & -902.3 \tiny{(-58.3,+58.3)} & -1126.1 \tiny{(-16.4,+14.8)} & -1166.3 \tiny{(-6.8,+9.1)} & -1168.8 \tiny{(-9.5,+7.3)} & -983.9 \tiny{(-34.5,+44.3)} & -1054.5 \tiny{(-31.5,+55.7)} \\ \cmidrule{1-8}
\multicolumn{1}{ l }{QMIX} & -1132.6 \tiny{(-6.6,+5.9)} & -796.9 \tiny{(-9.0,+12.7)} & -1080.0 \tiny{(-12.0,+8.8)} & -1030.4 \tiny{(-42.9,+58.2)} & -1020.5 \tiny{(-36.9,+44.2)} & -852.9 \tiny{(-24.0,+20.0)} & -945.6 \tiny{(-46.5,+64.7)} \\ \cmidrule{1-8}
\multicolumn{1}{ l }{IPPO} & -1133.2 \tiny{(-7.1,+8.4)} & -781.6 \tiny{(-18.0,+10.5)} & -1062.8 \tiny{(-10.1,+10.1)} & -1110.5 \tiny{(-16.3,+14.7)} & -1162.0 \tiny{(-27.8,+44.8)} & -915.8 \tiny{(-52.4,+104.8)} & -869.2 \tiny{(-10.9,+10.3)} \\ \cmidrule{1-8}
\multicolumn{1}{ l }{MAPPO} & -1116.9 \tiny{(-43.2,+79.0)} & -832.8 \tiny{(-123.3,+65.9)} & -1066.9 \tiny{(-23.2,+28.7)} & -1128.1 \tiny{(-26.2,+20.1)} & -1160.7 \tiny{(-12.0,+13.1)} & -823.9 \tiny{(-1.9,+1.8)} & -827.8 \tiny{(-7.0,+7.5)} \\
\bottomrule
\end{tabular}
}
\end{table}

\begin{table}
\centering
\noindent
\caption{(SpreadXY-4) Mean episodic returns for $p_{\textrm{dynamic}}$ at execution time.}
\vspace{0.1cm}
\resizebox{\linewidth}{!}{%
\begin{tabular}{c c c c c c c c }\toprule
\multicolumn{1}{c }{\textbf{}} & \multicolumn{7}{c }{\textbf{SpreadXY-4 ($p_\textrm{dynamic}$)}} \\  
\cmidrule(lr){2-8}
\multicolumn{1}{ l }{\textbf{Algorithm}} & \textbf{Obs.} & \textbf{Oracle} & \textbf{Masked j. obs.} & \textbf{MD} & \textbf{MD w/ masks} & \textbf{MARO} & \textbf{MARO w/ drop.} \\
\cmidrule{1-8}
\multicolumn{1}{ l }{IQL} & -1225.5 \tiny{(-4.2,+4.9)} & -902.3 \tiny{(-58.3,+58.3)} & -1114.9 \tiny{(-12.4,+6.7)} & -1161.7 \tiny{(-4.9,+3.8)} & -1165.9 \tiny{(-7.9,+4.1)} & -971.8 \tiny{(-27.0,+40.3)} & -1041.3 \tiny{(-32.3,+54.4)} \\ \cmidrule{1-8}
\multicolumn{1}{ l }{QMIX} & -1132.6 \tiny{(-6.6,+5.9)} & -796.9 \tiny{(-9.0,+12.7)} & -1069.5 \tiny{(-2.7,+5.0)} & -1016.4 \tiny{(-38.6,+58.4)} & -1006.9 \tiny{(-40.3,+43.3)} & -833.4 \tiny{(-25.9,+19.9)} & -932.2 \tiny{(-46.6,+70.9)} \\ \cmidrule{1-8}
\multicolumn{1}{ l }{IPPO} & -1133.2 \tiny{(-7.1,+8.4)} & -781.6 \tiny{(-18.0,+10.5)} & -1077.0 \tiny{(-8.2,+8.2)} & -1097.3 \tiny{(-18.2,+16.0)} & -1155.6 \tiny{(-26.5,+38.9)} & -904.6 \tiny{(-56.0,+101.7)} & -863.6 \tiny{(-6.9,+10.1)} \\ \cmidrule{1-8}
\multicolumn{1}{ l }{MAPPO} & -1116.9 \tiny{(-43.2,+79.0)} & -832.8 \tiny{(-123.3,+65.9)} & -1068.3 \tiny{(-2.7,+4.4)} & -1101.7 \tiny{(-32.2,+23.5)} & -1136.7 \tiny{(-13.5,+16.8)} & -813.0 \tiny{(-3.4,+2.9)} & -813.8 \tiny{(-2.6,+3.7)} \\
\bottomrule
\end{tabular}
}
\end{table}

\begin{figure}
    \centering
    \begin{subfigure}[b]{0.24\textwidth}
        \centering
        \includegraphics[width=0.97\linewidth]{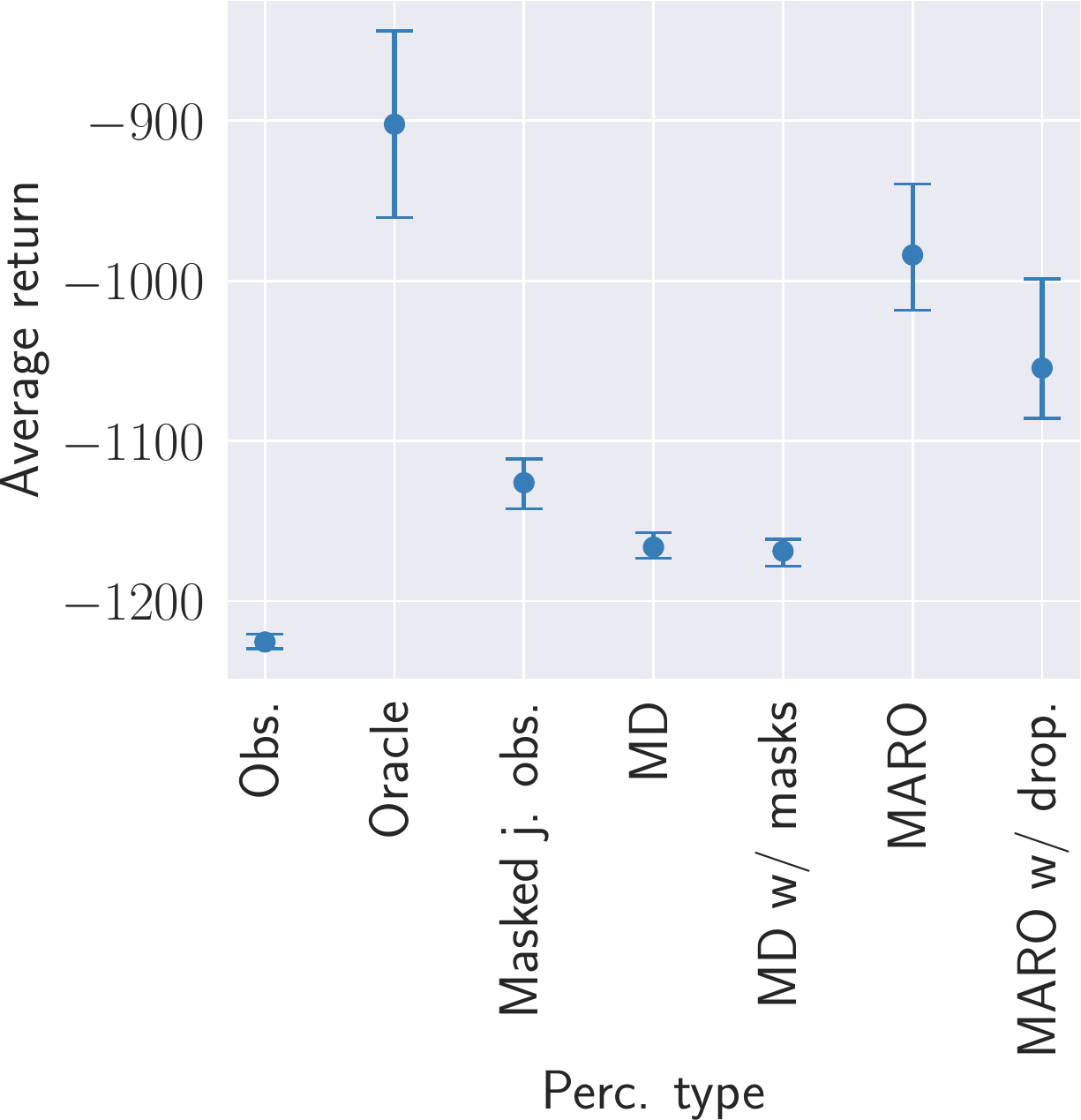}
        \caption{IQL.}
    \end{subfigure}
    \begin{subfigure}[b]{0.24\textwidth}
        \centering
        \includegraphics[width=0.97\linewidth]{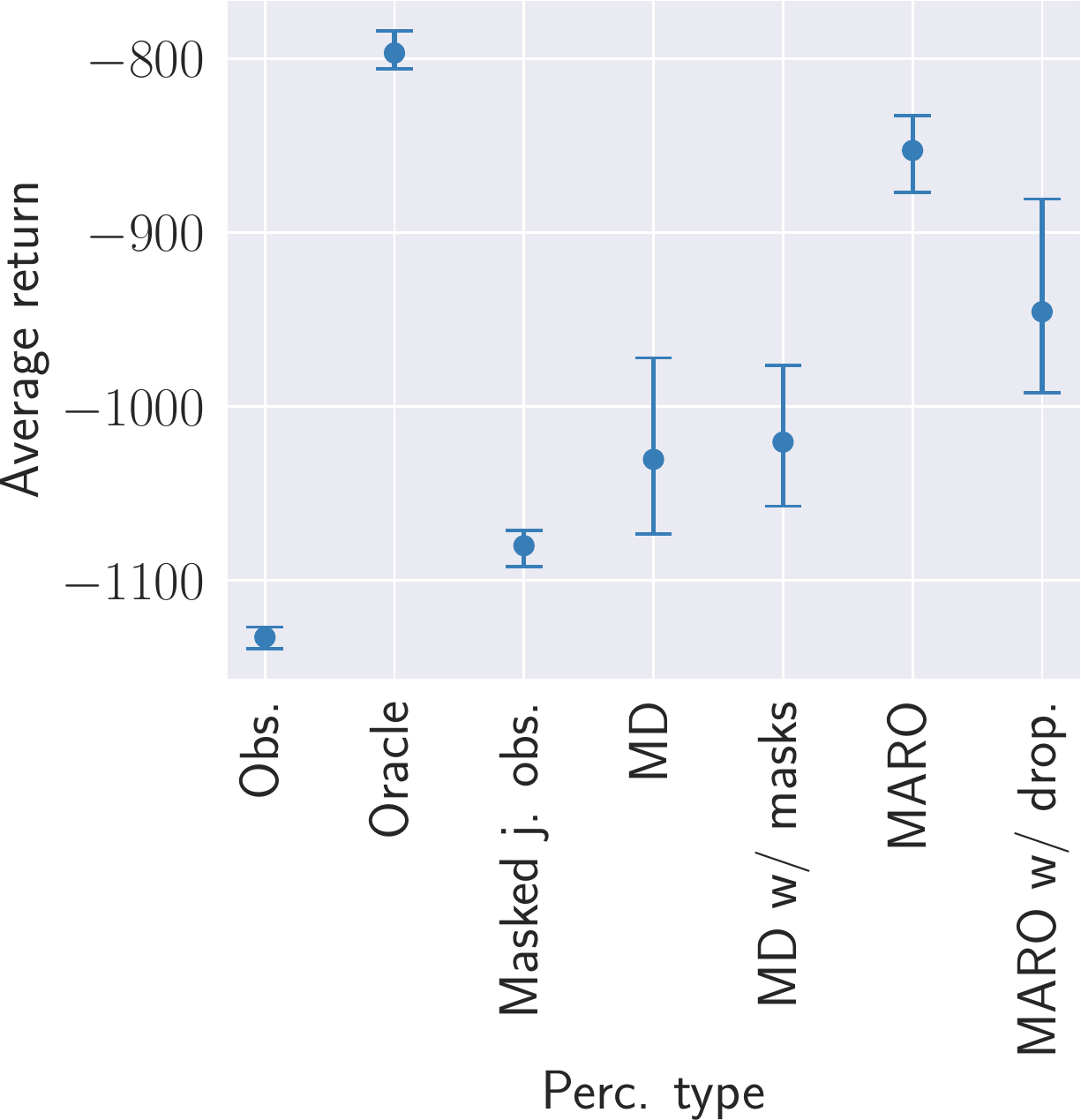}
        \caption{QMIX.}
    \end{subfigure}
    \begin{subfigure}[b]{0.24\textwidth}
        \centering
        \includegraphics[width=0.97\linewidth]{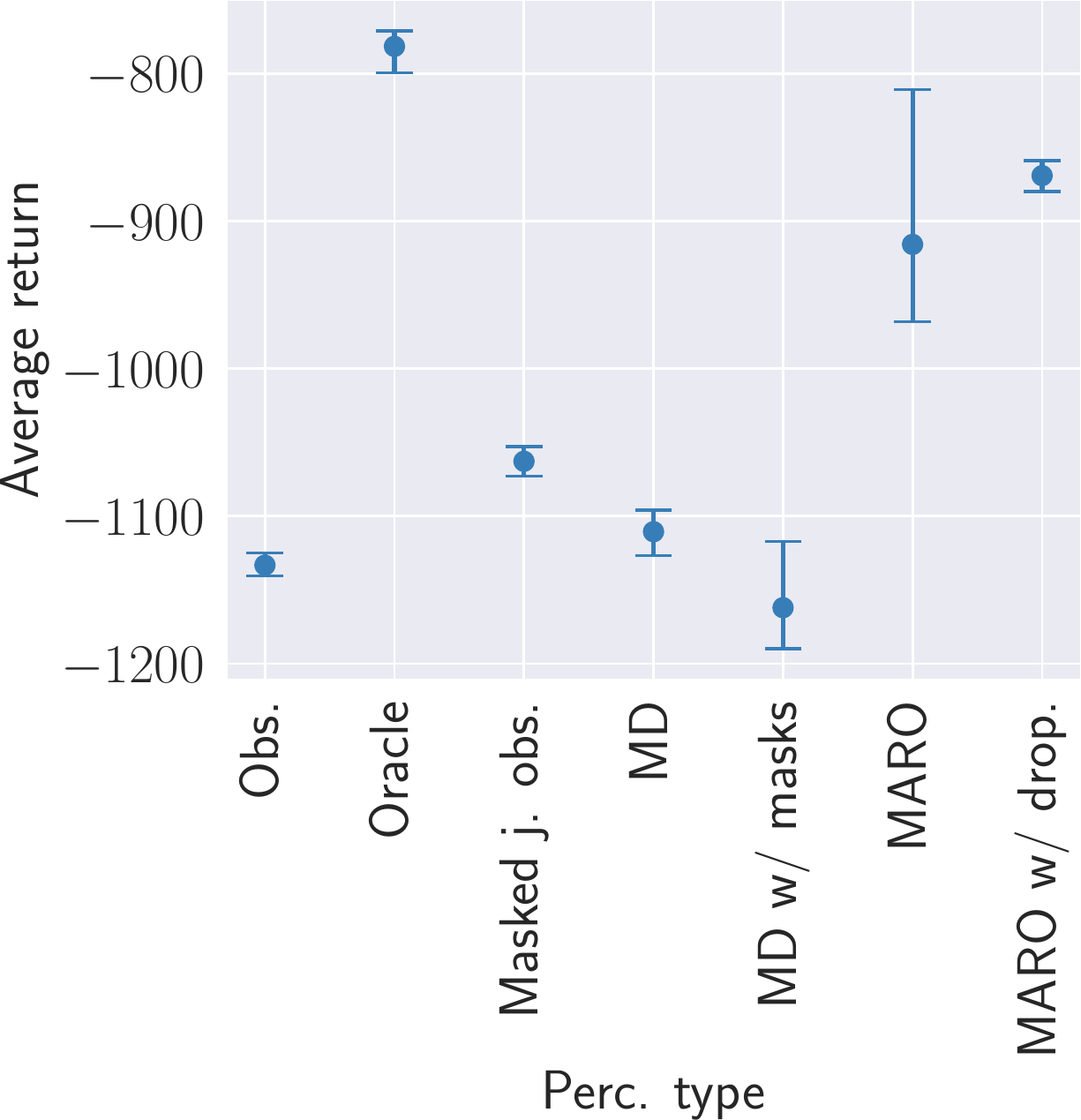}
        \caption{IPPO.}
    \end{subfigure}
    \begin{subfigure}[b]{0.24\textwidth}
        \centering
        \includegraphics[width=0.97\linewidth]{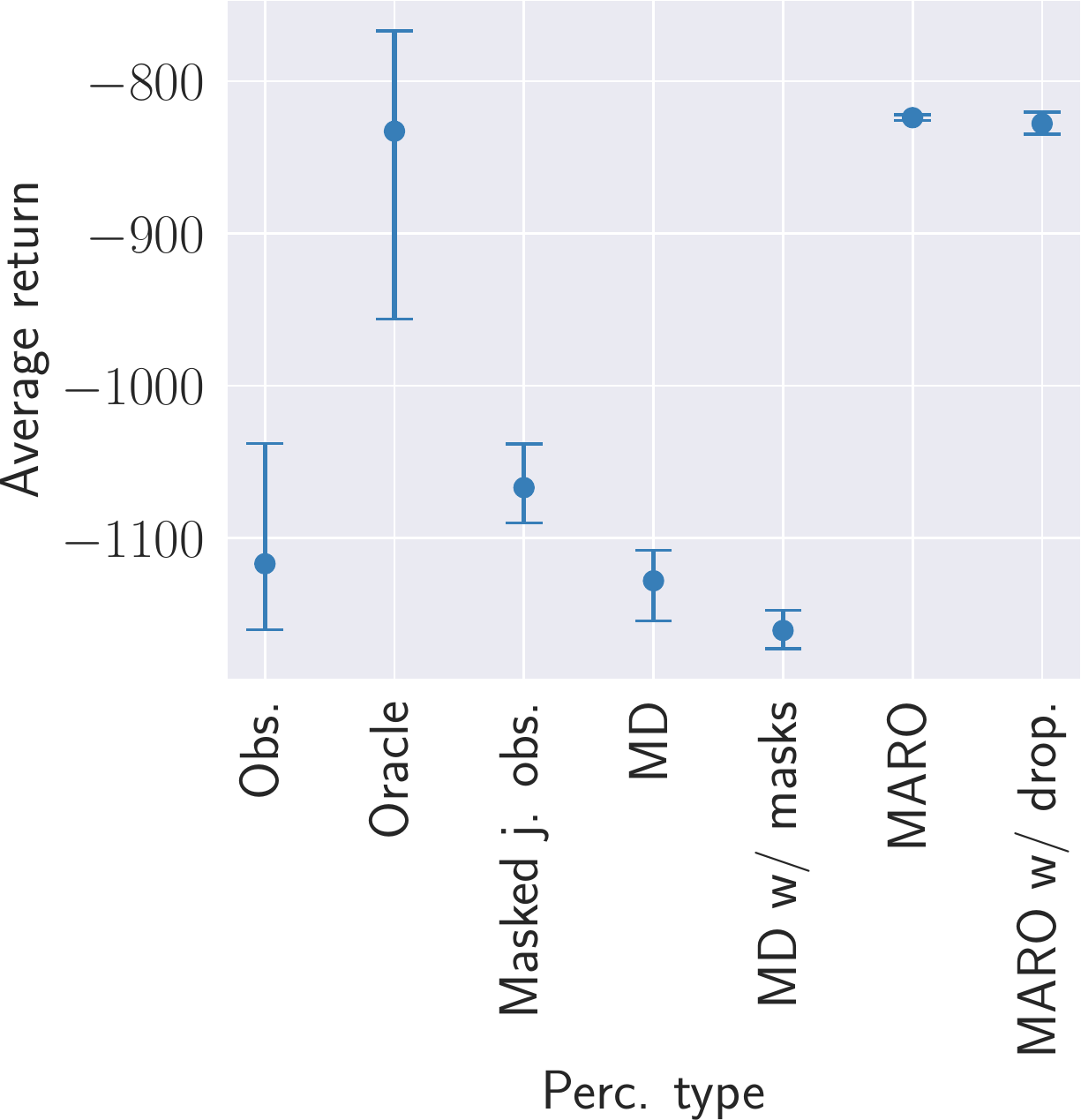}
        \caption{MAPPO.}
    \end{subfigure}
    \caption{(SpreadXY-4) Mean episodic returns for $p_\textrm{asymmetric}$ during training.}
\end{figure}

\begin{figure}
    \centering
    \begin{subfigure}[b]{0.24\textwidth}
        \centering
        \includegraphics[width=0.97\linewidth]{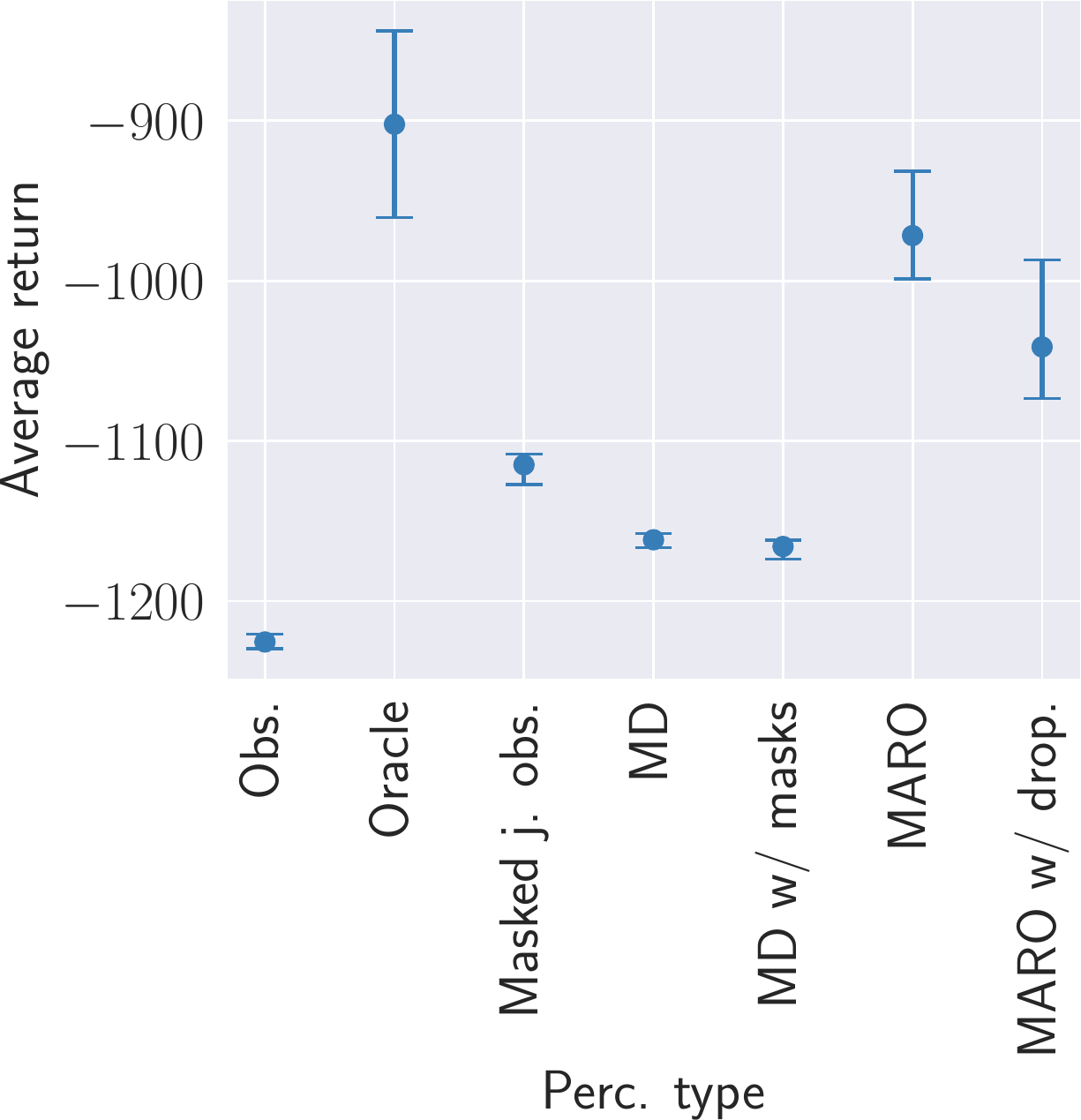}
        \caption{IQL.}
    \end{subfigure}
    \begin{subfigure}[b]{0.24\textwidth}
        \centering
        \includegraphics[width=0.97\linewidth]{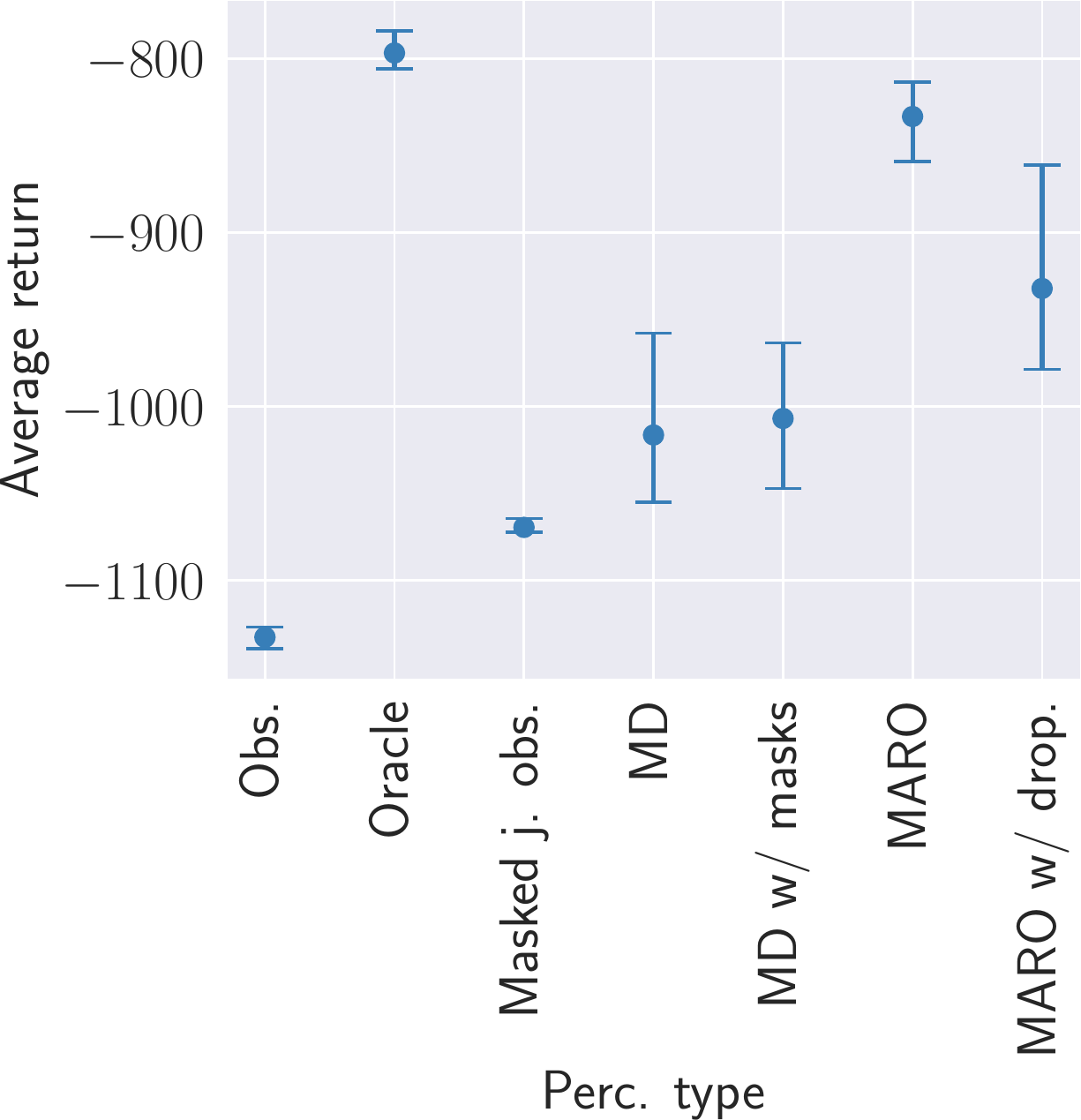}
        \caption{QMIX.}
    \end{subfigure}
    \begin{subfigure}[b]{0.24\textwidth}
        \centering
        \includegraphics[width=0.97\linewidth]{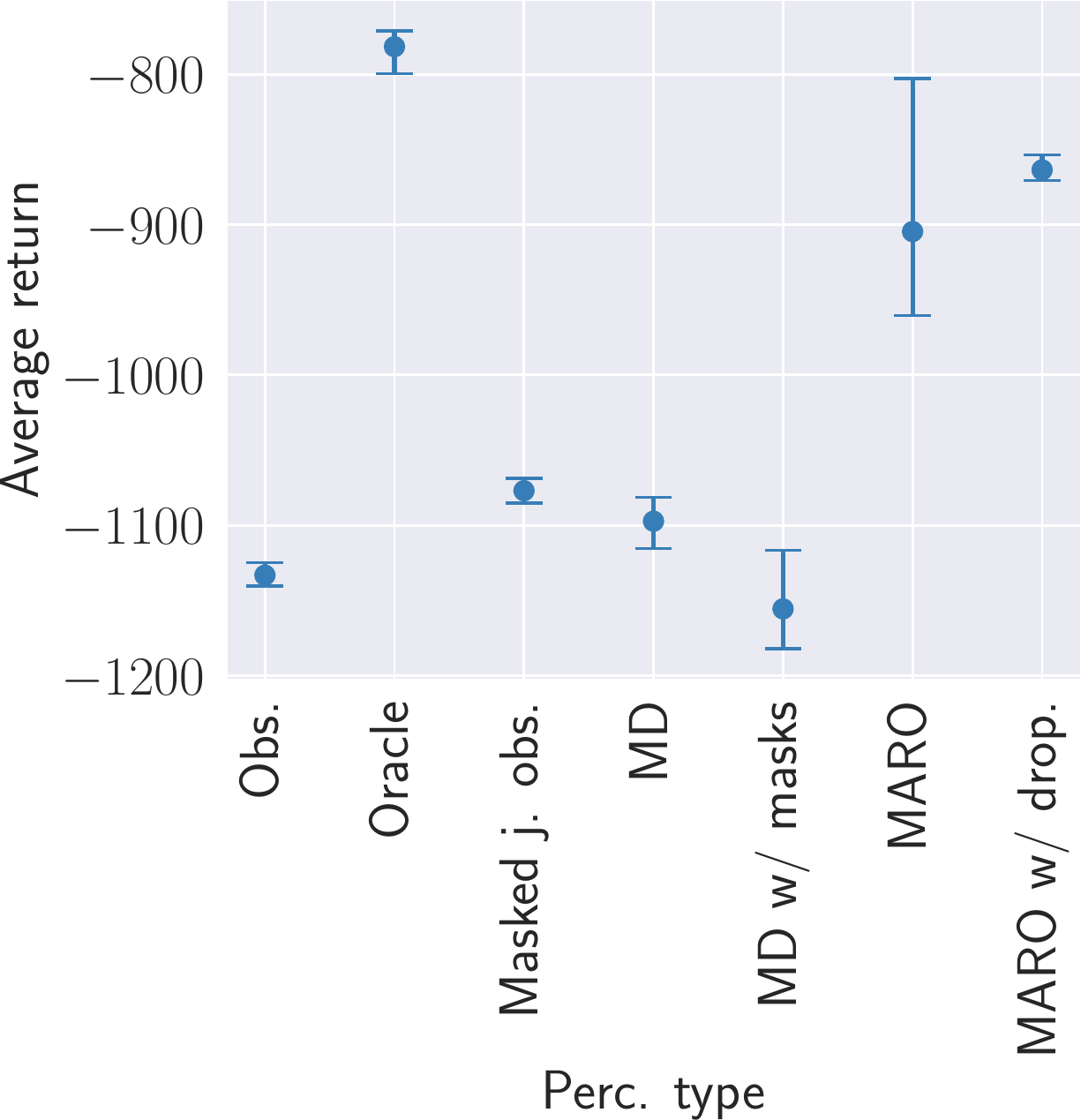}
        \caption{IPPO.}
    \end{subfigure}
    \begin{subfigure}[b]{0.24\textwidth}
        \centering
        \includegraphics[width=0.97\linewidth]{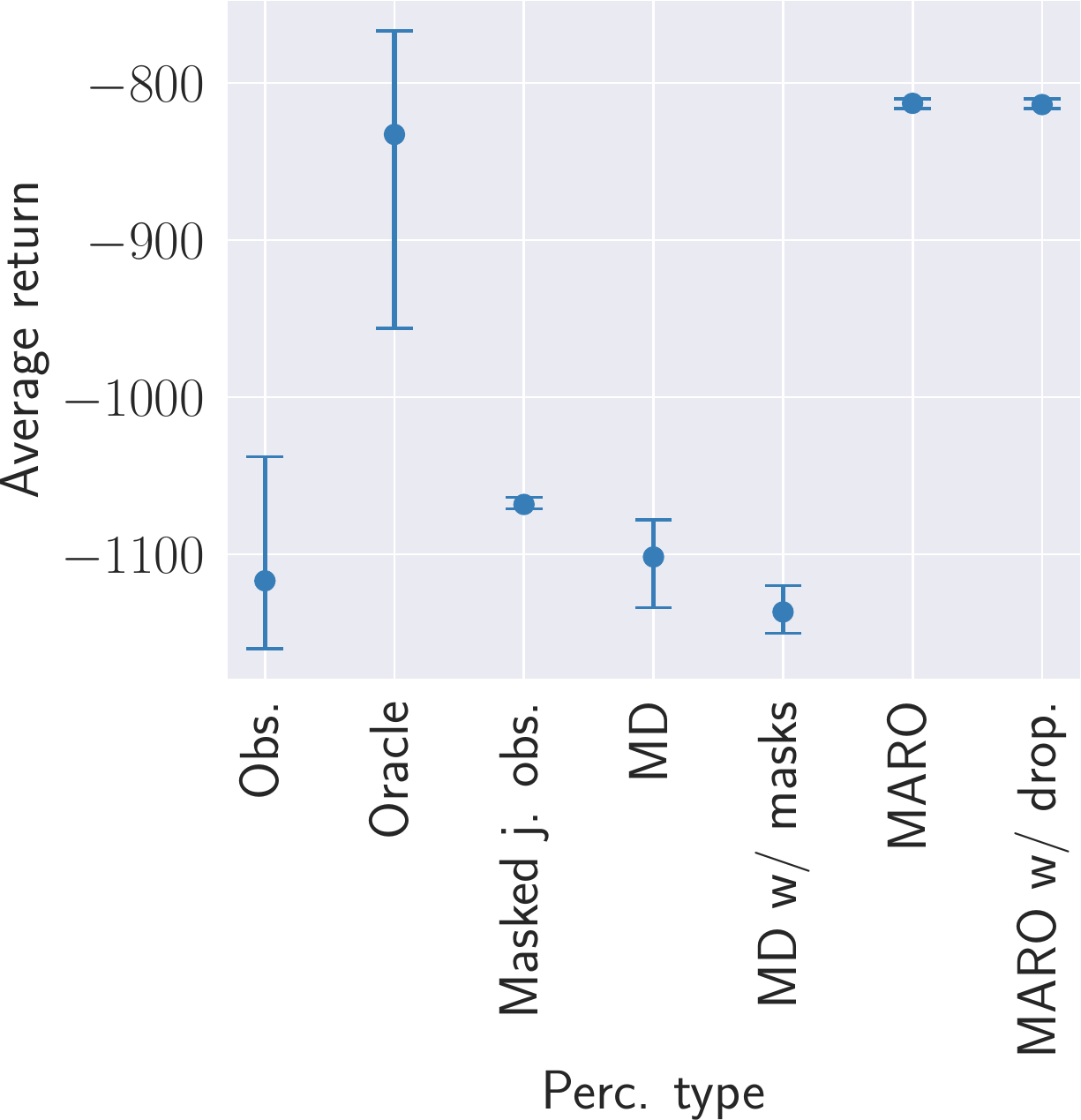}
        \caption{MAPPO.}
    \end{subfigure}
    \caption{(SpreadXY-4) Mean episodic returns for $p_\textrm{dynamic}$ during training.}
\end{figure}

\clearpage

\begin{table}
\centering
\noindent
\caption{(SpreadBlindFold) Mean episodic returns for $p_{\textrm{asymmetric}}$ at execution time.}
\vspace{0.1cm}
\resizebox{\linewidth}{!}{%
\begin{tabular}{c c c c c c c c }\toprule
\multicolumn{1}{c }{\textbf{}} & \multicolumn{7}{c }{\textbf{SpreadBlindFold ($p_\textrm{asymmetric}$)}} \\  
\cmidrule(lr){2-8}
\multicolumn{1}{ l }{\textbf{Algorithm}} & \textbf{Obs.} & \textbf{Oracle} & \textbf{Masked j. obs.} & \textbf{MD} & \textbf{MD w/ masks} & \textbf{MARO} & \textbf{MARO w/ drop.} \\
\cmidrule{1-8}
\multicolumn{1}{ l }{IQL} & -425.1 \tiny{(-1.1,+1.4)} & -395.4 \tiny{(-2.8,+2.5)} & -414.6 \tiny{(-6.4,+5.3)} & -402.7 \tiny{(-6.5,+3.4)} & -398.5 \tiny{(-1.6,+2.2)} & -400.5 \tiny{(-3.4,+3.0)} & -390.0 \tiny{(-0.7,+0.6)} \\ \cmidrule{1-8}
\multicolumn{1}{ l }{QMIX} & -416.1 \tiny{(-10.0,+7.7)} & -376.4 \tiny{(-4.7,+4.5)} & -407.3 \tiny{(-6.8,+6.6)} & -403.7 \tiny{(-8.4,+4.4)} & -401.0 \tiny{(-6.6,+5.0)} & -380.8 \tiny{(-2.1,+3.3)} & -373.5 \tiny{(-6.0,+3.6)} \\ \cmidrule{1-8}
\multicolumn{1}{ l }{IPPO} & -436.3 \tiny{(-74.5,+38.4)} & -407.5 \tiny{(-2.8,+1.6)} & -402.1 \tiny{(-1.2,+2.4)} & -447.7 \tiny{(-5.8,+5.5)} & -473.0 \tiny{(-11.7,+7.7)} & -403.4 \tiny{(-0.7,+1.3)} & -420.5 \tiny{(-9.9,+12.3)} \\ \cmidrule{1-8}
\multicolumn{1}{ l }{MAPPO} & -420.3 \tiny{(-0.4,+0.4)} & -404.5 \tiny{(-2.8,+2.0)} & -403.3 \tiny{(-1.7,+1.5)} & -409.6 \tiny{(-2.9,+3.5)} & -412.3 \tiny{(-2.1,+2.1)} & -401.4 \tiny{(-2.1,+1.8)} & -402.6 \tiny{(-2.3,+2.2)} \\
\bottomrule
\end{tabular}
}
\end{table}

\begin{table}
\centering
\noindent
\caption{(SpreadBlindFold) Mean episodic returns for $p_{\textrm{dynamic}}$ at execution time.}
\vspace{0.1cm}
\resizebox{\linewidth}{!}{%
\begin{tabular}{c c c c c c c c }\toprule
\multicolumn{1}{c }{\textbf{}} & \multicolumn{7}{c }{\textbf{SpreadBlindFold ($p_\textrm{dynamic}$)}} \\  
\cmidrule(lr){2-8}
\multicolumn{1}{ l }{\textbf{Algorithm}} & \textbf{Obs.} & \textbf{Oracle} & \textbf{Masked j. obs.} & \textbf{MD} & \textbf{MD w/ masks} & \textbf{MARO} & \textbf{MARO w/ drop.} \\
\cmidrule{1-8}
\multicolumn{1}{ l }{IQL} & -425.1 \tiny{(-1.1,+1.4)} & -395.4 \tiny{(-2.8,+2.5)} & -411.9 \tiny{(-6.4,+4.5)} & -403.0 \tiny{(-7.9,+7.5)} & -397.8 \tiny{(-4.6,+7.7)} & -401.4 \tiny{(-2.4,+4.4)} & -389.3 \tiny{(-1.5,+0.9)} \\ \cmidrule{1-8}
\multicolumn{1}{ l }{QMIX} & -416.1 \tiny{(-10.0,+7.7)} & -376.4 \tiny{(-4.7,+4.5)} & -404.6 \tiny{(-2.7,+2.0)} & -402.7 \tiny{(-6.7,+3.5)} & -398.6 \tiny{(-4.2,+4.5)} & -381.9 \tiny{(-5.0,+5.6)} & -372.0 \tiny{(-3.9,+2.7)} \\ \cmidrule{1-8}
\multicolumn{1}{ l }{IPPO} & -436.3 \tiny{(-74.5,+38.4)} & -407.5 \tiny{(-2.8,+1.6)} & -403.6 \tiny{(-1.5,+1.3)} & -447.3 \tiny{(-8.6,+8.7)} & -470.4 \tiny{(-5.0,+9.3)} & -404.4 \tiny{(-1.1,+1.1)} & -418.7 \tiny{(-7.6,+10.6)} \\ \cmidrule{1-8}
\multicolumn{1}{ l }{MAPPO} & -420.3 \tiny{(-0.4,+0.4)} & -404.5 \tiny{(-2.8,+2.0)} & -405.4 \tiny{(-2.0,+1.4)} & -408.1 \tiny{(-2.4,+3.7)} & -411.1 \tiny{(-1.9,+1.8)} & -404.1 \tiny{(-2.1,+2.9)} & -403.0 \tiny{(-2.0,+1.5)} \\
\bottomrule
\end{tabular}
}
\end{table}

\begin{figure}
    \centering
    \begin{subfigure}[b]{0.24\textwidth}
        \centering
        \includegraphics[width=0.97\linewidth]{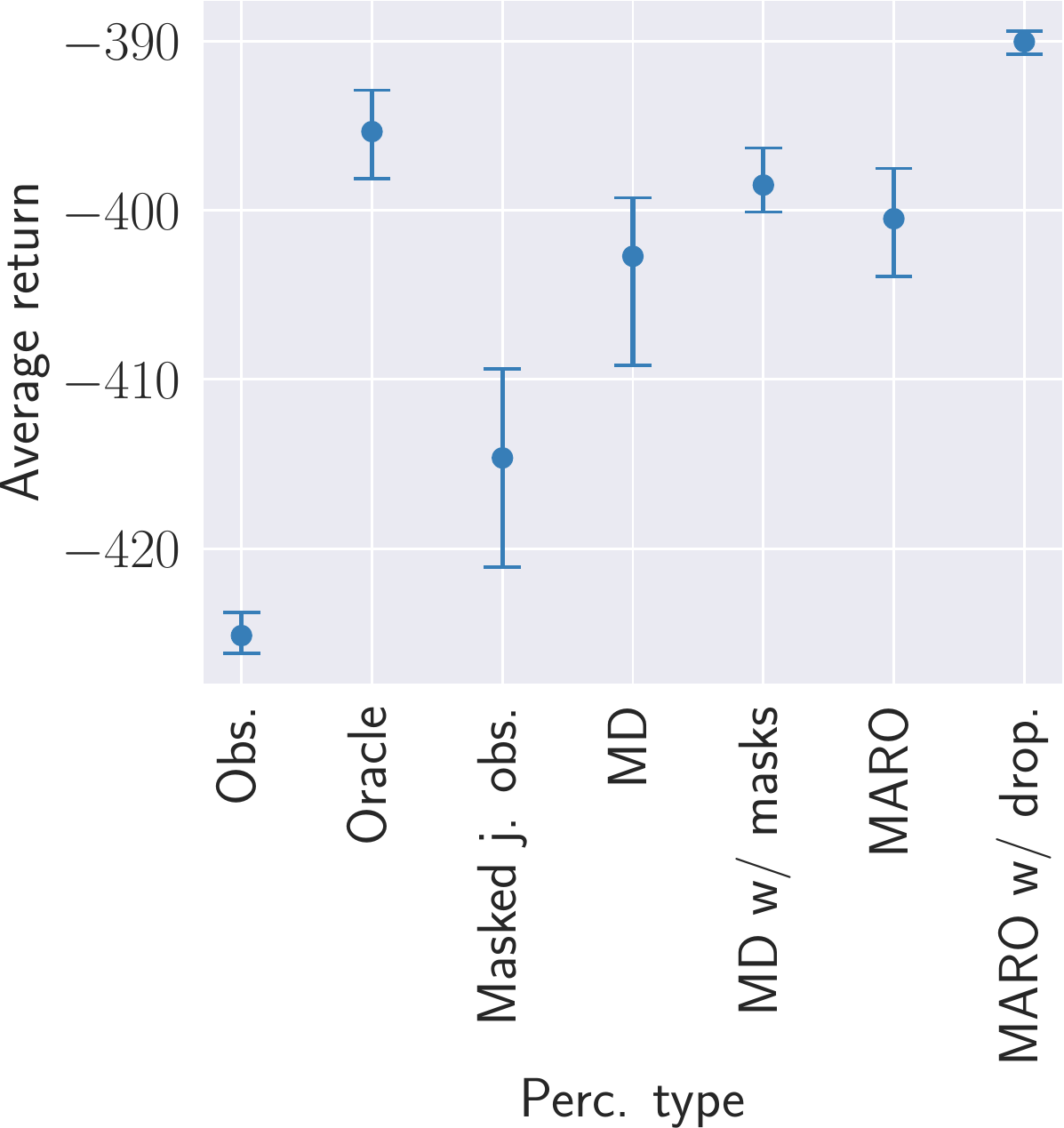}
        \caption{IQL.}
    \end{subfigure}
    \begin{subfigure}[b]{0.24\textwidth}
        \centering
        \includegraphics[width=0.97\linewidth]{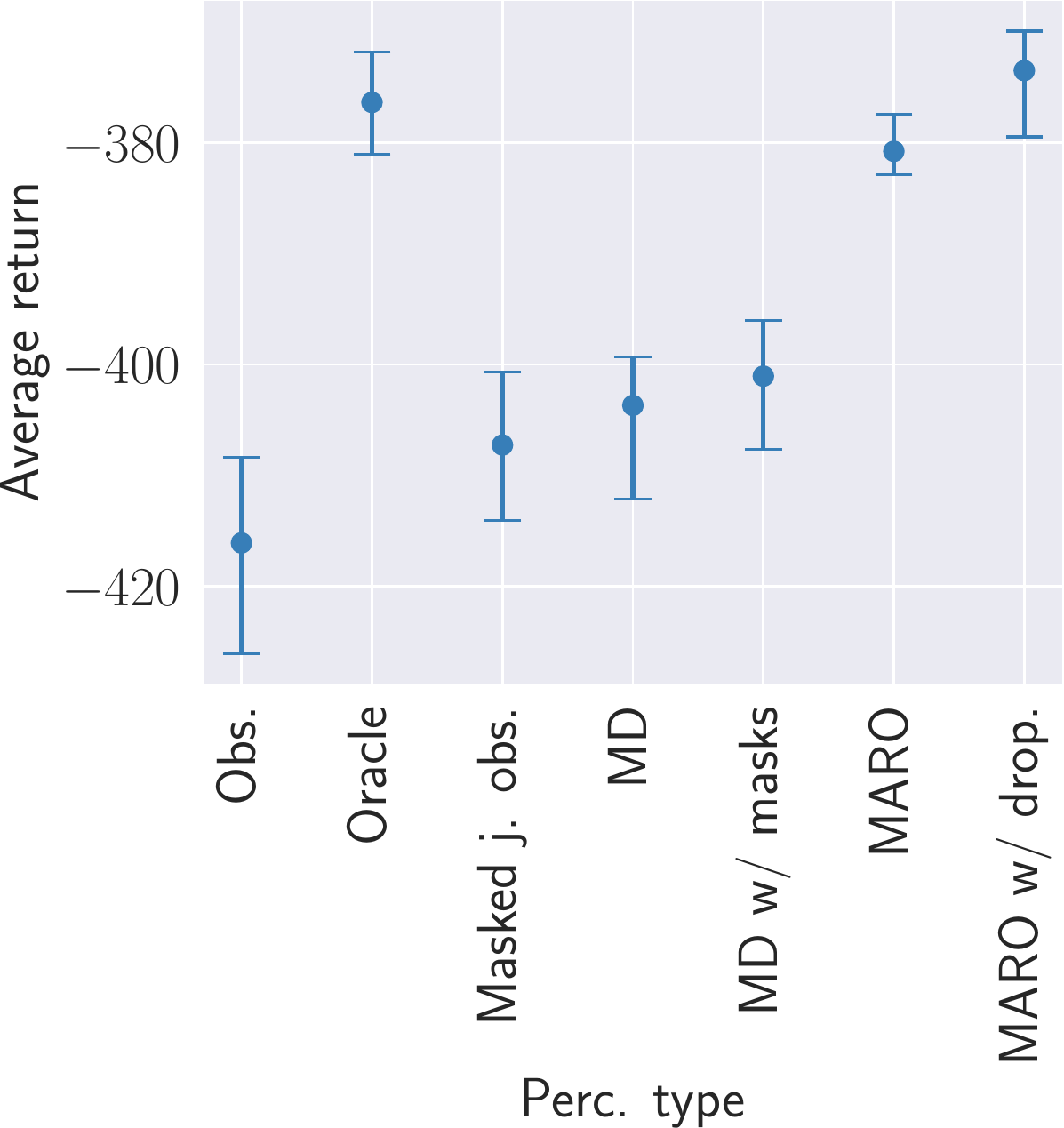}
        \caption{QMIX.}
    \end{subfigure}
    \begin{subfigure}[b]{0.24\textwidth}
        \centering
        \includegraphics[width=0.97\linewidth]{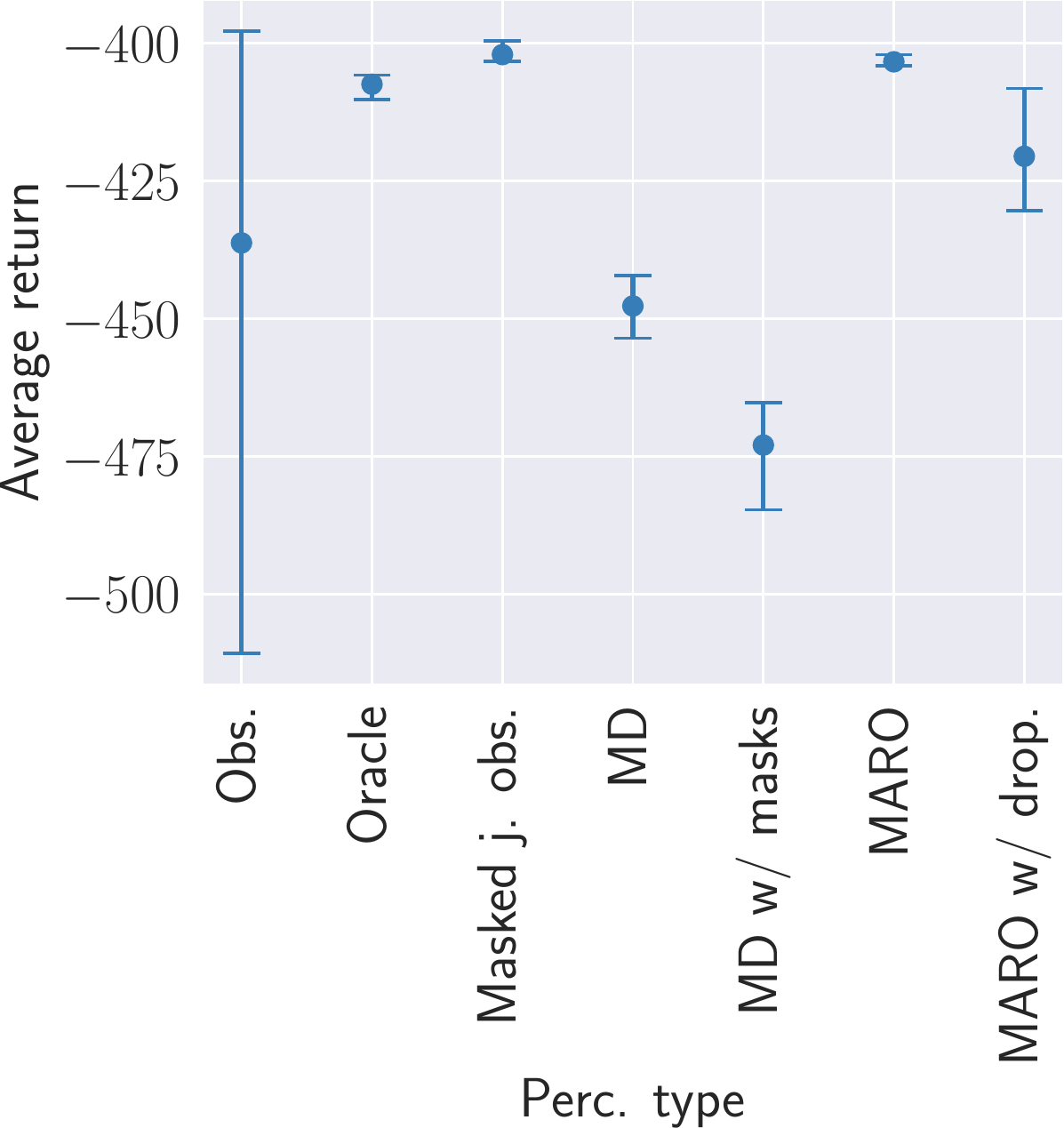}
        \caption{IPPO.}
    \end{subfigure}
    \begin{subfigure}[b]{0.24\textwidth}
        \centering
        \includegraphics[width=0.97\linewidth]{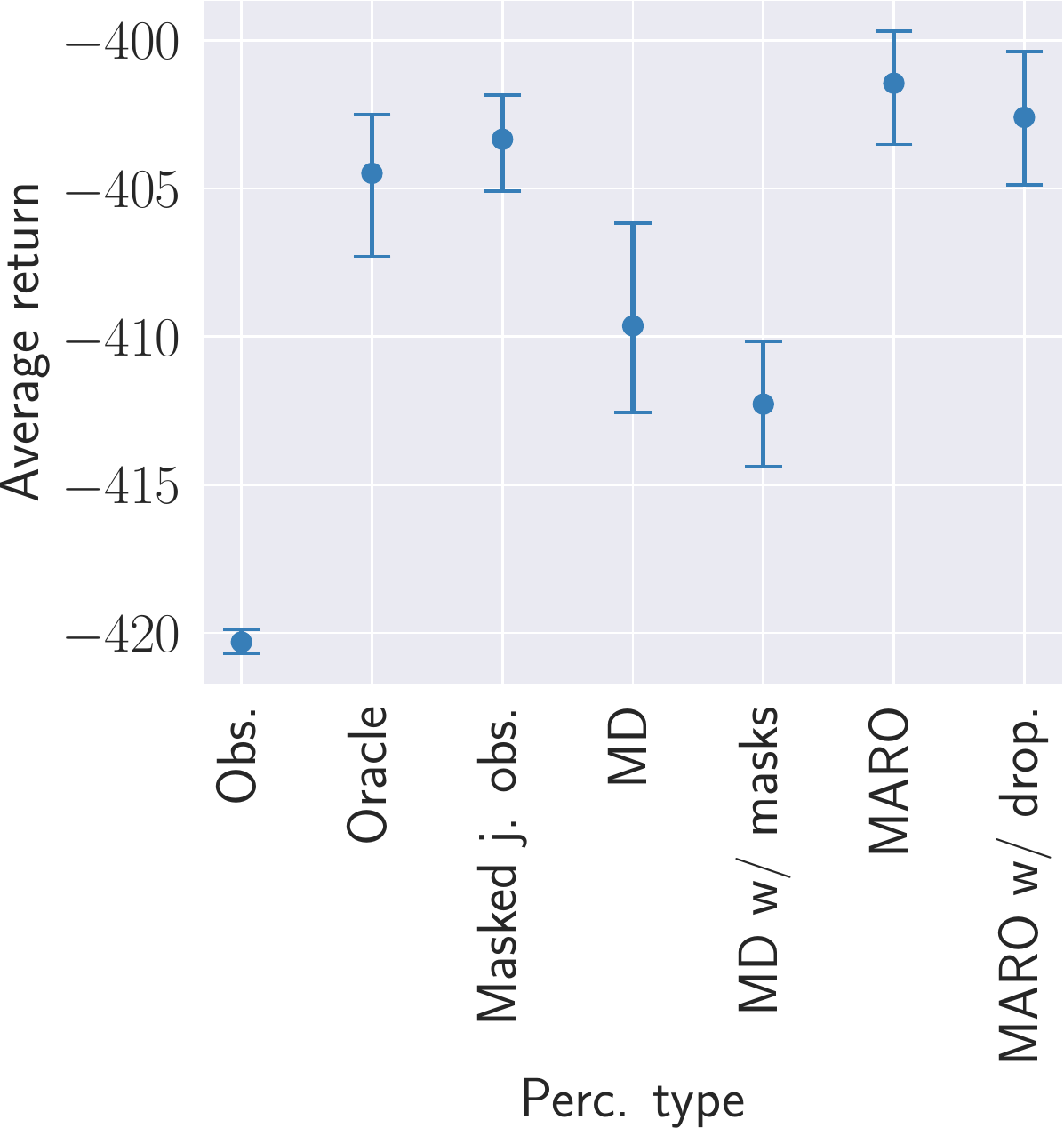}
        \caption{MAPPO.}
    \end{subfigure}
    \caption{(SpreadBlindFold) Mean episodic returns for $p_\textrm{asymmetric}$ during training.}
\end{figure}

\begin{figure}
    \centering
    \begin{subfigure}[b]{0.24\textwidth}
        \centering
        \includegraphics[width=0.97\linewidth]{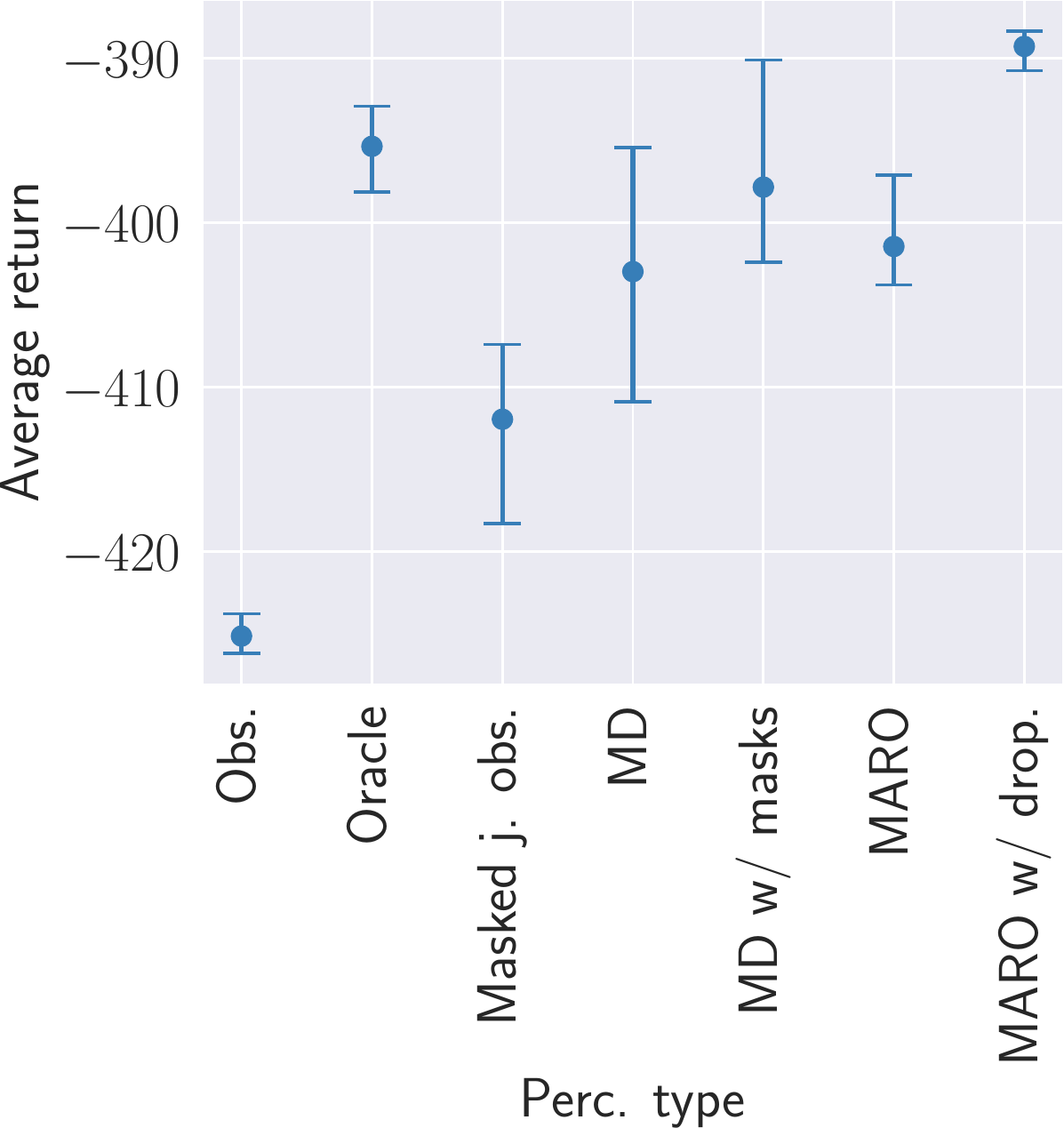}
        \caption{IQL.}
    \end{subfigure}
    \begin{subfigure}[b]{0.24\textwidth}
        \centering
        \includegraphics[width=0.97\linewidth]{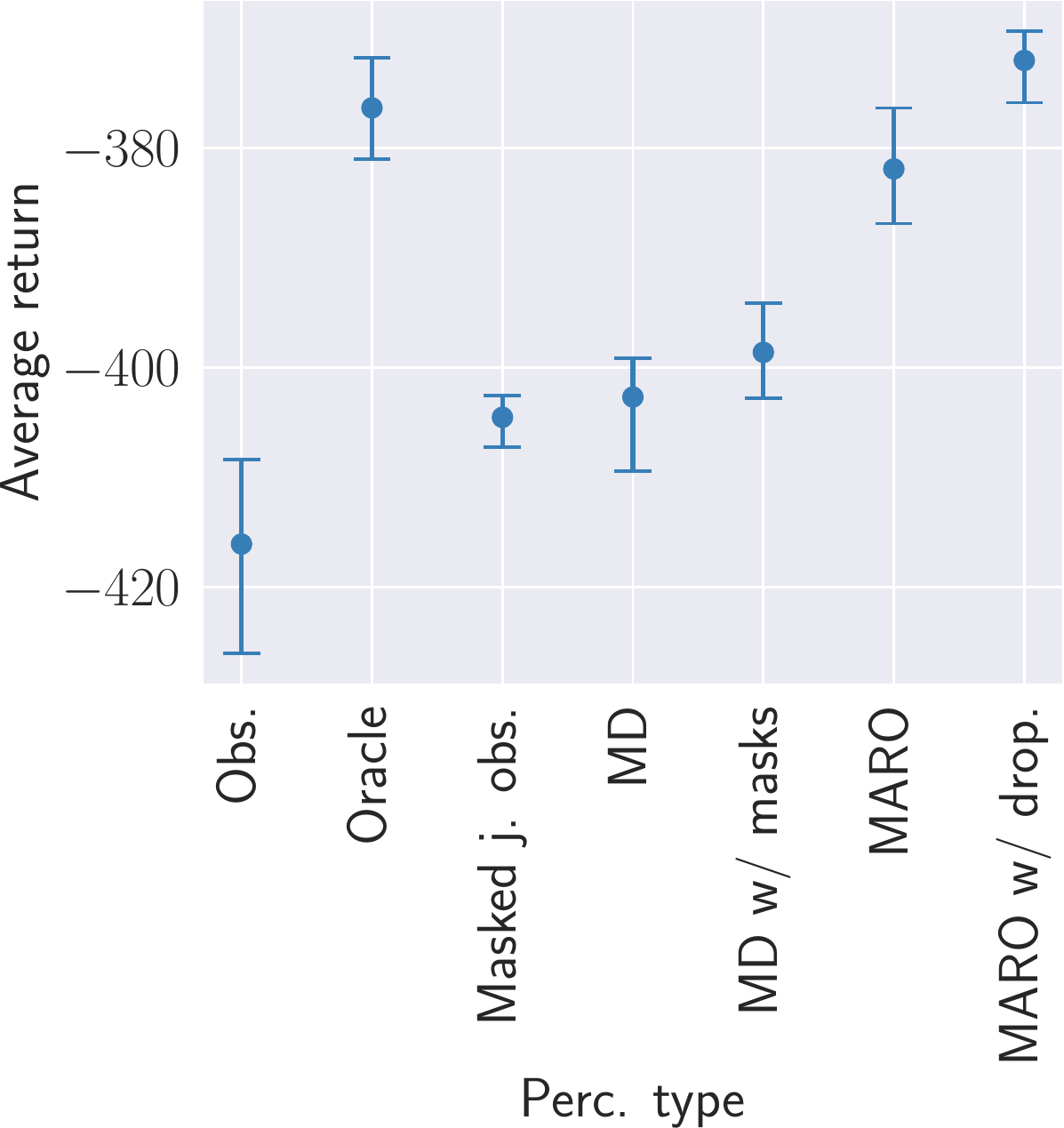}
        \caption{QMIX.}
    \end{subfigure}
    \begin{subfigure}[b]{0.24\textwidth}
        \centering
        \includegraphics[width=0.97\linewidth]{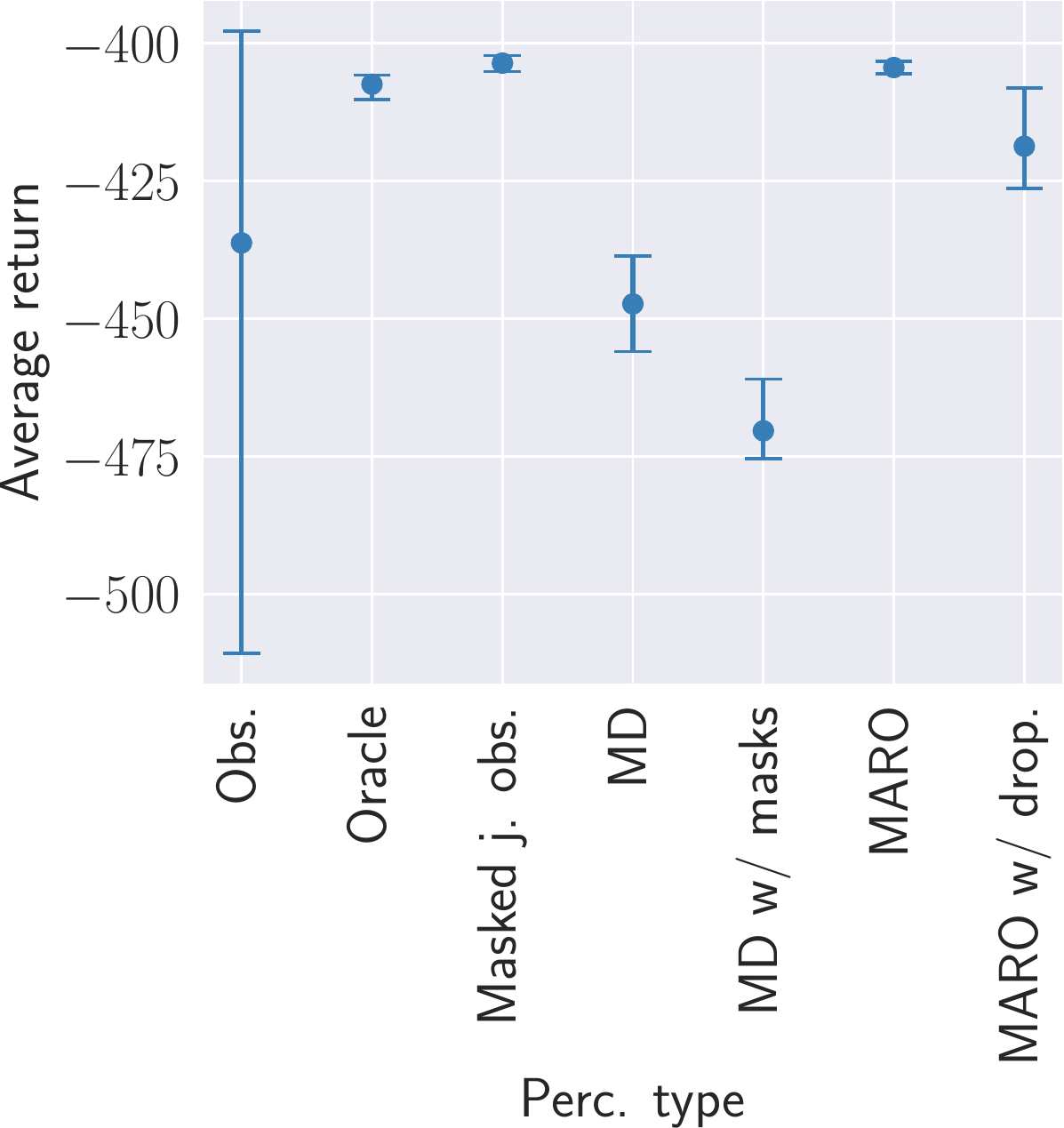}
        \caption{IPPO.}
    \end{subfigure}
    \begin{subfigure}[b]{0.24\textwidth}
        \centering
        \includegraphics[width=0.97\linewidth]{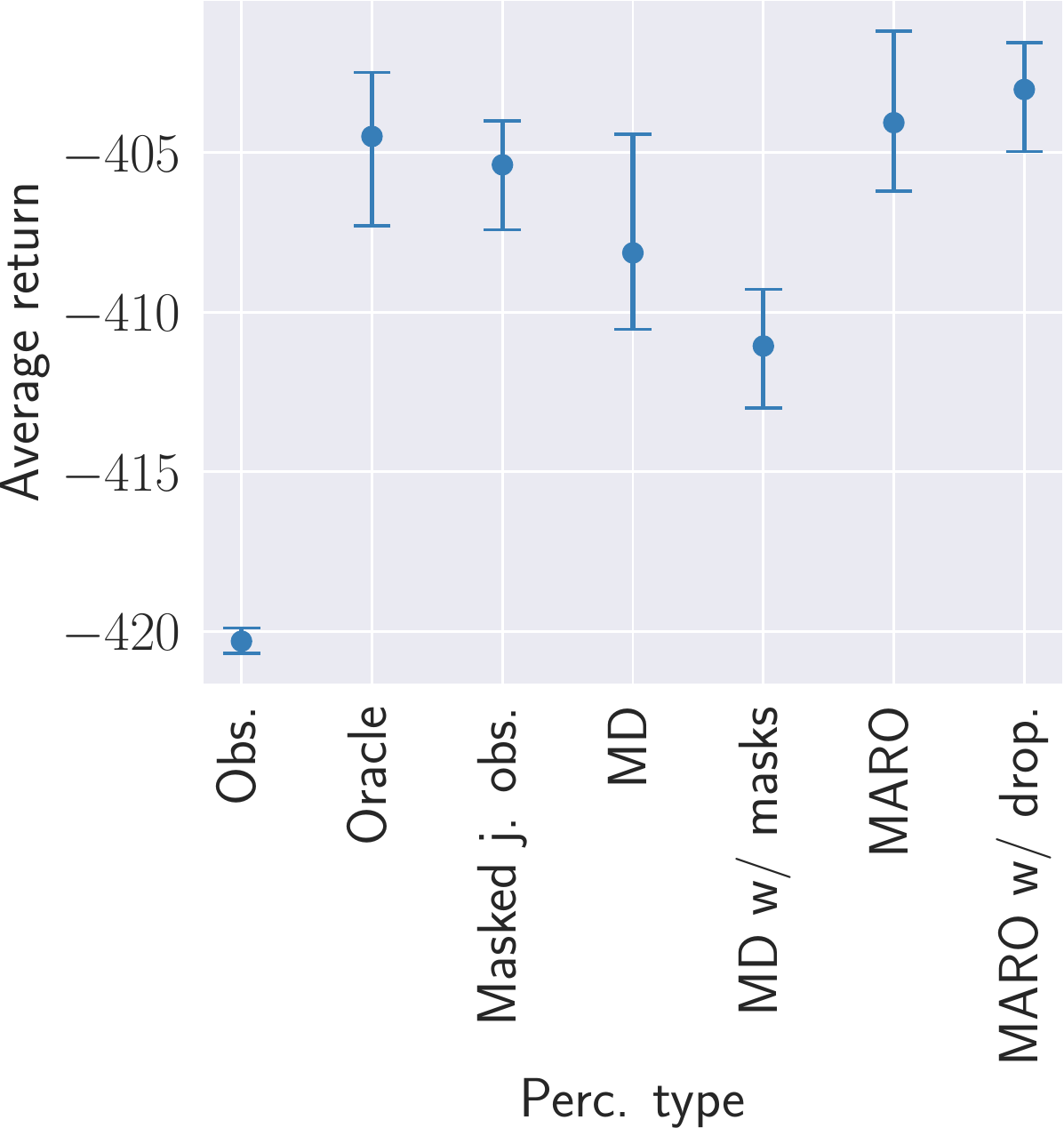}
        \caption{MAPPO.}
    \end{subfigure}
    \caption{(SpreadBlindFold) Mean episodic returns for $p_\textrm{dynamic}$ during training.}
\end{figure}

\clearpage

\begin{table}
\centering
\noindent
\caption{(Foraging-2s-15x15-2p-2f-coop-v2) Mean episodic returns for $p_{\textrm{asymmetric}}$ at execution time.}
\vspace{0.1cm}
\resizebox{\linewidth}{!}{%
\begin{tabular}{c c c c c c c c }\toprule
\multicolumn{1}{c }{\textbf{}} & \multicolumn{7}{c }{\textbf{Foraging-2s-15x15-2p-2f-coop-v2 ($p_\textrm{asymmetric}$)}} \\  
\cmidrule(lr){2-8}
\multicolumn{1}{ l }{\textbf{Algorithm}} & \textbf{Obs.} & \textbf{Oracle} & \textbf{Masked j. obs.} & \textbf{MD} & \textbf{MD w/ masks} & \textbf{MARO} & \textbf{MARO w/ drop.} \\
\cmidrule{1-8}
\multicolumn{1}{ l }{IQL} & 0.38 \tiny{(-0.03,+0.01)} & 0.48 \tiny{(-0.03,+0.02)} & 0.14 \tiny{(-0.02,+0.02)} & 0.52 \tiny{(-0.0,+0.0)} & 0.52 \tiny{(-0.02,+0.02)} & 0.34 \tiny{(-0.01,+0.02)} & 0.51 \tiny{(-0.01,+0.02)} \\ \cmidrule{1-8}
\multicolumn{1}{ l }{QMIX} & 0.55 \tiny{(-0.01,+0.0)} & 0.68 \tiny{(-0.02,+0.03)} & 0.19 \tiny{(-0.06,+0.05)} & 0.58 \tiny{(-0.02,+0.02)} & 0.59 \tiny{(-0.01,+0.01)} & 0.41 \tiny{(-0.01,+0.01)} & 0.59 \tiny{(-0.01,+0.01)} \\ \cmidrule{1-8}
\multicolumn{1}{ l }{IPPO} & 0.31 \tiny{(-0.0,+0.0)} & 0.44 \tiny{(-0.0,+0.0)} & 0.3 \tiny{(-0.01,+0.0)} & 0.03 \tiny{(-0.03,+0.05)} & 0.01 \tiny{(-0.01,+0.01)} & 0.36 \tiny{(-0.01,+0.01)} & 0.01 \tiny{(-0.0,+0.0)} \\ \cmidrule{1-8}
\multicolumn{1}{ l }{MAPPO} & 0.36 \tiny{(-0.0,+0.0)} & 0.45 \tiny{(-0.01,+0.01)} & 0.29 \tiny{(-0.02,+0.03)} & 0.02 \tiny{(-0.01,+0.02)} & 0.02 \tiny{(-0.02,+0.02)} & 0.36 \tiny{(-0.03,+0.05)} & 0.09 \tiny{(-0.08,+0.06)} \\
\bottomrule
\end{tabular}
}
\end{table}

\begin{table}
\centering
\noindent
\caption{(Foraging-2s-15x15-2p-2f-coop-v2) Mean episodic returns for $p_{\textrm{dynamic}}$ at execution time.}
\vspace{0.1cm}
\resizebox{\linewidth}{!}{%
\begin{tabular}{c c c c c c c c }\toprule
\multicolumn{1}{c }{\textbf{}} & \multicolumn{7}{c }{\textbf{Foraging-2s-15x15-2p-2f-coop-v2 ($p_\textrm{dynamic}$)}} \\  
\cmidrule(lr){2-8}
\multicolumn{1}{ l }{\textbf{Algorithm}} & \textbf{Obs.} & \textbf{Oracle} & \textbf{Masked j. obs.} & \textbf{MD} & \textbf{MD w/ masks} & \textbf{MARO} & \textbf{MARO w/ drop.} \\
\cmidrule{1-8}
\multicolumn{1}{ l }{IQL} & 0.38 \tiny{(-0.03,+0.01)} & 0.48 \tiny{(-0.03,+0.02)} & 0.14 \tiny{(-0.03,+0.02)} & 0.55 \tiny{(-0.02,+0.02)} & 0.53 \tiny{(-0.01,+0.02)} & 0.37 \tiny{(-0.01,+0.01)} & 0.54 \tiny{(-0.0,+0.0)} \\ \cmidrule{1-8}
\multicolumn{1}{ l }{QMIX} & 0.55 \tiny{(-0.01,+0.0)} & 0.68 \tiny{(-0.02,+0.03)} & 0.19 \tiny{(-0.06,+0.04)} & 0.6 \tiny{(-0.02,+0.02)} & 0.6 \tiny{(-0.02,+0.01)} & 0.47 \tiny{(-0.04,+0.02)} & 0.6 \tiny{(-0.01,+0.02)} \\ \cmidrule{1-8}
\multicolumn{1}{ l }{IPPO} & 0.31 \tiny{(-0.0,+0.0)} & 0.44 \tiny{(-0.0,+0.0)} & 0.31 \tiny{(-0.02,+0.01)} & 0.03 \tiny{(-0.03,+0.06)} & 0.02 \tiny{(-0.01,+0.02)} & 0.4 \tiny{(-0.03,+0.03)} & 0.01 \tiny{(-0.0,+0.0)} \\ \cmidrule{1-8}
\multicolumn{1}{ l }{MAPPO} & 0.36 \tiny{(-0.0,+0.0)} & 0.45 \tiny{(-0.01,+0.01)} & 0.3 \tiny{(-0.01,+0.03)} & 0.02 \tiny{(-0.02,+0.03)} & 0.02 \tiny{(-0.02,+0.03)} & 0.4 \tiny{(-0.01,+0.01)} & 0.1 \tiny{(-0.09,+0.07)} \\
\bottomrule
\end{tabular}
}
\end{table}

\begin{figure}
    \centering
    \begin{subfigure}[b]{0.24\textwidth}
        \centering
        \includegraphics[width=0.97\linewidth]{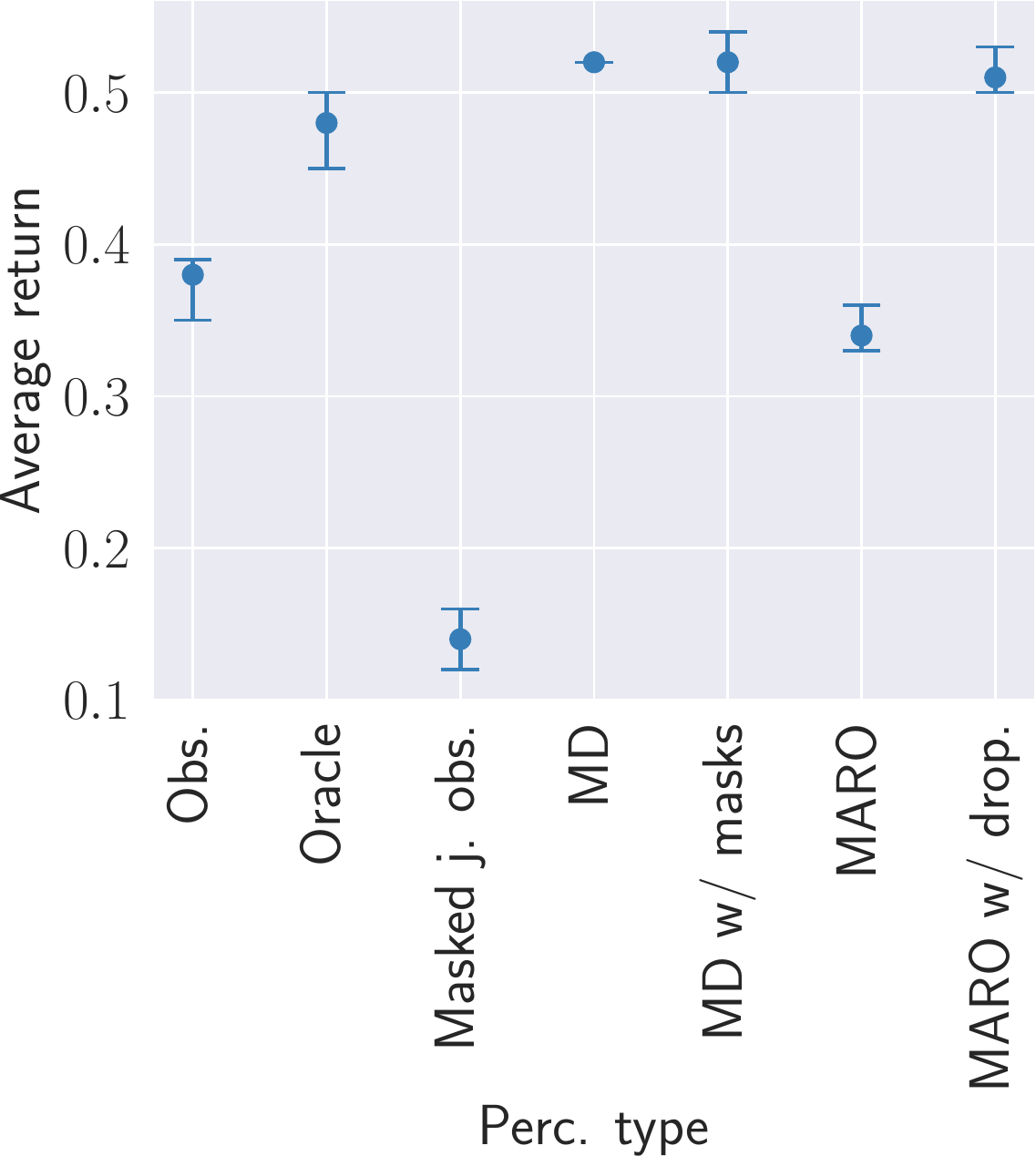}
        \caption{IQL.}
    \end{subfigure}
    \begin{subfigure}[b]{0.24\textwidth}
        \centering
        \includegraphics[width=0.97\linewidth]{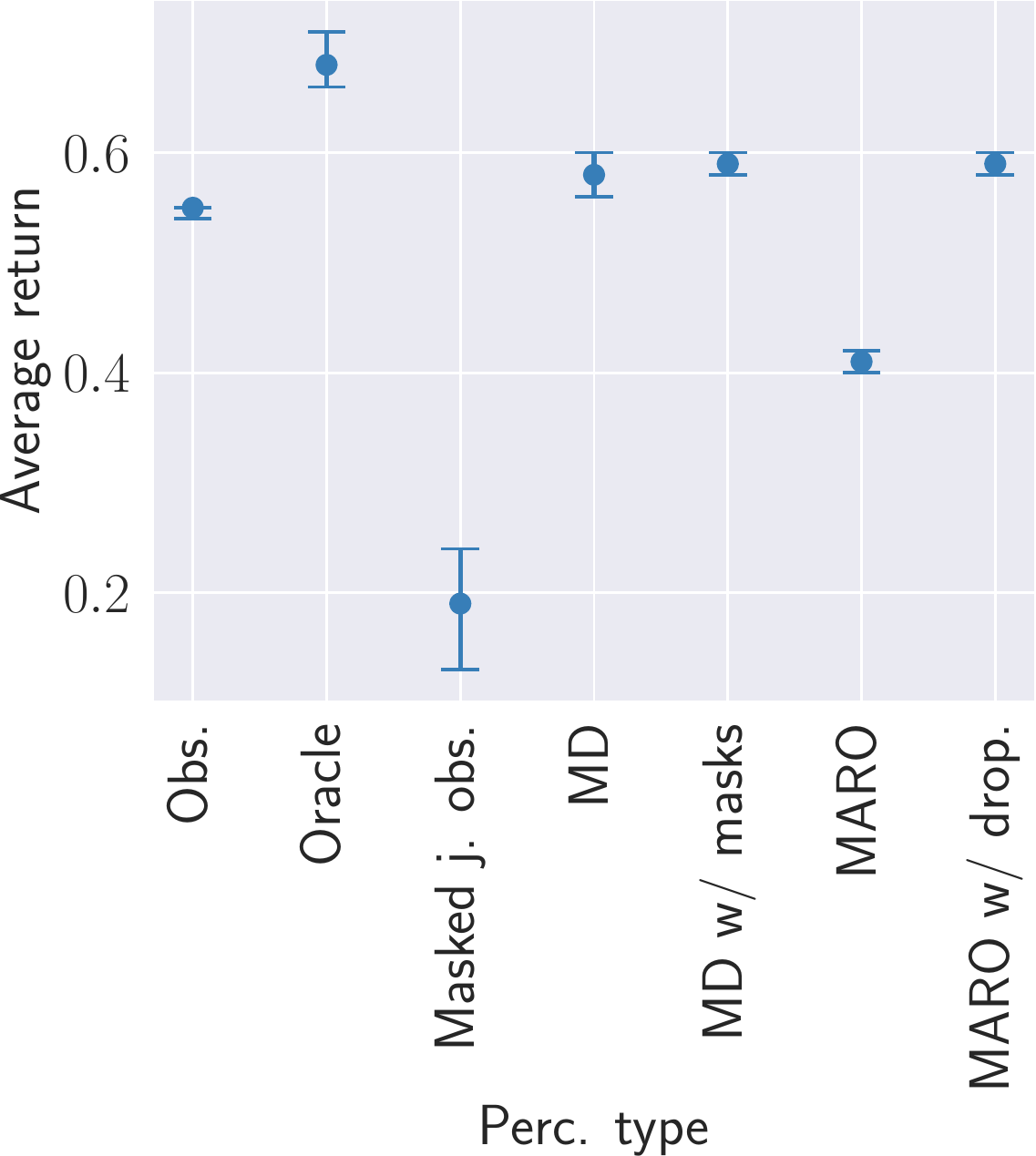}
        \caption{QMIX.}
    \end{subfigure}
    \begin{subfigure}[b]{0.24\textwidth}
        \centering
        \includegraphics[width=0.97\linewidth]{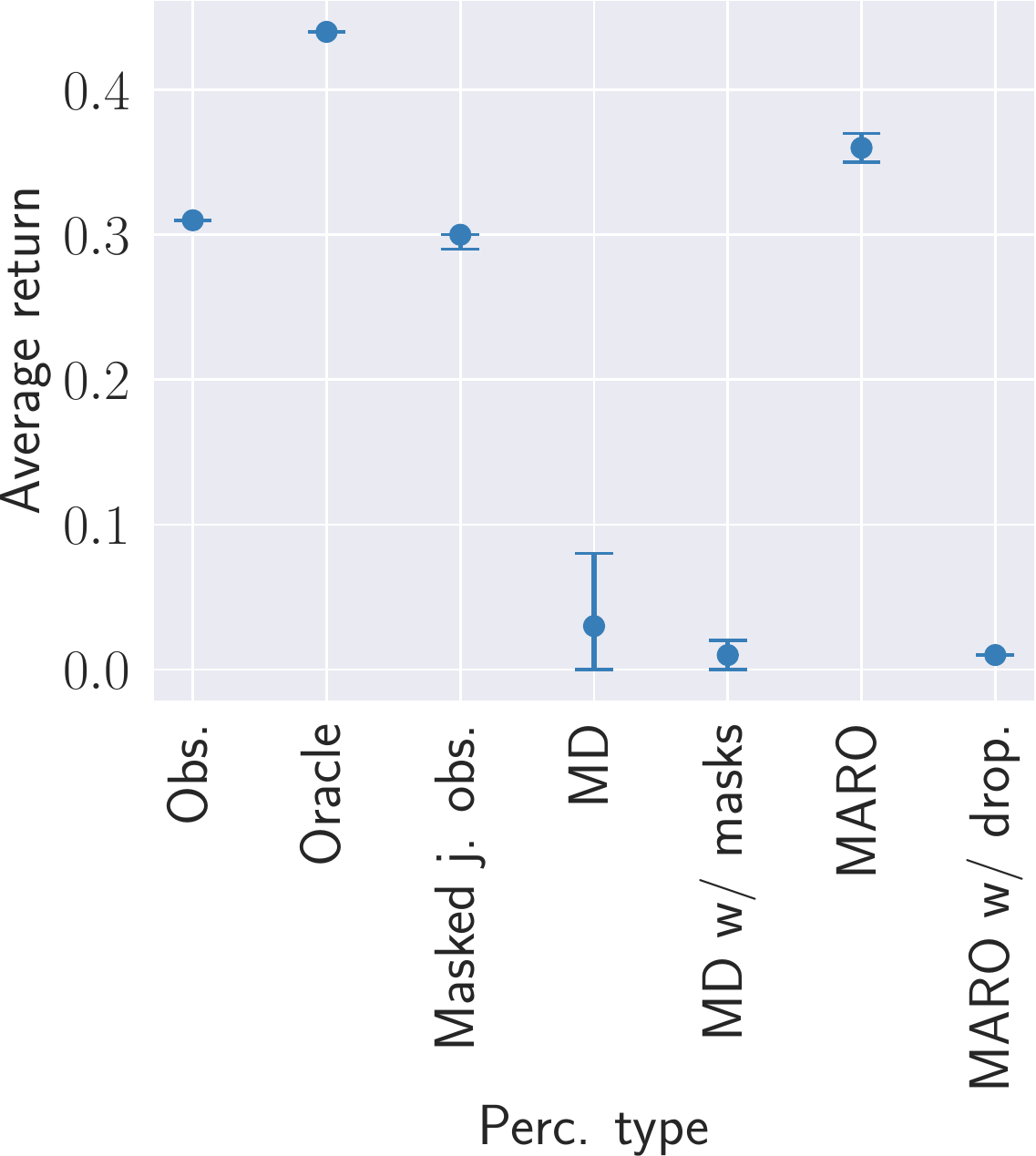}
        \caption{IPPO.}
    \end{subfigure}
    \begin{subfigure}[b]{0.24\textwidth}
        \centering
        \includegraphics[width=0.97\linewidth]{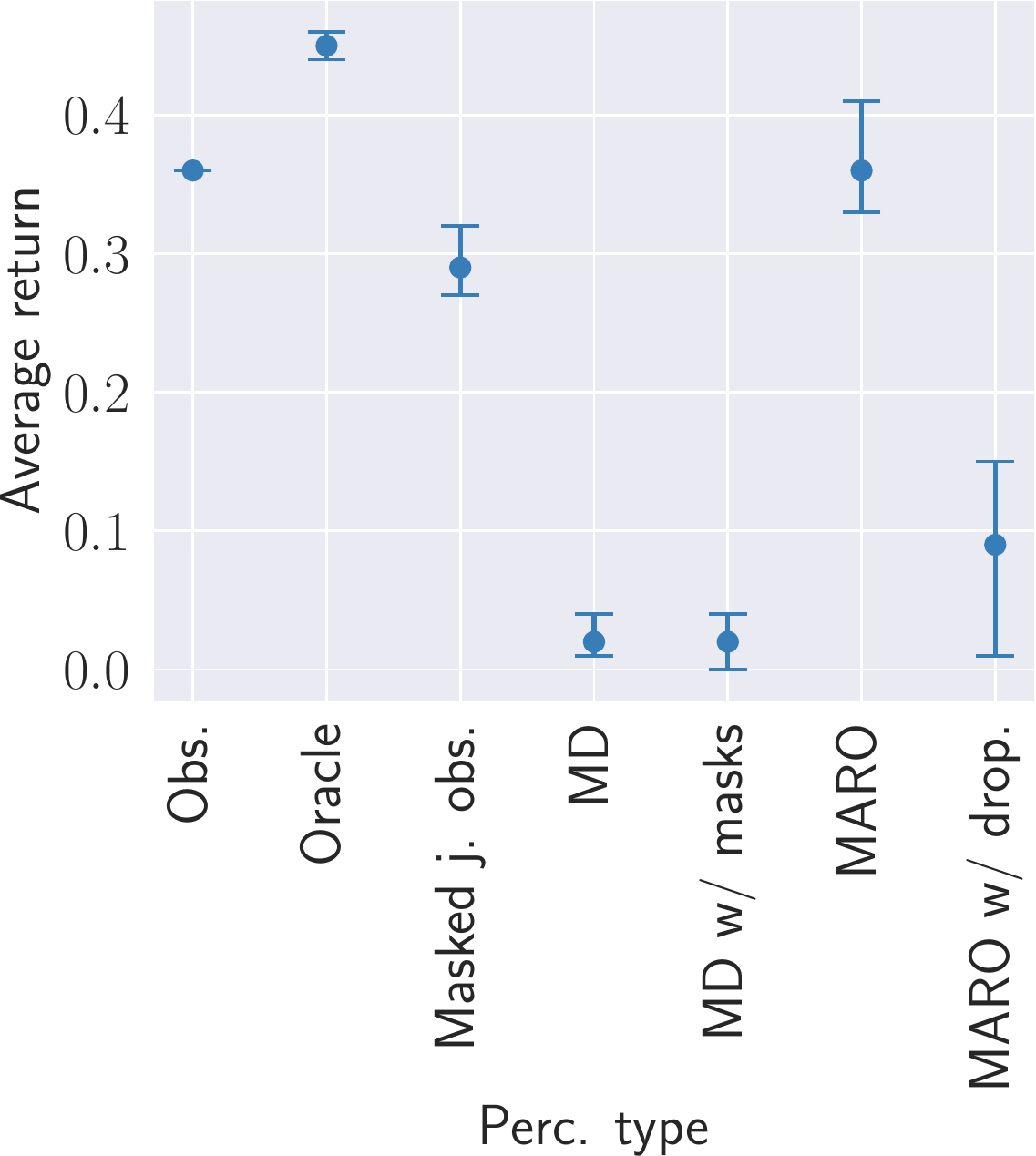}
        \caption{MAPPO.}
    \end{subfigure}
    \caption{(Foraging-2s-15x15-2p-2f-coop-v2) Mean episodic returns for $p_\textrm{asymmetric}$ during training.}
\end{figure}

\begin{figure}
    \centering
    \begin{subfigure}[b]{0.24\textwidth}
        \centering
        \includegraphics[width=0.97\linewidth]{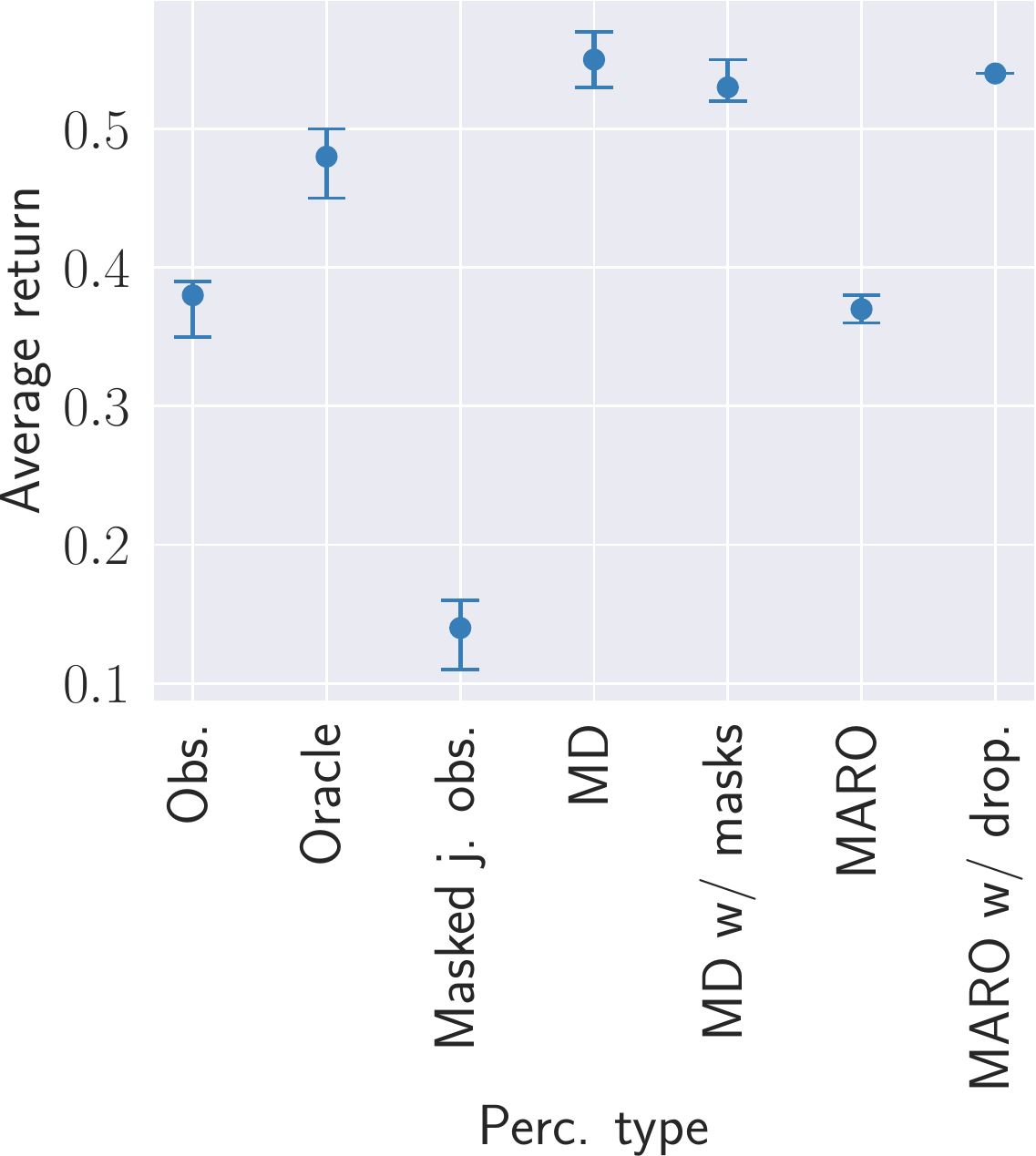}
        \caption{IQL.}
    \end{subfigure}
    \begin{subfigure}[b]{0.24\textwidth}
        \centering
        \includegraphics[width=0.97\linewidth]{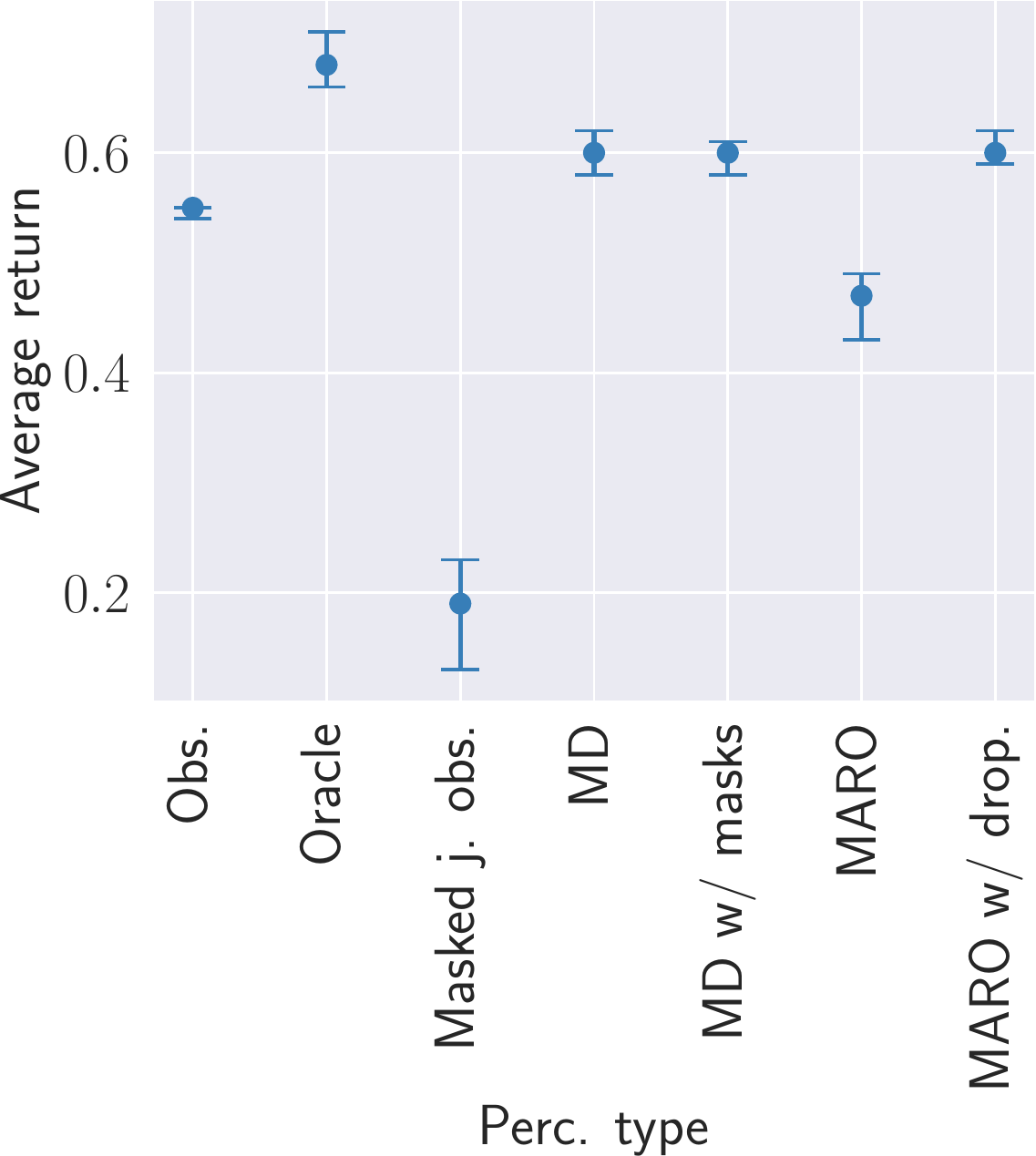}
        \caption{QMIX.}
    \end{subfigure}
    \begin{subfigure}[b]{0.24\textwidth}
        \centering
        \includegraphics[width=0.97\linewidth]{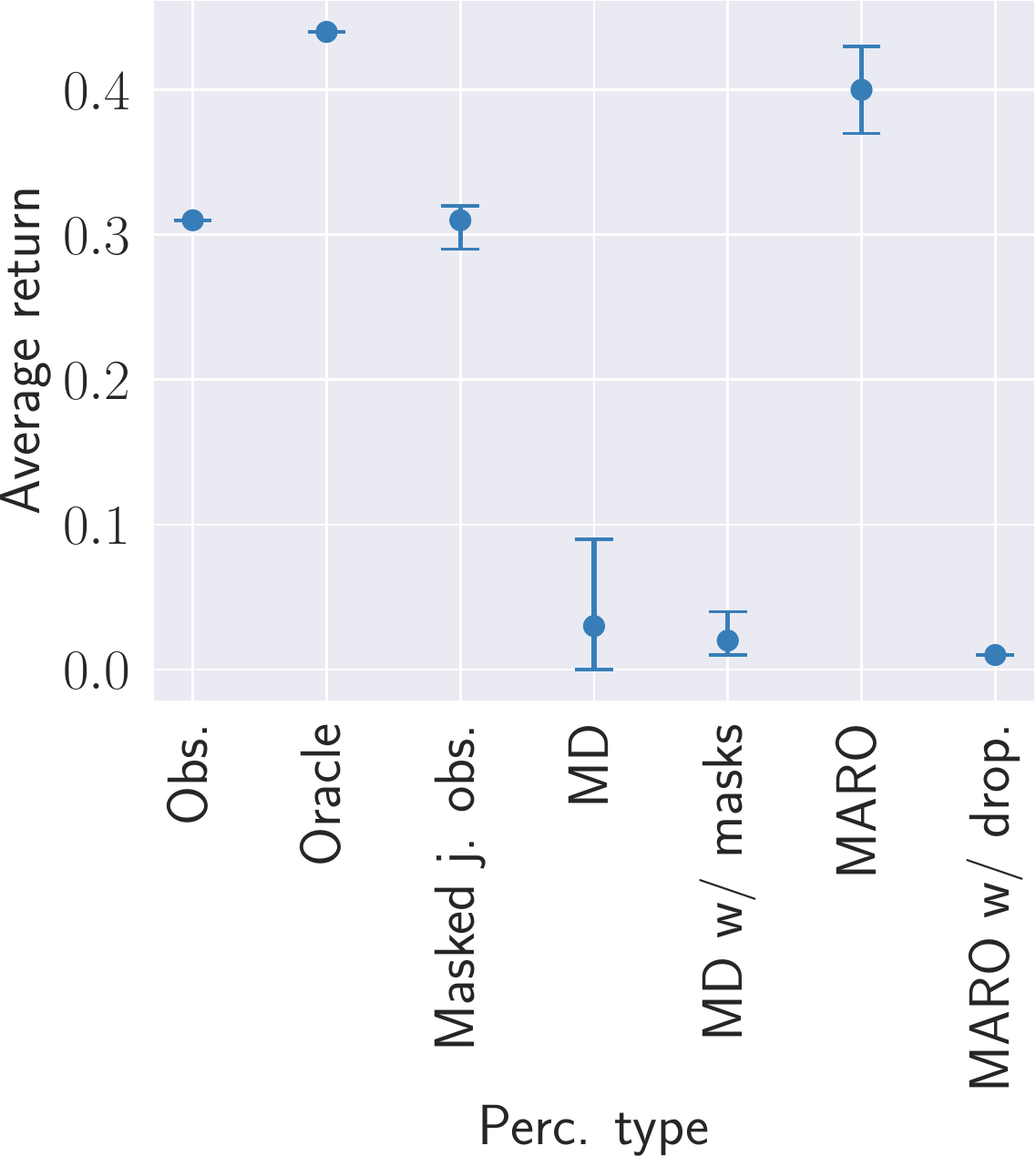}
        \caption{IPPO.}
    \end{subfigure}
    \begin{subfigure}[b]{0.24\textwidth}
        \centering
        \includegraphics[width=0.97\linewidth]{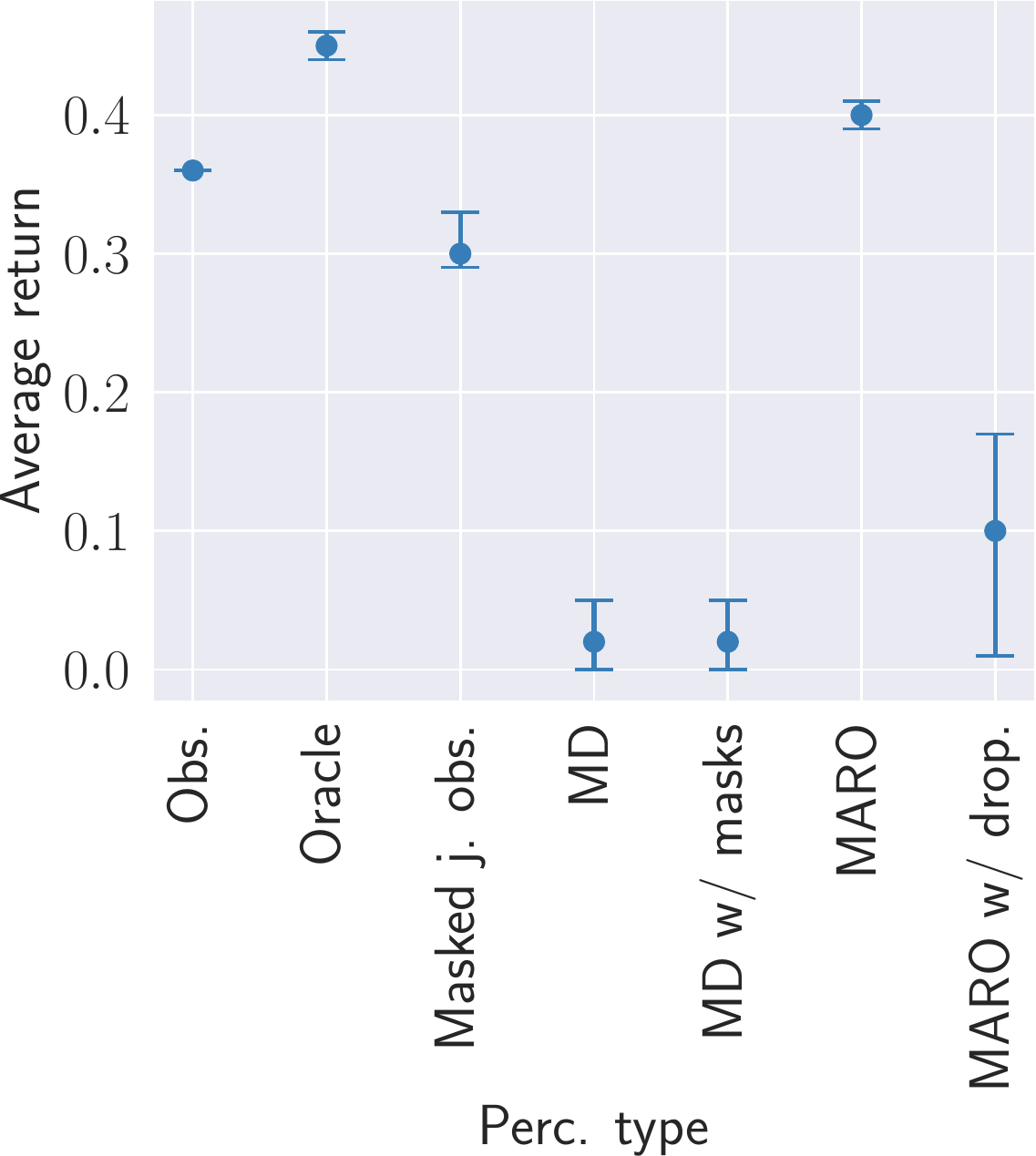}
        \caption{MAPPO.}
    \end{subfigure}
    \caption{(Foraging-2s-15x15-2p-2f-coop-v2) Mean episodic returns for $p_\textrm{dynamic}$ during training.}
\end{figure}


\clearpage
\subsubsection{Multi-agent Trajectory Prediction}
\label{appendix:experimental_evaluation:experimental_results:trajectory_prediction}

\begin{figure*}[h]
    \centering
    \includegraphics[width=\textwidth]{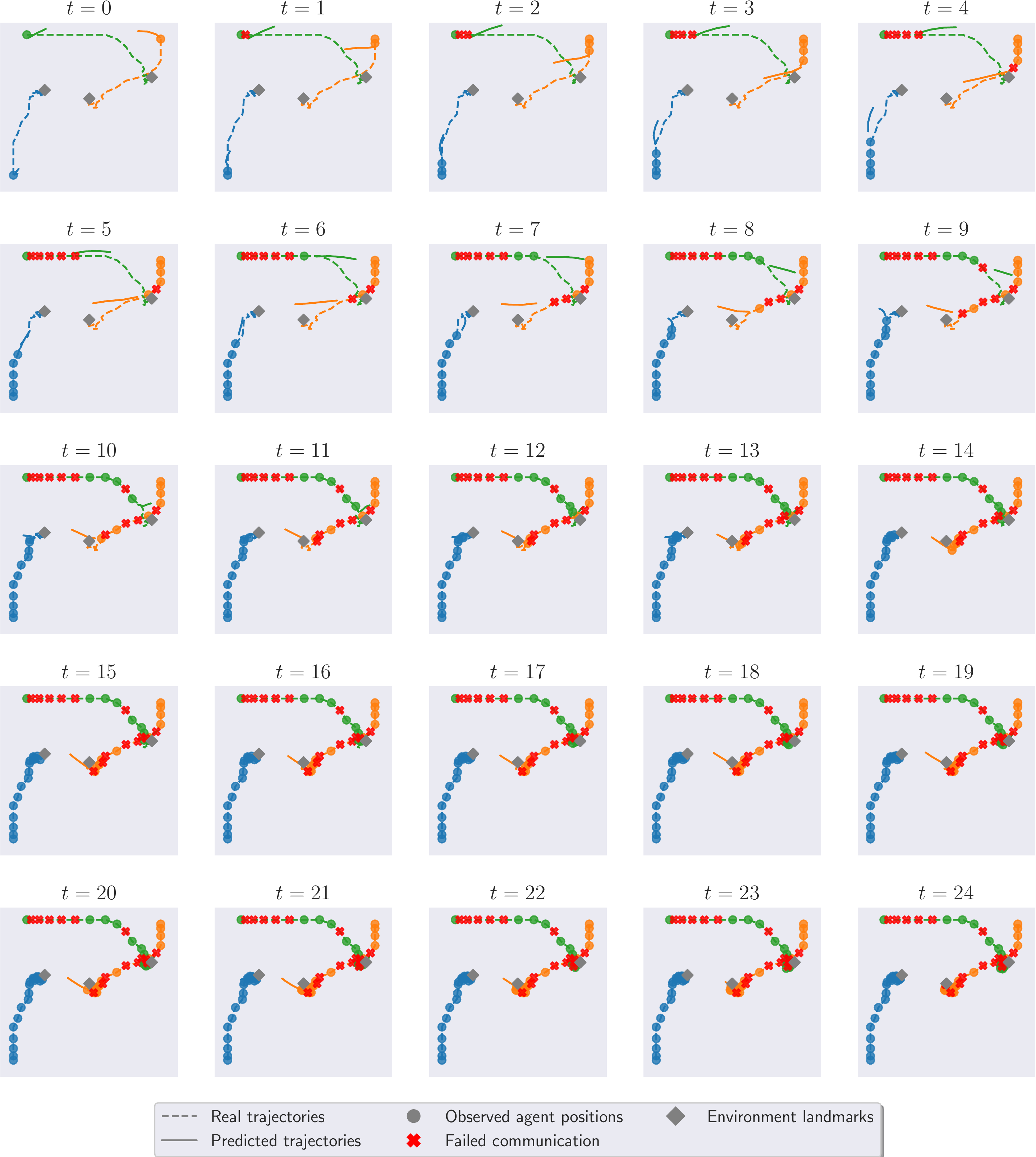}
    \caption{Trajectory prediction plots for the Spreadblindfold environment under the QMIX algorithm from the perspective of agent 0 (blue).}
    \label{appendix:experimental_evaluation:experimental_results:grid_traj_agent0}
\end{figure*}

\begin{figure*}[h]
    \centering
    \includegraphics[width=\textwidth]{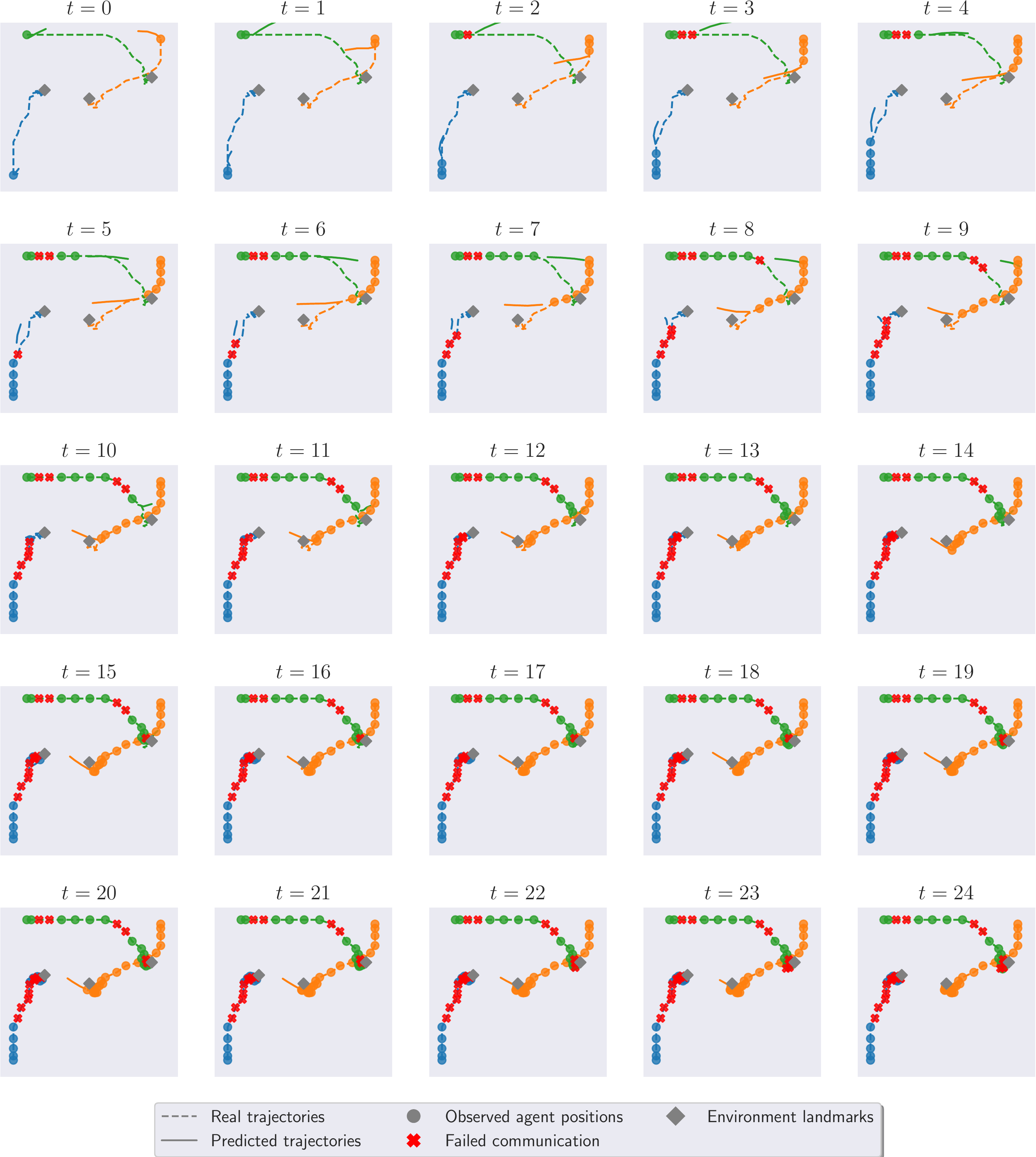}
    \caption{Trajectory prediction plots for the Spreadblindfold environment under the QMIX algorithm from the perspective of agent 1 (orange).}
    \label{appendix:experimental_evaluation:experimental_results:grid_traj_agent1}
\end{figure*}

\begin{figure*}[h]
    \centering
    \includegraphics[width=\textwidth]{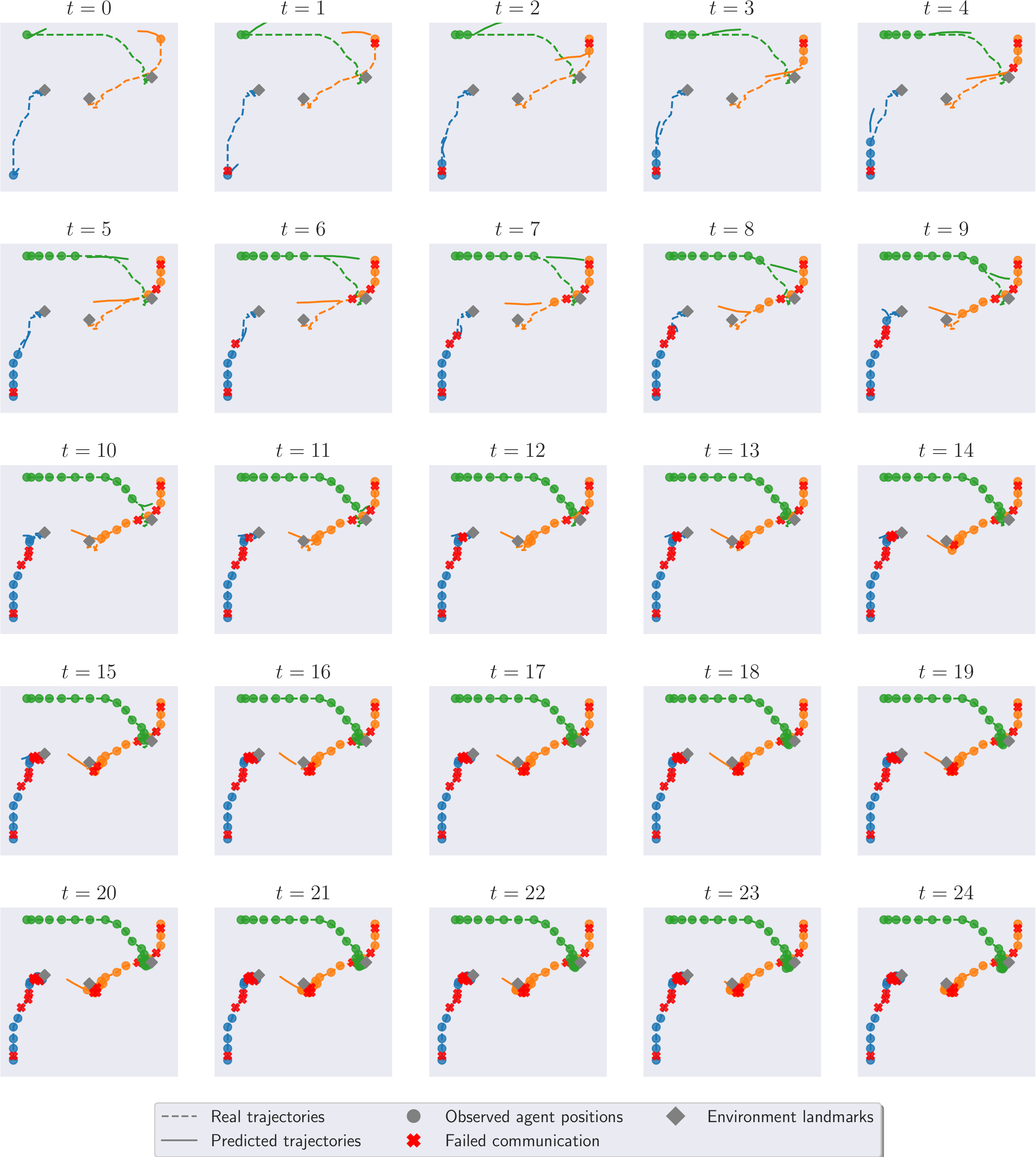}
    \caption{Trajectory prediction plots for the Spreadblindfold environment under the QMIX algorithm from the perspective of agent 2 (green).}
    \label{appendix:experimental_evaluation:experimental_results:grid_traj_agent2}
\end{figure*}

We display, in Figs.~\ref{appendix:experimental_evaluation:experimental_results:grid_traj_agent0}, \ref{appendix:experimental_evaluation:experimental_results:grid_traj_agent1} and \ref{appendix:experimental_evaluation:experimental_results:grid_traj_agent2}, an illustration of the trajectory predictions made by the predictive model from the perspective of each of the agents. The plots are computed, at each timestep and from the perspective of each agent, by computing the estimated trajectories of all agents for the next 4 timesteps. The 4-step ahead predictions are entirely computed using estimated quantities, i.e., real observations are not incorporated into the predictions and the predictive model works in a fully auto-regressive manner.

